\documentclass[review,3p,times]{elsarticle}
\usepackage{amsmath,amsthm,verbatim,amssymb,amsfonts,amscd,graphicx}

\usepackage[utf8]{inputenc}
\usepackage{longtable}
\usepackage{pgfplots}
\usepackage{multirow}
\usepackage{makecell}
\usepackage{enumitem}
\usepackage{xfrac}
\usepackage[title]{appendix}
\usepackage[english]{babel}
\usepackage{pdflscape}
\usepackage{hyperref}
\usepackage{forest}
\usepackage{tikz}
\usetikzlibrary{shapes,arrows}
\usepackage{pgf-pie}

\usepackage[nodisplayskipstretch]{setspace}

\setenumerate{itemsep=0.02cm, topsep=0.05cm}
\setlist[itemize]{itemsep=0.02cm, topsep=0.05cm}

\usepackage[tableposition=top]{caption}  
\usepackage{subcaption}
\usepackage[linesnumbered,ruled]{algorithm2e}

\SetCommentSty{mycommfont}

\def\R{\mathbb R}

\def\O{\mathcal O}

\newcommand{\eps}{\varepsilon}

\usepackage{titlesec}

\newlength{\esummtableheight}  
\newlength{\tsummtableheight}
\newlength{\tableheight}  
\newlength{\negspace}
\usepackage{array}
\tikzstyle{linesAll} = [thick, line join=round, mark phase=1, mark repeat=1]

\def\myMarkerLineWidth{0.25}
\def\myMarkSize{1.6}
\def\myMarkSizeD{0.2}
\tikzstyle{markerCirc} = [mark size=\myMarkSize pt, mark=*, mark options={solid, draw=black, line width=\myMarkerLineWidth pt}]
\tikzstyle{markerAst} = [mark size=\myMarkSize pt, mark=asterisk, mark options={solid, draw=black, line width=\myMarkerLineWidth pt}]
\tikzstyle{markerTria} = [mark size=\myMarkSize+2*\myMarkSizeD pt, mark=triangle*, mark options={solid, draw=black, line width=\myMarkerLineWidth pt}]
\tikzstyle{markerSquare} = [mark size=\myMarkSize-\myMarkSizeD pt, mark=square*, mark options={solid, draw=black, line
width=\myMarkerLineWidth pt}]
\tikzstyle{markerDiam} = [mark size=\myMarkSize+\myMarkSizeD pt, mark=diamond*, mark options={solid, draw=black, line
width=\myMarkerLineWidth pt}]
\tikzstyle{markerO} = [mark size=\myMarkSize pt, mark=o, mark options={solid, draw=black, line width=\myMarkerLineWidth pt}]
\tikzstyle{markerMercedes} = [mark size=\myMarkSize+4*\myMarkSizeD pt, mark=Mercedes star, mark options={solid, draw=black, line width=\myMarkerLineWidth pt}]
\tikzstyle{markerStick} = [mark size=\myMarkSize+4*\myMarkSizeD pt, mark=|, mark options={solid, draw=black, line width=\myMarkerLineWidth pt}]
\tikzstyle{plotStyle1} = [linesAll, markerCirc, color=red, solid]
\tikzstyle{plotStyle2} = [linesAll, markerAst, color=green, dashed]
\tikzstyle{plotStyle3} = [linesAll, markerTria, color=blue, dotted]
\tikzstyle{plotStyle4} = [linesAll, markerSquare, color=magenta, dashdotted]
\tikzstyle{plotStyle5} = [linesAll, markerDiam, color=pink, solid]
\tikzstyle{plotStyle6} = [linesAll, markerO, color=teal, dashed]
\tikzstyle{plotStyle7} = [linesAll, markerMercedes, color=violet, dotted]
\tikzstyle{plotStyle8} = [linesAll, markerStick, color=cyan, dashdotted]

\newtheorem{definition}{Definition}[section]

\pgfplotsset{compat=1.17}

\journal{Expert Systems with Applications}

\begin{document}

\begin{frontmatter}

\title{Comparative Analysis of Optimization Strategies for K-means Clustering in Big Data Contexts: A Review}

\author[a,b,c]{Ravil Mussabayev\corref{cor1}}
\ead{ravmus@uw.edu}

\cortext[cor1]{Corresponding author.}

\affiliation[a]{organization={Department of Mathematics, University of Washington},
            addressline={Padelford Hall C-138}, 
            city={Seattle},
            postcode={98195-4350}, 
            state={WA},
            country={USA}}


\affiliation[b]{organization={Satbayev University},
            addressline={Satbaev str. 22}, 
            city={Almaty},
            postcode={050013},
            country={Kazakhstan}}

\affiliation[c]{organization={Laboratory for Analysis and Modeling of Information Processes, Institute of Information and Computational Technologies},
            addressline={Pushkin str. 125}, 
            city={Almaty},
            postcode={050010}, 
            country={Kazakhstan}}

\author[b,c]{Rustam Mussabayev}
\ead{rustam@iict.kz}

\begin{abstract}
This paper presents a comparative analysis of different optimization techniques for the K-means algorithm in the context of big data. K-means is a widely used clustering algorithm, but it can suffer from scalability issues when dealing with large datasets. The paper explores different approaches to overcome these issues, including parallelization, approximation, and sampling methods. The authors evaluate the performance of various clustering techniques on a large number of benchmark datasets, comparing them according to the dominance criterion provided by the ``less is more'' approach (LIMA), i.e., simultaneously along the dimensions of speed, clustering quality, and simplicity. The results show that different techniques are more suitable for different types of datasets and provide insights into the trade-offs between speed and accuracy in K-means clustering for big data. Overall, the paper offers a comprehensive guide for practitioners and researchers on how to optimize K-means for big data applications.
\end{abstract}

\begin{keyword}
Big data \sep Clustering \sep Minimum sum-of-squares \sep Less-is-more approach (LIMA) \sep Divide and conquer algorithm \sep Decomposition \sep K-means \sep K-means++ \sep Global optimization \sep Partitioning-based clustering \sep Grid-based clustering \sep Hierarchical clustering \sep Density-based clustering \sep Spectral clustering \sep Model-based clustering \sep Sampling \sep Stream-based clustering \sep Parallelized and distributed clustering \sep Triangle inequality \sep Memory-efficient algorithms \sep Canopy clustering \sep Approximation techniques \sep Coresets \sep Hybrid clustering algorithms \sep Genetic algorithm-based clustering
\end{keyword}

\end{frontmatter}

\section{Introduction}

Big data refers to the vast amount of information that is generated and collected in various forms, such as text, images, videos, and sensor readings. This phenomenon is characterized by its large volume, high velocity, and wide variety~\cite{Mauro2015}. These factors make it difficult to manage, process, and analyze such datasets using traditional data processing tools and techniques. The exact threshold for what constitutes ``big'' data varies depending on the context and the available resources for processing and analysis.

For humans, ``big data'' refers to datasets that are impossible to process or analyze manually. Accordingly, for computers, big data represents a scenario where data cannot be processed using standard computational and algorithmic resources. Thus, big data generally refers to datasets that are prohibitively large, complex, or diverse to be handled using traditional data processing methods and average computing resources~\cite{Mussabayev2023}. In other possible situations, there may be either no difficulties with data processing or there are some technical difficulties, but processing is still possible. According to these two situations, the notions of small and large data can be defined, respectively~\cite{Mussabayev2023}. Consequently, big data refers to datasets of such immense size that their processing poses substantial technical challenges or becomes unfeasible when employing conventional methods and standard computing resources. Considering these challenges, it is essential to develop and implement optimized algorithms for a wide spectrum of downstream tasks. These algorithms should have enhanced efficiency and be able to process large datasets using relatively limited computing resources. Alternatively, they should be scalable to accommodate additional computing resources.

When addressing various applied problems, clustering is frequently employed as one of the fundamental approaches. The process of clustering, which groups similar items within a dataset, is integral to solving numerous applied tasks, including those related to big data processing, due to its efficiency and versatility. The importance of clustering is further amplified by the rapid expansion of digital data. For instance, its applications include anomaly detection to uncover irregular patterns~\cite{Tu2020}, customer segmentation for targeted marketing~\cite{Chen2018}, and the analysis of gene expression to decipher genetic information~\cite{Jiang2004}. Clustering is also widely used in image and video processing for various analytical purposes~\cite{Yeung1998}, and in the field of information retrieval to enhance search results~\cite{Djenouri2021}. Furthermore, clustering aids in natural language processing, enabling more effective machine understanding of human language~\cite{Alguliyev2019}, and in bioinformatics, facilitating the analysis of biological data~\cite{Deridder2013}. The method is equally important in network and traffic analysis, improving the management and security of data flow~\cite{Depaire2008}, and in social media analysis, where it helps in understanding complex interaction patterns~\cite{Zhao2011}. Additionally, it plays a crucial role in medical diagnosis by assisting in the identification of diseases based on patient data~\cite{Mittal2021}, in time series analysis for forecasting and detecting trends~\cite{Rakthanmanon2012}, and in pattern recognition and classification, contributing to the categorization of data into distinct groups~\cite{Deridder2013}. Lastly, clustering is fundamental in vector quantization and data compression, optimizing storage and transmission of digital content~\cite{Yin2015}.

Cluster analysis encompasses various models, among which the minimum sum-of-squares clustering (MSSC) is foundational and widely examined~\cite{Aloise2009}. This model addresses the challenge of determining $k$ centroids $C=\left(c_1,\ldots,c_k\right) \in \R^{n \times k}$ from $m$ data points $X={x_1, \ldots, x_m}$ within Euclidean space $\R^n$. The goal is to minimize the sum of squared Euclidean distances between each point $x_i$ and the nearest centroid $c_j$, as articulated in the equation:

\begin{equation}
\min\limits_{C} \ \ \ f\left(C,X\right)=\sum\limits_{i=1}^m \min_{j=1,\ldots,k} \| x_i - c_j \|^2
\label{eq:mssc}
\end{equation}
where $| \cdot |$ denotes the Euclidean norm. This equation~\eqref{eq:mssc} represents the objective function, or the sum-of-squared distances, determining the quality of a partition $X = X_1 \cup \ldots \cup X_k$ by how tightly data points are assigned to their closest centroids $c_j$. For arbitrary $k$ and $m$ values, the main challenge of producing high-quality solutions to the MSSC problem lies in its NP-hard nature~\cite{Aloise2009}, while big data contexts aggravate this problem much further.

In this work, we mainly focus on the type of big data with a large number of objects and a relatively small number of features. Otherwise, MSSC would fail to be an accurate clustering model due to the curse of dimensionality, i.e., Euclidean distances becoming less meaningful with the increase in the feature size.

Viewed as a global optimization problem, MSSC aims to divide a dataset into clusters, optimizing a single objective detailed in~\eqref{eq:mssc} that inherently enhances within-cluster similarity and between-cluster dissimilarity. This optimization criterion is fundamental in assessing the performance of clustering algorithms, with global minimizers offering a more precise reflection of a dataset's inherent clustering structure~\cite{Gribel2019}. However, the search for global minimizers is hindered by the objective function's significant non-convexity, making MSSC a complex yet crucial problem in cluster analysis. The non-convex landscape of the MSSC objective function becomes dramatically more complex as the number of data points increases, which is especially characteristic to big data conditions.

To address these MSSC challenges, which are exacerbated by big data contexts, several approaches have been proposed in the literature to explore the solution space and locate global minimizers which include, but are not limited to: gradient-based optimization techniques~\cite{Karmitsa2018}, stochastic optimization algorithms~\cite{Mussabayev2023}, metaheuristic search strategies~\cite{Gribel2019,Hansen2001}, and hybrid approaches~\cite{Mansueto2021}. Gradient-based techniques~\cite{Karmitsa2018} have a focus on providing a fast convergence to local minimizers but may get trapped in poor solutions due to the non-convex nature of the objective function. On the other hand, stochastic optimization algorithms~\cite{Mussabayev2023} incorporate randomness in the search process to escape local minima and explore a broader solution space. Metaheuristic search strategies~\cite{Gribel2019,Hansen2001} aim to balance exploration and exploitation in the search process. Hybrid approaches~\cite{Mansueto2021} are frequently used, as they can combine multiple different methods with the goal of not only leveraging their advantages but also acquiring new qualitative properties.

Despite the proposition of these numerous optimization techniques aimed at overcoming the high non-convexity challenge of MSSC, each method presents its unique advantages and limitations, indicating the absence of a universally applicable solution, especially in big data environments. Consequently, this underscores the need for continued investigation to devise methodologies that are both more effective and resilient in identifying global minimizers, particularly within the intersection of the MSSC area with the expansive field of big data.

In today's landscape, selecting the right clustering paradigm and tool for new big datasets is not a straightforward endeavor for practitioners. Without proper guidance, comprehending the strengths and constraints of existing clustering models within big data scenarios becomes challenging. Additionally, gaining insights into the range of appropriate big data clustering algorithms tailored for the specific dataset can prove to be daunting.

\begin{figure}[ht]
	\centering
	\begin{subfigure}[b]{0.75\textwidth}
		\centering
		\includegraphics[width=\linewidth]{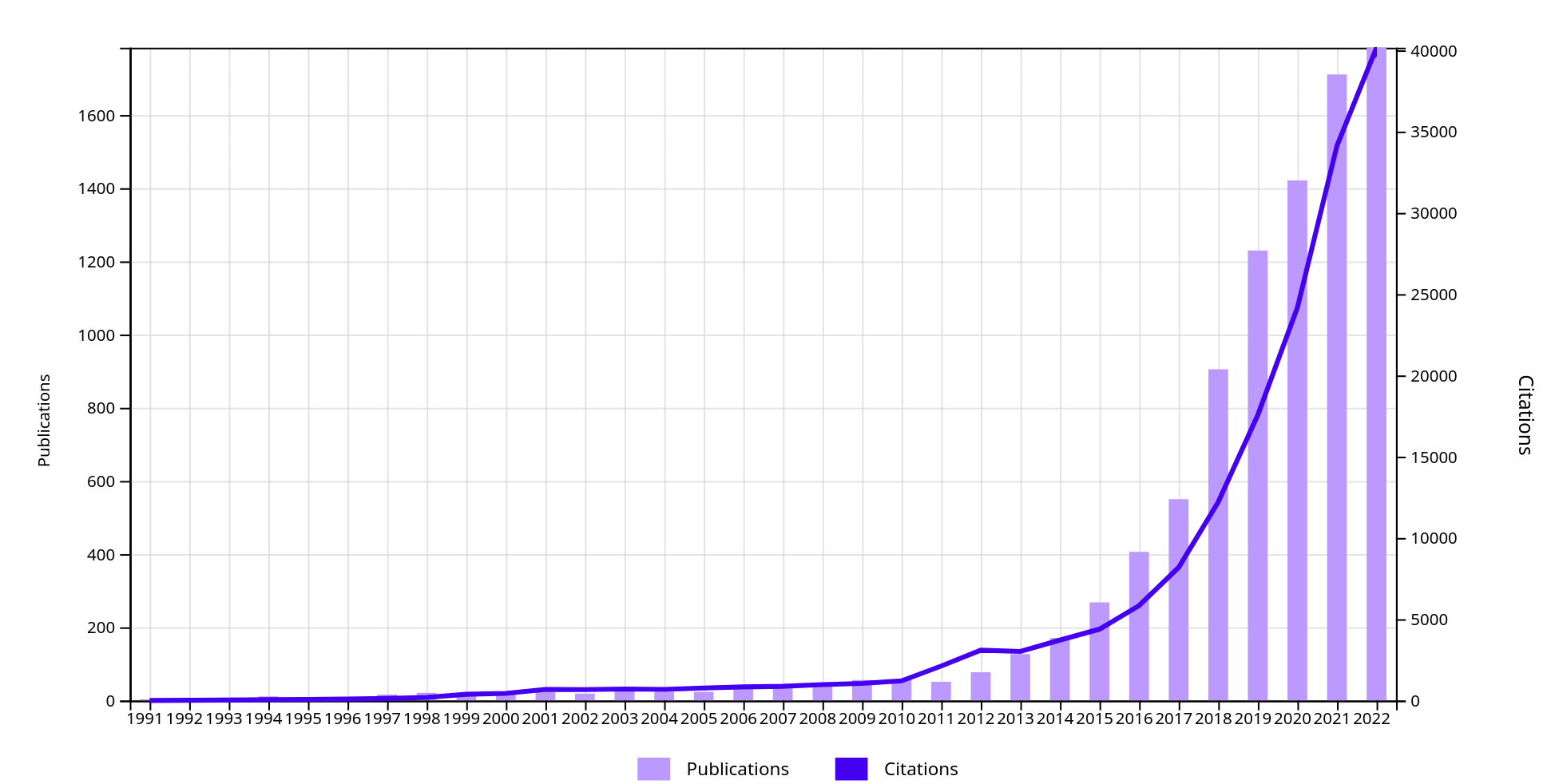}
	\end{subfigure}%
	\begin{subfigure}[b]{0.25\textwidth}
		\centering
		\includegraphics[scale=1.0]{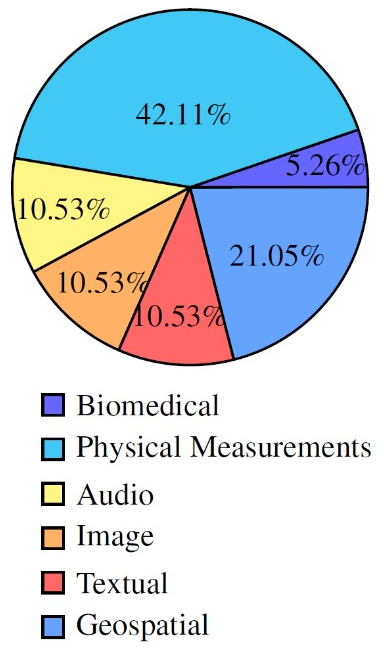}
	\end{subfigure}
	\caption{Trends in big data publications and citations over the years alongside common data types used in big data clustering.}
	\label{fig:pub_growth_data_pie_chart}
\end{figure}

In Figure~\ref{fig:pub_growth_data_pie_chart}, a publication growth graph shows the increase in the number of publications on the topic of big data clustering over time, highlighting its growing relevance. The statistics was collected via the Web of Science citation network by searching the Web of Science Core Collection by keywords ``clustering big data''.

The data types most commonly used for big data clustering are represented on the right side of Figure~\ref{fig:pub_growth_data_pie_chart}. Specifically, the pie chart shows the distribution of different types of data (e.g., text, images, videos, sensor readings) that are frequently used for big data clustering~\cite{Mahdi2021-scalalgs}, including the current study. The statistics for constructing the pie chart was based on the sample of the datasets used in our experimental section. The chart highlights the diversity of data that big data clustering techniques can handle.




The K-means algorithm (Lloyd’s algorithm~\cite{Lloyd1982}) is widely accepted as one of the simplest and most efficient clustering algorithms, which naturally minimizes the MSSC criterion~\eqref{eq:mssc}. It can be considered a classic method for solving the MSSC problem~\cite{Cuong2020}. When applied to small or large data, K-means can find high-quality local solutions in reasonable time~\cite{Jain2010}. K-means can be regarded as a universal clustering algorithm that is often used as a local search in more sophisticated clustering approaches.

While K-means offers a fast local search within the solution space of the MSSC problem, it is not directly applicable to big data due to its time complexity of $\O(m \cdot n \cdot k)$. Making a full pass through the dataset is necessary for K-means, which becomes prohibitive under big data conditions. However, K-means is highly adaptable, allowing optimization at nearly every step for big data processing and offering flexible integration with advanced clustering metaheuristics. Moreover, K-means can still be effectively utilized to accelerate other clustering models, such as density-based clustering~\cite{Bai2017} and spectral clustering~\cite{Filippone2008}. For a more detailed discussion on K-means, its steps, and opportunities for their optimization, we refer the reader to Section~\ref{sec:k_means}.


The minimum sum-of-squares clustering (MSSC) is a more formal and scientific name for the optimization problem~\eqref{eq:mssc}, meaning all methods related to MSSC that are not necessarily similar to K-means. However, within the broad literature, MSSC is also often referred to as the K-means clustering problem due to the popularity, simplicity, and efficiency of the K-means algorithm in solving it. Our primary interest lies in various enhancements and modifications of the K-means algorithm designed to optimize its application to large datasets, which we call ``K-means-like methods''. This paper specifically focuses on these advanced and optimized techniques for improving the standard K-means. While these may include new methods, they must either share some similarities with the standard K-means or employ it within a higher-level heuristic as a form of local search. A review of the methods that are entirely unrelated to K-means yet still address the minimum sum-of-squares clustering (MSSC) problem is of secondary interest to us.

Our work aims to examine various K-means-like methods proposed in existing literature to tackle the complex challenges of clustering big datasets. We will conduct a thorough survey of these methods, categorizing them into distinct groups. For each category, we will explore the latest advanced algorithms they encompass. Moreover, we will evaluate the strengths and limitations inherent in each approach. Towards the conclusion of this paper, a comprehensive experimental analysis will be performed. In our analysis, we adopt the dominance relation based on the ``less is more'' approach (LIMA)~\cite{Brimberg2023} as the principal metric for comparing various big data clustering algorithms. To the best of our knowledge, this study signifies the innovative use of this criterion for evaluating big data clustering algorithms.


The LIMA dominance criterion, as established by Brimberg et al.~\cite{Brimberg2023}, defines a comparative relation among a collection of algorithms. Specifically, it asserts that an algorithm is considered superior to another if it achieves equal or better performance in terms of average accuracy, execution speed, and simplicity. Here, simplicity is quantified by the cardinality of the set of used algorithmic components, and it is required that the superior algorithm strictly outperforms the other in at least one of these dimensions. The aspect of simplicity becomes critical in the context of big data. A comprehensive discussion on the LIMA dominance criterion is provided in Section~\ref{sec:lima_dominance}.

This article for the first time proposes practical recommendations and a flowchart intended to assist in selecting clustering algorithms for big data. We have also undertaken an extensive evaluation of the most cutting-edge K-means-like clustering algorithms for big data, available from the recent literature, using a large array of real-world datasets. These aspects of the current work distinguish it from our previous paper~\cite{Mussabayev2023} that introduced the Big-means clustering algorithm. Moreover, the differentiation stems from the fact that in~\cite{Mussabayev2023}, we aimed to juxtapose Big-means with a sophisticated state-of-the-art heuristic, LMBM-Clust~\cite{Karmitsa2018}, alongside other relatively elementary standard algorithms. Conversely, in the current study, our objective is to conduct an exhaustive survey and comparison of a diverse set of existing approaches for optimizing K-means-like algorithms in the context of big data.

In this paper, Big-means is merely one of several approaches under consideration, although it excels according to the LIMA dominance criterion. Also, it should be noted that our prior research used another parallelization strategy for Big-means, i.e., parallelizing internal K-means and K-means++ algorithms in each iteration. However, the present study uses a novel hybrid parallelization strategy for Big-means, which, according to our recent investigations, has proven to be more effective~\cite{Mussabayev2023-optpar, Mussabayev2023-optparext}.

Thus, the primary aim of this study is to offer a comprehensive review of the techniques for optimizing K-means and, secondly, other MSSC-related clustering approaches in big data environments. We reinforce our findings with extensive experiments and interpret the results using the LIMA dominance criterion. In doing so, we aspire to make a novel and significant contribution to the progress of research in the dynamic field of big data clustering.

Our paper has the following outline. Section~\ref{sec:challenges} describes the main challenges associated with the field of big data clustering. Section~\ref{sec:k_means} discusses K-means in more detail and sets the stage for the subsequent chapters. Section~\ref{sec:approaches} gives a an in-depth analysis of various techniques for optimizing K-means clustering and other clustering approaches related to MSSC in big data environments, exhibiting their notable representative algorithms. Section~\ref{sec:practice} delves into the direct practical conclusions based on the preceding analysis of big data clustering algorithms, providing more useful generalizations, reflection, and guidelines for practitioners. Section~\ref{sec:lima_dominance} presents the concept of LIMA dominance, which is the primary criterion for algorithm comparison in our study. Section~\ref{sec:experiments} conducts a thorough experimental evaluation of the most promising big data clustering algorithms on a multitude of large datasets. Section~\ref{sec:conclusion} draws final conclusions of this study and motivates future research directions. Appendix~\ref{app:details} provides comprehensive details of the conducted experiments.

\section{Problem Challenges} \label{sec:challenges}

Big data clustering is a complex task that faces several challenges that can be grouped into a few categories. One of these categories is related to the computational and practical aspects of clustering algorithms, such as scalability~\cite{Kruger2014-scaldev}, computational complexity~\cite{Mahdi2021-scalalgs}, simplicity~\cite{Mishra2019-fasthybrid}, as well as ease of implementation and parallelization~\cite{Alguliyev2021}.

The second category is concerned with the selection and application of clustering algorithms form the third category. Choosing an appropriate clustering algorithm and its parameters is crucial for effective data analysis, as different data types may require unique clustering methods. With the variety present in big data, it is particularly important to choose suitable clustering techniques to ensure successful insights.

The last category involves specific computational requirements and limitations, such as computational resources~\cite{Sinha2016-survey}, storage and retrieval~\cite{Alazzawe2022-access}, time-dependency, and real-time analysis~\cite{Hassani2016-streams}.


Scalability can be defined as the ability to find the balance between the quality of the obtained solution and the amount of processed information (computational cost) in response to changes in volume and configuration of input data~\cite{Mussabayev2023}. Traditional clustering algorithms that work well with small and large datasets may not be suitable for big data due to its increased computational complexity and memory requirements. Additionally, big data is often stored across multiple machines and storage systems, making it challenging to perform clustering analysis efficiently. To handle big data, clustering algorithms need to be designed to utilize distributed computing architectures and parallel processing~\cite{Kruger2014-scaldev}.

The computational complexity of traditional clustering algorithms can make analyzing big datasets difficult within a reasonable amount of time. This is especially true when there is a need for real-time or near-real-time analysis. As noted by Mahdi et al.~\cite{Mahdi2021-scalalgs}, the time required to perform clustering on large datasets can be prohibitive.

Modern clustering algorithms often prioritize complexity over simplicity, leading to increased time complexity and challenges in implementation across various computing architectures, including parallel systems~\cite{Mishra2019-fasthybrid}. The number and types of instructions used in these algorithms contribute to their complexity. This complexity, while potentially enhancing accuracy, tends to limit their practical use, especially for non-experts. The research trend towards these hybrid and complex algorithms might be a misleading scientific direction. Instead, there is a growing need to focus on developing simple yet effective algorithms that ensure wider applicability and accessibility in practical scenarios.

Parallelizability is essential in the context of clustering algorithms as it refers to an algorithm's capability to decompose the initial, large task into multiple smaller subtasks. This decomposition enables the concurrent (parallel) and independent processing of these subtasks by several processors or workers, shortening the overall processing time. The final solution to the original task is then derived by integrating the solutions of these smaller subtasks. Parallelization enhances an algorithm's efficiency and scalability by effectively decomposing tasks and enabling parallel processing, while also optimizing the use of available computing resources and speeding up data processing. As noted by Alguliyev et al.~\cite{Alguliyev2021}, this is a crucial consideration for any clustering algorithm, particularly those designed to handle big datasets.

Choosing the right clustering algorithm and its parameters is essential for each dataset. There are various clustering algorithms available, each with its own unique strengths and weaknesses. However, selecting the most suitable algorithm for a specific dataset can be challenging, particularly when dealing with large datasets. Different algorithms may have varying computational requirements, and some may not be scalable to large datasets. Furthermore, most advanced clustering algorithms involve a large number of hyperparameters but do not provide clear guidelines for their optimal selection. Performing an ordinary grid search may be unfeasible in big data conditions due to unacceptable time costs. Also, despite the usually better accuracy obtained by hybrid clustering algorithms, the necessity of optimizing a large number of hyperparameters is a negative factor that impedes their widespread adoption.

Clustering big data can require substantial computational resources, including processing power, memory, and storage~\cite{Sinha2016-survey}. Thus, to perform clustering analysis on big data, organizations may need to employ high-performance computing (HPC) technologies, including investing in specialized hardware or cloud computing services.

Storing and retrieving large datasets can present a challenge, particularly when real-time or near-real-time analysis is required. Big data clustering may necessitate specialized distributed storage and retrieval systems capable of handling large data volumes and providing quick access to data for analysis~\cite{Alazzawe2022-access}. Furthermore, some big datasets may be too large to fit into the RAM of a computing system.

Some big datasets may have a time-dependent nature, so the patterns and relationships in the data may change over time~\cite{Hassani2016-streams}. Clustering algorithms may need to be adjusted to handle the time-dependency of the data and ensure that the clustering results remain relevant. Real-time or near-real-time clustering of big data can be challenging because of the need for fast processing and the large volume of data. Developing real-time clustering algorithms that can handle big data is an active area of research~\cite{Hassani2016-streams}.

To summarize, all the above points show that the emergence of big data entails a set of unique challenges. To extract insights and value from vast datasets, there is a need for specialized clustering techniques (``true big data'' algorithms). These algorithms are specifically designed to address the big data issues and the inadequacy of classic approaches. Nowadays, this is an active area of research. In this paper, we will review the numerous attempts made in the literature to create big data clustering algorithms.


\section{K-means Algorithm} \label{sec:k_means}

Partitioning-based clustering methods, especially K-means~\cite{Lloyd1982}, excel in big data analysis by creating non-overlapping subsets where each data point is associated with a single cluster. These methods, which also include K-medoids~\cite{Kaufman1990_K_medoids} and CLARA~\cite{Kaufman1990_Clara}, are efficient and scalable, making them ideal for handling large datasets. They start with initial centroids and refine these through iterative optimization to reduce within-cluster variance. Despite the necessity to predetermine the number of clusters and their sensitivity to initial centroid placements, partitioning-based algorithms remain central to big data clustering tasks. Other classes like grid-based algorithms (STING~\cite{Wang1997}, CLIQUE~\cite{Agrawal1998}), hierarchical algorithms (Ward's~\cite{Ward1963}, CURE~\cite{Guha1998}), density-based methods (DBSCAN~\cite{Ester1996}, OPTICS~\cite{Ankerst1999}), spectral clustering techniques~\cite{Luxburg2007,Huang2020}, and model-based approaches (EM algorithm for Gaussian Mixture Models~\cite{Yang2012}, Hierarchical Dirichlet Processes~\cite{Teh2006}), each have their own strengths when applied to small datasets or adapted for use in a K-means-like pipeline. However, in big data contexts, partitioning-based algorithms are particularly significant due to their time complexity, namely K-means, underscoring their relevance to this study.

As highlighted in the introduction, K-means is regarded as one of the most popular methods for addressing the MSSC problem. Its popularity stems from the method's relatively high effectiveness, coupled with its simplicity and versatility. A pseudocode for the K-means algorithm is provided in Algorithm~\ref{alg:k_means}.

\begin{algorithm}
	\SetAlgoLined
	\KwResult{Determine optimal centroids $C$ and allocate dataset $X$ into clusters $Y$ using the K-means algorithm.}
	Initialize centroids $C$ uniformly at random in the convex hull of $X$\;
	Initialize the iteration counter $t = 0$\;
	\While{there is a change in centroids $C$ or $t < T$}{
		\For{every point $x$ in $X$}{
			Assign $x$ to its nearest centroid in $C$\;
		}
		\For{each centroid $c_i$ in $C$}{
			Recalculate $c_i$ to be the centroid (mean) of all points currently assigned to it\;
		}
		Increment iteration counter $t \leftarrow t + 1$\;
	}
	Allocate each point in $X$ to its closest centroid in $C$, forming cluster assignments $Y$.
	\caption{K-Means Clustering Algorithm}
	\label{alg:k_means}
\end{algorithm}

K-means is an iterative algorithm consisting of two alternating steps: assignment and update. The algorithm's input is an initial set of $k$ centroids $C = \left(c_1, \ldots, c_k\right)$ and some stopping criterion, which can be a maximum number of iterations or a tolerance on the distance between two consecutive solutions. In the assignment step, each point $x_i$ is assigned to the nearest centroid from $C$. In the update step, the centroids are updated by assigning the means of points in each cell $X_j$ of the resulting partition: $c_j=\frac{1}{\left|X_j\right|}\sum_{x_i\in X_j} x_i$. These two steps are repeated until the stopping criterion is met.

Due to its simple and straightforward structure, K-means is amenable to optimization at nearly every stage in general and specifically for big data. First, before running the K-means algorithm, it might be advisable to perform normalization or dimensionality reduction on the input data. Second, the choice of initial centroids (step 1 in Algorithm~\ref{alg:k_means}) is an important part of the K-means algorithm, which significantly impacts the accuracy of the resulting clustering solution~\cite{Franti2019}. The classical version of K-means algorithm initially assigns centroids to clusters using a uniform random initialization. However, it has been established that the clustering solution obtained with a random initialization of K-means may be significantly worse in terms of the objective function when compared to the optimal clustering solution. This means that relying solely on random initialization can lead to suboptimal clustering results~\cite{Hennig2016}. Therefore, improving initialization is another method for optimizing the performance of K-means.

The initialization issue of the classical K-means algorithm can be addressed through two main approaches. The first one involves selecting the initial solution carefully, as suggested by Franti et el.~\cite{Franti2019} and Ismkhan et el.~\cite{Ismkhan2018}. The other approach is embedding the K-means algorithm into a global optimization method as a local search procedure, as proposed by Franti et al.~\cite{Franti2019}. Thus, numerous search methods, both local~\cite{Hansen2001} and global~\cite{Mansueto2021}, have been proposed to offer improved solutions.

One of the most effective heuristics for K-means initialization is K-means++~\cite{Arthur2007}. K-means++ works as follows: the first cluster center $c_1$ is chosen randomly from all data points. Each subsequent cluster center $c_j$ is selected from the remaining data points with a probability proportional to its squared distance from the existing cluster centers. The pseudocode of K-means++ can be found in Algorithm~\ref{alg:k_means_pp}. While K-means++ initialization takes longer than a random initialization, it ensures that K-means requires fewer iterations to reach an optimum. Additionally, K-means++ often produces a smaller objective function value at the final solution compared to a naive random initialization. In \cite{Makarychev2020}, Makarychev et al. demonstrated that the expected cost of the solution produced by K-means++ is at most $5(\log k + 2)$ times the optimal solution's cost. In the meantime, the time complexity of the K-means++ initialization is $\O(m \cdot n \cdot k)$. Hence, it is also not directly applicable to big data. However, the K-means++ algorithm can be efficiently parallelized for multi-core processing of large data. In K-means and K-means++, parallelization can be used to compute the distances between centers and data points. In K-means++, it is also possible to parallelize the computation of the probabilities for selecting each data point as a center.


\begin{algorithm}
	\SetAlgoLined
	\KwResult{Determine optimal centroids $C$ utilizing the K-means++ strategy.}
	Randomly select the first centroid $c_1$ from dataset $X$\;
	\For{$i = 2$ {\bf to} $k$}{
		Compute $d(x_j)$ as the minimum distance from any point $x_j \in X$ to the nearest selected centroid among ${c_1,\ldots,c_{i-1}}$\;
		Select a new centroid $c_i = x_t \in X$ probabilistically, where the chance of choosing $x_t$ is proportional to $\frac{d(x_t)^2}{\sum_{x_j \in X} d(x_j)^2}$\;
	}
	Set $C = \{c_1, c_2, \ldots, c_k\}$ as the resulting set of centroids.
	\caption{K-Means++ Clustering Procedure}
	\label{alg:k_means_pp}
\end{algorithm}

A simpler but less efficient initialization method is Forgy~\cite{Forgy1965}, which also samples an initial solution from the original data points, but uniformly at random. Multi-start K-means~\cite{Franti2019} is yet another initialization method, which executes several random initializations and picks the best one in the end.

The main loop of the K-means algorithm (line 3 of Algorithm~\ref{alg:k_means}) can be parallelized both in a parallel multi-start manner and at the level of internal parallelization of steps within each iteration (distance calculations at step 5 of Algorithm~\ref{alg:k_means}). Also, the distance calculation step at line 5 of Algorithm~\ref{alg:k_means} can be optimized using the triangle inequality~\cite{Elkan2003,Moodi2022}.

During the operation of the K-means algorithm (line 4 of Algorithm~\ref{alg:k_means}), various ``divide and conquer'' techniques can also be applied, where initially the data are divided and distributed among parallel workers, solutions for small subtasks are obtained, which are then combined at the final stage to produce a solution for the entire task. Decomposition of input data into subtasks can be accomplished in multiple ways, such as simple uniformly random sampling~\cite{Mussabayev2023} or more advanced techniques like coresets~\cite{Bachem2018LWC} and canopies~\cite{McCallum2000}. The coreset method is known for its efficiency and accuracy in reducing the size of large datasets without losing important information.

During the parallelization process, advanced optimization methods can also be employed, such as collective or genetic algorithms, among others. In this approach, K-means is integrated as a local search procedure within higher-level heuristics such as MDEClust~\cite{Mansueto2021}, HG-means~\cite{Gribel2019}, or Big-means~\cite{Mussabayev2023}. K-means can be optimized by using High-Performance Computing (HPC) technologies such as vector computations based on SIMD technology and others implemented in Numba~\cite{Lam2015, Marowka2018}, among others.

\section{K-means Optimization Approaches} \label{sec:approaches}

The literature proposes numerous techniques to optimize K-means and address the challenges of MSSC in big data contexts, as discussed above. In this section, we will review these techniques and classify them into separate categories. For each category, we will consider its corresponding state-of-the-art algorithms. Also, we will analyze the strengths and weaknesses of each approach and conduct an extensive experimental analysis. The objective of this study is to offer a thorough review of the techniques used for optimizing K-means clustering and other clustering approaches related to MSSC in big data environments, as the MSSC concept is often used interchangeably with K-means clustering.

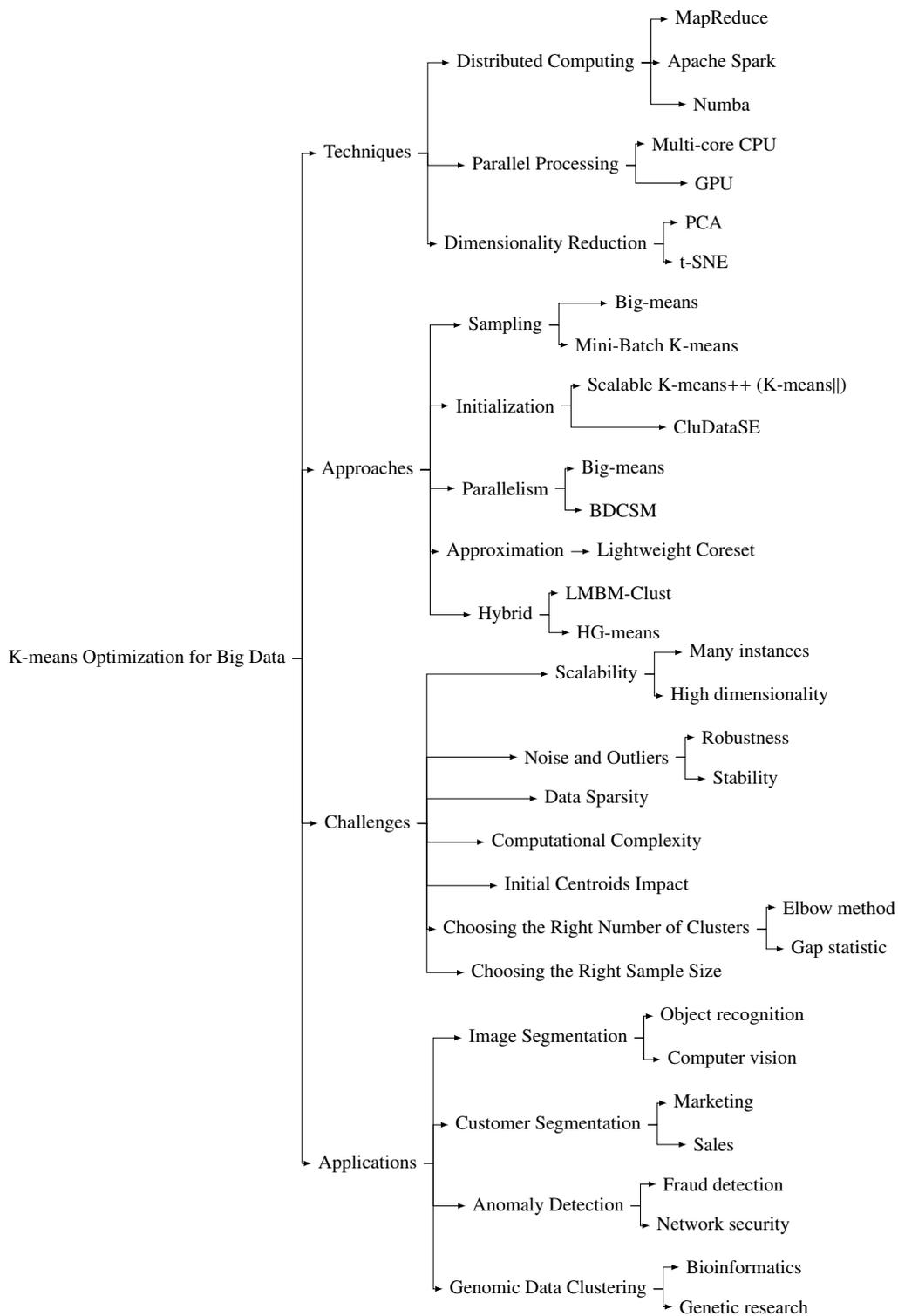
\begin{figure}
    \centering
    \resizebox{!}{20cm}{%
    \begin{forest}
    for tree={
        grow'=0,
        child anchor=west,
        parent anchor=east,
        edge path={
            \noexpand\path[\forestoption{edge}]
            (!u.parent anchor) -- +(5pt,0) |- (.child anchor)\forestoption{edge label};
        },
        l sep=10pt,
        edge={->,>=latex},
    }
    [K-means Optimization for Big Data
        [Techniques
            [Distributed Computing
                [MapReduce]
                [Apache Spark]
                [Numba]
            ]
            [Parallel Processing
                [Multi-core CPU]
                [GPU]
            ]
            [Dimensionality Reduction
                [PCA]
                [t-SNE]
            ]
        ]
        [Approaches
            [Sampling
                [Big-means]
                [Mini-Batch K-means]
            ]
            [Initialization
                [Scalable K-means++ (K-means$\|$)]
                [CluDataSE]
            ]
            [Parallelism
                [Big-means]
                [BDCSM]
            ]
            [Approximation
                [Lightweight Coreset]
            ]
            [Hybrid
                [LMBM-Clust]
                [HG-means]
            ]
        ]
        [Challenges
            [Scalability
                [Many instances]
                [High dimensionality]
            ]
            [Noise and Outliers
                [Robustness]
                [Stability]
            ]
            [Data Sparsity]
            [Computational Complexity]
            [Initial Centroids Impact]
            [Choosing the Right Number of Clusters
                [Elbow method]
                [Gap statistic]
            ]
            [Choosing the Right Sample Size]
        ]
        [Applications
            [Image Segmentation
                [Object recognition]
                [Computer vision]
            ]
            [Customer Segmentation
                [Marketing]
                [Sales]
            ]
            [Anomaly Detection
                [Fraud detection]
                [Network security]
            ]
            [Genomic Data Clustering
                [Bioinformatics]
                [Genetic research]
            ]
        ]
    ]
    \end{forest}
    }
    \caption{Ontological graph of the problem area and its main technologies}
    \label{fig:ontological_graph}
\end{figure}

Figure~\ref{fig:ontological_graph} shows an ontological graph that illustrates the problem area and its main technologies used in optimizing K-means for big data. It shows the relationships between different concepts and technologies, and how they contribute to the overall field.

\begin{figure}[ht]
    \centering
    \includegraphics[scale=0.6]{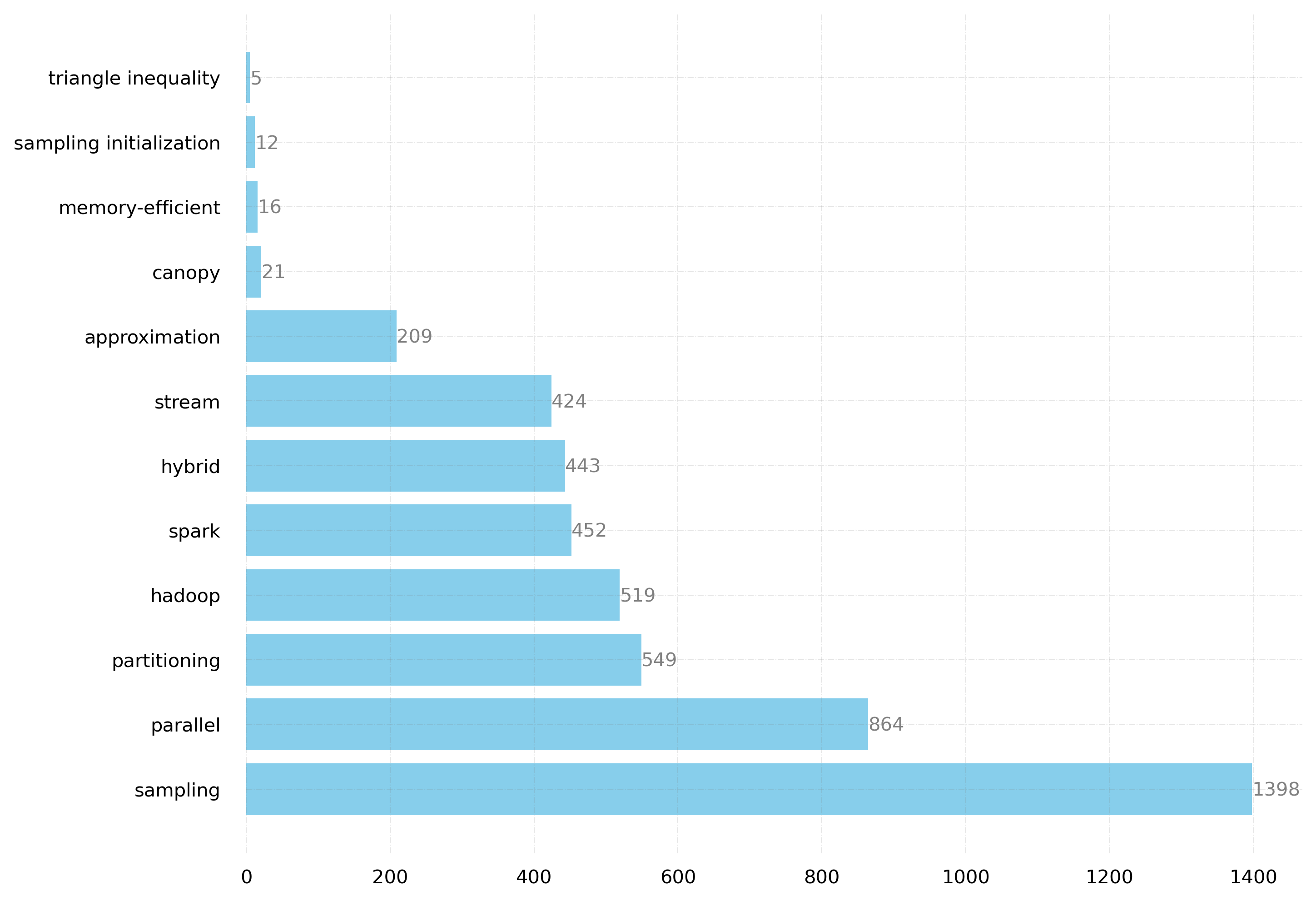}
    \caption{Histogram of most commonly used approaches and technologies in the field of big data clustering}
    \label{fig:appr_histogram}
\end{figure}

Figure~\ref{fig:appr_histogram} shows the frequency of different big data clustering approaches or technologies used in the literature. It can help readers understand which methods are most commonly used or discussed in the field. Each number in the histogram signifies the number of published papers with the given keyword contained in the title or the abstract. The statistics was obtained by searching the Web of Science Core Collection by keywords.

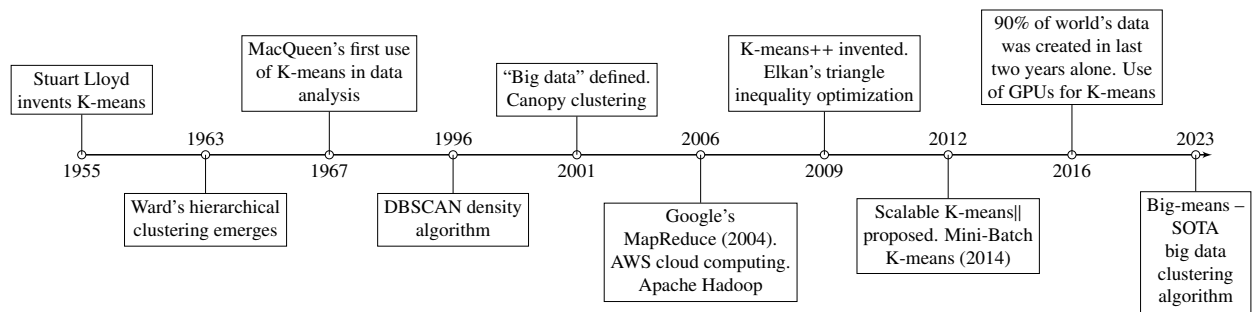
\begin{figure}[ht]
	\centering
	\def\start{1955}
	\def\stop{2023}
	\resizebox{\linewidth}{!}{
		\begin{tikzpicture}[x=0.3cm, y=0.3cm]
			\draw[line width=0.8pt, ->, >=latex'](0,0) -- (\stop-\start+1,0);
			
			\newcommand{\events}{\start/Stuart Lloyd\\invents K-means/4,1963/Ward's hierarchical\\clustering emerges/-4,1967/MacQueen's first use\\of K-means in data\\analysis/5,1996/DBSCAN density\\algorithm/-4,2001/``Big data'' defined.\\Canopy clustering/4,2006/Google's\\MapReduce (2004).\\AWS cloud computing.\\Apache Hadoop/-6,2009/K-means++ invented.\\Elkan's triangle\\inequality optimization/5,2012/Scalable K-means$||$\\proposed. Mini-Batch\\K-means (2014)/-5,2016/90\% of world's data\\was created in last\\two years alone. Use\\of GPUs for K-means/6,2023/Big-means --\\SOTA\\big data\\clustering\\algorithm/-6}
			
			\newcounter{total}
			\foreach \year/\desc in \events {
				\stepcounter{total}
			}
			
			\newcounter{index}
			\foreach \x/\desc/\ypos in \events {
					\pgfmathsetmacro{\xpos}{(\stop - \start) / (\thetotal - 1) * \theindex}
					\filldraw[fill=white] (\xpos,0) circle (0.25); 
					\node[draw, rectangle, align=center, fill=white, yshift=0pt] (box) at (\xpos, \ypos) {\desc}; 
					\draw (\xpos,0) -- (box); 
					\stepcounter{index}
					
					\ifdim\ypos pt < 0pt
					\node at (\xpos,1.0) {\x}; 
					\else
					\node at (\xpos,-1.0) {\x}; 
					\fi
				}
		\end{tikzpicture}
	}
	\caption{Timeline of milestones in K-means clustering optimization and other MSSC algorithms}
	\label{fig:timeline}
\end{figure}

Figure~\ref{fig:timeline} shows a timeline of key developments in K-means clustering optimization and other MSSC algorithms.

\subsection{Data decomposition}

This approach involves partitioning the data into smaller subsets and performing clustering on each subset independently. The clustering of each subset can be considered as a separate subproblem. When all the subproblems have been solved, their solutions are combined to form a solution for the original problem.

Decomposition can be performed in different ways:
\begin{itemize}
\item Partitioning. This can be achieved using techniques such as vertical partitioning or horizontal partitioning. In vertical partitioning, the data is partitioned based on the features, whereas in horizontal partitioning, the data is partitioned based on the instances. This approach can be effective when the data is too large to fit into memory or when the computation time is a bottleneck;
\item Sampling. This approach involves selecting a random subset of the data and clustering it. The resulting clusters can then be used as an approximation of the clusters that would be obtained from clustering the entire dataset. This approach can be effective when the data is uniformly distributed and the clusters are well separated;
\item Stream-based clustering. This approach is also called the online or incremental clustering. It involves clustering data points in small batches in real-time as they arrive, without needing to store the entire dataset in memory. Thus, clusters are updated as new batches arrive. This approach can be effective when the data is continuously streaming, and real-time clustering is required (the clusters need to be updated dynamically).
\end{itemize}
All three decomposition methods are tightly interrelated. Each method can be naturally transformed into another one.

Minibatch K-means~\cite{Sculley2010} is a basic algorithm that utilizes the sampling approach. In each iteration of the MiniBatch K-means algorithm, a minibatch (a random subset of the data) is selected for processing. Then this minibatch goes through the assignment and update steps, as in the standard K-means. However, rather than updating each centroid by averaging all data points assigned to it, the update is made using a fraction of the points in the minibatch. Specifically, the new centroid position is a weighted average between its old position and the mean of the new points assigned to it. The weights ensure more recent data has a larger influence on the centroid position, allowing the algorithm to better adapt to changes in the data distribution.

The algorithm continues to iterate, selecting new minibatches and repeating the assignment and update steps, until a stopping criterion is met. This could be a fixed number of iterations, or a point where the centroids no longer move significantly between iterations.

Thus, the Minibatch K-means algorithm consists of the following key steps:
\begin{enumerate}
\item Initialization. Randomly select $k$ points from the dataset to act as the initial cluster centroids;
\item Sampling. Draw a random minibatch $M$ of size $s$ from the dataset;
\item Assignment. Assign each data point in the minibatch to its nearest centroid;
\item Update. Update the centroids by calculating the mean of all data points assigned to each centroid;
\item Repeat. Repeat steps 2 -- 4 until the centroids no longer move significantly.
\end{enumerate}

The pseudocode for the Minibarch K-means algorithm is showed in Algorithm~\ref{alg:minibatch_kmeans}.

\begin{algorithm}
\SetAlgoLined
\KwResult{Determine centroids $C$ and allocate dataset $X$ to clusters $Y$ via the Minibatch K-means method.}
Randomly select $k$ centroids $C$ from dataset $X$\;
Initialize iteration counter $t = 0$\;
\While{$t < T$}{
Select a minibatch $M$ of $m$ samples randomly from $X$\;
\For{each sample $x$ in $M$}{
Link $x$ to its closest centroid in $C$\;
}
\For{every centroid $c$ in $C$}{
Recalculate $c$ as the average of all samples linked to $c$\;
}
Increment iteration counter $t \leftarrow t + 1$\;
}
Map each sample in $X$ to its closest centroid in $C$, forming $Y$\;
\caption{Minibatch K-means Clustering Approach}
\label{alg:minibatch_kmeans}
\end{algorithm}

In each iteration of Minibatch K-means, the centroids are updated only using the points of an incoming minibatch, making the algorithm more memory-efficient and scalable than traditional K-means. However, the quality of the solution may be drastically affected by the randomness of the minibatch selection, and lots of random initializations may be required to obtain an optimal result. The Big-means algorithm~\cite{Mussabayev2023} addresses these limitations by implementing a stable and effective K-means++ initialization for the initial sampled subset and degenerate clusters. It also introduces an advanced centroid update criterion based on the optimal objective function value obtained from all processed samples. Furthermore, Big-means deepens the exploration of each new sample by employing a K-means local search strategy. All these modifications significantly enhance the stability and precision of the clustering outcomes~\cite{Mussabayev2023}.

The Big-means algorithm represents a synthesis of three key approaches to data decomposition. This simple method begins by sampling a subset $S$ from the dataset $X$, containing $s$ data points, a small fraction of the total $m$. It leverages the K-means++ algorithm for the initial selection of centroids $C$ from $S$. Subsequently, it employs a dynamic clustering process, utilizing K-means with the most effective centroid set found across previous samples. This iterative refinement, guided by the ``keep the best'' strategy, continually updates to the incumbent solution, based on the sum-of-squares criterion evaluated on current samples.

Unique to Big-means is its strategy for handling empty clusters: it rejuvenates these by reapplying K-means++ to produce new potential centers, thus avoiding the pitfalls of traditional methods and enhancing the solution's quality by broadening the search for minimizing the objective function. Another pivotal feature of Big-means is the ``shaking procedure'', which generates new samples to perturb the current centroid solution, thereby infusing diversity and adaptability into the clustering process. By treating the dataset as a point cloud in the $n$-dimensional space, each sample acts as a sparse representation, injecting fresh perspectives into each iteration.

Upon reaching a predetermined time limit or processed sample count, Big-means assigns all data points to the best obtained centroid configuration $C$. However, this step can be either omitted or adjusted according to the specific requirements of an application area. The algorithm's iteration time complexity stands at $\O(s \cdot n \cdot k)$, offering a significant speed advantage for big datasets, especially when compared to the conventional K-means or K-means++ algorithms.

Big-means is designed to marry the swift execution of K-means with high-quality clustering outcomes through intelligent successive refinements across random data subsets, adeptly managing empty clusters. This iterative, sample-based approach not only ensures scalability for big data but also mitigates the risk of settling into suboptimal solutions by continually refreshing the search space and centroids. The algorithm's stochastic optimization approach, through iterative random sampling, ventures into diverse objective function landscapes, enhancing the likelihood of discovering superior configurations than possible with a single K-means execution. Big-means exemplifies a blend of deterministic and probabilistic strategies to achieve a balance between exploration and exploitation in the solution space. This balance is crucial for the algorithm's ability to navigate through and beyond local optima, towards a globally optimal clustering arrangement.

The research in~\cite{Mussabayev2023} benchmarks Big-means against established clustering algorithms, revealing its superior performance in accuracy and speed across various dataset sizes, thereby setting a new standard for ``true big data'' clustering methodologies. Algorithm~\ref{alg:big_means} encapsulates the Big-means algorithm.

\begin{algorithm}
\SetAlgoLined
\KwResult{Finalize centroids $C$ and allocate dataset $X$ into clusters $Y$ utilizing the Big-means technique.}
Mark all $k$ centroids $C$ as uninitialized\;
Set the best objective function value $\hat{f}$ to infinity\;
Initialize iteration counter $t$ to 0\;
\While{$t < T$}{
Randomly select a sample $S$ from $X$ with size $s$\;
\For{each centroid $c$ in $C$}{
\If{$c$ is linked with an empty cluster}{
\text{Refresh $c$ by applying K-means++ to $S$}\;
}
}
Calculate updated centroids $C_{\text{new}}$ by executing K-means on $S$, starting with $C$\;
\If{the objective $f(C_{\text{new}}, S)$ is less than $\hat{f}$}{
Update $C$ to $C_{\text{new}}$\;
Set $\hat{f}$ to $f(C_{\text{new}}, S)$\;
}
Increment $t$ by 1\;
}
Assign each data point in $X$ to its nearest centroid in $C$, forming $Y$\;
\caption{Big-means Clustering Algorithm}
\label{alg:big_means}
\end{algorithm}

%
%

By iterating through samples and refining centroid positions, Big-means efficiently addresses the computational complexity challenge by reducing computational demands while preventing stagnation at inferior solutions, a common issue inherent to traditional approaches. By treating each sample as an incoming time-dependent data chunk, Big-means can also tackle the challenge of real-time analysis in big data. Big-means can accept non-overlapping samples, facilitating the distribution of input data across separate computing nodes, thereby addressing the challenge of data storage and retrieval within big data environments. Importantly, these advantages of Big-means do not increase the algorithm's time complexity, nor do they compromise the simplicity of its implementation.

Alguliyev et al. took a different approach in their Big Data Clustering on a Single Machine (BDCSM) algorithm in~\cite{Alguliyev2021}. BDCSM partitions large datasets into chunks, clusters them in parallel using K-means, and aggregates the resulting cluster centers into a final pool. Then, the algorithm clusters the pool using K-means. The pseudocode for the BDCSM algorithm is presented in Algorithm~\ref{alg:bdscm}. BDCSM is a prominent instance of partition-based parallel clustering algorithms. While BDCSM is a straightforward algorithm that leverages parallel processing to accelerate clustering and partially addresses the challenge of distributed storage for big data, it fails to address the initialization issue, resulting in reduced clustering accuracy (effectiveness) despite its gains in efficiency.

The empirical evaluation of BDCSM in the original paper showed its superiority over the classical K-means algorithm~\cite{Alguliyev2021}. However, this evaluation did not demonstrate any superiority of BDCSM to other more advanced algorithms for clustering large datasets.

\begin{algorithm}
\SetAlgoLined
\KwResult{Compute the final centroids $C$ and cluster assignments $Y$ for a dataset $X$ using the BDCSM algorithm.}
Partition the dataset $X$ into chunks of size $p$\;
\For{each chunk $C_i$}{
    Cluster $C_i$ using K-means to obtain centroids $C_{i, \text{new}}$\;
    Add $C_{i, \text{new}}$ to the pool of centroids $P$\;
}
Cluster the pool $P$ using K-means to obtain final centroids $C_{\text{final}}$\;
Assign each point in $X$ to the nearest centroid in $C_{\text{final}}$ to obtain final cluster assignments $Y$\;
\caption{BDCSM Clustering}
\label{alg:bdscm}
\end{algorithm}


Notable examples of online clustering algorithms include online K-means~\cite{Cohen2021-online} and online spectral clustering~\cite{Ning2010-incremental}. The pseudocode for the online K-means clustering algorithm is provided in Algorithm~\ref{alg:online_kmeans}. Online K-means updates the cluster centroids incrementally as each new data point arrives, without the need to access the entire dataset. This gives the Online K-means algorithm an advantage over traditional K-means in terms of processing real-time data and scalability to large datasets. However, it still shares several limitations with Minibatch K-means and BDCSM, including centroid initialization issues, convergence to local optima, and ignoring degenerate clusters. Furthermore, Online K-means cannot process entire chunks of incoming real-time data, making the final clustering outcome highly sensitive to the sequence of incoming data points.

\begin{algorithm}
	\SetAlgoLined
	\KwResult{Compute the final centroids $C$ and cluster assignments $Y$ for a streaming dataset $X$ using the Online K-means algorithm.}
	Initialize $C$ with $k$ centroids chosen from the first batch of data\;
	\While{new data point $x$ arrives}{
		Find the nearest centroid $c_i$ in $C$ to $x$\;
		Update $c_i$ by incorporating $x$ into the calculation:
		$c_i = (c_i \cdot n_i + x) / (n_i + 1)$\;
		Increment the count $n_i$ for centroid $c_i$\;
	}
	Optionally, periodically re-evaluate and adjust centroids with a batch process\;
	Assign each point in $X$ to the nearest centroid in $C$ to obtain final cluster assignments $Y$\;
	\caption{Online K-means Clustering}
	\label{alg:online_kmeans}
\end{algorithm}

With sufficient computational resources, data decomposition enables the application of advanced parallel and distributed techniques to individual subproblems, as explored in the next subsection, unlocking vast opportunities for more efficient processing and analysis.

\subsection{Parallelization and distributed computing}

Parallelization and distributed computing are two common techniques for optimizing K-means-like methods in big data settings~\cite{Alguliyev2021}. Most commonly, parallelization involves breaking the data into smaller subsets and clustering them simultaneously on multiple processors. The results are then combined to generate the final clusters. This approach can be useful when the data is too large to fit into memory or when the computation time is a bottleneck. Alternatively, big data can be distributed across multiple machines. In this case, clustering is performed in a distributed manner, using frameworks such as Apache Hadoop or Apache Spark. This approach can be effective when the data is too large to fit on a single machine. By distributing the data and computations, the workload can be divided among multiple machines, reducing the processing time and allowing for scalability.

Apart from the BDCSM algorithm discussed above, paper~\cite{MohamedAymen2019} proposes the ``Sampling, Triangle inequality and MapReduce'' (STiMR) K-means algorithm, which leverages the MapReduce framework at every stage of the clustering process. STiMR represents a typical approach in the class of parallel and distributed clustering algorithms, employing the $map$ and $reduce$ operations on groups of input data points. Specifically, it leverages MapReduce to draw a random sample from the big dataset using reservoir sampling, cluster the sample using K-means with the triangle inequality, and assign the data in the original dataset to the obtained centroids. Technically, the STiMR K-means algorithm is an accelerated version of applying K-means to a single random sample, with the added benefit of MapReduce parallelization at every stage.


However, the logic of some other big data clustering algorithms allows for other types of parallelization. For instance, independent or somewhat dependent loops inside a clustering algorithm can be naturally parallelized. Big-means algorithms~\cite{Mussabayev2023} is a good example that is amenable to this kind of parallelization. Specifically, each execution of the loop at line 4 in Algorithm~\ref{alg:big_means} can be assigned to a separate parallel worker. These workers can be totally independent from one another (competitive parallelization) or share information about the best sample solutions (collective parallelization). The former mode promotes exploration among the initial solutions, while the latter ensures a thorough exploitation of the best initial solution. Additionally, one can consider a hybrid parallelization mode, in which for some number of iterations the competitive mode is employed until being switched to the collective mode. The hybrid mode is expected to combine both benefits of the competitive and collective parallelization schemes, achieving an optimal trade-off between them.

Throughout our experimental phase, we leveraged the Numba library for Python to parallelize only those specific parts of the big data clustering algorithms that were explicitly highlighted as parallelizable within their original research papers. Numba~\cite{Lam2015, Marowka2018} is a key instrument in high-performance computing, featuring optimization capabilities such as parallelization, multi-threading, and vectorization. These features are core strategies in performance optimization, transforming the execution speed of Python functions, loops, and numerical computations. Numba's dynamic generation of optimized machine code for both CPUs and GPUs further contributes to this performance boost, converging Python's usability and the speed of lower-level languages.


While the MapReduce framework may offer advantages in terms of scalability and fault tolerance, particularly for very large datasets, we decided not to use it in our experiments. Our goal was to ensure that all algorithms being tested were parallelized under equal conditions, in order to enable fair and accurate comparison of their performance in terms of parallelizability. Thus, we parallelized all algorithms using the Numba library, which allowed us to control and optimize the parallelization process for each individual algorithm.

To maximize memory efficiency in MSSC clustering, the next subsection describes how algorithms may employ more frugal data structures to represent the dataset, which are then either directly passed to K-means for iterative refinement or in some other way adapted to yield an MSSC solution.


\subsection{Memory-efficient algorithms}

Memory-efficient clustering algorithms are designed to handle datasets that are too large to fit into memory. These algorithms typically take advantage of some form of data summarization or use disk-based or online methods to manage memory usage. Often, memory-efficient data structures are employed to reduce the memory footprint. Memory-efficient clustering algorithms can be effective when the memory available is limited. They can be incredibly useful for big data applications where datasets may consist of millions or billions of data points, far beyond what traditional clustering algorithms can handle.

Here are a few examples of memory-efficient clustering algorithms and techniques:
\begin{enumerate}
\item BIRCH~\cite{Zhang1996} (Balanced Iterative Reducing and Clustering using Hierarchies): BIRCH is an efficient clustering method designed for large datasets that may not fit into memory. It creates a memory-efficient, hierarchical data structure called a CF (Clustering Feature) Tree, which holds summarized information about the data, rather than storing all data points. The CF Tree is built in a single pass through the data, where each data point is either integrated into an existing cluster or used to create a new one, based on a distance threshold. By storing and manipulating summaries of data points, rather than the entire dataset, BIRCH is able to perform clustering operations more efficiently, making it particularly well-suited for handling big data. BIRCH can indirectly solve the MSSC problem by using the centroids of the micro-clusters formed by the leaf nodes of its CF Tree as initial seeds for the K-means algorithm, which refines these centroids to minimize the within-cluster sum of squared distances. This two-step approach combines BIRCH's efficiency in handling large datasets with the precision of MSSC algorithms in optimizing cluster centroids;

\item DBSCAN~\cite{Ester1996} with Incremental Updates: Variations of the popular DBSCAN algorithm have been developed that incrementally update the clustering structure with the arrival of new data points, reducing the memory required to store the entire dataset at once. Thus, these algorithms can apply a density-based clustering to portions of a dataset one at a time, without the need to process the whole dataset at once. This version of DBSCAN can be adapted to the MSSC paradigm by using it as preprocessing step on each newly arriving data portion, and passing the obtained centroids as initialization to the K-means clustering on the data portion;

\item CluStream~\cite{Aggarwal2003}: The CluStream algorithm effectively clusters evolving data streams by creating and updating lightweight microclusters in real-time with each incoming data point. These microclusters, each summarized by a ``Clustering Feature'' vector, dynamically evolve, allowing for the creation of new microclusters, deletion of outdated ones, or merging of close clusters, providing an adaptive response to changes in data distribution. Periodically, these microclusters are consolidated into macroclusters using the K-means algorithm, which offer a longer-term, comprehensive view of the data stream's overall clustering structure;

\item Canopy Clustering~\cite{McCallum2000}: The Canopy Clustering algorithm is a pre-clustering technique used to speed up other clustering algorithms by reducing the size of the dataset. The algorithm works by scanning through the data and creating canopies (overlapping groups of data points) using cheap distance measures, which form a first approximation of the true clusters. These canopies are then used to partition the data, reducing the computational cost for a subsequent run of a more expensive clustering algorithm like K-means or hierarchical clustering. Canopy clustering is often used as a preprocessing step for the K-means algorithm or a hierarchical clustering algorithm.
\end{enumerate}

By employing memory-efficient algorithms, it becomes possible to perform clustering on big datasets. However, these memory-efficient techniques often require careful tuning and consideration of their parameters, and might make assumptions or approximations that could limit their effectiveness depending on the specific application.

Several memory-efficient hierarchical clustering algorithms have been developed that are scalable to handle large datasets. Although these algorithms do not employ K-means and cannot be classified as MSSC algorithms in a direct sense, they indirectly achieve the minimization of the sum-of-squares through their operations. While BIRCH (Balanced Iterative Reducing and Clustering using Hierarchies)~\cite{Zhang1996} constructs a tree-like data structure to represent data points and merges similar clusters, CURE~\cite{Guha1998} (Clustering Using Representatives) takes a different approach: it selects a fixed number of well-scattered points to represent each cluster.



\begin{algorithm}
\SetAlgoLined
\KwResult{Compute the final centroids $C$ and cluster assignments $A$ for a dataset $X$ using the CURE algorithm.}
\textbf{Initialization:}\\
Take a random sample of size $s$ from $X$ and partition into $p = s / (f * k * q)$ partitions, each of size $q * k$\;

\For{each partition}{
    Apply hierarchical clustering to form $q$ clusters\;
    Select $c$ representative points for each cluster, move them $\alpha$ closer to the geometric center, and add the clusters with representatives to a pool $P$\;
}

Apply hierarchical clustering to the pool $P$, considering the distance between two clusters as the minimum distance between their representative points, until $k$ clusters are obtained\;

$A \leftarrow \text{Assign each point in } X \text{ to nearest cluster centroid in } C$\;
\caption{CURE Clustering}
\label{alg:cure}
\end{algorithm}

BIRCH (Balanced Iterative Reducing and Clustering using Hierarchies)~\cite{Zhang1996} is a hierarchical clustering algorithm that is well-suited for large data. It constructs a Clustering Feature (CF) tree to efficiently summarize the distribution of a dataset. The CF tree summarizes the information of each subcluster, including the number of points in the subcluster, the sum of the points' values, and the sum of squares of the points' values. The Clustering Tree groups together subclusters with similar characteristics and assigns each data point to the nearest subcluster by traversing the tree from the root to the leaf that contains the closest subcluster. The CF tree allows BIRCH to summarize the dataset's distribution in a compact way, avoiding processing each data point multiple times. The Clustering Tree enables BIRCH to handle large datasets efficiently by reducing the number of computations needed to cluster the data.

Despite its strengths, BIRCH has some limitations. One of these is its assumption that each subcluster has a spherical shape, which can limit its ability to handle datasets with irregular or complex structures. Nevertheless, this can be considered as a general drawback inherent to all the MSSC algorithms and the MSSC paradigm itself. Additionally, BIRCH requires careful parameter selection, such as the maximum number of CF entries and the threshold for merging subclusters, to produce high-quality clusters. Finally, BIRCH may struggle with datasets that have a large number of subclusters, as it can become computationally expensive to traverse the Clustering Tree. Studies have demonstrated that CURE is computationally superior to BIRCH in big data conditions~\cite{Guha1998}. As a result, we chose not to include BIRCH in our experiments. 

CURE (Clustering Using REpresentatives)~\cite{Guha1998} is a hierarchical clustering algorithm designed with the intention of overcoming the limitations of existing clustering algorithms like K-means and hierarchical clustering, particularly in their ability to handle large datasets and to identify clusters of arbitrary shapes and sizes. It takes a more robust approach to cluster analysis by using multiple representative points instead of a single centroid or medoid to characterize clusters. This way, it captures the shape and variability of a cluster more effectively. The pseudocode of the BIRCH algorithm is presented in Algorithm~\ref{alg:cure}.

The CURE algorithm begins by taking a random sample of size $s$ from the dataset $X$. This sample is partitioned into $p = s / (f \cdot k \cdot q)$ partitions, each of size $q \cdot k$. Hierarchical clustering is then applied to each partition to form $q$ clusters. Note that each partition is processed independently, making the algorithm amenable to embarrassingly parallel implementation. For each cluster, the $c$ representative points, which are the points furthest from the centroid, are selected and moved towards the geometric center of the cluster by a factor of $\alpha$. Each cluster, along with its adjusted representative points, is added to a pool $P$. Hierarchical clustering is then applied to the pool $P$, where the distance between two clusters is defined as the minimum distance between their representative points. This process continues until the desired number of $k$ clusters is reached. Each data point in the original dataset $X$ is then assigned to the closest cluster centroid. The output is the set of $k$ clusters and the assignments of data points to these clusters.

The outcome is a set of clusters that can be of different sizes and shapes, making CURE a more flexible and robust option compared to traditional hierarchical clustering methods. However, as it involves computation of distances between pairs of representative points, the complexity can be high, making it computationally intensive for large datasets.

\subsection{Canopy Clustering}

The Canopy Clustering algorithm, as outlined in~\cite{McCallum2000}, serves as a preliminary step for more complex clustering algorithms. It aims to streamline the dataset size, enhancing the speed of subsequent clustering processes.

The crux of the Canopy Clustering algorithm is the creation of ``canopies''. The algorithm begins by randomly choosing a point in the dataset, which then becomes the center of a new canopy. From there, the algorithm measures the distance from the center point to every other point in the dataset. Two thresholds are predefined for this process: $T1$ and $T2$, where $T1 > T2$. All points within the $T1$ distance from the center point are considered part of the canopy. But only those points within the $T2$ distance cannot become centers of other canopies. Therefore, a point can belong to multiple canopies, but canopies have unique center points. This process of canopy creation continues until all points in the dataset have either been used as the center of a canopy or are within $T2$ distance of an existing canopy center. The pseudocode of the canopy clustering algorithm is provided in Algorithm~\ref{alg:canopy}.

The result is a set of overlapping canopies that give a rough initial grouping of the data points. More computationally intensive clustering, such as K-means, can then be applied to the points within each canopy separately, significantly reducing the computation time. This is particularly valuable when the dataset is large, as it allows for more sophisticated clustering algorithms to be used without an excessive computational cost. Also, the canopy method can be particularly useful when dealing with text data, as its low computational requirements and allowance for overlap make it well-suited to handling the high dimensionality and ambiguity that are characteristic of text data.

Despite these advantages, Canopy Clustering is not without its drawbacks. The quality of the canopies, and hence of the final clustering result, is highly dependent on the chosen distance measure and the values of the $T1$ and $T2$ thresholds. The latter, in particular, can be difficult to set correctly, as they require a good understanding of the data's scale and distribution. Furthermore, since the canopy clustering only provides an initial partitioning of the data, it needs to be used in conjunction with another clustering algorithm to obtain a final clustering result. This means that the overall clustering process may still be quite computationally intensive, especially if the subsequent clustering algorithm is a complex one. Finally, the algorithm does not guarantee optimal clustering, and the final clustering result can be affected by the order in which the data points are processed.

Xia et al. proposed a parallel adaptive Canopy-K-means clustering algorithm for big data, leveraging the MapReduce cloud computing model~\cite{Xia2020}. While it promises to manage the complexities of large datasets with a unique adaptive process for selecting the similarity threshold, there is ambiguity in its description. Despite their claims of a MapReduce implementation, our attempts to reproduce their algorithm were thwarted by this unclear adaptive procedure, which was the key step in their algorithm. Consequently, we unable to successfully replicate or verify the assertions made in the paper.

\begin{algorithm}
\SetAlgoLined
\KwResult{Compute the canopies $C$ for a dataset $X$ using the Canopy Clustering algorithm.}
\textbf{Initialization:}
Initialize an empty set of canopies $C$\;
\While{$X$ is not empty}{
Randomly select a point $x$ from $X$\;
Create a new canopy $c$ with center $x$\;
\For{each point $y$ in $X$}{
    \If{distance$(x, y) < T1$}{
        Add $y$ to $c$;
    }
}
\If{distance$(x, y) < T2$}{
Remove $y$ from $X$\;
}
Add $c$ to $C$;
}
\caption{Canopy Clustering}
\label{alg:canopy}
\end{algorithm}

Beyond hardware acceleration and memory efficiency, K-means optimization can be achieved through analytical refinement of the underlying mathematical operations, as explored in the next subsection.

\subsection{Triangle inequality}


During K-means clustering, it can be observed that not all points alter their cluster membership. By using some logic to identify and label these points as fixed, the algorithm can avoid the need to recalculate cluster distances for them. This optimization can significantly reduce computation time. The triangle inequality is widely employed to identify points that remain fixed to their respective centroids across iterations~\cite{Elkan2003}.

Moodi et al. introduced an optimized version of the K-means algorithm that utilizes the triangle inequality to reduce processing time~\cite{Moodi2022}. Their approach applies the A-means algorithm according to Early Classification (EC), which stabilizes points to their respective clusters based on the probability of membership change in subsequent iterations. This reduces the computational time required for assigning points to clusters, especially for large datasets. However, significant movement of centroids in some iteration can lead to inaccurate results. To address this issue, the authors proposed a method to reintroduce stabilized points into computation if their distances from the centroids exceed the cluster radius. This ensures that the approach achieves significant computational savings without compromising accuracy. We have included this algorithm into our experiments under the name IK-means. The pseudocode of the IK-means algorithm is shown in Algorithm~\ref{alg:ik_means}.


\begin{algorithm}
	\SetAlgoLined
	\KwResult{Compute the final centroids $C$ for a dataset $X$ using the IK-means algorithm.}
	\textbf{Initialization:}\\
	Initialize all $k$ centroids $C \in \R^{k \times n}$ randomly from $X$\;
	Initialize the assignment vector $A \in \R^m$ with zeros\;
	Initialize the cluster radii $R \in \R^k$ with zeros\;
	Initialize the centroid shifts $S \in \R^k$ with zeros\;
	Initialize the exclusion vector $E \in \R^m$ with zeros\;
	\For{$t = 1$ \KwTo $T$}{
		\If{$t \mod E = 0$}{
			\For{$i = 1$ \KwTo $m$}{
				\If{$E[i] = 1$ \textbf{and} $\|C[A[i]] - X[i]\| > R[A[i]]$}{
					$E[i] \leftarrow 0$\;
				}
			}
		}
		\For{$i = 1$ \KwTo $m$}{
			\If{$E[i] = 0$}{
				Compute distances $d$ to all centroids $C$\;
				Sort $d$ and get the indices of the two nearest centroids $idx1$, $idx2$\;
				$A[i] \leftarrow idx1$\;
				\If{$t \geq 2$ \textbf{and} $\|d[idx1]^2 - d[idx2]^2\| > S[idx1] + S[idx2]$}{
					$E[i] \leftarrow 1$\;
				}
			}
		}
		\If{$\sum E = m$}{
			\textbf{break}\;
		}
		\For{$j = 1$ \KwTo $k$}{
			Compute new centroid $C_{\text{new}}[j]$ as the mean of $X$ where $A = j$\;
			Compute new radius $R[j]$ as the maximum distance from $C_{\text{new}}[j]$ to $X$ where $A = j$\;
			Compute centroid shift $S[j]$ as $\|C[j] - C_{\text{new}}[j]\|$\;
			$C[j] \leftarrow C_{\text{new}}[j]$\;
		}
	}
	\caption{IK-means Clustering}
	\label{alg:ik_means}
\end{algorithm}

The IK-means algorithm reduces the computation of the K-means algorithm by reducing the number of points that need to be reassigned to clusters in each iteration. This is achieved by introducing the concept of exclusion, where points that are very likely to remain in their current cluster are excluded from further consideration in the algorithm. The exclusion is based on the triangle inequality, which can help estimate whether a point is likely to change its cluster assignment or not, thus reducing the number of distance computations in the algorithm.

The triangle inequality can also be leveraged to accelerate the K-means local search within the Big-means algorithm, resulting in a substantial increase in efficiency. This demonstrates that Big-means is a highly versatile algorithm within the realm of MSSC algorithms, capable of applying optimization at almost every step and addressing nearly all challenges of big data clustering, showcasing its all-round excellence.

The following subsection explores another avenue for optimizing K-means: improving initialization efficiency through strategic sampling.




\subsection{Sampling-based initialization}


Initializing the K-means algorithm on a small random subset of the data and then applying it to the entire dataset is a powerful strategy for optimizing K-means in the big data context. This approach addresses scalability challenges, enabling the use of more advanced initialization methods for improved clustering performance.

Density-based algorithms, like DBSCAN or OPTICS, form clusters based on areas of high data point density, separated by areas of low density. They can discover clusters of arbitrary shapes and sizes, and are capable of handling noise and outliers. They are generally less sensitive to the initialization issue, unlike K-means, which can converge to different solutions depending on the initial centroids. Density-based methods can be slower and more computationally intensive, especially on large datasets. They also often require tuning parameters related to the density concept, which can be difficult in practice. However, applying a density-based algorithm, such as DBSCAN or OPTICS, to a smaller random sample from the dataset may be computationally justified.

Since a random sample is usually relatively small, more computationally complex algorithms can be applied to it. For instance, paper~\cite{Dierckens2017} proposed a new clustering algorithm for big data that leverages the idea of applying a density-based algorithm, DBSCAN, to a small random sample to initialize the global clustering process. The resulting centroids from the sample are used to initialize the K-means algorithm on the entire dataset, which can lead to a faster convergence and higher quality clusters than initializing K-means randomly. A representative algorithm employing this approach is the K-means Clustering Data Science and Engineering Solution, abbreviated as CluDataSE. Its pseudocode is shown in Algorithm~\ref{alg:clu_data_se}. CluDataSE provides a way to balance accuracy and efficiency in clustering large datasets.

\begin{algorithm}
\SetAlgoLined
\KwResult{Compute the final centroids and cluster assignments for a dataset $X$ using the CluDataSE algorithm.}
\textbf{Initialization:}\\
Choose a random sample of size $s$ from $X$\;
Create an empty centroid list\;
\While{number of centroids $< k$}{
    Apply DBSCAN with $eps$ and $min\_pts$ to the sample\;
    Add the DBSCAN components to the centroid list\;
}
Apply K-means++ on the centroid list to reduce the number of centroids\;
Apply K-means on $X$ using the reduced centroids to get the final cluster assignments and centroids\;
\caption{CluDataSE Clustering}
\label{alg:clu_data_se}
\end{algorithm}

The multi-start K-means algorithm~\cite{Franti2019} is a variant of the K-means algorithm that aims to improve the quality of the produced clusters by running the K-means algorithm multiple times with different initializations. The algorithm randomly initializes the cluster centers multiple times and runs the K-means algorithm on each initialization. The final set of clusters is selected from among the various runs based on the lowest sum of squared distances between the data points and their respective cluster centers.

The Forgy K-means algorithm~\cite{Pena1999} is a simple and efficient variant of the K-means algorithm. Unlike random initialization, where centroids are chosen as random locations in the space, Forgy initialization randomly selects $k$ data points from the dataset to serve as the initial centroids. This approach ensures that the initial cluster centers are actual data points, making them more representative of the dataset and potentially leading to faster convergence of the algorithm. This selection strategy leads to the initial cluster centers that are more likely to be close to the true cluster centers since randomly picked data points tend to be close to regions with high density. Meanwhile, using random space locations can sometimes lead to initial centroids that are not representative of the dataset or that are too far away from the true clusters.

The Big-means algorithm exemplifies the use of K-means++ initialization on a random subsample of the dataset. The algorithm then employs an iterative process to improve the current best set of centroids using local search on various subsequent subsamples, thereby reducing the dependence of final results on the accuracy of initialization. As a result, different initialization schemes can also be used on the first sample in Big-means.

The following subsection reveals that sampling has far-reaching benefits, extending beyond initialization to enhance K-means iterations themselves.

\subsection{Approximation techniques}

Approximation techniques aim to create a compressed or simplified representation of the original dataset while preserving its essential structure and properties. These techniques allow clustering algorithms to efficiently process large datasets while striving to minimize any loss in accuracy.

Lightweight coreset~\cite{Bachem2018LWC} is a technique used to generate a smaller, weighted subset of the original dataset that preserves its essential properties. Coresets can be utilized to improve the computational and memory efficiency of big data clustering algorithms by reducing the number of data points processed. They are useful in big data clustering as they enhance scalability and can be adapted to different clustering algorithms. Despite these advantages, coresets have some limitations, such as a slight loss of accuracy due to approximation, complexity in constructing them, sensitivity to the data distribution, and the need for domain expertise in parameter tuning for optimal performance. Overall, coresets provide a valuable approach for handling large-scale clustering tasks while addressing resource constraints.

As a promising technique for big data applications, a clustering algorithm based on applying the K-means++ seeding to a lightweight coreset was included into experiments under the name LW-coreset. The pseudocode for this method is provided in Algorithm~\ref{alg:lwcore_clustering}.


\begin{algorithm}
\SetAlgoLined
\KwResult{Compute a lightweight coreset $C$ for a dataset $X$, apply K-means and return the cluster centroids $C_k$ and the point-to-cluster assignments $A$.}
\textbf{Initialization:}\\
Compute the mean $\mu$ of the dataset $X$\;
Compute the sum of squared distances $D = \sum_{x' \in X} \|x' - \mu \|^2$\;
\For{each $x \in X$}{
    Compute the probability $q(x) = \frac{1}{2|X|} + \frac{1}{2} \frac{\| x - \mu \|^2}{D}$\;
}
Set $C$ to an empty set\;
\For{$i$ in range $s$}{
    Sample a point $x$ from $X$ according to the probabilities $q(x)$, without replacement\;
    Add $x$ to $C$\;
}
\textbf{K-means++ on sampled coreset:}\\
Initialize $k$ centroids $C_k$ from $C$ using K-means++\;
\While{centroids $C_k$ change}{
    Assign each point in $C$ to the closest centroid, forming $k$ clusters\;
    Update each centroid to be the mean of the points assigned to its cluster\;
}
Assign each point $x_i \in X$ to the closest cluster centroid $c_j \in C_k$\;
\Return $C_k, \quad A$
\caption{LW-Coreset: Clustering using Lightweight Coreset}
\label{alg:lwcore_clustering}
\end{algorithm}

Another efficient yet effective approximation method, employed by the Big-means algorithm, involves taking a simple uniform sample from the available data $X$. This approach has a constant time complexity of $\O(1)$ and does not require the calculation of joint probability distributions. Additionally, under a mild assumption on the dataset structure, simple uniform sampling is theoretically guaranteed to produce high-quality solutions, as demonstrated in \cite{Huang2023-simple}. This method is optimal according to the ``less is more'' principle, making it the fastest possible sampling approach computationally.

The next subsection explores the final frontier of K-means optimization, harnessing its power within advanced global optimization metaheuristics or synergizing it with other clustering approaches and meta-ideas.





\subsection{Hybrid approaches}

Hybrid clustering algorithms often combine multiple techniques or algorithms to improve clustering performance, such as using genetic operations to improve initial centroid selection or combining K-means with hierarchical clustering. The choice of the combination of approaches depends on the specific characteristics of the data and the computational resources available. While these approaches can improve clustering quality, they often incur high computational costs due to their increased complexity.


Karmitsa et al. proposed LMBM-Clust in~\cite{Karmitsa2018}, an incremental clustering method designed for very large datasets with a large number of attributes and data points. The method combines the limited memory bundle method (LMBM) \cite{Haarala2007} as the local search with a method for generating starting centers proposed by Ordin et al. in~\cite{Ordin2015}. Experimental results show that LMBM-Clust produces good results in terms of both accuracy and time, only losing to K-means++ in terms of time. The method also requires fewer distance function evaluations than other incremental approaches. However, one major drawback of LMBM-Clust is its high complexity and a large number of hyperparameters. The original implementation of LMBM-Clust provided by the authors was unable to cluster a significant portion of large real-world datasets in our experimental evaluation due to the out-of-memory error~\cite{Mussabayev2023}.

Gribel et al proposed a hybrid clustering algorithm called HG-means in~\cite{Gribel2019}, which combines the multi-start K-means approach with genetic operations such as mutation, crossover, and selection to select new initial centroids. The algorithm was shown to produce high-quality clusters and to be faster than many existing algorithms, with the exception of LMBM-Clust. However, HG-means requires clustering the full dataset in every iteration, which makes the algorithm unsuitable for big data due to its high computational cost.

Mansueto et al.~\cite{Mansueto2021} proposed a Memetic Differential Evolution Clustering (MDEClust) algorithm for the MSSC problem. The algorithm combines the global optimization framework of differential evolution with the local search procedure of K-means clustering. The MDEClust algorithm evolves a population of solutions through crossover, mutation, and local refinement via K-means. The algorithm ensures diversity and avoids premature convergence by employing an innovative crossover operation based on differential evolution and a roulette wheel selection in the mutation process. The MDEClust algorithm has been shown to be efficient in generating high-quality solutions, especially for datasets with a large number of clusters and high dimensionality. However, the algorithm's performance can be sensitive to the choice of initial population and parameter settings, including population diversity tolerance, maximum number of iterations, crossover parameter, and mutation parameter. While MDEClust outperformed HG-means in terms of clustering accuracy, scalability remains a challenge as the computational effort increases significantly, especially when the mutation operator is used. Additionally, neither MDEClust nor HG-means claim to handle clustering datasets with a large number of input objects $m$.

In the context of big data, where datasets can be very large and high-dimensional, hybrid clustering algorithms may not be suitable due to their high computational complexity. These algorithms may require extensive memory usage or excessive processing time, making them impractical for large datasets. Furthermore, hybrid algorithms often involve a large number of hyperparameters that need to be carefully tuned, which can be difficult and time-consuming.

In summary, while hybrid clustering algorithms can improve clustering performance, their high computational complexity and extensive hyperparameters make them difficult to apply to large datasets. As a result, simpler and more scalable clustering algorithms, such as K-means and its variants, are often preferred for big data clustering. In particular, we did not consider any of the advanced hybrid MSSC clustering algorithms in our experimental study.

\subsection{Summary of clustering techniques}

In summary, there are various ways to address the challenges of applying K-means-like techniques to big data. Sampling approaches, like Minibatch K-means, use a random subset of the data for the computation of centroids, thus making the clustering process more efficient. Algorithms like BDCSM use partitioning, where the dataset is divided into chunks, each chunk is clustered separately, and the results are aggregated. Other approaches leverage parallelization and distributed computing to distribute the workload across multiple processors or machines. Some methods, such as IK-means, exploit the properties of geometric spaces, like the triangle inequality, to reduce the computation time. The Big-means algorithm combines both the sampling and parallelization approaches in a unique way that maximizes not only efficiency, but the accuracy as well.

However, each of the considered methods has its own trade-offs and potential challenges, so it is important to choose the right approach based on the specific requirements and constraints of the given clustering task, characteristics of the data, and the computational resources available. Evaluating different approaches and selecting the one that best fits the problem at hand is crucial for achieving accurate and efficient clustering results. In Section~\ref{sec:experiments}, we provide an empirical comparison of these methods to understand their performance and efficacy in clustering big datasets.

The LIMA number is a measure of algorithm complexity based on the number of distinct operations in the algorithm~\cite{Brimberg2023}. In this paper, we use the LIMA number as a measure of complexity for several clustering algorithms, as shown in Table~\ref{tab:lima_numbers}. Notably, despite having lowest LIMA numbers, Big-means and BDCSM algorithms yield quite accurate clustering results for big data, suggesting their effectiveness. On the other hand, the IK-means algorithm, despite having the highest LIMA number, exploits the properties of geometric space (triangle inequality) to efficiently reduce the computation time.

\begin{table}[ht]
	\centering
	\begin{tabular}{c|c|c|p{7.5cm}}
		Algorithm & LIMA number & Input parameters & Algorithmic ingredients \\ \hline
		BDCSM & 5 & 1 ($p$) & Partitioning, clustering, centroid pooling, final clustering, final assignment \\
		Big-means & 6 & 2 ($s$, $T$) & Random sampling, conditional reinitialization, centroid update, condition checking and updating, iteration, final assignment \\
		Minibatch K-means & 6 & 2 ($T$, $m$) & Initialization, random sampling, assignment, centroid update, iteration, final assignment \\
		K-means++ & 6 & 1 ($T$) & Random selection, distance calculation, probabilistic selection, iteration, cluster assignment, and centroid update \\
		CURE & 7 & 6 ($s$, $f$, $k$, $q$, $c$, $\alpha$) & Random sampling, partitioning, hierarchical clustering, representative selection, geometric transformation, iteration, and cluster assignment \\
		CluDataSE & 7 & 3 ($s$, $eps$, $min\_pts$) & Random sampling, density-based clustering, cluster center reduction, k-means clustering, iteration, parameter adjustment, and cluster assignment \\
		LW-Coreset & 7 & 1 ($s$) & Mean calculation, distance computation, probability computation, sampling, centroid initialization, centroid update, and assignment \\
		IK-means & 8 & 0 & Initialization, distance computation, exclusion check, assignment, centroid reinitialization, centroid update, radius update, convergence check \\
	\end{tabular}
	\caption{LIMA numbers of the considered algorithms}
	\label{tab:lima_numbers}
\end{table}

While scalable clustering algorithms provide important tools for handling big data, they also highlight the need for further research and development in this area. Future research should address the ongoing challenges of big data, which include dealing with high-dimensional and multi-view data, handling data sparsity and imbalanced data, and improving the quality of clusters in the presence of noise and outliers. It should also consider simplifying the parameter tuning process and enhancing the reproducibility of clustering results.

\section{Generalizations, Reflections, and Practical Advices} \label{sec:practice}

As the scale and complexity of data increase, it becomes increasingly complex to choose the right big data clustering algorithm, as the traditional choices of clustering algorithms can struggle to work effectively and efficiently at this scale. Therefore, one of the aims of this review paper is to guide and support practitioners in understanding, selecting, and implementing suitable big data clustering methods for their datasets.

\subsection{The imperative of scalability}

A recurring theme in the discussed techniques is the imperative for scalability, as big data presents inherent challenges in volume, velocity, and variety. The sheer size and speed of data generation demand algorithms that are both efficient and scalable, making the use of algorithms like Big-means a necessity.

\subsection{Complexity and its trade-offs}

Big data is not just about size; it is about complexity too. High dimensionality, varying data distributions, presence of noise and outliers, all these factors contribute to the complexity. While techniques like the density-based clustering (DBSCAN), CURE, and the EM algorithm~\cite{Dempster1977-em} can handle complex distributions, there is always a trade-off to consider between computational complexity, speed, and accuracy. Often, MSSC remains as the only feasible paradigm to use in big data conditions.

\subsection{Trade-offs in big data clustering}

It is essential to understand that there is no all-encompassing algorithm for big data clustering under the MSSC paradigm. The right algorithm depends on the specific characteristics and requirements of the task at hand. For instance, Big-means is well-suited for large and big datasets when both speed and accuracy are essential. On the other hand, hierarchical (Ward's method~\cite{Ward1963}) or evolutionary (MDEClust) algorithms might be better when accuracy is prioritized and the dataset is small to moderate in size.

The choice of clustering algorithm often involves trade-offs between computational efficiency, accuracy, ability to handle complex distributions, and resilience to noise or outliers. For example, Big-means, Minibatch K-means, and BDCSM are fast, but may not handle complex data distributions. Similarly, while Ward's method and MDEClust may offer higher quality clusters, they can be computationally intensive and more challenging to tune.

\subsection{Understanding algorithmic strengths and weaknesses}

Big-means offers both effectiveness and efficiency in big data clustering, addressing most MSSC challenges. However, it assumes convex and isotropic clusters, which may not always hold in real-world data. Moreover, while it does not explicitly handle noise and outliers, random sampling in the algorithm partially mitigates this limitation.

Hierarchical clustering, specifically Ward's method, provides an interpretable dendrogram and does not require the number of clusters to be predefined. However, it can be extremely computationally expensive for big data.

Density-based clustering, such as DBSCAN, can discover clusters of arbitrary shape and is resilient to outliers. It can be a powerful tool for datasets with complex distributions but tuning its parameters ($eps$ and $MinPts$) can be challenging. Also, its computational complexity can be as high as quadratic with respect to the overall number of data points.

Evolutionary clustering algorithms like MDEClust have the potential to reveal insightful clusters through multiple generations of solutions, but they are quite computationally demanding and may not be practical for large-scale data.

HG-means, as a hybrid algorithm, tries to balance the strengths of hierarchical and K-means clustering but inherits some of their weaknesses too, including the sensitivity to initial centroids.

\subsection{Guiding the choice of a big data clustering algorithm}

\begin{figure}
    \centering
    \resizebox{0.8\linewidth}{!}{
    \begin{tikzpicture}[
      decision/.style={diamond, draw, fill=blue!20, text width=5em, text badly centered, node distance=3.5cm, inner sep=0pt},
      block/.style={rectangle, draw, fill=blue!20, text width=9em, text centered, rounded corners, minimum height=4em},
      line/.style={draw, -latex'}
    ]
    
    \node [block] (init) {Start};
    \node [decision, below of=init] (dataSize) {Data Size?};
    \node [block, right of=dataSize, node distance=5cm] (largeData) {Big-means};
    \node [block, left of=dataSize, node distance=5cm] (smallData) {Big-means, Hierarchical (Ward's), Density-based (DBSCAN), Evolutionary (MDEClust)};
    \node [decision, below of=dataSize] (numClusters) {Number of Clusters?};
    \node [block, right of=numClusters, node distance=5cm] (manyClusters) {Big-means, HG-means};
    \node [block, left of=numClusters, node distance=5cm] (fewClusters) {Big-means, Hierarchical (Ward's), Density-based (DBSCAN), Evolutionary (MDEClust)};
    \node [decision, below of=numClusters] (dataDist) {Data Distribution?};
    \node [block, right of=dataDist, node distance=5cm] (complexDist) {Density-based (DBSCAN), CURE, EM algorithm};
    \node [block, left of=dataDist, node distance=5cm] (simpleDist) {Big-means, K-means++};
    \node [decision, below of=dataDist] (speedAcc) {Speed vs Accuracy?};
    \node [block, right of=speedAcc, node distance=5cm] (fastAlg) {Big-means, Forgy K-means, Minibatch K-means, LW-coreset, BDCSM};
    \node [block, left of=speedAcc, node distance=5cm] (accurateAlg) {Big-means, Hierarchical (Ward's), Density-based (DBSCAN), Evolutionary (MDEClust)};
    \node [block, below of=speedAcc, node distance=5cm] (end) {End};
    
    \path [line] (init) -- (dataSize);
    \path [line] (dataSize) -- node [midway, above, sloped] {Large, Big} (largeData);
    \path [line] (dataSize) -- node [midway, above, sloped] {Small} (smallData);
    \path [line] (dataSize) -- (numClusters);
    \path [line] (numClusters) -- node [midway, above, sloped] {Many} (manyClusters);
    \path [line] (numClusters) -- node [midway, above, sloped] {Few} (fewClusters);
    \path [line] (numClusters) -- (dataDist);
    \path [line] (dataDist) -- node [midway, above, sloped] {Complex} (complexDist);
    \path [line] (dataDist) -- node [midway, above, sloped] {Simple} (simpleDist);
    \path [line] (dataDist) -- (speedAcc);
    \path [line] (speedAcc) -- node [midway, above, sloped] {Fast} (fastAlg);
    \path [line] (speedAcc) -- node [midway, above, sloped] {Accurate} (accurateAlg);
    \path [line] (speedAcc) -- (end);
    \end{tikzpicture}
	}
    \caption{Flowchart of big data clustering algorithm selection}
    \label{fig:alg_select_flowchart}
\end{figure}
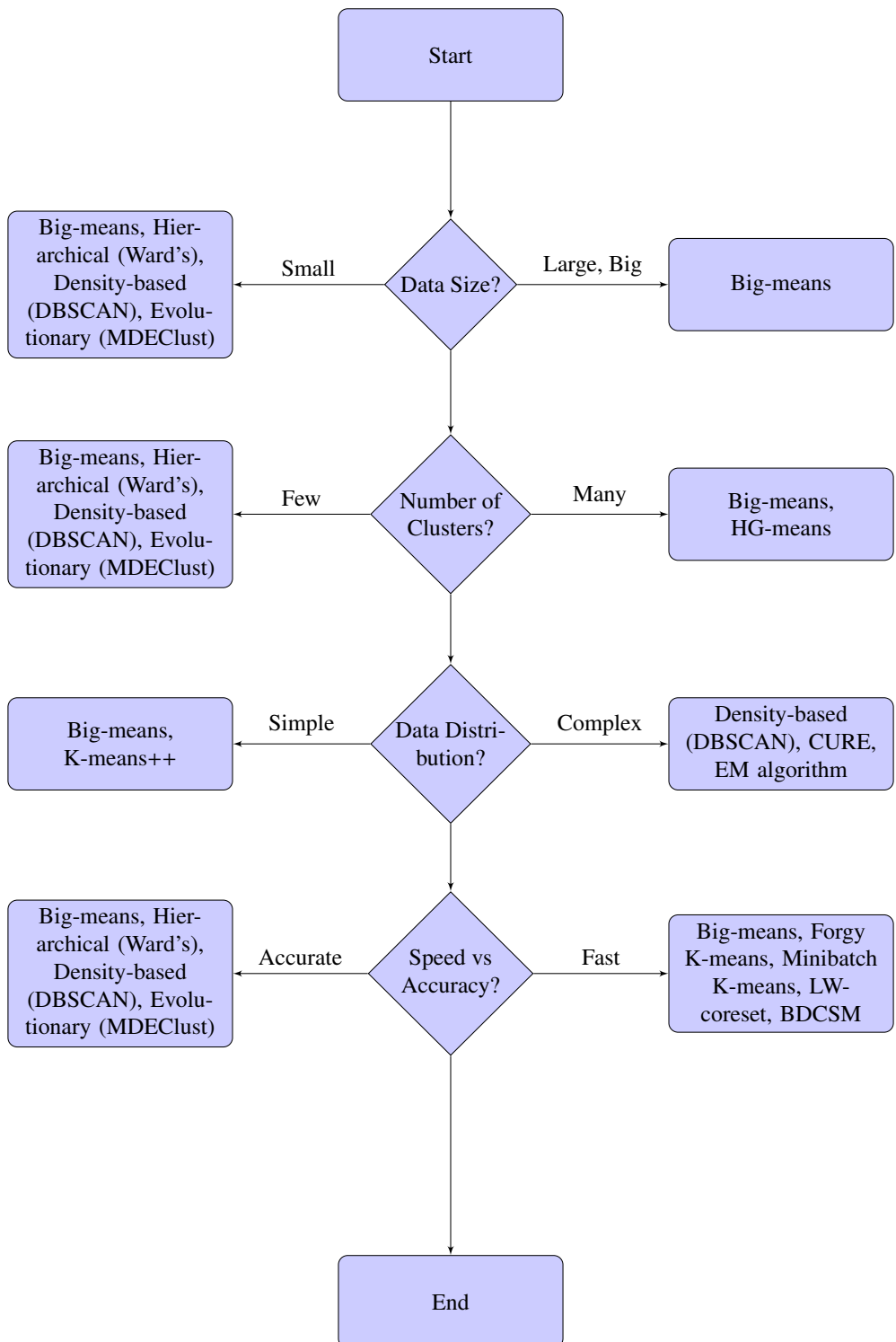

We introduce a flowchart (Figure~\ref{fig:alg_select_flowchart}), a practical tool for selecting an initial clustering algorithm based on data size, number of clusters, distribution, and speed-accuracy trade-off, serving as a valuable aid in big data analysis. However, it is important to note that these are guidelines and not definitive rules. Depending on the specific circumstances, it may be necessary to experiment with multiple algorithms or a combination of them. The final decision may require experimentation and fine-tuning, based on the specifics of a task.

The decision should also be guided by the data characteristics and the specific requirements of the task. For large and big datasets, Big-means offers a rapid clustering solution with high accuracy, making it a suitable choice when speed and accuracy are equally important. For a small to moderate-sized dataset where accuracy and quality of clusters are paramount, hierarchical clustering (Ward's method), DBSCAN, or MDEClust may be preferable. For complex distributions, DBSCAN, CURE, or the EM algorithm would be ideal, while simpler distributions could be efficiently handled by Big-means or K-means++.

Our hope is that the presented flowchart-based approach to the selection of clustering algorithms can be a practical tool to guide data scientists in the selection of the most suitable algorithm for their specific application and dataset.

\subsection{The role of experimental validation (experimentation and iteration)}

It is crucial to validate the choice of a clustering algorithm through experimentation. Apply the chosen algorithm to a subset of your data, analyze its performance and appropriateness for your task, and adjust accordingly. This iterative process allows to refine the choice and find the most suitable algorithm for the specific task.


\subsection{Other review papers}

Recent review papers on scalable and parallel big data clustering methods offer valuable insights into the field's current state. Notably, Mehdi et al.'s review paper \cite{Mahdi2021-scalalgs} surveyed 101 algorithms and found that most have complex implementations. Moreover, the authors highlight a shift towards deploying algorithms on cloud-based infrastructure, rather than developing new practical methods. This underscores the importance of efficient and scalable clustering approaches in big data analytics.

The survey in~\cite{Saeed2020} concluded that few existing methods support the variety and velocity characteristics of big data. In other words, most of the available parallel clustering algorithms do not handle real-time data, focusing on a single, mostly numerical, type of data. This limits their capability to process big data \cite{Dafir2020}.

Based on the detailed discussion of different parallel algorithms, Mohebi et al.~\cite{Mohebi2016} concluded that the ﬁeld of parallel big data clustering is still young and open for new research. They argued that parallel data processing can help improve the clustering time of large datasets, but it may degrade the quality and performance. Therefore, the main concern is to achieve a reasonable trade-off between quality and speed in the context of big data.

\subsection{Summary}

In summary, the complexity of big data clustering requires a flexible, informed approach. By understanding the strengths, weaknesses, and trade-offs of different algorithms and techniques, and by adopting an experimental, iterative mindset, practitioners can effectively navigate the challenges of big data clustering.

\section{LIMA Dominance Criterion} \label{sec:lima_dominance}

The ``Less is More'' Approach (LIMA) is a principle that emphasizes the power of simplicity in algorithm design~\cite{Brimberg2023}. It suggests that simpler algorithms can often yield more effective results than their more complex counterparts. Despite its successful application in various fields, it is seldomly used for solving optimization problems.

In heuristic search design, LIMA fits well as a heuristic is usually defined by several characteristics, including speed, accuracy, simplicity, flexibility, effectiveness, and robustness. However, there is no formal approach to consider these attributes. Therefore, a mathematical framework was developed to formally incorporate one of these attributes, namely, simplicity, alongside the commonly used ones, speed (CPU time) and accuracy (solution quality). This resulted in the LIMA approach, which provides a bridge connecting Optimization to Artificial Intelligence and Machine Learning.

The usual way to compare different methods to solve an optimization problem is to assess the relative solution quality (accuracy) and algorithm efficiency (speed). The LIMA approach adds a comparison of the simplicity of each method to the mix.

\begin{definition}
	The LIMA number of an algorithm $A$ used to solve an optimization problem $P$ on a data set $D$ is the cardinality of the set of ingredients $U$ used by $A$.
\end{definition}

\begin{definition}
	Algorithm $B$ is LIMA-dominant over algorithm $A$ in solving problem $P$ on data set $D$ if it satisfies the following conditions:
	\begin{itemize}
		\item accuracy: the objective function $f_B \leq f_A$,
		\item speed: the running time $t_B \leq t_A$,
		\item simplicity: the LIMA numbers satisfy $|U_B| \leq |U_A|$,
		\item At least one of the above conditions is satisfied strictly.
	\end{itemize}
\end{definition}

The LIMA-dominance relation could be too strong if applied to each run and each test instance. One way forward is to introduce the concept of weak or average LIMA-dominance, i.e., after many runs on the same problem, we verify that average values satisfy $f_B \leq f_A$ and $t_B \leq t_A$.

The problem can be considered as a challenging three-objective optimization problem defined by accuracy, speed, and simplicity. There are several ways of tackling a multi-objective problem including weighted multi-objective, goal programming, and lexicographic, among others. If one wants to stress attribute simplicity, the aim would be to solve the following problem. First, find a subset of ingredients $U'$ used by algorithm $B$ that is as small as possible, but still allows $B$ to produce solutions that are at least as good as those produced by a state-of-the-art algorithm $A$, and in the same or less time. If $B$ can do this with fewer ingredients than $A$, then $B$ is said to dominate $A$ in the LIMA sense. As noted above the ‘total’ dominance implied may be replaced by a less stringent requirement such as ‘average’ dominance or dominance by statistical hypothesis testing.

\section{Experimental Evaluation} \label{sec:experiments}

\subsection{Implementation details}

The computational experiments were conducted on a system utilizing Ubuntu 22.04 64-bit, powered by an AMD EPYC 7663 processor and equipped with 1.46 TB of RAM. Our preliminary experiments showed that using 8 cores struck the optimal trade-off between the gained speedup and the communication overhead between parallel workers. The computational environment was bolstered by Python 3.10.11, NumPy 1.24.3, and Numba 0.57.0. Numba~\cite{Marowka2018}, a key component of our software stack, significantly enhanced Python's execution speed and parallel computing capabilities by compiling Python code into machine code on the fly and enabling multi-processor execution.

We assessed the algorithms' performance using the same 19 publicly accessible datasets used in~\cite{Mussabayev2023}, where their detailed descriptions and respective URLs can be found. Additionally, normalization was applied to four datasets, increasing the total to 23 datasets for our analysis. These datasets consisted entirely of numerical features, devoid of any missing data, and varied extensively in size—from datasets with as few as two attributes to those with up to 5,000 attributes. The datasets also varied in the number of instances, from thousands (the smallest being 7,797) to tens of millions (the largest being 10,500,000). Their deliberate imbalance and diversity in object and feature counts facilitated a comprehensive evaluation of the algorithms across a wide range of dataset sizes. By employing the same datasets and methodology as Karmitsa et al.~\cite{Karmitsa2018}, we could draw direct comparisons between our findings and previous results.

Each dataset was subjected to a series of experiments by clustering with each algorithm $n_{exec}$ times for cluster sizes of 2, 3, 5, 10, 15, 20, 25, with some datasets additionally clustered into 4 clusters. This resulted in a total of 29,464 experiments. The outcomes of these experiments were analyzed based on the relative error $\eps$ and the CPU time $t$ consumed (measured in seconds). For each algorithm $A$, dataset $X$, and cluster count $k$, the relative error is quantified as $\eps = 100 \cdot (f - f^*) / f^*$, where $f^*$ represents the optimal objective function value found for dataset $X$ using $k$ clusters. We relied on $f^*$ values as reported in~\cite{Karmitsa2018} and our own optimal values for larger datasets not covered in~\cite{Karmitsa2018}, indicated by asterisks. For the Big-means algorithm, the CPU time $t$ was recorded at the point of the last incumbent solution update $C$. Since a hybrid parallel version of Big-means was utilized~\cite{Mussabayev2023-optparext}, $t$ reflects the moment the last incumbent solution $C_w$ was updated by the worker $w$ that attained the best objective function value $f(C_w, S_w)$ for some sample $S_w$.

The maximum allowable CPU time for Big-means was set at $T$ seconds, with the clustering of each sample ceasing either after 300 iterations or when the relative change between successive objective function values fell below $10^{-4}$. These parameters were chosen based on empirical testing to optimize the balance between computational speed and clustering accuracy for Big-means. In the K-means++ algorithm, the selection of the next centroid was based on identifying the most optimal one from three candidates.

Each series has a minimum, median, and maximum resulting values of relative accuracy and time, which are calculated across $n_{exec}$ executions of the algorithm on configuration $(X, k)$. The means of these metrics across the values of $k$ for each dataset are displayed in the corresponding columns of Table~\ref{tab:result_e}, Table~\ref{tab:result_t}, and the appendices. The means in the final rows of these tables highlight overall performance across datasets. The best results for each metric and dataset pair are bolded, indicating top algorithm performance. Success is indicated when an algorithm's median performance on a series of executions for a value of $k$ outperforms or matches the best result among all the algorithms for this series.


The determination of the sample size $s$ for Big-means on each dataset followed a heuristic approach, adjusting $s$ to a point where further changes neither enhanced nor diminished the quality of the objective function outcomes. A total of 8 CPUs were utilized across all parallel algorithms tested.

\begin{landscape}

\begin{table}[!htbp]%
\centering%
\captionsetup{font=scriptsize}
\caption{Relative clustering accuracies ($\epsilon$, in percentage) resulting from the comparison of the selected clustering algorithms.}%
\label{tab:result_e}%
\resizebox{!}{\esummtableheight}{%
\begin{tabular}{l|cccc|cccc|cccc|cccc}
	\hline
	\multirow{2}{*}{Dataset} & \multicolumn{4}{p{3cm}}{\mbox{Big-means}}& \multicolumn{4}{p{3cm}}{\mbox{IK-means}}& \multicolumn{4}{p{3cm}}{\mbox{BDCSM}}& \multicolumn{4}{p{3cm}}{\mbox{Minibatch K-means}} \\
	\cline{2-17}
	& \#Succ & Min & Median & Max & \#Succ & Min & Median & Max & \#Succ & Min & Median & Max & \#Succ & Min & Median & Max  \\
	\hline
	CORD-19 Embeddings & 2/7 & 0.02 & \textbf{0.08} & \textbf{0.18} & 0/7 & 0.94 & 1.95 & 5.41 & 0/7 & 0.6 & 1.84 & 4.81 & 0/7 & 0.41 & 0.81 & 1.81 \\
	HEPMASS & 2/7 & \textbf{0.02} & \textbf{0.12} & \textbf{0.18} & 0/7 & 1.58 & 2.24 & 4.18 & 0/7 & 0.89 & 1.61 & 3.51 & 0/7 & 0.53 & 1.06 & 2.22 \\
	US Census Data 1990 & 4/7 & \textbf{0.34} & \textbf{1.24} & \textbf{3.19} & 1/7 & 4.63 & 84.2 & 308.78 & 0/7 & 15.32 & 90.94 & 279.53 & 1/7 & 0.9 & 2.31 & 6.96 \\
	Gisette & 0/7 & -0.46 & -0.39 & -0.3 & 0/7 & -0.04 & 0.34 & 1.03 & 0/7 & -0.47 & -0.42 & -0.32 & 0/7 & -0.25 & -0.08 & 0.39 \\
	Music Analysis & 2/7 & 0.4 & 0.83 & \textbf{1.87} & 0/7 & 4.84 & 11.55 & 22.61 & 0/7 & 1.5 & 4.47 & 31.86 & 0/7 & 0.58 & 2.43 & 12.0 \\
	Protein Homology & 1/7 & 0.49 & 0.97 & \textbf{1.75} & 0/7 & 73.47 & 169.83 & 205.07 & 0/7 & 6.38 & 21.85 & 46.47 & 0/7 & 56.02 & 101.04 & 176.63 \\
	MiniBooNE Particle Identification & 3/7 & \textbf{-0.08} & \textbf{0.01} & 823309.91 & 0/7 & 40846.28 & 111022.79 & 311782.51 & 0/7 & 2.62 & 2.84 & 110987.1 & 0/7 & 2810404.23 & 2826338.32 & 2831962.18 \\
	MiniBooNE Particle Identification (normalized) & 0/7 & 0.29 & 0.59 & \textbf{1.44} & 0/7 & 73.84 & 182.29 & 246.15 & 0/7 & 2.59 & 8.99 & 1252.0 & 0/7 & 136.31 & 467.9 & 537.05 \\
	MFCCs for Speech Emotion Recognition & 2/7 & 0.1 & \textbf{0.36} & \textbf{1.03} & 0/7 & 3.87 & 11.07 & 25.01 & 0/7 & 1.79 & 19.9 & 36.15 & 0/7 & 0.6 & 2.35 & 9.78 \\
	ISOLET & 6/7 & \textbf{0.0} & \textbf{0.3} & \textbf{0.65} & 0/7 & 1.24 & 2.68 & 7.54 & 0/7 & 0.37 & 1.02 & 2.59 & 0/7 & 0.4 & 1.75 & 3.2 \\
	Sensorless Drive Diagnosis & 6/7 & \textbf{-0.43} & \textbf{-0.21} & \textbf{6.23} & 0/7 & 242.17 & 672.88 & 706.19 & 0/7 & 154.89 & 182.9 & 224.55 & 0/7 & 630.63 & 674.68 & 694.06 \\
	Sensorless Drive Diagnosis (normalized) & 6/7 & \textbf{0.27} & \textbf{1.12} & \textbf{4.15} & 0/7 & 2.52 & 8.26 & 38.27 & 0/7 & 3.19 & 10.66 & 52.86 & 0/7 & 0.53 & 4.05 & 10.37 \\
	Online News Popularity & 5/7 & \textbf{0.35} & \textbf{1.51} & \textbf{5.79} & 0/7 & 30.14 & 95.91 & 229.91 & 0/7 & 13.2 & 33.77 & 121.0 & 0/7 & 8.5 & 16.5 & 37.86 \\
	Gas Sensor Array Drift & 5/7 & \textbf{-0.02} & \textbf{0.66} & \textbf{3.92} & 0/7 & 22.26 & 77.74 & 192.13 & 0/7 & 8.58 & 25.99 & 45.77 & 0/7 & 3.06 & 11.0 & 26.79 \\
	3D Road Network & 0/7 & 0.04 & 0.24 & 0.97 & 0/7 & 17.08 & 109.16 & 407.92 & 0/7 & 6.0 & 40.57 & 141.57 & 0/7 & 0.64 & 2.76 & 10.79 \\
	Skin Segmentation & 5/7 & \textbf{-0.1} & \textbf{0.93} & \textbf{4.23} & 0/7 & 11.16 & 32.25 & 77.69 & 0/7 & 5.34 & 20.97 & 82.67 & 0/7 & 0.18 & 3.86 & 13.63 \\
	KEGG Metabolic Relation Network (Directed) & 4/7 & \textbf{-0.32} & \textbf{0.28} & \textbf{4.74} & 0/7 & 1225.29 & 2024.58 & 2324.8 & 0/7 & 94.23 & 96.41 & 109.05 & 0/7 & 1627.77 & 1859.2 & 1980.12 \\
	Shuttle Control & 7/8 & \textbf{-0.15} & \textbf{2.58} & \textbf{47.68} & 0/8 & 369.96 & 558.74 & 958.42 & 0/8 & 136.81 & 174.37 & 239.82 & 0/8 & 1116.94 & 1230.13 & 1300.53 \\
	Shuttle Control (normalized) & 4/8 & 0.61 & \textbf{1.65} & \textbf{4.63} & 0/8 & 2.99 & 24.07 & 79.56 & 0/8 & 8.72 & 27.68 & 111.97 & 1/8 & 0.95 & 4.91 & 18.38 \\
	EEG Eye State & 4/8 & \textbf{0.53} & \textbf{0.56} & \textbf{42.33} & 0/8 & 93292.5 & 706326.79 & 1350191.25 & 0/8 & 3.85 & 937883.36 & 1020813.95 & 0/8 & 2438049.8 & 2593558.8 & 2639777.84 \\
	EEG Eye State (normalized) & 8/8 & \textbf{-0.06} & \textbf{-0.01} & \textbf{19.56} & 0/8 & 170.58 & 572.7 & 692.01 & 0/8 & 159.97 & 572.33 & 691.21 & 0/8 & 904.72 & 969.67 & 983.49 \\
	Pla85900 & 4/7 & 0.06 & \textbf{0.18} & \textbf{0.67} & 0/7 & 4.61 & 12.9 & 39.93 & 0/7 & 3.4 & 11.56 & 45.87 & 0/7 & 1.14 & 3.42 & 8.06 \\
	D15112 & 4/7 & 0.08 & \textbf{0.19} & \textbf{0.62} & 0/7 & 6.55 & 17.84 & 47.33 & 0/7 & 0.38 & 1.59 & 7.84 & 0/7 & 1.15 & 3.48 & 10.32 \\
	\hline
	Overall Results & 84/165 & \textbf{0.09} & \textbf{0.6} & 35802.84 & 1/165 & 5930.8 & 35740.12 & 72547.55 & 0/165 & 27.4 & 40836.31 & 49362.25 & 2/165 & 228388.95 & 235880.88 & 238155.85 \\ \hline
\end{tabular}%
}

\vspace{10pt}

\resizebox{!}{\esummtableheight}{%
\begin{tabular}{l|cccc|cccc|cccc|cccc}
	\hline
	\multirow{2}{*}{Dataset} & \multicolumn{4}{p{3cm}}{\mbox{K-means++}}& \multicolumn{4}{p{3cm}}{\mbox{CURE}}& \multicolumn{4}{p{3cm}}{\mbox{CluDataSE}}& \multicolumn{4}{p{3cm}}{\mbox{LW-coreset}} \\
	\cline{2-17}
	& \#Succ & Min & Median & Max & \#Succ & Min & Median & Max & \#Succ & Min & Median & Max & \#Succ & Min & Median & Max  \\
	\hline
	CORD-19 Embeddings & 3/7 & \textbf{-0.01} & 0.13 & 1.44 & 0/7 & 20.59 & 28.29 & 35.69 & 3/7 & -0.0 & 0.14 & 0.83 & 0/7 & 0.04 & 0.25 & 0.76 \\
	HEPMASS & 2/7 & 0.04 & 0.21 & 0.69 & 0/7 & 24.12 & 26.32 & 29.19 & 3/7 & 0.08 & 0.17 & 0.73 & 0/7 & 0.13 & 0.55 & 1.01 \\
	US Census Data 1990 & 1/7 & 1.4 & 4.95 & 84.97 & 0/7 & 17.55 & 24.92 & 33.33 & 1/7 & 4.1 & 9.59 & 27.51 & 0/7 & 9.11 & 118.25 & 245.89 \\
	Gisette & 2/7 & \textbf{-0.52} & \textbf{-0.48} & \textbf{-0.39} & 0/7 & 8.87 & 9.51 & 9.85 & 5/7 & \textbf{-0.52} & \textbf{-0.48} & \textbf{-0.39} & 0/7 & -0.36 & -0.27 & -0.11 \\
	Music Analysis & 4/7 & \textbf{-0.03} & \textbf{0.55} & 7.52 & 0/7 & 26.74 & 44.31 & 79.79 & 2/7 & 0.0 & 2.19 & 2.96 & 0/7 & 0.53 & 1.88 & 10.36 \\
	Protein Homology & 6/7 & \textbf{0.02} & \textbf{0.4} & 15.15 & 0/7 & 216.6 & 243.48 & 286.52 & 2/7 & 14.85 & 14.9 & 15.01 & 0/7 & 0.44 & 14.68 & 15.54 \\
	MiniBooNE Particle Identification & 4/7 & -0.06 & 0.48 & \textbf{24.53} & 0/7 & 2832329.21 & 2832343.36 & 2832361.67 & 2/7 & 5.71 & 40994.64 & 40994.98 & 0/7 & 1.11 & 4.11 & 28.27 \\
	MiniBooNE Particle Identification (normalized) & 5/7 & \textbf{-0.05} & 0.73 & 101.48 & 0/7 & 22.75 & 33.94 & 155.39 & 4/7 & 0.09 & \textbf{0.56} & 2.07 & 0/7 & 0.39 & 2.25 & 4.4 \\
	MFCCs for Speech Emotion Recognition & 3/7 & \textbf{0.0} & 0.72 & 2.34 & 0/7 & 22.72 & 39.4 & 71.09 & 4/7 & 0.75 & 1.25 & 3.22 & 0/7 & 0.37 & 1.75 & 4.5 \\
	ISOLET & 1/7 & 0.02 & 0.63 & 2.15 & 0/7 & 25.42 & 38.99 & 48.75 & 0/7 & 0.35 & 1.34 & 2.31 & 0/7 & 0.35 & 1.45 & 3.73 \\
	Sensorless Drive Diagnosis & 1/7 & -0.38 & 10.76 & 63.17 & 0/7 & 739.52 & 755.39 & 778.76 & 0/7 & 122.72 & 162.38 & 224.34 & 0/7 & 21.94 & 128.63 & 145.29 \\
	Sensorless Drive Diagnosis (normalized) & 1/7 & 0.47 & 4.27 & 22.13 & 0/7 & 14.23 & 47.3 & 97.82 & 0/7 & 1.25 & 6.06 & 26.66 & 0/7 & 2.65 & 11.54 & 57.73 \\
	Online News Popularity & 1/7 & 1.07 & 6.71 & 29.74 & 0/7 & 81.69 & 137.86 & 210.37 & 2/7 & 9.29 & 17.71 & 30.4 & 0/7 & 1.86 & 16.81 & 77.2 \\
	Gas Sensor Array Drift & 2/7 & 0.32 & 3.5 & 22.34 & 0/7 & 57.44 & 82.36 & 135.38 & 1/7 & 8.08 & 16.14 & 26.17 & 0/7 & 1.63 & 8.12 & 28.57 \\
	3D Road Network & 7/7 & \textbf{-0.0} & \textbf{0.01} & \textbf{0.54} & 0/7 & 44.47 & 100.47 & 215.66 & 5/7 & 0.23 & 0.23 & 4.01 & 0/7 & 0.78 & 4.68 & 12.73 \\
	Skin Segmentation & 1/7 & 0.29 & 5.24 & 17.35 & 0/7 & 27.42 & 95.62 & 165.49 & 2/7 & 6.68 & 15.73 & 23.83 & 0/7 & 4.46 & 15.72 & 34.25 \\
	KEGG Metabolic Relation Network (Directed) & 2/7 & 0.22 & 4.84 & 66.24 & 0/7 & 2817.93 & 2944.16 & 3240.08 & 1/7 & 94.27 & 94.27 & 95.61 & 0/7 & 11.19 & 64.11 & 77.88 \\
	Shuttle Control & 1/8 & 2.76 & 16.01 & 73.7 & 0/8 & 1336.47 & 1339.01 & 1341.97 & 0/8 & 139.1 & 189.52 & 232.24 & 0/8 & 8.08 & 45.09 & 182.03 \\
	Shuttle Control (normalized) & 1/8 & 1.03 & 7.28 & 38.15 & 0/8 & 33.8 & 88.43 & 293.11 & 3/8 & \textbf{0.39} & 3.98 & 51.82 & 0/8 & 7.31 & 32.76 & 126.54 \\
	EEG Eye State & 3/8 & 0.54 & 4.3 & 51.77 & 0/8 & 2641365.35 & 2641430.58 & 2641645.46 & 1/8 & 854977.62 & 1020813.06 & 1020813.77 & 0/8 & 1.29 & 124.33 & 35642.74 \\
	EEG Eye State (normalized) & 0/8 & \textbf{-0.06} & 23.01 & 51.12 & 0/8 & 980.93 & 993.82 & 1009.74 & 0/8 & 462.46 & 572.78 & 620.43 & 0/8 & 59.44 & 305.97 & 564.1 \\
	Pla85900 & 2/7 & \textbf{-0.01} & 0.45 & 2.09 & 0/7 & 17.49 & 40.43 & 71.38 & 2/7 & 0.0 & 0.34 & 1.99 & 0/7 & 0.2 & 1.02 & 3.21 \\
	D15112 & 3/7 & \textbf{0.01} & 0.74 & 3.21 & 0/7 & 13.76 & 33.03 & 62.42 & 2/7 & 0.5 & 2.85 & 4.79 & 0/7 & 0.51 & 2.21 & 5.43 \\
	\hline
	Overall Results & 56/165 & 0.31 & 4.15 & \textbf{29.63} & 0/165 & 238271.52 & 238300.91 & 238364.3 & 45/165 & 37210.78 & 46213.89 & 46226.32 & 0/165 & 5.8 & 39.39 & 1620.52 \\ \hline
\end{tabular}%
}
\end{table}

\newpage

\begin{table}[!htbp]%
\centering%
\captionsetup{font=scriptsize}
\caption{Total clustering times ($t$, in seconds) resulting from the comparison of the selected clustering algorithms.}%
\label{tab:result_t}%
\resizebox{!}{\tsummtableheight}{%
\begin{tabular}{l|cccc|cccc|cccc|cccc}
	\hline
	\multirow{2}{*}{Dataset} & \multicolumn{4}{p{3cm}}{\mbox{Big-means}}& \multicolumn{4}{p{3cm}}{\mbox{IK-means}}& \multicolumn{4}{p{3cm}}{\mbox{BDCSM}}& \multicolumn{4}{p{3cm}}{\mbox{Minibatch K-means}} \\
	\cline{2-17}
	& \#Succ & Min & Median & Max & \#Succ & Min & Median & Max & \#Succ & Min & Median & Max & \#Succ & Min & Median & Max  \\
	\hline
	CORD-19 Embeddings & 2/7 & 15.14 & 26.27 & 38.34 & 0/7 & 31.44 & 32.08 & 34.0 & 1/7 & 61.42 & 74.85 & 96.2 & 4/7 & \textbf{9.62} & \textbf{15.7} & \textbf{20.46} \\
	HEPMASS & 0/7 & \textbf{5.05} & 19.84 & 27.32 & 0/7 & 110.12 & 112.59 & 117.47 & 2/7 & 32.22 & 35.01 & 37.59 & 0/7 & 10.36 & 12.49 & 16.72 \\
	US Census Data 1990 & 2/7 & \textbf{0.39} & 2.3 & \textbf{3.04} & 0/7 & 26.46 & 28.34 & 30.31 & 2/7 & 4.07 & 4.45 & 5.3 & 3/7 & 0.75 & \textbf{1.78} & 3.38 \\
	Gisette & 0/7 & 15.75 & 19.6 & 26.08 & 4/7 & 6.44 & 6.8 & \textbf{7.42} & 0/7 & 21.4 & 34.54 & 68.3 & 3/7 & \textbf{3.43} & \textbf{6.57} & 11.64 \\
	Music Analysis & 0/7 & 1.53 & 5.29 & 8.66 & 0/7 & 4.39 & 4.68 & 5.0 & 1/7 & 5.4 & 6.9 & 9.94 & 5/7 & \textbf{0.64} & \textbf{1.34} & \textbf{2.75} \\
	Protein Homology & 0/7 & 1.62 & 3.12 & 5.27 & 1/7 & 1.29 & 1.3 & 2.01 & 0/7 & 4.85 & 7.88 & 11.72 & 4/7 & \textbf{0.62} & \textbf{0.89} & \textbf{1.44} \\
	MiniBooNE Particle Identification & 0/7 & 2.37 & 4.12 & 6.32 & 2/7 & 0.91 & \textbf{0.93} & \textbf{1.37} & 0/7 & 8.15 & 12.24 & 18.94 & 2/7 & \textbf{0.71} & 1.12 & 3.02 \\
	MiniBooNE Particle Identification (normalized) & 0/7 & 0.31 & 0.8 & 1.29 & 0/7 & 4.83 & 6.04 & 7.44 & 2/7 & 0.81 & 1.2 & 1.63 & 4/7 & \textbf{0.18} & \textbf{0.3} & \textbf{0.61} \\
	MFCCs for Speech Emotion Recognition & 0/7 & 0.26 & 0.74 & 1.28 & 0/7 & 0.62 & 0.71 & 1.04 & 3/7 & 0.73 & 0.98 & 1.29 & 3/7 & \textbf{0.2} & \textbf{0.29} & \textbf{0.47} \\
	ISOLET & 0/7 & 0.98 & 3.31 & 4.75 & 3/7 & 0.36 & 0.38 & \textbf{0.44} & 1/7 & 0.41 & 0.63 & 1.32 & 3/7 & \textbf{0.22} & \textbf{0.3} & 0.46 \\
	Sensorless Drive Diagnosis & 0/7 & 0.73 & 1.4 & 2.48 & 0/7 & 0.51 & 0.53 & 0.75 & 0/7 & 1.25 & 2.12 & 3.96 & 0/7 & 0.25 & 0.38 & 0.72 \\
	Sensorless Drive Diagnosis (normalized) & 0/7 & 0.05 & 0.23 & 0.33 & 0/7 & 0.7 & 0.88 & 1.27 & 1/7 & 0.1 & 0.14 & 0.23 & 0/7 & 0.06 & 0.14 & 0.33 \\
	Online News Popularity & 0/7 & 0.14 & 0.49 & 0.87 & 0/7 & 0.3 & 0.31 & 0.48 & 1/7 & 0.51 & 0.83 & 1.21 & 3/7 & 0.12 & \textbf{0.18} & \textbf{0.29} \\
	Gas Sensor Array Drift & 0/7 & 0.42 & 1.3 & 2.09 & 5/7 & 0.16 & \textbf{0.16} & \textbf{0.21} & 1/7 & 0.25 & 0.61 & 1.29 & 1/7 & \textbf{0.14} & 0.27 & 0.48 \\
	3D Road Network & 0/7 & 0.16 & 0.46 & 1.33 & 0/7 & 2.17 & 2.76 & 3.61 & 0/7 & 1.89 & 2.28 & 3.1 & 0/7 & 0.42 & 0.62 & 1.03 \\
	Skin Segmentation & 0/7 & 0.03 & 0.16 & 0.21 & 0/7 & 1.05 & 1.06 & 1.56 & 0/7 & 0.06 & 0.08 & 0.11 & 0/7 & 0.06 & 0.15 & 0.26 \\
	KEGG Metabolic Relation Network (Directed) & 0/7 & 0.27 & 0.8 & 1.15 & 0/7 & 0.34 & 0.36 & 0.89 & 0/7 & 1.2 & 1.61 & 2.08 & 0/7 & 0.27 & 0.43 & \textbf{0.84} \\
	Shuttle Control & 0/8 & 0.26 & 0.96 & 1.47 & 0/8 & 0.28 & 0.31 & 0.49 & 0/8 & 0.09 & 0.17 & 0.35 & 0/8 & 0.18 & 0.26 & 0.46 \\
	Shuttle Control (normalized) & 0/8 & 0.04 & 0.26 & 0.39 & 0/8 & 0.6 & 0.84 & 1.24 & 0/8 & 0.02 & 0.02 & 0.03 & 0/8 & 0.04 & 0.09 & 0.2 \\
	EEG Eye State & 0/8 & 0.22 & 0.94 & 1.46 & 0/8 & 0.08 & 0.08 & 0.1 & 0/8 & 0.07 & 0.12 & 0.19 & 0/8 & 0.06 & 0.09 & 0.14 \\
	EEG Eye State (normalized) & 0/8 & 0.17 & 0.71 & 0.96 & 0/8 & 0.83 & 1.26 & 1.97 & 0/8 & 0.06 & 0.11 & 0.21 & 0/8 & 0.05 & 0.09 & 0.17 \\
	Pla85900 & 0/7 & 0.07 & 0.9 & 1.48 & 0/7 & 0.34 & 0.35 & 0.62 & 0/7 & 0.05 & 0.08 & 0.13 & 0/7 & 0.09 & 0.16 & 0.29 \\
	D15112 & 0/7 & 0.15 & 1.0 & 1.44 & 0/7 & 0.07 & 0.09 & 0.09 & 4/7 & \textbf{0.01} & \textbf{0.01} & \textbf{0.02} & 0/7 & 0.07 & 0.11 & 0.16 \\
	\hline
	Overall Results & 4/165 & 2.0 & 4.13 & 6.0 & 15/165 & 8.45 & 8.82 & 9.56 & 19/165 & 6.31 & 8.12 & 11.53 & 35/165 & \textbf{1.24} & \textbf{1.9} & \textbf{2.89} \\ \hline
\end{tabular}%
}

\vspace{10pt}

\resizebox{!}{\tsummtableheight}{%
\begin{tabular}{l|cccc|cccc|cccc|cccc}
	\hline
	\multirow{2}{*}{Dataset} & \multicolumn{4}{p{3cm}}{\mbox{K-means++}}& \multicolumn{4}{p{3cm}}{\mbox{CURE}}& \multicolumn{4}{p{3cm}}{\mbox{CluDataSE}}& \multicolumn{4}{p{3cm}}{\mbox{LW-coreset}} \\
	\cline{2-17}
	& \#Succ & Min & Median & Max & \#Succ & Min & Median & Max & \#Succ & Min & Median & Max & \#Succ & Min & Median & Max  \\
	\hline
	CORD-19 Embeddings & 0/7 & 464.35 & 815.42 & 1536.13 & 0/7 & 33.29 & 38.02 & 51.0 & 0/7 & 335.89 & 693.94 & 1360.43 & 0/7 & 23.66 & 33.47 & 52.41 \\
	HEPMASS & 0/7 & 312.36 & 590.65 & 929.22 & 0/7 & 47.1 & 49.38 & 56.46 & 0/7 & 155.94 & 412.73 & 1263.65 & 5/7 & 6.34 & \textbf{7.22} & \textbf{8.88} \\
	US Census Data 1990 & 0/7 & 49.42 & 73.35 & 132.69 & 0/7 & 3.36 & 3.7 & 4.77 & 0/7 & 30.64 & 54.71 & 165.78 & 0/7 & 3.22 & 3.34 & 4.72 \\
	Gisette & 0/7 & 39.19 & 63.33 & 105.25 & 0/7 & 23.87 & 25.3 & 34.47 & 0/7 & 51.86 & 82.55 & 136.48 & 0/7 & 23.28 & 33.97 & 53.87 \\
	Music Analysis & 0/7 & 51.27 & 79.46 & 216.08 & 0/7 & 3.36 & 3.69 & 5.17 & 0/7 & 60.2 & 99.23 & 184.61 & 1/7 & 2.11 & 3.37 & 5.89 \\
	Protein Homology & 1/7 & 7.86 & 14.41 & 26.7 & 0/7 & 87.88 & 93.78 & 107.19 & 0/7 & 20.36 & 26.57 & 37.94 & 1/7 & 3.12 & 4.35 & 10.36 \\
	MiniBooNE Particle Identification & 1/7 & 3.33 & 7.96 & 14.53 & 0/7 & 217.68 & 243.64 & 272.31 & 0/7 & 11.9 & 15.23 & 18.55 & 2/7 & 0.99 & 1.82 & 3.18 \\
	MiniBooNE Particle Identification (normalized) & 0/7 & 4.41 & 6.91 & 16.33 & 0/7 & 6.53 & 7.04 & 7.93 & 0/7 & 5.11 & 8.31 & 14.57 & 1/7 & 0.26 & 0.4 & 0.62 \\
	MFCCs for Speech Emotion Recognition & 0/7 & 2.17 & 3.89 & 6.68 & 0/7 & 7.86 & 8.99 & 10.31 & 0/7 & 2.88 & 5.42 & 8.42 & 1/7 & 0.23 & 0.44 & 0.82 \\
	ISOLET & 0/7 & 1.28 & 1.92 & 4.12 & 0/7 & 1.87 & 1.94 & 2.2 & 0/7 & 1.19 & 2.07 & 3.98 & 0/7 & 0.55 & 0.77 & 1.44 \\
	Sensorless Drive Diagnosis & 0/7 & 0.82 & 1.42 & 2.77 & 0/7 & 4.5 & 4.82 & 5.62 & 0/7 & 1.73 & 2.69 & 4.71 & 7/7 & \textbf{0.12} & \textbf{0.19} & \textbf{0.34} \\
	Sensorless Drive Diagnosis (normalized) & 0/7 & 0.43 & 0.81 & 2.09 & 0/7 & 1.18 & 1.35 & 1.59 & 0/7 & 0.34 & 0.78 & 1.84 & 6/7 & \textbf{0.03} & \textbf{0.05} & \textbf{0.09} \\
	Online News Popularity & 0/7 & 0.52 & 0.79 & 1.92 & 0/7 & 6.16 & 6.73 & 7.77 & 0/7 & 1.45 & 2.05 & 4.19 & 3/7 & \textbf{0.1} & \textbf{0.18} & 0.43 \\
	Gas Sensor Array Drift & 0/7 & 0.33 & 0.55 & 1.11 & 0/7 & 6.86 & 8.12 & 11.61 & 0/7 & 0.63 & 1.19 & 1.9 & 0/7 & 0.19 & 0.36 & 0.81 \\
	3D Road Network & 0/7 & 1.36 & 5.39 & 9.37 & 0/7 & 14.51 & 16.58 & 19.05 & 0/7 & 10.66 & 11.14 & 12.9 & 7/7 & \textbf{0.04} & \textbf{0.07} & \textbf{0.13} \\
	Skin Segmentation & 0/7 & 0.17 & 0.29 & 0.8 & 0/7 & 4.46 & 4.9 & 5.57 & 0/7 & 0.24 & 0.43 & 0.76 & 7/7 & \textbf{0.01} & \textbf{0.02} & \textbf{0.03} \\
	KEGG Metabolic Relation Network (Directed) & 0/7 & 0.21 & 0.36 & 1.28 & 0/7 & 4.81 & 5.44 & 13.67 & 0/7 & 1.76 & 2.02 & 2.49 & 7/7 & \textbf{0.05} & \textbf{0.08} & 2.53 \\
	Shuttle Control & 0/8 & 0.04 & 0.1 & 0.22 & 0/8 & 6.35 & 7.06 & 7.83 & 0/8 & 0.45 & 0.58 & 0.76 & 8/8 & \textbf{0.01} & \textbf{0.02} & \textbf{0.04} \\
	Shuttle Control (normalized) & 0/8 & 0.05 & 0.09 & 0.16 & 0/8 & 0.57 & 0.6 & 0.72 & 0/8 & 0.04 & 0.07 & 0.15 & 8/8 & \textbf{0.01} & \textbf{0.01} & \textbf{0.01} \\
	EEG Eye State & 1/8 & 0.05 & 0.09 & 0.24 & 0/8 & 3.12 & 3.44 & 4.13 & 0/8 & 0.9 & 0.96 & 1.1 & 7/8 & \textbf{0.02} & \textbf{0.03} & \textbf{0.08} \\
	EEG Eye State (normalized) & 0/8 & 0.05 & 0.11 & 0.27 & 0/8 & 3.36 & 3.54 & 3.96 & 0/8 & 0.46 & 0.74 & 1.02 & 8/8 & \textbf{0.02} & \textbf{0.04} & \textbf{0.08} \\
	Pla85900 & 0/7 & 0.14 & 0.3 & 0.75 & 0/7 & 160.17 & 176.9 & 204.49 & 0/7 & 0.15 & 0.29 & 0.75 & 7/7 & \textbf{0.02} & \textbf{0.03} & \textbf{0.08} \\
	D15112 & 0/7 & 0.02 & 0.03 & 0.08 & 0/7 & 11.13 & 12.25 & 13.56 & 0/7 & 0.06 & 0.08 & 0.11 & 3/7 & \textbf{0.01} & \textbf{0.01} & \textbf{0.02} \\
	\hline
	Overall Results & 3/165 & 40.86 & 72.51 & 130.82 & 0/165 & 28.7 & 31.62 & 37.02 & 0/165 & 30.21 & 61.9 & 140.31 & 89/165 & 2.8 & 3.92 & 6.39 \\ \hline
\end{tabular}%
}
\end{table}

\end{landscape}

Appendix~\ref{app:details} provides detailed information on all experiments conducted. In the tables of Appendix~\ref{app:details} showing clustering details, $n_d$ denotes the total number of distance function evaluations performed by an algorithm across different executions and all choices of $k$, while $n_s$ stands for the total number of processed samples by the Big-means algorithm.


\subsection{Analysis of obtained results}

In terms of clustering accuracy $\eps$, averaged across all the datasets, only the following three algorithms exhibited satisfactory results: Big-means, K-means++, and LW-coreset. Notably, Big-means was the top performer, showing a substantial lead in overall effectiveness. In fact, it was the only algorithm that achieved the average accuracy gap of less than $1\%$. The remaining algorithms yielded less consistent outcomes, particularly when applied to specific datasets, such as ``MiniBooNE Particle Identification'' and ``EEG Eye State'' datasets. These datasets are marked by an elevated susceptibility to clustering instability, where even minor deviations in the initial centroid positions yield significantly deteriorated final optima.

Big-means attained the best accuracy results on the small and large datasets, while achieving better or comparable accuracy results for the biggest datasets. The ``Max'' statistic of Big-means has the lowest values across the majority of datasets, and this reveals that Big-means possesses the highest stability with respect to a large number of executions. We attribute this property to the iterative sampling nature of Big-means, especially to its shaking effect that allows Big-means to avoid the valleys of bad local minima.

Regarding the clustering time $t$, the K-means++, CURE, CluDataSE and BDCSM algorithms exhibited the slowest performance. By looking at the obtained time scores of these algorithms for the biggest datasets in the list, one can easily note the inability of these algorithms to efficiently cluster datasets with a huge number of entries. The Minibatch K-means and LW-coreset were the fastest on average, with Minibatch K-means leading in the big datasets, and LW-coreset leading in the small and large datasets.

Adopting the trade-off between the clustering accuracy and time as a criterion, the Big-means algorithm appears as an obvious leader. Big-means achieved the top average accuracy score, while exhibiting the best or comparable time results among the most accurate algorithms. It is interesting to note that although Minibatch K-means was the fastest algorithm by far, it also produced one of the most inaccurate average results. This observation justifies our belief that although simplicity of an algorithm is crucial for the ease of implementation, parallelization, and processing speed, excessive simplicity can undermine the algorithm's accuracy and effectiveness.

Thus, in developing new algorithms, it is vital to strike the right balance between the aspects of effectiveness (accuracy), efficiency (time), and simplicity (implementation and time). As we mentioned earlier, the LIMA dominance criterion~\cite{Brimberg2023} is a measure that takes into account all these aspects. In this study, we adopt LIMA dominance as the final measure of the obtained experimental results. Among the algorithms that obtained the best accuracy scores (Big-means, K-means++, and LW-coreset), Big-means stands out as LIMA-dominant:

$$
\text{Big-means}(0.6, 4.13, 6) \ \underset{LIMA}{>} \ \text{K-means++}(4.15, 72.51, 6)
$$
$$
\text{Big-means}(0.6, 4.13, 6) \ \underset{LIMA}{>} \ \text{LW-coreset}(39.39, 3.92, 7)
$$
Here the notation $A(accuracy, \ time, \ LIMA \ number)$ is used to represent the performance of an algorithm, and the last LIMA comparison was made assuming that the time results of $4.13$ and $3.92$ seconds were comparable.

\section{Conclusion and Future Research} \label{sec:conclusion}

The findings of our study highlight the challenges involved in solving the NP-hard MSSC problem in big data conditions using K-means-like methods. Based on the comprehensive review of the available approaches for optimizing K-means, it is apparent that only very few ones can be applied to real big data. In our experimental findings, Big-means~\cite{Mussabayev2023} emerged as a standout big data clustering algorithm. It outperformed traditional and other tested MSSC methods at all data sizes (small, large, and big), achieving a perfect balance of accuracy, speed, and simplicity, as proven by the LIMA dominance criterion~\cite{Brimberg2023}. The inherent simplicity of Big-means challenges the notion that more intricate hybrid methodologies are required to achieve superior clustering outcomes. While Big-means does necessitate a well-informed selection for its sample size parameter $s$, it has the potential to serve as a definitive algorithm within the realm of MSSC algorithms based on K-means~\cite{Mussabayev2023}. Nevertheless, its efficacy on big data could be further enhanced through the integration of advanced metaheuristics.

The LW-coreset algorithm for coreset construction, which necessitates a minimum of two passes over the complete dataset, proves infeasible under big data constraints. Similarly, many alternative approaches either demand multiple dataset passes or utilize intricate concepts and metaheuristics, resulting in high computational demands. These methods often grapple with memory overflow issues or exorbitant processing times~\cite{Mussabayev2023}. While sampling-based clustering techniques such as BDCSM, Minibatch K-means, and CluDataSE expedite processing time through sampling, their final clustering accuracy remains suboptimal. Singularly applying other methods intrinsic to big data clustering algorithms, like parallelization, triangle inequality, employing streamlined data structures, or hybridization, fails to yield satisfactory outcomes on big datasets. We posit that these strategies truly shine when integrated as enhancements to a big data clustering algorithm rooted in foundational principles like the ``less is more'' and decomposition approaches~\cite{Mussabayev2023}. Such principles are quintessential for genuine big data clustering algorithms.

Other recent surveys of scalable and parallel big data clustering methods emphasize that most of the existing algorithms have high-complexity implementations. Furthermore, few existing methods have shown support for the velocity and variety characteristics of big data. Most parallel clustering algorithms proposed in the literature do not handle real-time data and focus on a single, mostly numerical, type of data.

In this study, based on our research findings and critical insights, we provide guidelines for selecting the optimal big data clustering algorithm. We anticipate that the accompanying infographic and the algorithm selection flowchart will serve as useful tools for practitioners struggling with emergent complexities inherent to big data clustering.

Our research underscores the pressing need for innovative MSSC algorithms adept at addressing the NP-hard problem in the context of big data. These solutions should be scalable, efficient, and streamlined to maximize their practical relevance. They ought to transcend the limitations inherent to current methodologies, such as scalability issues, the demand for multiple dataset passes, memory limitations, and the constraints of processing real-time data across multiple modalities. Upcoming research endeavors should prioritize the design of these novel algorithms, while also benchmarking them against existing approaches like the Big-means algorithm.

With the continuous advancement in computational power and data generation, the landscape of big data clustering continues to evolve. Techniques such as online clustering, where the algorithm adapts to data arriving in a stream, and ensemble (consensus) clustering, which combines multiple clustering results, present exciting avenues for future research and development. Additionally, we see untapped potential in the Big-means algorithm, which could be further enhanced by integrating modern metaheuristics like Variable Neighborhood Search (VNS), opening up new possibilities for improvement~\cite{Mladenovic1997}. We also aim to develop methods for automatically selecting the best parameters for big data clustering algorithms, including sample size and parallelization strategy, to optimize performance. Finding ways to combine state-of-the-art deep learning techniques with big data clustering approaches is another exiting future research direction.

\section*{Acknowledgements}

This research was funded by the Committee of Science of the Ministry of Science and Higher Education of the Republic of Kazakhstan (Grant No. BR21882268).

\begin{appendices}

\section{Details of Experiments with Real-World Datasets} \label{app:details}

\begin{landscape}

\subsection{CORD-19 Embeddings}
Dimensions: $m$ = 599616, $n$ = 768.
\par
Description: COVID-19 Open Research Dataset (CORD-19) is a resource of more than half a million scholarly articles about COVID-19, SARS-CoV-2, and related coronaviruses represended as embeddings in vectorized form.

\begin{table}[!htbp]
\centering

\caption{Summary of the results with CORD-19 Embeddings ($\times10^{9}$)}
\label{TabResultsD1}
\small
\resizebox{!}{\tableheight}{
\begin{tabular}{|l|l|llll|llll|llll|llll|}
\hline
\multicolumn{1}{|c|}{\multirow{3}{*}{$k$}} & \multicolumn{1}{c|}{\multirow{3}{*}{$f^*$}} & \multicolumn{4}{c|}{Big-means} & \multicolumn{4}{c|}{IK-means} & \multicolumn{4}{c|}{BDCSM} & \multicolumn{4}{c|}{Minibatch K-means} \\ \cline{3-18}
\multicolumn{1}{|c|}{} & \multicolumn{1}{c|}{} & \multicolumn{2}{c|}{$\varepsilon$} & \multicolumn{2}{c|}{$t$} & \multicolumn{2}{c|}{$\varepsilon$} & \multicolumn{2}{c|}{$t$} & \multicolumn{2}{c|}{$\varepsilon$} & \multicolumn{2}{c|}{$t$} & \multicolumn{2}{c|}{$\varepsilon$} & \multicolumn{2}{c|}{$t$} \\ \cline{3-18}
\multicolumn{1}{|c|}{} & \multicolumn{1}{c|}{} & \multicolumn{1}{c|}{med} & \multicolumn{1}{c|}{std} & \multicolumn{1}{c|}{med} & \multicolumn{1}{c|}{std} & \multicolumn{1}{c|}{med} & \multicolumn{1}{c|}{std} & \multicolumn{1}{c|}{med} & \multicolumn{1}{c|}{std} & \multicolumn{1}{c|}{med} & \multicolumn{1}{c|}{std} & \multicolumn{1}{c|}{med} & \multicolumn{1}{c|}{std} & \multicolumn{1}{c|}{med} & \multicolumn{1}{c|}{std} & \multicolumn{1}{c|}{med} & \multicolumn{1}{c|}{std} \\ \hline
2 & 2.03893$^*$ & 0.01 & 0.0 & 27.73 & 9.28 & 2.2 & 3.8 & 6.79 & 2.7 & 0.0 & 0.0 & 2.33 & 0.16 & 0.0 & 0.01 & 10.63 & 3.82 \\
3 & 1.9093$^*$ & 0.01 & 0.0 & 21.5 & 10.04 & 0.92 & 2.81 & 11.11 & 0.37 & 0.06 & 2.25 & 10.08 & 1.57 & 0.07 & 0.66 & 9.61 & 1.53 \\
5 & 1.77676$^*$ & 0.02 & 0.06 & 12.39 & 11.69 & 2.48 & 1.2 & 15.88 & 0.39 & 2.26 & 3.52 & 15.64 & 3.0 & 0.86 & 1.15 & 26.38 & 7.26 \\
10 & 1.62555$^*$ & 0.04 & 0.05 & 27.17 & 7.29 & 1.76 & 1.12 & 28.32 & 0.72 & 4.17 & 1.7 & 60.12 & 11.24 & 0.94 & 0.48 & 6.13 & 1.27 \\
15 & 1.55295$^*$ & 0.12 & 0.07 & 34.61 & 4.6 & 1.89 & 0.43 & 41.53 & 0.62 & 1.62 & 0.55 & 102.28 & 11.81 & 1.2 & 0.29 & 17.94 & 1.87 \\
20 & 1.49987$^*$ & 0.23 & 0.09 & 29.21 & 6.57 & 2.31 & 0.45 & 54.1 & 0.49 & 2.88 & 1.3 & 152.44 & 34.08 & 1.37 & 0.3 & 29.46 & 8.28 \\
25 & 1.46394$^*$ & 0.12 & 0.12 & 31.3 & 7.94 & 2.07 & 0.5 & 66.86 & 0.57 & 1.92 & 0.79 & 181.09 & 17.81 & 1.26 & 0.32 & 9.75 & 1.26 \\
\hline
\multicolumn{2}{|c|}{Mean:} & \textbf{0.08} & & \textbf{26.27} & & \textbf{1.95} & & \textbf{32.08} & & \textbf{1.84} & & \textbf{74.85} & & \textbf{0.81} & & \textbf{15.7} & \\ \hline
\end{tabular}
}

\medskip

\small
\resizebox{!}{\tableheight}{
\begin{tabular}{|l|l|llll|llll|llll|llll|}
\hline
\multicolumn{1}{|c|}{\multirow{3}{*}{$k$}} & \multicolumn{1}{c|}{\multirow{3}{*}{$f^*$}} & \multicolumn{4}{c|}{K-means++} & \multicolumn{4}{c|}{CURE} & \multicolumn{4}{c|}{CluDataSE} & \multicolumn{4}{c|}{LW-coreset} \\ \cline{3-18}
\multicolumn{1}{|c|}{} & \multicolumn{1}{c|}{} & \multicolumn{2}{c|}{$\varepsilon$} & \multicolumn{2}{c|}{$t$} & \multicolumn{2}{c|}{$\varepsilon$} & \multicolumn{2}{c|}{$t$} & \multicolumn{2}{c|}{$\varepsilon$} & \multicolumn{2}{c|}{$t$} & \multicolumn{2}{c|}{$\varepsilon$} & \multicolumn{2}{c|}{$t$} \\ \cline{3-18}
\multicolumn{1}{|c|}{} & \multicolumn{1}{c|}{} & \multicolumn{1}{c|}{med} & \multicolumn{1}{c|}{std} & \multicolumn{1}{c|}{med} & \multicolumn{1}{c|}{std} & \multicolumn{1}{c|}{med} & \multicolumn{1}{c|}{std} & \multicolumn{1}{c|}{med} & \multicolumn{1}{c|}{std} & \multicolumn{1}{c|}{med} & \multicolumn{1}{c|}{std} & \multicolumn{1}{c|}{med} & \multicolumn{1}{c|}{std} & \multicolumn{1}{c|}{med} & \multicolumn{1}{c|}{std} & \multicolumn{1}{c|}{med} & \multicolumn{1}{c|}{std} \\ \hline
2 & 2.03893$^*$ & 0.0 & 0.0 & 19.99 & 2.07 & 19.78 & 0.08 & 38.17 & 17.75 & 0.0 & 0.0 & 15.96 & 6.54 & 0.01 & 0.0 & 10.01 & 5.94 \\
3 & 1.9093$^*$ & 0.01 & 1.99 & 55.72 & 16.15 & 27.29 & 0.76 & 27.68 & 7.65 & -0.0 & 0.0 & 49.67 & 1.85 & 0.02 & 0.01 & 12.57 & 3.9 \\
5 & 1.77676$^*$ & 0.13 & 1.09 & 139.77 & 41.48 & 33.96 & 2.34 & 16.99 & 0.69 & 0.13 & 1.08 & 86.02 & 49.91 & 0.02 & 0.34 & 15.3 & 1.89 \\
10 & 1.62555$^*$ & 0.43 & 0.21 & 675.22 & 259.77 & 33.34 & 6.75 & 23.42 & 0.64 & 0.02 & 0.26 & 533.11 & 289.38 & 0.15 & 0.23 & 26.38 & 8.46 \\
15 & 1.55295$^*$ & 0.23 & 0.57 & 1072.21 & 438.69 & 35.74 & 9.6 & 30.67 & 1.04 & 0.52 & 0.32 & 1084.2 & 770.22 & 0.58 & 0.46 & 51.16 & 17.82 \\
20 & 1.49987$^*$ & 0.12 & 0.2 & 1167.53 & 737.09 & 23.87 & 9.32 & 40.98 & 8.34 & 0.29 & 0.15 & 1265.77 & 565.41 & 0.43 & 0.3 & 43.91 & 15.6 \\
25 & 1.46394$^*$ & -0.01 & 0.08 & 2577.51 & 813.09 & 24.06 & 6.6 & 88.2 & 4.27 & 0.01 & 0.27 & 1822.84 & 878.27 & 0.54 & 0.34 & 74.97 & 12.38 \\
\hline
\multicolumn{2}{|c|}{Mean:} & \textbf{0.13} & & \textbf{815.42} & & \textbf{28.29} & & \textbf{38.02} & & \textbf{0.14} & & \textbf{693.94} & & \textbf{0.25} & & \textbf{33.47} & \\ \hline
\end{tabular}
}

\bigskip

\caption{Clustering details with CORD-19 Embeddings}
\label{TabDetailsD1}
\resizebox{\linewidth}{!}{
\begin{tabular}{|l|l|lllll|l|l|ll|l|ll|lllll|ll|}
\hline
\multicolumn{1}{|c|}{\multirow{2}{*}{$k$}} & \multicolumn{1}{c|}{\multirow{2}{*}{$n_{exec}$}} & \multicolumn{5}{c|}{Big-means} & \multicolumn{1}{c|}{IK-means} & \multicolumn{1}{c|}{BDCSM} & \multicolumn{2}{c|}{Minibatch K-means} & \multicolumn{1}{c|}{K-means++} & \multicolumn{2}{c|}{CURE} & \multicolumn{5}{c|}{CluDataSE} & \multicolumn{2}{c|}{LW-coreset} \\ \cline{3-21}
\multicolumn{1}{|c|}{} & \multicolumn{1}{c|}{} & \multicolumn{1}{c|}{$s$} & \multicolumn{1}{c|}{$n_{s}$} & \multicolumn{1}{c|}{$T_1$} & \multicolumn{1}{c|}{$T_2$} & \multicolumn{1}{c|}{$n_{d}$} & \multicolumn{1}{c|}{$n_{d}$} & \multicolumn{1}{c|}{$n_{d}$} & \multicolumn{1}{c|}{$n_{s}$} & \multicolumn{1}{c|}{$n_{d}$} & \multicolumn{1}{c|}{$n_{d}$} & \multicolumn{1}{c|}{$s$} & \multicolumn{1}{c|}{$n_{d}$} & \multicolumn{1}{c|}{$s$} & \multicolumn{1}{c|}{$eps$} & \multicolumn{1}{c|}{$min\_pts$} & \multicolumn{1}{c|}{$sf$} & \multicolumn{1}{c|}{$n_{d}$} & \multicolumn{1}{c|}{$s$} & \multicolumn{1}{c|}{$n_{d}$} \\
\hline
2 & 7 & 32000 & 591 & 37.33 & 2.67 & 1.7E+08 & 4.8E+06 & 1.2E+07 & 42 & 2.7E+06 & 1.4E+07 & 32000 & 7.1E+07 & 32000 & 62.5 & 16 & 0.5 & 1.0E+09 & 32000 & 3.0E+06 \\
3 & 7 & 32000 & 381 & 32.0 & 8.0 & 2.2E+08 & 7.9E+06 & 5.1E+07 & 68 & 6.5E+06 & 5.6E+07 & 32000 & 4.2E+07 & 32000 & 62.5 & 16 & 0.5 & 1.1E+09 & 32000 & 5.8E+06 \\
5 & 7 & 32000 & 142 & 21.33 & 18.67 & 2.8E+08 & 1.3E+07 & 1.1E+08 & 81 & 1.3E+07 & 1.6E+08 & 32000 & 3.6E+07 & 32000 & 62.5 & 16 & 0.5 & 1.1E+09 & 32000 & 9.8E+06 \\
10 & 7 & 32000 & 184 & 24.0 & 16.0 & 3.6E+08 & 2.5E+07 & 4.4E+08 & 71 & 2.3E+07 & 8.6E+08 & 32000 & 5.3E+07 & 32000 & 62.5 & 16 & 0.5 & 1.7E+09 & 32000 & 2.3E+07 \\
15 & 7 & 32000 & 110 & 26.67 & 13.33 & 3.7E+08 & 3.7E+07 & 7.3E+08 & 73 & 3.5E+07 & 1.4E+09 & 32000 & 7.6E+07 & 32000 & 62.5 & 16 & 0.5 & 2.5E+09 & 32000 & 6.7E+07 \\
20 & 7 & 32000 & 42 & 8.0 & 32.0 & 3.5E+08 & 4.9E+07 & 1.0E+09 & 86 & 5.5E+07 & 1.5E+09 & 32000 & 9.8E+07 & 32000 & 62.5 & 16 & 0.5 & 2.9E+09 & 32000 & 4.5E+07 \\
25 & 7 & 32000 & 24 & 32.0 & 8.0 & 3.7E+08 & 6.2E+07 & 1.4E+09 & 75 & 6.0E+07 & 3.7E+09 & 32000 & 1.2E+08 & 32000 & 62.5 & 16 & 0.5 & 3.6E+09 & 32000 & 9.9E+07 \\
\hline
\end{tabular}
}

\end{table}

\newpage


\subsection{HEPMASS}
Dimensions: $m$ = 10500000, $n$ = 27.
\par
Description: The data set contains the 28 normalized features of physical particles that can be used for discovering the exotic ones in the field of high-energy physics.

\begin{table}[!htbp]
\centering

\caption{Summary of the results with HEPMASS ($\times10^{8}$)}
\label{TabResultsD2}
\small
\resizebox{!}{\tableheight}{
\begin{tabular}{|l|l|llll|llll|llll|llll|}
\hline
\multicolumn{1}{|c|}{\multirow{3}{*}{$k$}} & \multicolumn{1}{c|}{\multirow{3}{*}{$f^*$}} & \multicolumn{4}{c|}{Big-means} & \multicolumn{4}{c|}{IK-means} & \multicolumn{4}{c|}{BDCSM} & \multicolumn{4}{c|}{Minibatch K-means} \\ \cline{3-18}
\multicolumn{1}{|c|}{} & \multicolumn{1}{c|}{} & \multicolumn{2}{c|}{$\varepsilon$} & \multicolumn{2}{c|}{$t$} & \multicolumn{2}{c|}{$\varepsilon$} & \multicolumn{2}{c|}{$t$} & \multicolumn{2}{c|}{$\varepsilon$} & \multicolumn{2}{c|}{$t$} & \multicolumn{2}{c|}{$\varepsilon$} & \multicolumn{2}{c|}{$t$} \\ \cline{3-18}
\multicolumn{1}{|c|}{} & \multicolumn{1}{c|}{} & \multicolumn{1}{c|}{med} & \multicolumn{1}{c|}{std} & \multicolumn{1}{c|}{med} & \multicolumn{1}{c|}{std} & \multicolumn{1}{c|}{med} & \multicolumn{1}{c|}{std} & \multicolumn{1}{c|}{med} & \multicolumn{1}{c|}{std} & \multicolumn{1}{c|}{med} & \multicolumn{1}{c|}{std} & \multicolumn{1}{c|}{med} & \multicolumn{1}{c|}{std} & \multicolumn{1}{c|}{med} & \multicolumn{1}{c|}{std} & \multicolumn{1}{c|}{med} & \multicolumn{1}{c|}{std} \\ \hline
2 & 2.48889$^*$ & 0.0 & 0.0 & 12.72 & 7.67 & 0.08 & 2.52 & 24.35 & 2.1 & 0.0 & 0.0 & 2.91 & 0.28 & 0.01 & 0.03 & 9.73 & 2.21 \\
3 & 2.36789$^*$ & 0.01 & 0.01 & 12.73 & 7.17 & 1.93 & 0.49 & 41.72 & 1.48 & 0.38 & 0.79 & 5.25 & 0.48 & 1.29 & 0.58 & 10.96 & 1.57 \\
5 & 2.21106$^*$ & 0.33 & 0.16 & 22.27 & 5.92 & 1.88 & 0.66 & 61.15 & 2.16 & 3.1 & 2.43 & 9.07 & 0.81 & 1.26 & 0.76 & 14.15 & 2.19 \\
10 & 2.00353$^*$ & 0.02 & 0.05 & 23.13 & 5.73 & 3.13 & 0.79 & 112.1 & 3.44 & 1.26 & 1.23 & 27.19 & 1.88 & 1.2 & 1.09 & 12.84 & 2.58 \\
15 & 1.89922$^*$ & 0.18 & 0.08 & 19.67 & 6.86 & 2.46 & 0.38 & 143.8 & 1.8 & 1.63 & 0.77 & 51.17 & 1.16 & 1.12 & 0.52 & 8.23 & 0.72 \\
20 & 1.82904$^*$ & 0.17 & 0.06 & 24.68 & 10.35 & 2.81 & 0.71 & 183.13 & 4.07 & 2.53 & 0.54 & 70.66 & 5.14 & 1.21 & 0.43 & 20.9 & 4.47 \\
25 & 1.77524$^*$ & 0.13 & 0.08 & 23.64 & 10.68 & 3.37 & 0.56 & 221.85 & 3.29 & 2.4 & 0.66 & 78.79 & 3.16 & 1.3 & 0.35 & 10.62 & 0.75 \\
\hline
\multicolumn{2}{|c|}{Mean:} & \textbf{0.12} & & \textbf{19.84} & & \textbf{2.24} & & \textbf{112.59} & & \textbf{1.61} & & \textbf{35.01} & & \textbf{1.06} & & \textbf{12.49} & \\ \hline
\end{tabular}
}

\medskip

\small
\resizebox{!}{\tableheight}{
\begin{tabular}{|l|l|llll|llll|llll|llll|}
\hline
\multicolumn{1}{|c|}{\multirow{3}{*}{$k$}} & \multicolumn{1}{c|}{\multirow{3}{*}{$f^*$}} & \multicolumn{4}{c|}{K-means++} & \multicolumn{4}{c|}{CURE} & \multicolumn{4}{c|}{CluDataSE} & \multicolumn{4}{c|}{LW-coreset} \\ \cline{3-18}
\multicolumn{1}{|c|}{} & \multicolumn{1}{c|}{} & \multicolumn{2}{c|}{$\varepsilon$} & \multicolumn{2}{c|}{$t$} & \multicolumn{2}{c|}{$\varepsilon$} & \multicolumn{2}{c|}{$t$} & \multicolumn{2}{c|}{$\varepsilon$} & \multicolumn{2}{c|}{$t$} & \multicolumn{2}{c|}{$\varepsilon$} & \multicolumn{2}{c|}{$t$} \\ \cline{3-18}
\multicolumn{1}{|c|}{} & \multicolumn{1}{c|}{} & \multicolumn{1}{c|}{med} & \multicolumn{1}{c|}{std} & \multicolumn{1}{c|}{med} & \multicolumn{1}{c|}{std} & \multicolumn{1}{c|}{med} & \multicolumn{1}{c|}{std} & \multicolumn{1}{c|}{med} & \multicolumn{1}{c|}{std} & \multicolumn{1}{c|}{med} & \multicolumn{1}{c|}{std} & \multicolumn{1}{c|}{med} & \multicolumn{1}{c|}{std} & \multicolumn{1}{c|}{med} & \multicolumn{1}{c|}{std} & \multicolumn{1}{c|}{med} & \multicolumn{1}{c|}{std} \\ \hline
2 & 2.48889$^*$ & 0.0 & 0.0 & 31.37 & 20.57 & 13.0 & 0.63 & 28.88 & 3.87 & 0.0 & 0.0 & 13.83 & 3.86 & 0.0 & 0.0 & 5.59 & 0.21 \\
3 & 2.36789$^*$ & 0.0 & 0.7 & 39.93 & 20.92 & 17.13 & 1.1 & 22.64 & 6.19 & 0.0 & 0.56 & 25.57 & 12.63 & 1.25 & 0.56 & 5.79 & 0.16 \\
5 & 2.21106$^*$ & 0.32 & 0.19 & 90.69 & 23.6 & 22.48 & 1.36 & 24.59 & 2.9 & 0.32 & 0.36 & 56.38 & 8.13 & 0.82 & 0.26 & 5.55 & 0.28 \\
10 & 2.00353$^*$ & 0.28 & 0.3 & 431.3 & 198.04 & 29.31 & 2.31 & 35.22 & 2.15 & 0.42 & 0.38 & 237.68 & 82.07 & 0.9 & 0.38 & 7.33 & 1.76 \\
15 & 1.89922$^*$ & 0.33 & 0.13 & 770.54 & 220.21 & 32.29 & 2.8 & 45.82 & 1.26 & 0.18 & 0.12 & 562.15 & 456.89 & 0.31 & 0.49 & 7.49 & 1.59 \\
20 & 1.82904$^*$ & 0.26 & 0.14 & 1386.78 & 552.29 & 34.49 & 1.47 & 61.31 & 2.58 & 0.07 & 0.18 & 820.05 & 1400.54 & 0.25 & 0.34 & 8.8 & 0.63 \\
25 & 1.77524$^*$ & 0.25 & 0.23 & 1383.96 & 473.33 & 35.52 & 1.88 & 127.2 & 3.64 & 0.23 & 0.23 & 1173.44 & 520.57 & 0.31 & 0.24 & 10.01 & 1.23 \\
\hline
\multicolumn{2}{|c|}{Mean:} & \textbf{0.21} & & \textbf{590.65} & & \textbf{26.32} & & \textbf{49.38} & & \textbf{0.17} & & \textbf{412.73} & & \textbf{0.55} & & \textbf{7.22} & \\ \hline
\end{tabular}
}

\bigskip

\caption{Clustering details with HEPMASS}
\label{TabDetailsD2}
\resizebox{\linewidth}{!}{
\begin{tabular}{|l|l|lllll|l|l|ll|l|ll|lllll|ll|}
\hline
\multicolumn{1}{|c|}{\multirow{2}{*}{$k$}} & \multicolumn{1}{c|}{\multirow{2}{*}{$n_{exec}$}} & \multicolumn{5}{c|}{Big-means} & \multicolumn{1}{c|}{IK-means} & \multicolumn{1}{c|}{BDCSM} & \multicolumn{2}{c|}{Minibatch K-means} & \multicolumn{1}{c|}{K-means++} & \multicolumn{2}{c|}{CURE} & \multicolumn{5}{c|}{CluDataSE} & \multicolumn{2}{c|}{LW-coreset} \\ \cline{3-21}
\multicolumn{1}{|c|}{} & \multicolumn{1}{c|}{} & \multicolumn{1}{c|}{$s$} & \multicolumn{1}{c|}{$n_{s}$} & \multicolumn{1}{c|}{$T_1$} & \multicolumn{1}{c|}{$T_2$} & \multicolumn{1}{c|}{$n_{d}$} & \multicolumn{1}{c|}{$n_{d}$} & \multicolumn{1}{c|}{$n_{d}$} & \multicolumn{1}{c|}{$n_{s}$} & \multicolumn{1}{c|}{$n_{d}$} & \multicolumn{1}{c|}{$n_{d}$} & \multicolumn{1}{c|}{$s$} & \multicolumn{1}{c|}{$n_{d}$} & \multicolumn{1}{c|}{$s$} & \multicolumn{1}{c|}{$eps$} & \multicolumn{1}{c|}{$min\_pts$} & \multicolumn{1}{c|}{$sf$} & \multicolumn{1}{c|}{$n_{d}$} & \multicolumn{1}{c|}{$s$} & \multicolumn{1}{c|}{$n_{d}$} \\
\hline
2 & 7 & 64000 & 108 & 17.0 & 13.0 & 1.1E+08 & 9.7E+07 & 4.1E+08 & 332 & 4.2E+07 & 7.1E+08 & 64000 & 1.9E+08 & 64000 & 5.0 & 16 & 0.5 & 4.6E+09 & 64000 & 4.4E+07 \\
3 & 7 & 64000 & 92 & 7.0 & 23.0 & 2.0E+08 & 1.5E+08 & 1.2E+09 & 394 & 7.6E+07 & 1.5E+09 & 64000 & 1.4E+08 & 64000 & 5.0 & 16 & 0.5 & 5.1E+09 & 64000 & 6.0E+07 \\
5 & 7 & 64000 & 213 & 7.0 & 23.0 & 3.7E+08 & 2.4E+08 & 2.7E+09 & 515 & 1.6E+08 & 4.2E+09 & 64000 & 1.4E+08 & 64000 & 5.0 & 16 & 0.5 & 6.9E+09 & 64000 & 8.6E+07 \\
10 & 7 & 64000 & 186 & 16.0 & 14.0 & 8.0E+08 & 4.8E+08 & 1.1E+10 & 525 & 3.4E+08 & 2.4E+10 & 64000 & 2.1E+08 & 64000 & 5.0 & 16 & 0.5 & 1.8E+10 & 64000 & 1.9E+08 \\
15 & 7 & 64000 & 160 & 9.0 & 21.0 & 1.4E+09 & 7.2E+08 & 2.2E+10 & 572 & 5.5E+08 & 4.4E+10 & 64000 & 2.9E+08 & 64000 & 5.0 & 16 & 0.5 & 3.8E+10 & 64000 & 2.9E+08 \\
20 & 7 & 64000 & 225 & 28.0 & 2.0 & 2.0E+09 & 9.6E+08 & 2.9E+10 & 546 & 7.0E+08 & 8.3E+10 & 64000 & 3.8E+08 & 64000 & 5.0 & 16 & 0.5 & 5.5E+10 & 64000 & 4.0E+08 \\
25 & 7 & 64000 & 196 & 22.0 & 8.0 & 2.6E+09 & 1.2E+09 & 3.7E+10 & 511 & 8.2E+08 & 8.9E+10 & 64000 & 4.7E+08 & 64000 & 5.0 & 16 & 0.5 & 8.3E+10 & 64000 & 5.2E+08 \\
\hline
\end{tabular}
}

\end{table}

\newpage


\newpage

\subsection{US Census Data 1990}
Dimensions: $m$ = 2458285, $n$ = 68.
\par
Description: The data set was obtained from the (U.S. Department of Commerce) Census Bureau website and contains a one percent sample of the Public Use Microdata Samples (PUMS) person records drawn from the entire 1990 U.S. census sample.

\begin{table}[!htbp]
\centering

\caption{Summary of the results with US Census Data 1990 ($\times10^{8}$)}
\label{TabResultsD3}
\small
\resizebox{!}{\tableheight}{
\begin{tabular}{|l|l|llll|llll|llll|llll|}
\hline
\multicolumn{1}{|c|}{\multirow{3}{*}{$k$}} & \multicolumn{1}{c|}{\multirow{3}{*}{$f^*$}} & \multicolumn{4}{c|}{Big-means} & \multicolumn{4}{c|}{IK-means} & \multicolumn{4}{c|}{BDCSM} & \multicolumn{4}{c|}{Minibatch K-means} \\ \cline{3-18}
\multicolumn{1}{|c|}{} & \multicolumn{1}{c|}{} & \multicolumn{2}{c|}{$\varepsilon$} & \multicolumn{2}{c|}{$t$} & \multicolumn{2}{c|}{$\varepsilon$} & \multicolumn{2}{c|}{$t$} & \multicolumn{2}{c|}{$\varepsilon$} & \multicolumn{2}{c|}{$t$} & \multicolumn{2}{c|}{$\varepsilon$} & \multicolumn{2}{c|}{$t$} \\ \cline{3-18}
\multicolumn{1}{|c|}{} & \multicolumn{1}{c|}{} & \multicolumn{1}{c|}{med} & \multicolumn{1}{c|}{std} & \multicolumn{1}{c|}{med} & \multicolumn{1}{c|}{std} & \multicolumn{1}{c|}{med} & \multicolumn{1}{c|}{std} & \multicolumn{1}{c|}{med} & \multicolumn{1}{c|}{std} & \multicolumn{1}{c|}{med} & \multicolumn{1}{c|}{std} & \multicolumn{1}{c|}{med} & \multicolumn{1}{c|}{std} & \multicolumn{1}{c|}{med} & \multicolumn{1}{c|}{std} & \multicolumn{1}{c|}{med} & \multicolumn{1}{c|}{std} \\ \hline
2 & 18.39812$^*$ & 0.2 & 0.14 & 2.33 & 0.84 & 0.0 & 120.22 & 4.69 & 0.16 & 0.0 & 0.0 & 0.37 & 0.02 & 0.0 & 0.01 & 0.57 & 0.37 \\
3 & 6.1591$^*$ & 0.08 & 0.04 & 1.9 & 0.92 & 162.97 & 59.74 & 6.87 & 0.38 & 170.1 & 43.26 & 0.65 & 0.02 & 0.0 & 0.0 & 0.6 & 0.41 \\
5 & 3.35214$^*$ & 0.13 & 0.03 & 1.85 & 0.7 & 357.16 & 173.52 & 13.82 & 1.71 & 393.91 & 168.91 & 1.56 & 0.12 & 0.02 & 2.68 & 2.22 & 1.15 \\
10 & 2.36352$^*$ & 1.96 & 2.01 & 2.85 & 0.69 & 17.01 & 183.68 & 25.1 & 0.97 & 21.65 & 204.93 & 3.5 & 0.15 & 4.19 & 3.21 & 1.91 & 0.84 \\
15 & 2.04097$^*$ & 2.03 & 0.96 & 2.56 & 0.62 & 18.3 & 131.91 & 36.61 & 1.18 & 18.31 & 190.83 & 5.64 & 0.23 & 3.91 & 1.4 & 1.8 & 0.81 \\
20 & 1.81278$^*$ & 2.14 & 1.15 & 1.81 & 0.86 & 16.58 & 7.27 & 48.33 & 0.94 & 17.2 & 6.57 & 7.88 & 0.68 & 3.96 & 1.9 & 2.32 & 0.85 \\
25 & 1.64602$^*$ & 2.14 & 0.96 & 2.8 & 0.88 & 17.34 & 7.63 & 62.95 & 1.4 & 15.42 & 8.2 & 11.54 & 0.71 & 4.06 & 1.45 & 3.08 & 1.15 \\
\hline
\multicolumn{2}{|c|}{Mean:} & \textbf{1.24} & & \textbf{2.3} & & \textbf{84.2} & & \textbf{28.34} & & \textbf{90.94} & & \textbf{4.45} & & \textbf{2.31} & & \textbf{1.78} & \\ \hline
\end{tabular}
}

\medskip

\small
\resizebox{!}{\tableheight}{
\begin{tabular}{|l|l|llll|llll|llll|llll|}
\hline
\multicolumn{1}{|c|}{\multirow{3}{*}{$k$}} & \multicolumn{1}{c|}{\multirow{3}{*}{$f^*$}} & \multicolumn{4}{c|}{K-means++} & \multicolumn{4}{c|}{CURE} & \multicolumn{4}{c|}{CluDataSE} & \multicolumn{4}{c|}{LW-coreset} \\ \cline{3-18}
\multicolumn{1}{|c|}{} & \multicolumn{1}{c|}{} & \multicolumn{2}{c|}{$\varepsilon$} & \multicolumn{2}{c|}{$t$} & \multicolumn{2}{c|}{$\varepsilon$} & \multicolumn{2}{c|}{$t$} & \multicolumn{2}{c|}{$\varepsilon$} & \multicolumn{2}{c|}{$t$} & \multicolumn{2}{c|}{$\varepsilon$} & \multicolumn{2}{c|}{$t$} \\ \cline{3-18}
\multicolumn{1}{|c|}{} & \multicolumn{1}{c|}{} & \multicolumn{1}{c|}{med} & \multicolumn{1}{c|}{std} & \multicolumn{1}{c|}{med} & \multicolumn{1}{c|}{std} & \multicolumn{1}{c|}{med} & \multicolumn{1}{c|}{std} & \multicolumn{1}{c|}{med} & \multicolumn{1}{c|}{std} & \multicolumn{1}{c|}{med} & \multicolumn{1}{c|}{std} & \multicolumn{1}{c|}{med} & \multicolumn{1}{c|}{std} & \multicolumn{1}{c|}{med} & \multicolumn{1}{c|}{std} & \multicolumn{1}{c|}{med} & \multicolumn{1}{c|}{std} \\ \hline
2 & 18.39812$^*$ & 0.0 & 0.0 & 4.23 & 0.21 & 1.74 & 0.97 & 2.54 & 0.11 & 0.0 & 0.0 & 0.84 & 0.3 & 0.03 & 0.04 & 3.68 & 0.12 \\
3 & 6.1591$^*$ & 0.0 & 80.36 & 6.39 & 0.93 & 2.55 & 0.33 & 2.62 & 0.1 & 0.0 & 0.0 & 0.88 & 0.19 & 0.04 & 59.35 & 3.66 & 0.24 \\
5 & 3.35214$^*$ & 4.14 & 76.79 & 15.81 & 6.59 & 17.26 & 2.82 & 3.19 & 0.46 & 16.02 & 25.24 & 6.05 & 6.89 & 78.97 & 90.73 & 3.38 & 0.87 \\
10 & 2.36352$^*$ & 7.53 & 3.45 & 54.0 & 16.78 & 38.82 & 12.91 & 4.95 & 0.25 & 10.59 & 11.32 & 27.84 & 18.88 & 138.83 & 112.26 & 3.39 & 0.14 \\
15 & 2.04097$^*$ & 7.27 & 4.26 & 83.82 & 29.54 & 35.16 & 5.86 & 5.86 & 0.48 & 11.82 & 3.36 & 60.96 & 48.11 & 171.1 & 68.72 & 3.3 & 0.27 \\
20 & 1.81278$^*$ & 7.05 & 3.82 & 139.92 & 32.13 & 42.25 & 4.36 & 3.74 & 0.74 & 13.93 & 3.49 & 108.2 & 100.25 & 203.75 & 77.55 & 3.08 & 0.98 \\
25 & 1.64602$^*$ & 8.66 & 4.16 & 209.28 & 58.04 & 36.64 & 6.21 & 2.98 & 0.3 & 14.76 & 4.07 & 178.2 & 51.25 & 235.02 & 72.35 & 2.88 & 0.55 \\
\hline
\multicolumn{2}{|c|}{Mean:} & \textbf{4.95} & & \textbf{73.35} & & \textbf{24.92} & & \textbf{3.7} & & \textbf{9.59} & & \textbf{54.71} & & \textbf{118.25} & & \textbf{3.34} & \\ \hline
\end{tabular}
}

\bigskip

\caption{Clustering details with US Census Data 1990}
\label{TabDetailsD3}
\resizebox{\linewidth}{!}{
\begin{tabular}{|l|l|lllll|l|l|ll|l|ll|lllll|ll|}
\hline
\multicolumn{1}{|c|}{\multirow{2}{*}{$k$}} & \multicolumn{1}{c|}{\multirow{2}{*}{$n_{exec}$}} & \multicolumn{5}{c|}{Big-means} & \multicolumn{1}{c|}{IK-means} & \multicolumn{1}{c|}{BDCSM} & \multicolumn{2}{c|}{Minibatch K-means} & \multicolumn{1}{c|}{K-means++} & \multicolumn{2}{c|}{CURE} & \multicolumn{5}{c|}{CluDataSE} & \multicolumn{2}{c|}{LW-coreset} \\ \cline{3-21}
\multicolumn{1}{|c|}{} & \multicolumn{1}{c|}{} & \multicolumn{1}{c|}{$s$} & \multicolumn{1}{c|}{$n_{s}$} & \multicolumn{1}{c|}{$T_1$} & \multicolumn{1}{c|}{$T_2$} & \multicolumn{1}{c|}{$n_{d}$} & \multicolumn{1}{c|}{$n_{d}$} & \multicolumn{1}{c|}{$n_{d}$} & \multicolumn{1}{c|}{$n_{s}$} & \multicolumn{1}{c|}{$n_{d}$} & \multicolumn{1}{c|}{$n_{d}$} & \multicolumn{1}{c|}{$s$} & \multicolumn{1}{c|}{$n_{d}$} & \multicolumn{1}{c|}{$s$} & \multicolumn{1}{c|}{$eps$} & \multicolumn{1}{c|}{$min\_pts$} & \multicolumn{1}{c|}{$sf$} & \multicolumn{1}{c|}{$n_{d}$} & \multicolumn{1}{c|}{$s$} & \multicolumn{1}{c|}{$n_{d}$} \\
\hline
2 & 20 & 6000 & 219 & 0.2 & 2.8 & 1.3E+07 & 2.0E+07 & 1.9E+07 & 18 & 2.2E+05 & 1.5E+07 & 6000 & 2.2E+07 & 6000 & 4.5 & 16 & 0.5 & 4.6E+07 & 6000 & 9.9E+06 \\
3 & 20 & 6000 & 188 & 2.1 & 0.9 & 1.9E+07 & 2.9E+07 & 4.4E+07 & 34 & 6.0E+05 & 2.2E+07 & 6000 & 2.2E+07 & 6000 & 4.5 & 16 & 0.5 & 5.1E+07 & 6000 & 1.2E+07 \\
5 & 20 & 6000 & 194 & 0.6 & 2.4 & 3.5E+07 & 5.6E+07 & 1.5E+08 & 350 & 1.0E+07 & 1.6E+08 & 6000 & 2.6E+07 & 6000 & 4.5 & 16 & 0.5 & 1.5E+08 & 6000 & 1.7E+07 \\
10 & 20 & 6000 & 238 & 2.4 & 0.6 & 8.8E+07 & 1.1E+08 & 5.1E+08 & 306 & 1.8E+07 & 9.1E+08 & 6000 & 4.0E+07 & 6000 & 4.5 & 16 & 0.5 & 6.5E+08 & 6000 & 3.0E+07 \\
15 & 20 & 6000 & 186 & 1.9 & 1.1 & 1.3E+08 & 1.6E+08 & 9.2E+08 & 300 & 2.7E+07 & 1.3E+09 & 6000 & 5.3E+07 & 6000 & 4.5 & 16 & 0.5 & 1.5E+09 & 6000 & 4.3E+07 \\
20 & 20 & 6000 & 66 & 0.1 & 2.9 & 1.8E+08 & 2.1E+08 & 1.3E+09 & 422 & 5.1E+07 & 2.7E+09 & 6000 & 6.4E+07 & 6000 & 4.5 & 16 & 0.5 & 2.5E+09 & 6000 & 5.6E+07 \\
25 & 20 & 6000 & 124 & 2.5 & 0.5 & 2.2E+08 & 2.6E+08 & 1.7E+09 & 433 & 6.5E+07 & 4.0E+09 & 6000 & 7.6E+07 & 6000 & 4.5 & 16 & 0.5 & 4.0E+09 & 6000 & 6.8E+07 \\
\hline
\end{tabular}
}

\end{table}

\newpage


\subsection{Gisette}
Dimensions: $m$ = 13500, $n$ = 5000.
\par
Description: patterns for handwritten digit recognition problem.

\begin{table}[!htbp]
\centering

\caption{Summary of the results with Gisette ($\times10^{12}$)}
\label{TabResultsD4}
\small
\resizebox{!}{\tableheight}{
\begin{tabular}{|l|l|llll|llll|llll|llll|}
\hline
\multicolumn{1}{|c|}{\multirow{3}{*}{$k$}} & \multicolumn{1}{c|}{\multirow{3}{*}{$f^*$}} & \multicolumn{4}{c|}{Big-means} & \multicolumn{4}{c|}{IK-means} & \multicolumn{4}{c|}{BDCSM} & \multicolumn{4}{c|}{Minibatch K-means} \\ \cline{3-18}
\multicolumn{1}{|c|}{} & \multicolumn{1}{c|}{} & \multicolumn{2}{c|}{$\varepsilon$} & \multicolumn{2}{c|}{$t$} & \multicolumn{2}{c|}{$\varepsilon$} & \multicolumn{2}{c|}{$t$} & \multicolumn{2}{c|}{$\varepsilon$} & \multicolumn{2}{c|}{$t$} & \multicolumn{2}{c|}{$\varepsilon$} & \multicolumn{2}{c|}{$t$} \\ \cline{3-18}
\multicolumn{1}{|c|}{} & \multicolumn{1}{c|}{} & \multicolumn{1}{c|}{med} & \multicolumn{1}{c|}{std} & \multicolumn{1}{c|}{med} & \multicolumn{1}{c|}{std} & \multicolumn{1}{c|}{med} & \multicolumn{1}{c|}{std} & \multicolumn{1}{c|}{med} & \multicolumn{1}{c|}{std} & \multicolumn{1}{c|}{med} & \multicolumn{1}{c|}{std} & \multicolumn{1}{c|}{med} & \multicolumn{1}{c|}{std} & \multicolumn{1}{c|}{med} & \multicolumn{1}{c|}{std} & \multicolumn{1}{c|}{med} & \multicolumn{1}{c|}{std} \\ \hline
2 & 4.19944 & 0.01 & 0.01 & 3.2 & 0.73 & 0.62 & 0.41 & 1.03 & 0.03 & 0.01 & 0.0 & 2.51 & 0.67 & 0.18 & 0.33 & 2.97 & 0.9 \\
3 & 4.11596 & 0.02 & 0.03 & 6.61 & 5.76 & 0.74 & 0.3 & 1.53 & 0.09 & 0.01 & 0.0 & 9.41 & 2.82 & 0.26 & 0.23 & 6.66 & 2.41 \\
5 & 4.02303 & 0.06 & 0.05 & 7.24 & 1.25 & 0.64 & 0.26 & 2.5 & 0.18 & 0.04 & 0.04 & 14.6 & 9.06 & 0.34 & 0.14 & 8.37 & 2.4 \\
10 & 3.87672 & 0.15 & 0.05 & 17.71 & 2.27 & 0.84 & 0.32 & 5.23 & 0.21 & 0.07 & 0.06 & 28.14 & 14.24 & 0.54 & 0.16 & 5.83 & 2.13 \\
15 & 3.81766 & -0.31 & 0.05 & 25.4 & 3.22 & 0.62 & 0.27 & 10.31 & 0.63 & -0.37 & 0.05 & 47.56 & 13.94 & 0.04 & 0.16 & 6.72 & 1.66 \\
20 & 3.81436 & -1.64 & 0.05 & 32.27 & 2.63 & -0.86 & 0.29 & 11.7 & 0.24 & -1.67 & 0.05 & 66.77 & 30.02 & -1.24 & 0.13 & 11.01 & 5.81 \\
25 & 3.74937 & -1.05 & 0.07 & 44.76 & 4.2 & -0.22 & 0.33 & 15.34 & 0.6 & -1.01 & 0.08 & 72.81 & 16.45 & -0.66 & 0.1 & 4.46 & 1.04 \\
\hline
\multicolumn{2}{|c|}{Mean:} & \textbf{-0.39} & & \textbf{19.6} & & \textbf{0.34} & & \textbf{6.8} & & \textbf{-0.42} & & \textbf{34.54} & & \textbf{-0.08} & & \textbf{6.57} & \\ \hline
\end{tabular}
}

\medskip

\small
\resizebox{!}{\tableheight}{
\begin{tabular}{|l|l|llll|llll|llll|llll|}
\hline
\multicolumn{1}{|c|}{\multirow{3}{*}{$k$}} & \multicolumn{1}{c|}{\multirow{3}{*}{$f^*$}} & \multicolumn{4}{c|}{K-means++} & \multicolumn{4}{c|}{CURE} & \multicolumn{4}{c|}{CluDataSE} & \multicolumn{4}{c|}{LW-coreset} \\ \cline{3-18}
\multicolumn{1}{|c|}{} & \multicolumn{1}{c|}{} & \multicolumn{2}{c|}{$\varepsilon$} & \multicolumn{2}{c|}{$t$} & \multicolumn{2}{c|}{$\varepsilon$} & \multicolumn{2}{c|}{$t$} & \multicolumn{2}{c|}{$\varepsilon$} & \multicolumn{2}{c|}{$t$} & \multicolumn{2}{c|}{$\varepsilon$} & \multicolumn{2}{c|}{$t$} \\ \cline{3-18}
\multicolumn{1}{|c|}{} & \multicolumn{1}{c|}{} & \multicolumn{1}{c|}{med} & \multicolumn{1}{c|}{std} & \multicolumn{1}{c|}{med} & \multicolumn{1}{c|}{std} & \multicolumn{1}{c|}{med} & \multicolumn{1}{c|}{std} & \multicolumn{1}{c|}{med} & \multicolumn{1}{c|}{std} & \multicolumn{1}{c|}{med} & \multicolumn{1}{c|}{std} & \multicolumn{1}{c|}{med} & \multicolumn{1}{c|}{std} & \multicolumn{1}{c|}{med} & \multicolumn{1}{c|}{std} & \multicolumn{1}{c|}{med} & \multicolumn{1}{c|}{std} \\ \hline
2 & 4.19944 & 0.0 & 0.0 & 4.98 & 1.61 & 4.12 & 0.13 & 11.12 & 0.28 & -0.0 & 0.0 & 13.71 & 1.7 & 0.02 & 0.0 & 5.1 & 0.94 \\
3 & 4.11596 & 0.0 & 0.0 & 11.81 & 3.56 & 5.92 & 0.19 & 11.4 & 1.98 & 0.0 & 0.0 & 15.85 & 2.65 & 0.04 & 0.01 & 11.95 & 3.95 \\
5 & 4.02303 & 0.01 & 0.05 & 26.7 & 7.79 & 7.96 & 0.4 & 14.5 & 10.47 & 0.01 & 0.02 & 54.1 & 24.21 & 0.09 & 0.04 & 18.54 & 4.57 \\
10 & 3.87672 & 0.04 & 0.07 & 58.23 & 20.81 & 11.03 & 0.42 & 31.14 & 1.42 & 0.07 & 0.07 & 88.15 & 27.77 & 0.23 & 0.07 & 36.59 & 15.63 \\
15 & 3.81766 & -0.45 & 0.04 & 92.51 & 32.23 & 12.02 & 0.18 & 38.75 & 4.8 & -0.46 & 0.03 & 87.11 & 22.65 & -0.16 & 0.08 & 45.05 & 12.38 \\
20 & 3.81436 & -1.76 & 0.04 & 115.48 & 26.47 & 11.89 & 0.22 & 39.23 & 1.36 & -1.78 & 0.05 & 139.05 & 29.58 & -1.44 & 0.09 & 52.36 & 11.42 \\
25 & 3.74937 & -1.17 & 0.04 & 133.57 & 42.05 & 13.62 & 0.25 & 30.97 & 0.64 & -1.21 & 0.07 & 179.89 & 54.19 & -0.7 & 0.15 & 68.2 & 12.7 \\
\hline
\multicolumn{2}{|c|}{Mean:} & \textbf{-0.48} & & \textbf{63.33} & & \textbf{9.51} & & \textbf{25.3} & & \textbf{-0.48} & & \textbf{82.55} & & \textbf{-0.27} & & \textbf{33.97} & \\ \hline
\end{tabular}
}

\bigskip

\caption{Clustering details with Gisette}
\label{TabDetailsD4}
\resizebox{\linewidth}{!}{
\begin{tabular}{|l|l|lllll|l|l|ll|l|ll|lllll|ll|}
\hline
\multicolumn{1}{|c|}{\multirow{2}{*}{$k$}} & \multicolumn{1}{c|}{\multirow{2}{*}{$n_{exec}$}} & \multicolumn{5}{c|}{Big-means} & \multicolumn{1}{c|}{IK-means} & \multicolumn{1}{c|}{BDCSM} & \multicolumn{2}{c|}{Minibatch K-means} & \multicolumn{1}{c|}{K-means++} & \multicolumn{2}{c|}{CURE} & \multicolumn{5}{c|}{CluDataSE} & \multicolumn{2}{c|}{LW-coreset} \\ \cline{3-21}
\multicolumn{1}{|c|}{} & \multicolumn{1}{c|}{} & \multicolumn{1}{c|}{$s$} & \multicolumn{1}{c|}{$n_{s}$} & \multicolumn{1}{c|}{$T_1$} & \multicolumn{1}{c|}{$T_2$} & \multicolumn{1}{c|}{$n_{d}$} & \multicolumn{1}{c|}{$n_{d}$} & \multicolumn{1}{c|}{$n_{d}$} & \multicolumn{1}{c|}{$n_{s}$} & \multicolumn{1}{c|}{$n_{d}$} & \multicolumn{1}{c|}{$n_{d}$} & \multicolumn{1}{c|}{$s$} & \multicolumn{1}{c|}{$n_{d}$} & \multicolumn{1}{c|}{$s$} & \multicolumn{1}{c|}{$eps$} & \multicolumn{1}{c|}{$min\_pts$} & \multicolumn{1}{c|}{$sf$} & \multicolumn{1}{c|}{$n_{d}$} & \multicolumn{1}{c|}{$s$} & \multicolumn{1}{c|}{$n_{d}$} \\
\hline
2 & 15 & 10000 & 9 & 4.5 & 0.5 & 3.7E+06 & 1.1E+05 & 4.5E+05 & 26 & 5.2E+05 & 7.0E+05 & 10000 & 5.5E+06 & 10000 & 22000.0 & 16 & 0.5 & 1.0E+08 & 10000 & 5.1E+05 \\
3 & 15 & 10000 & 9 & 1.83 & 3.17 & 4.2E+06 & 1.6E+05 & 1.8E+06 & 24 & 7.2E+05 & 1.8E+06 & 10000 & 5.5E+06 & 10000 & 22000.0 & 16 & 0.5 & 1.0E+08 & 10000 & 1.8E+06 \\
5 & 15 & 10000 & 6 & 3.17 & 1.83 & 6.3E+06 & 2.7E+05 & 2.9E+06 & 31 & 1.6E+06 & 4.6E+06 & 10000 & 7.9E+06 & 10000 & 22000.0 & 16 & 0.5 & 1.1E+08 & 10000 & 2.8E+06 \\
10 & 15 & 10000 & 6 & 3.83 & 1.17 & 1.6E+07 & 5.4E+05 & 5.8E+06 & 26 & 2.6E+06 & 1.1E+07 & 10000 & 1.5E+07 & 10000 & 22000.0 & 16 & 0.5 & 1.2E+08 & 10000 & 6.4E+06 \\
15 & 15 & 10000 & 5 & 3.0 & 2.0 & 2.6E+07 & 8.1E+05 & 9.7E+06 & 30 & 4.5E+06 & 1.7E+07 & 10000 & 2.3E+07 & 10000 & 22000.0 & 16 & 0.5 & 1.2E+08 & 10000 & 7.4E+06 \\
20 & 15 & 10000 & 6 & 2.33 & 2.67 & 3.4E+07 & 1.1E+06 & 1.4E+07 & 28 & 5.6E+06 & 2.2E+07 & 10000 & 2.6E+07 & 10000 & 22000.0 & 16 & 0.5 & 1.3E+08 & 10000 & 9.3E+06 \\
25 & 15 & 10000 & 7 & 4.67 & 0.33 & 4.3E+07 & 1.4E+06 & 1.5E+07 & 30 & 7.5E+06 & 2.4E+07 & 10000 & 2.1E+07 & 10000 & 22000.0 & 16 & 0.5 & 1.3E+08 & 10000 & 1.2E+07 \\
\hline
\end{tabular}
}

\end{table}

\newpage


\newpage

\subsection{Music Analysis}
Dimensions: $m$ = 106574, $n$ = 518.
\par
Description: a dataset for music analysis which contains different spectral and statistical attributes for each music track.

\begin{table}[!htbp]
\centering

\caption{Summary of the results with Music Analysis ($\times10^{11}$)}
\label{TabResultsD5}
\small
\resizebox{!}{\tableheight}{
\begin{tabular}{|l|l|llll|llll|llll|llll|}
\hline
\multicolumn{1}{|c|}{\multirow{3}{*}{$k$}} & \multicolumn{1}{c|}{\multirow{3}{*}{$f^*$}} & \multicolumn{4}{c|}{Big-means} & \multicolumn{4}{c|}{IK-means} & \multicolumn{4}{c|}{BDCSM} & \multicolumn{4}{c|}{Minibatch K-means} \\ \cline{3-18}
\multicolumn{1}{|c|}{} & \multicolumn{1}{c|}{} & \multicolumn{2}{c|}{$\varepsilon$} & \multicolumn{2}{c|}{$t$} & \multicolumn{2}{c|}{$\varepsilon$} & \multicolumn{2}{c|}{$t$} & \multicolumn{2}{c|}{$\varepsilon$} & \multicolumn{2}{c|}{$t$} & \multicolumn{2}{c|}{$\varepsilon$} & \multicolumn{2}{c|}{$t$} \\ \cline{3-18}
\multicolumn{1}{|c|}{} & \multicolumn{1}{c|}{} & \multicolumn{1}{c|}{med} & \multicolumn{1}{c|}{std} & \multicolumn{1}{c|}{med} & \multicolumn{1}{c|}{std} & \multicolumn{1}{c|}{med} & \multicolumn{1}{c|}{std} & \multicolumn{1}{c|}{med} & \multicolumn{1}{c|}{std} & \multicolumn{1}{c|}{med} & \multicolumn{1}{c|}{std} & \multicolumn{1}{c|}{med} & \multicolumn{1}{c|}{std} & \multicolumn{1}{c|}{med} & \multicolumn{1}{c|}{std} & \multicolumn{1}{c|}{med} & \multicolumn{1}{c|}{std} \\ \hline
2 & 5.00474$^*$ & 0.08 & 0.03 & 4.94 & 1.89 & 0.81 & 2.88 & 0.84 & 0.03 & 0.42 & 0.99 & 0.35 & 0.04 & 0.35 & 5.59 & 0.89 & 0.24 \\
3 & 3.83748$^*$ & 0.16 & 0.06 & 4.71 & 1.92 & 11.57 & 6.72 & 1.19 & 0.01 & 1.96 & 15.02 & 0.89 & 0.27 & 0.76 & 5.54 & 0.52 & 0.11 \\
5 & 2.74249$^*$ & 0.24 & 0.72 & 4.55 & 1.96 & 13.33 & 9.42 & 2.05 & 0.11 & 4.98 & 19.55 & 1.7 & 0.34 & 2.79 & 3.36 & 0.77 & 0.68 \\
10 & 1.87296$^*$ & 0.56 & 0.65 & 4.58 & 2.31 & 10.45 & 3.55 & 3.51 & 0.07 & 6.1 & 6.31 & 5.69 & 1.63 & 3.41 & 1.69 & 3.72 & 1.69 \\
15 & 1.54422$^*$ & 1.1 & 0.36 & 4.5 & 2.18 & 15.3 & 4.28 & 5.21 & 0.05 & 6.15 & 4.85 & 9.86 & 1.54 & 3.32 & 1.25 & 1.52 & 0.69 \\
20 & 1.35315$^*$ & 1.66 & 0.48 & 6.56 & 2.11 & 13.44 & 4.42 & 11.43 & 1.07 & 5.74 & 2.91 & 13.65 & 2.44 & 3.14 & 0.81 & 0.68 & 0.13 \\
25 & 1.22622$^*$ & 1.99 & 0.75 & 7.23 & 2.27 & 15.97 & 3.13 & 8.5 & 0.04 & 5.95 & 3.17 & 16.19 & 2.45 & 3.27 & 0.94 & 1.24 & 0.42 \\
\hline
\multicolumn{2}{|c|}{Mean:} & \textbf{0.83} & & \textbf{5.29} & & \textbf{11.55} & & \textbf{4.68} & & \textbf{4.47} & & \textbf{6.9} & & \textbf{2.43} & & \textbf{1.34} & \\ \hline
\end{tabular}
}

\medskip

\small
\resizebox{!}{\tableheight}{
\begin{tabular}{|l|l|llll|llll|llll|llll|}
\hline
\multicolumn{1}{|c|}{\multirow{3}{*}{$k$}} & \multicolumn{1}{c|}{\multirow{3}{*}{$f^*$}} & \multicolumn{4}{c|}{K-means++} & \multicolumn{4}{c|}{CURE} & \multicolumn{4}{c|}{CluDataSE} & \multicolumn{4}{c|}{LW-coreset} \\ \cline{3-18}
\multicolumn{1}{|c|}{} & \multicolumn{1}{c|}{} & \multicolumn{2}{c|}{$\varepsilon$} & \multicolumn{2}{c|}{$t$} & \multicolumn{2}{c|}{$\varepsilon$} & \multicolumn{2}{c|}{$t$} & \multicolumn{2}{c|}{$\varepsilon$} & \multicolumn{2}{c|}{$t$} & \multicolumn{2}{c|}{$\varepsilon$} & \multicolumn{2}{c|}{$t$} \\ \cline{3-18}
\multicolumn{1}{|c|}{} & \multicolumn{1}{c|}{} & \multicolumn{1}{c|}{med} & \multicolumn{1}{c|}{std} & \multicolumn{1}{c|}{med} & \multicolumn{1}{c|}{std} & \multicolumn{1}{c|}{med} & \multicolumn{1}{c|}{std} & \multicolumn{1}{c|}{med} & \multicolumn{1}{c|}{std} & \multicolumn{1}{c|}{med} & \multicolumn{1}{c|}{std} & \multicolumn{1}{c|}{med} & \multicolumn{1}{c|}{std} & \multicolumn{1}{c|}{med} & \multicolumn{1}{c|}{std} & \multicolumn{1}{c|}{med} & \multicolumn{1}{c|}{std} \\ \hline
2 & 5.00474$^*$ & -0.0 & 10.4 & 3.72 & 0.48 & 30.96 & 9.36 & 3.07 & 0.18 & -0.0 & 0.0 & 3.0 & 0.1 & 0.04 & 9.28 & 1.43 & 0.55 \\
3 & 3.83748$^*$ & -0.0 & 4.15 & 5.23 & 4.12 & 52.97 & 9.72 & 2.71 & 0.09 & 9.06 & 1.97 & 9.82 & 1.22 & 0.09 & 4.33 & 1.37 & 0.11 \\
5 & 2.74249$^*$ & 1.12 & 1.4 & 16.08 & 3.78 & 39.73 & 22.57 & 3.16 & 0.26 & 2.23 & 1.42 & 19.0 & 4.32 & 2.39 & 1.85 & 1.74 & 0.22 \\
10 & 1.87296$^*$ & 1.56 & 2.04 & 61.44 & 25.53 & 42.8 & 17.53 & 4.08 & 1.13 & 1.67 & 0.5 & 60.01 & 12.01 & 1.62 & 2.68 & 2.91 & 0.82 \\
15 & 1.54422$^*$ & 0.29 & 0.52 & 106.37 & 43.67 & 40.91 & 11.44 & 4.34 & 0.1 & 0.65 & 0.36 & 153.38 & 38.67 & 2.68 & 1.96 & 3.89 & 0.8 \\
20 & 1.35315$^*$ & 0.18 & 0.7 & 161.72 & 104.7 & 46.31 & 5.99 & 4.01 & 0.42 & 1.04 & 0.44 & 177.15 & 68.8 & 2.78 & 2.12 & 5.39 & 1.99 \\
25 & 1.22622$^*$ & 0.74 & 0.51 & 201.65 & 80.42 & 56.47 & 13.0 & 4.45 & 0.9 & 0.7 & 0.54 & 272.22 & 77.43 & 3.57 & 1.28 & 6.85 & 2.06 \\
\hline
\multicolumn{2}{|c|}{Mean:} & \textbf{0.55} & & \textbf{79.46} & & \textbf{44.31} & & \textbf{3.69} & & \textbf{2.19} & & \textbf{99.23} & & \textbf{1.88} & & \textbf{3.37} & \\ \hline
\end{tabular}
}

\bigskip

\caption{Clustering details with Music Analysis}
\label{TabDetailsD5}
\resizebox{\linewidth}{!}{
\begin{tabular}{|l|l|lllll|l|l|ll|l|ll|lllll|ll|}
\hline
\multicolumn{1}{|c|}{\multirow{2}{*}{$k$}} & \multicolumn{1}{c|}{\multirow{2}{*}{$n_{exec}$}} & \multicolumn{5}{c|}{Big-means} & \multicolumn{1}{c|}{IK-means} & \multicolumn{1}{c|}{BDCSM} & \multicolumn{2}{c|}{Minibatch K-means} & \multicolumn{1}{c|}{K-means++} & \multicolumn{2}{c|}{CURE} & \multicolumn{5}{c|}{CluDataSE} & \multicolumn{2}{c|}{LW-coreset} \\ \cline{3-21}
\multicolumn{1}{|c|}{} & \multicolumn{1}{c|}{} & \multicolumn{1}{c|}{$s$} & \multicolumn{1}{c|}{$n_{s}$} & \multicolumn{1}{c|}{$T_1$} & \multicolumn{1}{c|}{$T_2$} & \multicolumn{1}{c|}{$n_{d}$} & \multicolumn{1}{c|}{$n_{d}$} & \multicolumn{1}{c|}{$n_{d}$} & \multicolumn{1}{c|}{$n_{s}$} & \multicolumn{1}{c|}{$n_{d}$} & \multicolumn{1}{c|}{$n_{d}$} & \multicolumn{1}{c|}{$s$} & \multicolumn{1}{c|}{$n_{d}$} & \multicolumn{1}{c|}{$s$} & \multicolumn{1}{c|}{$eps$} & \multicolumn{1}{c|}{$min\_pts$} & \multicolumn{1}{c|}{$sf$} & \multicolumn{1}{c|}{$n_{d}$} & \multicolumn{1}{c|}{$s$} & \multicolumn{1}{c|}{$n_{d}$} \\
\hline
2 & 20 & 6000 & 1264 & 1.33 & 6.67 & 9.0E+07 & 8.5E+05 & 3.5E+06 & 38 & 4.5E+05 & 4.9E+06 & 6000 & 1.0E+07 & 6000 & 500.0 & 16 & 0.5 & 4.1E+07 & 6000 & 6.3E+05 \\
3 & 20 & 6000 & 656 & 1.6 & 6.4 & 1.0E+08 & 1.3E+06 & 8.4E+06 & 33 & 5.9E+05 & 7.5E+06 & 6000 & 8.9E+06 & 6000 & 500.0 & 16 & 0.5 & 5.4E+07 & 6000 & 8.7E+05 \\
5 & 20 & 6000 & 336 & 1.33 & 6.67 & 1.1E+08 & 2.1E+06 & 1.7E+07 & 45 & 1.4E+06 & 2.9E+07 & 6000 & 8.6E+06 & 6000 & 500.0 & 16 & 0.5 & 7.3E+07 & 6000 & 1.9E+06 \\
10 & 20 & 6000 & 85 & 6.13 & 1.87 & 1.2E+08 & 4.3E+06 & 6.1E+07 & 43 & 2.6E+06 & 1.2E+08 & 6000 & 1.1E+07 & 6000 & 500.0 & 16 & 0.5 & 1.6E+08 & 6000 & 4.8E+06 \\
15 & 20 & 6000 & 36 & 4.53 & 3.47 & 1.2E+08 & 6.4E+06 & 1.1E+08 & 43 & 3.9E+06 & 2.1E+08 & 6000 & 1.3E+07 & 6000 & 500.0 & 16 & 0.5 & 3.6E+08 & 6000 & 7.3E+06 \\
20 & 20 & 6000 & 36 & 0.53 & 7.47 & 1.1E+08 & 8.5E+06 & 1.5E+08 & 43 & 5.2E+06 & 3.2E+08 & 6000 & 1.2E+07 & 6000 & 500.0 & 16 & 0.5 & 4.0E+08 & 6000 & 1.1E+07 \\
25 & 20 & 6000 & 22 & 0.27 & 7.73 & 1.2E+08 & 1.1E+07 & 2.0E+08 & 32 & 4.8E+06 & 4.0E+08 & 6000 & 1.1E+07 & 6000 & 500.0 & 16 & 0.5 & 6.1E+08 & 6000 & 1.5E+07 \\
\hline
\end{tabular}
}

\end{table}

\newpage


\subsection{Protein Homology}
Dimensions: $m$ = 145751, $n$ = 74.
\par
Description: a data set for protein homology prediction which contains a features describing the match (e.g. the score of a sequence alignment) between the native protein sequence and the sequence that is tested for homology.

\begin{table}[!htbp]
\centering

\caption{Summary of the results with Protein Homology ($\times10^{11}$)}
\label{TabResultsD6}
\small
\resizebox{!}{\tableheight}{
\begin{tabular}{|l|l|llll|llll|llll|llll|}
\hline
\multicolumn{1}{|c|}{\multirow{3}{*}{$k$}} & \multicolumn{1}{c|}{\multirow{3}{*}{$f^*$}} & \multicolumn{4}{c|}{Big-means} & \multicolumn{4}{c|}{IK-means} & \multicolumn{4}{c|}{BDCSM} & \multicolumn{4}{c|}{Minibatch K-means} \\ \cline{3-18}
\multicolumn{1}{|c|}{} & \multicolumn{1}{c|}{} & \multicolumn{2}{c|}{$\varepsilon$} & \multicolumn{2}{c|}{$t$} & \multicolumn{2}{c|}{$\varepsilon$} & \multicolumn{2}{c|}{$t$} & \multicolumn{2}{c|}{$\varepsilon$} & \multicolumn{2}{c|}{$t$} & \multicolumn{2}{c|}{$\varepsilon$} & \multicolumn{2}{c|}{$t$} \\ \cline{3-18}
\multicolumn{1}{|c|}{} & \multicolumn{1}{c|}{} & \multicolumn{1}{c|}{med} & \multicolumn{1}{c|}{std} & \multicolumn{1}{c|}{med} & \multicolumn{1}{c|}{std} & \multicolumn{1}{c|}{med} & \multicolumn{1}{c|}{std} & \multicolumn{1}{c|}{med} & \multicolumn{1}{c|}{std} & \multicolumn{1}{c|}{med} & \multicolumn{1}{c|}{std} & \multicolumn{1}{c|}{med} & \multicolumn{1}{c|}{std} & \multicolumn{1}{c|}{med} & \multicolumn{1}{c|}{std} & \multicolumn{1}{c|}{med} & \multicolumn{1}{c|}{std} \\ \hline
2 & 15.20433$^*$ & 1.87 & 0.01 & 1.3 & 1.05 & 2.65 & 0.93 & 0.26 & 0.55 & 1.83 & 0.0 & 0.33 & 0.05 & 2.26 & 2.62 & 0.57 & 0.19 \\
3 & 8.07129$^*$ & 0.79 & 0.37 & 3.1 & 1.0 & 66.78 & 9.6 & 0.39 & 0.1 & 0.02 & 53.0 & 0.9 & 0.2 & 66.34 & 5.0 & 0.59 & 0.14 \\
5 & 5.30537$^*$ & 0.93 & 0.6 & 2.12 & 0.83 & 130.11 & 13.43 & 0.6 & 0.14 & 19.35 & 9.58 & 1.88 & 0.33 & 118.23 & 9.23 & 0.77 & 0.2 \\
10 & 3.3767$^*$ & 0.19 & 0.19 & 3.34 & 0.63 & 223.7 & 38.77 & 1.28 & 0.09 & 32.58 & 6.99 & 5.12 & 1.63 & 165.18 & 42.26 & 0.93 & 0.34 \\
15 & 2.86473$^*$ & 0.99 & 0.37 & 3.22 & 1.12 & 263.65 & 50.83 & 1.66 & 0.28 & 31.35 & 5.8 & 9.95 & 1.77 & 115.05 & 66.0 & 0.97 & 0.28 \\
20 & 2.5732$^*$ & 1.16 & 0.41 & 4.45 & 1.78 & 262.05 & 72.07 & 2.17 & 0.13 & 34.3 & 11.71 & 15.6 & 3.87 & 170.19 & 88.61 & 1.15 & 0.29 \\
25 & 2.38539$^*$ & 0.87 & 0.47 & 4.34 & 0.82 & 239.88 & 85.73 & 2.77 & 0.07 & 33.53 & 13.46 & 21.4 & 5.09 & 70.04 & 50.31 & 1.24 & 0.23 \\
\hline
\multicolumn{2}{|c|}{Mean:} & \textbf{0.97} & & \textbf{3.12} & & \textbf{169.83} & & \textbf{1.3} & & \textbf{21.85} & & \textbf{7.88} & & \textbf{101.04} & & \textbf{0.89} & \\ \hline
\end{tabular}
}

\medskip

\small
\resizebox{!}{\tableheight}{
\begin{tabular}{|l|l|llll|llll|llll|llll|}
\hline
\multicolumn{1}{|c|}{\multirow{3}{*}{$k$}} & \multicolumn{1}{c|}{\multirow{3}{*}{$f^*$}} & \multicolumn{4}{c|}{K-means++} & \multicolumn{4}{c|}{CURE} & \multicolumn{4}{c|}{CluDataSE} & \multicolumn{4}{c|}{LW-coreset} \\ \cline{3-18}
\multicolumn{1}{|c|}{} & \multicolumn{1}{c|}{} & \multicolumn{2}{c|}{$\varepsilon$} & \multicolumn{2}{c|}{$t$} & \multicolumn{2}{c|}{$\varepsilon$} & \multicolumn{2}{c|}{$t$} & \multicolumn{2}{c|}{$\varepsilon$} & \multicolumn{2}{c|}{$t$} & \multicolumn{2}{c|}{$\varepsilon$} & \multicolumn{2}{c|}{$t$} \\ \cline{3-18}
\multicolumn{1}{|c|}{} & \multicolumn{1}{c|}{} & \multicolumn{1}{c|}{med} & \multicolumn{1}{c|}{std} & \multicolumn{1}{c|}{med} & \multicolumn{1}{c|}{std} & \multicolumn{1}{c|}{med} & \multicolumn{1}{c|}{std} & \multicolumn{1}{c|}{med} & \multicolumn{1}{c|}{std} & \multicolumn{1}{c|}{med} & \multicolumn{1}{c|}{std} & \multicolumn{1}{c|}{med} & \multicolumn{1}{c|}{std} & \multicolumn{1}{c|}{med} & \multicolumn{1}{c|}{std} & \multicolumn{1}{c|}{med} & \multicolumn{1}{c|}{std} \\ \hline
2 & 15.20433$^*$ & 0.0 & 0.89 & 0.21 & 0.91 & 63.07 & 29.67 & 233.4 & 17.83 & 1.82 & 0.0 & 3.12 & 0.43 & 0.0 & 0.0 & 0.3 & 4.47 \\
3 & 8.07129$^*$ & 0.0 & 31.4 & 0.74 & 0.33 & 91.91 & 43.08 & 104.94 & 6.41 & 0.0 & 0.0 & 4.37 & 0.14 & 0.01 & 0.0 & 0.37 & 0.03 \\
5 & 5.30537$^*$ & 0.0 & 0.19 & 2.47 & 0.5 & 160.61 & 12.34 & 57.45 & 3.71 & 0.0 & 0.0 & 6.37 & 0.32 & 0.02 & 0.01 & 0.95 & 0.22 \\
10 & 3.3767$^*$ & -0.0 & 7.15 & 9.49 & 2.16 & 282.51 & 13.15 & 50.3 & 1.84 & 18.12 & 0.0 & 17.29 & 1.63 & 18.17 & 6.32 & 3.31 & 0.88 \\
15 & 2.86473$^*$ & 0.21 & 0.95 & 20.55 & 7.25 & 333.95 & 15.32 & 56.5 & 2.45 & 23.94 & 0.03 & 38.66 & 7.86 & 24.04 & 8.04 & 6.42 & 1.5 \\
20 & 2.5732$^*$ & 0.96 & 0.46 & 32.45 & 7.61 & 370.47 & 9.61 & 71.5 & 1.66 & 28.56 & 0.18 & 49.28 & 5.73 & 28.36 & 13.44 & 8.65 & 3.79 \\
25 & 2.38539$^*$ & 1.62 & 0.74 & 34.93 & 14.68 & 401.85 & 26.31 & 82.36 & 2.69 & 31.85 & 0.16 & 66.93 & 16.05 & 32.14 & 14.23 & 10.44 & 1.9 \\
\hline
\multicolumn{2}{|c|}{Mean:} & \textbf{0.4} & & \textbf{14.41} & & \textbf{243.48} & & \textbf{93.78} & & \textbf{14.9} & & \textbf{26.57} & & \textbf{14.68} & & \textbf{4.35} & \\ \hline
\end{tabular}
}

\bigskip

\caption{Clustering details with Protein Homology}
\label{TabDetailsD6}
\resizebox{\linewidth}{!}{
\begin{tabular}{|l|l|lllll|l|l|ll|l|ll|lllll|ll|}
\hline
\multicolumn{1}{|c|}{\multirow{2}{*}{$k$}} & \multicolumn{1}{c|}{\multirow{2}{*}{$n_{exec}$}} & \multicolumn{5}{c|}{Big-means} & \multicolumn{1}{c|}{IK-means} & \multicolumn{1}{c|}{BDCSM} & \multicolumn{2}{c|}{Minibatch K-means} & \multicolumn{1}{c|}{K-means++} & \multicolumn{2}{c|}{CURE} & \multicolumn{5}{c|}{CluDataSE} & \multicolumn{2}{c|}{LW-coreset} \\ \cline{3-21}
\multicolumn{1}{|c|}{} & \multicolumn{1}{c|}{} & \multicolumn{1}{c|}{$s$} & \multicolumn{1}{c|}{$n_{s}$} & \multicolumn{1}{c|}{$T_1$} & \multicolumn{1}{c|}{$T_2$} & \multicolumn{1}{c|}{$n_{d}$} & \multicolumn{1}{c|}{$n_{d}$} & \multicolumn{1}{c|}{$n_{d}$} & \multicolumn{1}{c|}{$n_{s}$} & \multicolumn{1}{c|}{$n_{d}$} & \multicolumn{1}{c|}{$n_{d}$} & \multicolumn{1}{c|}{$s$} & \multicolumn{1}{c|}{$n_{d}$} & \multicolumn{1}{c|}{$s$} & \multicolumn{1}{c|}{$eps$} & \multicolumn{1}{c|}{$min\_pts$} & \multicolumn{1}{c|}{$sf$} & \multicolumn{1}{c|}{$n_{d}$} & \multicolumn{1}{c|}{$s$} & \multicolumn{1}{c|}{$n_{d}$} \\
\hline
2 & 15 & 56000 & 230 & 3.27 & 0.23 & 2.3E+08 & 1.2E+06 & 5.4E+06 & 19 & 2.1E+06 & 2.0E+06 & 56000 & 3.7E+08 & 56000 & 1000.0 & 16 & 0.5 & 3.1E+09 & 56000 & 1.0E+06 \\
3 & 15 & 56000 & 403 & 2.57 & 0.93 & 2.6E+08 & 1.7E+06 & 1.8E+07 & 27 & 4.5E+06 & 1.1E+07 & 56000 & 2.5E+08 & 56000 & 1000.0 & 16 & 0.5 & 3.2E+09 & 56000 & 4.1E+06 \\
5 & 15 & 56000 & 126 & 0.93 & 2.57 & 2.9E+08 & 2.9E+06 & 4.2E+07 & 24 & 6.7E+06 & 4.1E+07 & 56000 & 1.7E+08 & 56000 & 1000.0 & 16 & 0.5 & 3.2E+09 & 56000 & 1.6E+07 \\
10 & 15 & 56000 & 22 & 0.23 & 3.27 & 2.2E+08 & 5.8E+06 & 1.7E+08 & 28 & 1.6E+07 & 1.7E+08 & 56000 & 1.5E+08 & 56000 & 1000.0 & 16 & 0.5 & 3.4E+09 & 56000 & 6.2E+07 \\
15 & 15 & 56000 & 5 & 0.23 & 3.27 & 2.9E+08 & 8.7E+06 & 3.2E+08 & 30 & 2.5E+07 & 4.1E+08 & 56000 & 1.7E+08 & 56000 & 1000.0 & 16 & 0.5 & 3.8E+09 & 56000 & 1.3E+08 \\
20 & 15 & 56000 & 6 & 1.17 & 2.33 & 3.9E+08 & 1.2E+07 & 4.9E+08 & 28 & 3.1E+07 & 6.1E+08 & 56000 & 1.9E+08 & 56000 & 1000.0 & 16 & 0.5 & 4.0E+09 & 56000 & 1.6E+08 \\
25 & 15 & 56000 & 6 & 3.15 & 0.35 & 4.5E+08 & 1.5E+07 & 7.1E+08 & 37 & 5.2E+07 & 6.6E+08 & 56000 & 2.1E+08 & 56000 & 1000.0 & 16 & 0.5 & 4.4E+09 & 56000 & 2.1E+08 \\
\hline
\end{tabular}
}

\end{table}

\newpage


\newpage

\subsection{MiniBooNE Particle Identification}
Dimensions: $m$ = 130064, $n$ = 50.
\par
Description: a data set for distinguishing electron neutrinos (signal) from muon neutrinos (background) which contains different particle variables for each event.

\begin{table}[!htbp]
\centering

\caption{Summary of the results with MiniBooNE Particle Identification ($\times10^{10}$)}
\label{TabResultsD7}
\small
\resizebox{!}{\tableheight}{
\begin{tabular}{|l|l|llll|llll|llll|llll|}
\hline
\multicolumn{1}{|c|}{\multirow{3}{*}{$k$}} & \multicolumn{1}{c|}{\multirow{3}{*}{$f^*$}} & \multicolumn{4}{c|}{Big-means} & \multicolumn{4}{c|}{IK-means} & \multicolumn{4}{c|}{BDCSM} & \multicolumn{4}{c|}{Minibatch K-means} \\ \cline{3-18}
\multicolumn{1}{|c|}{} & \multicolumn{1}{c|}{} & \multicolumn{2}{c|}{$\varepsilon$} & \multicolumn{2}{c|}{$t$} & \multicolumn{2}{c|}{$\varepsilon$} & \multicolumn{2}{c|}{$t$} & \multicolumn{2}{c|}{$\varepsilon$} & \multicolumn{2}{c|}{$t$} & \multicolumn{2}{c|}{$\varepsilon$} & \multicolumn{2}{c|}{$t$} \\ \cline{3-18}
\multicolumn{1}{|c|}{} & \multicolumn{1}{c|}{} & \multicolumn{1}{c|}{med} & \multicolumn{1}{c|}{std} & \multicolumn{1}{c|}{med} & \multicolumn{1}{c|}{std} & \multicolumn{1}{c|}{med} & \multicolumn{1}{c|}{std} & \multicolumn{1}{c|}{med} & \multicolumn{1}{c|}{std} & \multicolumn{1}{c|}{med} & \multicolumn{1}{c|}{std} & \multicolumn{1}{c|}{med} & \multicolumn{1}{c|}{std} & \multicolumn{1}{c|}{med} & \multicolumn{1}{c|}{std} & \multicolumn{1}{c|}{med} & \multicolumn{1}{c|}{std} \\ \hline
2 & 8.92236 & 0.0 & 0.0 & 1.08 & 0.95 & 286908.11 & 284.03 & 0.19 & 0.0 & 0.0 & 143134.9 & 0.17 & 0.09 & 286911.79 & 8.34 & 1.17 & 0.45 \\
3 & 5.22601 & 0.0 & 0.0 & 2.0 & 0.81 & 489832.22 & 195901.07 & 0.27 & 0.05 & 0.0 & 122192.54 & 0.32 & 0.04 & 489836.85 & 279.94 & 0.59 & 0.55 \\
5 & 1.82252 & 0.01 & 0.0 & 2.36 & 0.63 & 127.27 & 561815.57 & 0.43 & 0.12 & 0.0 & 58.26 & 1.21 & 0.45 & 1404170.45 & 1381.36 & 1.08 & 0.5 \\
10 & 0.9092 & 0.03 & 0.01 & 4.48 & 0.98 & 43.32 & 92.24 & 0.83 & 0.21 & 0.0 & 0.41 & 8.5 & 3.28 & 2812016.23 & 5271.38 & 1.05 & 0.79 \\
15 & 0.63506 & 0.15 & 0.56 & 6.07 & 1.62 & 80.4 & 45.66 & 1.19 & 0.15 & 3.88 & 0.78 & 17.39 & 4.57 & 4026475.67 & 7873.09 & 1.96 & 1.5 \\
20 & 0.50863 & 0.1 & 0.19 & 5.9 & 1.53 & 87.89 & 30.75 & 1.61 & 0.12 & 7.05 & 0.59 & 23.62 & 6.11 & 5018494.92 & 10938.68 & 0.9 & 0.14 \\
25 & 0.44425 & -0.24 & 1437586.3 & 6.94 & 1.58 & 80.34 & 28.55 & 1.95 & 0.35 & 8.93 & 0.33 & 34.44 & 6.15 & 5746462.36 & 23826.26 & 1.1 & 0.24 \\
\hline
\multicolumn{2}{|c|}{Mean:} & \textbf{0.01} & & \textbf{4.12} & & \textbf{111022.79} & & \textbf{0.93} & & \textbf{2.84} & & \textbf{12.24} & & \textbf{2826338.32} & & \textbf{1.12} & \\ \hline
\end{tabular}
}

\medskip

\small
\resizebox{!}{\tableheight}{
\begin{tabular}{|l|l|llll|llll|llll|llll|}
\hline
\multicolumn{1}{|c|}{\multirow{3}{*}{$k$}} & \multicolumn{1}{c|}{\multirow{3}{*}{$f^*$}} & \multicolumn{4}{c|}{K-means++} & \multicolumn{4}{c|}{CURE} & \multicolumn{4}{c|}{CluDataSE} & \multicolumn{4}{c|}{LW-coreset} \\ \cline{3-18}
\multicolumn{1}{|c|}{} & \multicolumn{1}{c|}{} & \multicolumn{2}{c|}{$\varepsilon$} & \multicolumn{2}{c|}{$t$} & \multicolumn{2}{c|}{$\varepsilon$} & \multicolumn{2}{c|}{$t$} & \multicolumn{2}{c|}{$\varepsilon$} & \multicolumn{2}{c|}{$t$} & \multicolumn{2}{c|}{$\varepsilon$} & \multicolumn{2}{c|}{$t$} \\ \cline{3-18}
\multicolumn{1}{|c|}{} & \multicolumn{1}{c|}{} & \multicolumn{1}{c|}{med} & \multicolumn{1}{c|}{std} & \multicolumn{1}{c|}{med} & \multicolumn{1}{c|}{std} & \multicolumn{1}{c|}{med} & \multicolumn{1}{c|}{std} & \multicolumn{1}{c|}{med} & \multicolumn{1}{c|}{std} & \multicolumn{1}{c|}{med} & \multicolumn{1}{c|}{std} & \multicolumn{1}{c|}{med} & \multicolumn{1}{c|}{std} & \multicolumn{1}{c|}{med} & \multicolumn{1}{c|}{std} & \multicolumn{1}{c|}{med} & \multicolumn{1}{c|}{std} \\ \hline
2 & 8.92236 & 0.0 & 0.0 & 0.09 & 0.0 & 286928.91 & 0.13 & 426.12 & 19.95 & 286920.93 & 126881.21 & 0.7 & 0.09 & 0.0 & 0.0 & 0.09 & 0.02 \\
3 & 5.22601 & 0.0 & 5.41 & 0.33 & 0.11 & 489868.34 & 1.68 & 310.33 & 18.45 & 21.68 & 0.0 & 0.9 & 0.06 & 0.02 & 8.67 & 0.15 & 0.02 \\
5 & 1.82252 & 0.0 & 39.7 & 1.07 & 0.37 & 1404812.42 & 14.0 & 249.76 & 19.01 & 0.0 & 0.0 & 1.93 & 0.13 & 0.11 & 48.17 & 0.3 & 0.07 \\
10 & 0.9092 & 0.0 & 2.6 & 7.38 & 3.83 & 2816035.19 & 10.77 & 205.28 & 12.66 & 0.0 & 0.0 & 15.09 & 0.64 & 1.99 & 3.26 & 1.33 & 0.44 \\
15 & 0.63506 & 2.32 & 1.61 & 10.35 & 5.52 & 4031654.69 & 13.36 & 183.52 & 11.52 & 3.88 & 0.59 & 20.44 & 2.15 & 4.61 & 2.04 & 3.0 & 1.05 \\
20 & 0.50863 & 0.97 & 3.36 & 16.24 & 5.54 & 5033808.44 & 8.95 & 165.96 & 12.4 & 7.05 & 0.55 & 28.46 & 5.26 & 9.88 & 2.41 & 3.1 & 0.69 \\
25 & 0.44425 & 0.04 & 4.21 & 20.29 & 7.22 & 5763295.53 & 11.78 & 164.5 & 10.23 & 8.94 & 0.2 & 39.09 & 6.08 & 12.18 & 3.28 & 4.77 & 1.68 \\
\hline
\multicolumn{2}{|c|}{Mean:} & \textbf{0.48} & & \textbf{7.96} & & \textbf{2832343.36} & & \textbf{243.64} & & \textbf{40994.64} & & \textbf{15.23} & & \textbf{4.11} & & \textbf{1.82} & \\ \hline
\end{tabular}
}

\bigskip

\caption{Clustering details with MiniBooNE Particle Identification}
\label{TabDetailsD7}
\resizebox{\linewidth}{!}{
\begin{tabular}{|l|l|lllll|l|l|ll|l|ll|lllll|ll|}
\hline
\multicolumn{1}{|c|}{\multirow{2}{*}{$k$}} & \multicolumn{1}{c|}{\multirow{2}{*}{$n_{exec}$}} & \multicolumn{5}{c|}{Big-means} & \multicolumn{1}{c|}{IK-means} & \multicolumn{1}{c|}{BDCSM} & \multicolumn{2}{c|}{Minibatch K-means} & \multicolumn{1}{c|}{K-means++} & \multicolumn{2}{c|}{CURE} & \multicolumn{5}{c|}{CluDataSE} & \multicolumn{2}{c|}{LW-coreset} \\ \cline{3-21}
\multicolumn{1}{|c|}{} & \multicolumn{1}{c|}{} & \multicolumn{1}{c|}{$s$} & \multicolumn{1}{c|}{$n_{s}$} & \multicolumn{1}{c|}{$T_1$} & \multicolumn{1}{c|}{$T_2$} & \multicolumn{1}{c|}{$n_{d}$} & \multicolumn{1}{c|}{$n_{d}$} & \multicolumn{1}{c|}{$n_{d}$} & \multicolumn{1}{c|}{$n_{s}$} & \multicolumn{1}{c|}{$n_{d}$} & \multicolumn{1}{c|}{$n_{d}$} & \multicolumn{1}{c|}{$s$} & \multicolumn{1}{c|}{$n_{d}$} & \multicolumn{1}{c|}{$s$} & \multicolumn{1}{c|}{$eps$} & \multicolumn{1}{c|}{$min\_pts$} & \multicolumn{1}{c|}{$sf$} & \multicolumn{1}{c|}{$n_{d}$} & \multicolumn{1}{c|}{$s$} & \multicolumn{1}{c|}{$n_{d}$} \\
\hline
2 & 15 & 130000 & 100 & 2.5 & 0.5 & 1.5E+08 & 1.0E+06 & 2.9E+06 & 32 & 8.3E+06 & 7.8E+05 & 40000 & 4.4E+08 & 40000 & 70.0 & 16 & 0.5 & 1.6E+09 & 40000 & 7.6E+05 \\
3 & 15 & 130000 & 170 & 2.5 & 0.5 & 2.4E+08 & 1.6E+06 & 8.6E+06 & 15 & 5.8E+06 & 7.0E+06 & 40000 & 3.8E+08 & 40000 & 70.0 & 16 & 0.5 & 1.6E+09 & 40000 & 2.3E+06 \\
5 & 15 & 130000 & 73 & 0.2 & 2.8 & 2.5E+08 & 2.6E+06 & 3.6E+07 & 23 & 1.5E+07 & 2.9E+07 & 40000 & 3.3E+08 & 40000 & 70.0 & 16 & 0.5 & 1.6E+09 & 40000 & 6.7E+06 \\
10 & 15 & 130000 & 14 & 2.0 & 1.0 & 3.5E+08 & 5.2E+06 & 2.6E+08 & 19 & 2.5E+07 & 2.1E+08 & 40000 & 3.0E+08 & 40000 & 70.0 & 16 & 0.5 & 2.0E+09 & 40000 & 3.9E+07 \\
15 & 15 & 130000 & 9 & 0.5 & 2.5 & 6.1E+08 & 7.8E+06 & 5.5E+08 & 19 & 3.7E+07 & 2.9E+08 & 40000 & 3.1E+08 & 40000 & 70.0 & 16 & 0.5 & 2.2E+09 & 40000 & 8.9E+07 \\
20 & 15 & 130000 & 5 & 2.9 & 0.1 & 7.9E+08 & 1.0E+07 & 7.4E+08 & 19 & 4.9E+07 & 5.0E+08 & 40000 & 3.2E+08 & 40000 & 70.0 & 16 & 0.5 & 2.4E+09 & 40000 & 9.6E+07 \\
25 & 15 & 130000 & 4 & 1.7 & 1.3 & 1.1E+09 & 1.3E+07 & 1.1E+09 & 26 & 8.4E+07 & 6.4E+08 & 40000 & 3.3E+08 & 40000 & 70.0 & 16 & 0.5 & 2.8E+09 & 40000 & 1.5E+08 \\
\hline
\end{tabular}
}

\end{table}

\newpage


\subsection{MiniBooNE Particle Identification (normalized)}
Dimensions: $m$ = 130064, $n$ = 50.
\par
Description: a data set for distinguishing electron neutrinos (signal) from muon neutrinos (background) which contains different particle variables for each event. Min-max scaling was used for normalization of data set values for better clusterization.

\begin{table}[!htbp]
\centering

\caption{Summary of the results with MiniBooNE Particle Identification (normalized) ($\times10^{2}$)}
\label{TabResultsD8}
\small
\resizebox{!}{\tableheight}{
\begin{tabular}{|l|l|llll|llll|llll|llll|}
\hline
\multicolumn{1}{|c|}{\multirow{3}{*}{$k$}} & \multicolumn{1}{c|}{\multirow{3}{*}{$f^*$}} & \multicolumn{4}{c|}{Big-means} & \multicolumn{4}{c|}{IK-means} & \multicolumn{4}{c|}{BDCSM} & \multicolumn{4}{c|}{Minibatch K-means} \\ \cline{3-18}
\multicolumn{1}{|c|}{} & \multicolumn{1}{c|}{} & \multicolumn{2}{c|}{$\varepsilon$} & \multicolumn{2}{c|}{$t$} & \multicolumn{2}{c|}{$\varepsilon$} & \multicolumn{2}{c|}{$t$} & \multicolumn{2}{c|}{$\varepsilon$} & \multicolumn{2}{c|}{$t$} & \multicolumn{2}{c|}{$\varepsilon$} & \multicolumn{2}{c|}{$t$} \\ \cline{3-18}
\multicolumn{1}{|c|}{} & \multicolumn{1}{c|}{} & \multicolumn{1}{c|}{med} & \multicolumn{1}{c|}{std} & \multicolumn{1}{c|}{med} & \multicolumn{1}{c|}{std} & \multicolumn{1}{c|}{med} & \multicolumn{1}{c|}{std} & \multicolumn{1}{c|}{med} & \multicolumn{1}{c|}{std} & \multicolumn{1}{c|}{med} & \multicolumn{1}{c|}{std} & \multicolumn{1}{c|}{med} & \multicolumn{1}{c|}{std} & \multicolumn{1}{c|}{med} & \multicolumn{1}{c|}{std} & \multicolumn{1}{c|}{med} & \multicolumn{1}{c|}{std} \\ \hline
2 & 28.01938$^*$ & 0.02 & 0.01 & 0.43 & 0.25 & 684.56 & 38.55 & 0.57 & 0.14 & 4.17 & 1864.82 & 0.06 & 0.02 & 686.34 & 26.05 & 0.14 & 0.04 \\
3 & 19.85673$^*$ & 0.03 & 0.02 & 0.68 & 0.23 & 580.41 & 468.8 & 0.76 & 0.13 & 9.47 & 21.74 & 0.08 & 0.02 & 947.47 & 151.64 & 0.18 & 0.09 \\
5 & 12.10267$^*$ & 0.09 & 0.05 & 0.76 & 0.21 & -0.0 & 7.46 & 1.48 & 0.24 & 13.95 & 16.86 & 0.21 & 0.04 & 1625.43 & 703.04 & 0.4 & 0.35 \\
10 & 8.57382$^*$ & 0.56 & 0.76 & 0.68 & 0.25 & 3.12 & 1.17 & 4.89 & 0.88 & 8.42 & 5.08 & 0.87 & 0.19 & 3.77 & 85.05 & 0.28 & 0.13 \\
15 & 7.24131$^*$ & 0.78 & 0.35 & 0.79 & 0.25 & 2.66 & 1.75 & 7.67 & 1.07 & 8.68 & 4.79 & 1.86 & 0.35 & 3.4 & 1.25 & 0.3 & 0.06 \\
20 & 6.30493$^*$ & 1.33 & 0.52 & 1.1 & 0.33 & 2.98 & 2.23 & 11.2 & 0.87 & 10.07 & 4.49 & 2.4 & 0.49 & 4.13 & 1.26 & 0.3 & 0.05 \\
25 & 5.71335$^*$ & 1.35 & 0.51 & 1.18 & 0.41 & 2.32 & 0.75 & 15.74 & 1.29 & 8.14 & 3.58 & 2.92 & 0.49 & 4.74 & 1.27 & 0.48 & 0.1 \\
\hline
\multicolumn{2}{|c|}{Mean:} & \textbf{0.59} & & \textbf{0.8} & & \textbf{182.29} & & \textbf{6.04} & & \textbf{8.99} & & \textbf{1.2} & & \textbf{467.9} & & \textbf{0.3} & \\ \hline
\end{tabular}
}

\medskip

\small
\resizebox{!}{\tableheight}{
\begin{tabular}{|l|l|llll|llll|llll|llll|}
\hline
\multicolumn{1}{|c|}{\multirow{3}{*}{$k$}} & \multicolumn{1}{c|}{\multirow{3}{*}{$f^*$}} & \multicolumn{4}{c|}{K-means++} & \multicolumn{4}{c|}{CURE} & \multicolumn{4}{c|}{CluDataSE} & \multicolumn{4}{c|}{LW-coreset} \\ \cline{3-18}
\multicolumn{1}{|c|}{} & \multicolumn{1}{c|}{} & \multicolumn{2}{c|}{$\varepsilon$} & \multicolumn{2}{c|}{$t$} & \multicolumn{2}{c|}{$\varepsilon$} & \multicolumn{2}{c|}{$t$} & \multicolumn{2}{c|}{$\varepsilon$} & \multicolumn{2}{c|}{$t$} & \multicolumn{2}{c|}{$\varepsilon$} & \multicolumn{2}{c|}{$t$} \\ \cline{3-18}
\multicolumn{1}{|c|}{} & \multicolumn{1}{c|}{} & \multicolumn{1}{c|}{med} & \multicolumn{1}{c|}{std} & \multicolumn{1}{c|}{med} & \multicolumn{1}{c|}{std} & \multicolumn{1}{c|}{med} & \multicolumn{1}{c|}{std} & \multicolumn{1}{c|}{med} & \multicolumn{1}{c|}{std} & \multicolumn{1}{c|}{med} & \multicolumn{1}{c|}{std} & \multicolumn{1}{c|}{med} & \multicolumn{1}{c|}{std} & \multicolumn{1}{c|}{med} & \multicolumn{1}{c|}{std} & \multicolumn{1}{c|}{med} & \multicolumn{1}{c|}{std} \\ \hline
2 & 28.01938$^*$ & 0.0 & 149.2 & 0.08 & 0.06 & 3.09 & 319.04 & 4.0 & 0.2 & 0.0 & 0.0 & 0.15 & 0.01 & 0.01 & 0.01 & 0.12 & 0.01 \\
3 & 19.85673$^*$ & 3.49 & 3.49 & 0.48 & 0.11 & 19.17 & 6.86 & 3.97 & 0.48 & -0.0 & 0.0 & 0.53 & 0.06 & 7.02 & 3.34 & 0.14 & 0.02 \\
5 & 12.10267$^*$ & -0.0 & 2.67 & 0.95 & 0.41 & 40.53 & 17.34 & 4.44 & 0.18 & -0.0 & 0.83 & 1.16 & 0.17 & 0.14 & 1.91 & 0.16 & 0.03 \\
10 & 8.57382$^*$ & 0.31 & 0.98 & 4.45 & 2.1 & 37.64 & 14.94 & 6.72 & 0.44 & 2.48 & 0.93 & 6.0 & 1.03 & 1.15 & 1.43 & 0.4 & 0.1 \\
15 & 7.24131$^*$ & 0.28 & 0.83 & 8.47 & 3.67 & 43.43 & 7.36 & 8.59 & 0.33 & 0.54 & 1.63 & 11.12 & 5.71 & 1.3 & 0.76 & 0.54 & 0.13 \\
20 & 6.30493$^*$ & 0.83 & 0.57 & 15.36 & 6.38 & 43.98 & 7.18 & 12.33 & 0.58 & 0.6 & 0.61 & 15.32 & 4.37 & 2.62 & 1.23 & 0.61 & 0.22 \\
25 & 5.71335$^*$ & 0.23 & 0.24 & 18.57 & 9.43 & 49.78 & 8.67 & 9.26 & 0.49 & 0.29 & 0.16 & 23.9 & 6.86 & 3.51 & 1.21 & 0.85 & 0.21 \\
\hline
\multicolumn{2}{|c|}{Mean:} & \textbf{0.73} & & \textbf{6.91} & & \textbf{33.94} & & \textbf{7.04} & & \textbf{0.56} & & \textbf{8.31} & & \textbf{2.25} & & \textbf{0.4} & \\ \hline
\end{tabular}
}

\bigskip

\caption{Clustering details with MiniBooNE Particle Identification (normalized)}
\label{TabDetailsD8}
\resizebox{\linewidth}{!}{
\begin{tabular}{|l|l|lllll|l|l|ll|l|ll|lllll|ll|}
\hline
\multicolumn{1}{|c|}{\multirow{2}{*}{$k$}} & \multicolumn{1}{c|}{\multirow{2}{*}{$n_{exec}$}} & \multicolumn{5}{c|}{Big-means} & \multicolumn{1}{c|}{IK-means} & \multicolumn{1}{c|}{BDCSM} & \multicolumn{2}{c|}{Minibatch K-means} & \multicolumn{1}{c|}{K-means++} & \multicolumn{2}{c|}{CURE} & \multicolumn{5}{c|}{CluDataSE} & \multicolumn{2}{c|}{LW-coreset} \\ \cline{3-21}
\multicolumn{1}{|c|}{} & \multicolumn{1}{c|}{} & \multicolumn{1}{c|}{$s$} & \multicolumn{1}{c|}{$n_{s}$} & \multicolumn{1}{c|}{$T_1$} & \multicolumn{1}{c|}{$T_2$} & \multicolumn{1}{c|}{$n_{d}$} & \multicolumn{1}{c|}{$n_{d}$} & \multicolumn{1}{c|}{$n_{d}$} & \multicolumn{1}{c|}{$n_{s}$} & \multicolumn{1}{c|}{$n_{d}$} & \multicolumn{1}{c|}{$n_{d}$} & \multicolumn{1}{c|}{$s$} & \multicolumn{1}{c|}{$n_{d}$} & \multicolumn{1}{c|}{$s$} & \multicolumn{1}{c|}{$eps$} & \multicolumn{1}{c|}{$min\_pts$} & \multicolumn{1}{c|}{$sf$} & \multicolumn{1}{c|}{$n_{d}$} & \multicolumn{1}{c|}{$s$} & \multicolumn{1}{c|}{$n_{d}$} \\
\hline
2 & 20 & 12000 & 418 & 0.03 & 0.97 & 5.0E+07 & 1.9E+06 & 4.1E+06 & 14 & 3.5E+05 & 7.8E+05 & 12000 & 4.0E+07 & 12000 & 0.03 & 16 & 0.5 & 1.4E+08 & 12000 & 5.9E+05 \\
3 & 20 & 12000 & 451 & 0.03 & 0.97 & 7.8E+07 & 2.9E+06 & 7.2E+06 & 22 & 7.9E+05 & 8.2E+06 & 12000 & 3.3E+07 & 12000 & 0.03 & 16 & 0.5 & 1.5E+08 & 12000 & 1.2E+06 \\
5 & 20 & 12000 & 298 & 0.17 & 0.83 & 1.0E+08 & 5.9E+06 & 1.8E+07 & 74 & 4.4E+06 & 2.3E+07 & 12000 & 3.0E+07 & 12000 & 0.03 & 16 & 0.5 & 1.7E+08 & 12000 & 2.6E+06 \\
10 & 20 & 12000 & 50 & 0.67 & 0.33 & 1.1E+08 & 1.9E+07 & 7.9E+07 & 47 & 5.6E+06 & 1.2E+08 & 12000 & 3.3E+07 & 12000 & 0.03 & 16 & 0.5 & 3.0E+08 & 12000 & 7.0E+06 \\
15 & 20 & 12000 & 24 & 0.87 & 0.13 & 1.3E+08 & 3.6E+07 & 1.6E+08 & 42 & 7.5E+06 & 2.4E+08 & 12000 & 3.9E+07 & 12000 & 0.03 & 16 & 0.5 & 4.4E+08 & 12000 & 1.6E+07 \\
20 & 20 & 12000 & 12 & 0.23 & 0.77 & 1.1E+08 & 5.4E+07 & 2.3E+08 & 47 & 1.1E+07 & 4.7E+08 & 12000 & 4.5E+07 & 12000 & 0.03 & 16 & 0.5 & 6.0E+08 & 12000 & 1.8E+07 \\
25 & 20 & 12000 & 10 & 0.73 & 0.27 & 1.4E+08 & 7.2E+07 & 3.1E+08 & 44 & 1.3E+07 & 5.7E+08 & 12000 & 4.1E+07 & 12000 & 0.03 & 16 & 0.5 & 8.9E+08 & 12000 & 2.7E+07 \\
\hline
\end{tabular}
}

\end{table}

\newpage


\newpage

\subsection{MFCCs for Speech Emotion Recognition}
Dimensions: $m$ = 85134, $n$ = 58.
\par
Description: a data set for predicting females and males speech emotions based on Mel Frequency Cepstral Coefficients (MFCCs) values.

\begin{table}[!htbp]
\centering

\caption{Summary of the results with MFCCs for Speech Emotion Recognition ($\times10^{9}$)}
\label{TabResultsD9}
\small
\resizebox{!}{\tableheight}{
\begin{tabular}{|l|l|llll|llll|llll|llll|}
\hline
\multicolumn{1}{|c|}{\multirow{3}{*}{$k$}} & \multicolumn{1}{c|}{\multirow{3}{*}{$f^*$}} & \multicolumn{4}{c|}{Big-means} & \multicolumn{4}{c|}{IK-means} & \multicolumn{4}{c|}{BDCSM} & \multicolumn{4}{c|}{Minibatch K-means} \\ \cline{3-18}
\multicolumn{1}{|c|}{} & \multicolumn{1}{c|}{} & \multicolumn{2}{c|}{$\varepsilon$} & \multicolumn{2}{c|}{$t$} & \multicolumn{2}{c|}{$\varepsilon$} & \multicolumn{2}{c|}{$t$} & \multicolumn{2}{c|}{$\varepsilon$} & \multicolumn{2}{c|}{$t$} & \multicolumn{2}{c|}{$\varepsilon$} & \multicolumn{2}{c|}{$t$} \\ \cline{3-18}
\multicolumn{1}{|c|}{} & \multicolumn{1}{c|}{} & \multicolumn{1}{c|}{med} & \multicolumn{1}{c|}{std} & \multicolumn{1}{c|}{med} & \multicolumn{1}{c|}{std} & \multicolumn{1}{c|}{med} & \multicolumn{1}{c|}{std} & \multicolumn{1}{c|}{med} & \multicolumn{1}{c|}{std} & \multicolumn{1}{c|}{med} & \multicolumn{1}{c|}{std} & \multicolumn{1}{c|}{med} & \multicolumn{1}{c|}{std} & \multicolumn{1}{c|}{med} & \multicolumn{1}{c|}{std} & \multicolumn{1}{c|}{med} & \multicolumn{1}{c|}{std} \\ \hline
2 & 0.74513$^*$ & 0.04 & 0.02 & 0.67 & 0.29 & 0.77 & 4.71 & 0.13 & 0.01 & 0.0 & 0.0 & 0.05 & 0.01 & 0.13 & 2.48 & 0.16 & 0.07 \\
3 & 0.50215$^*$ & 0.04 & 0.05 & 0.38 & 0.23 & 3.93 & 8.01 & 0.18 & 0.03 & 69.81 & 36.42 & 0.07 & 0.01 & 0.2 & 4.53 & 0.23 & 0.08 \\
5 & 0.3456$^*$ & 0.05 & 0.03 & 0.47 & 0.22 & 10.13 & 5.12 & 0.29 & 0.04 & 25.73 & 27.26 & 0.19 & 0.04 & 2.1 & 3.28 & 0.25 & 0.07 \\
10 & 0.21763$^*$ & 0.13 & 0.04 & 0.7 & 0.24 & 13.59 & 7.17 & 0.57 & 0.07 & 8.23 & 5.86 & 0.7 & 0.14 & 3.89 & 2.45 & 0.35 & 0.07 \\
15 & 0.17608$^*$ & 0.47 & 0.65 & 0.85 & 0.25 & 14.35 & 4.47 & 0.82 & 0.13 & 8.91 & 4.99 & 1.07 & 0.22 & 3.29 & 1.27 & 0.29 & 0.05 \\
20 & 0.15383$^*$ & 0.8 & 0.49 & 0.8 & 0.32 & 19.41 & 6.93 & 1.3 & 0.15 & 13.61 & 4.85 & 2.07 & 0.31 & 3.55 & 1.38 & 0.38 & 0.07 \\
25 & 0.14109$^*$ & 0.96 & 0.5 & 1.29 & 0.48 & 15.31 & 4.71 & 1.66 & 0.3 & 13.04 & 4.11 & 2.68 & 0.33 & 3.32 & 0.79 & 0.35 & 0.07 \\
\hline
\multicolumn{2}{|c|}{Mean:} & \textbf{0.36} & & \textbf{0.74} & & \textbf{11.07} & & \textbf{0.71} & & \textbf{19.9} & & \textbf{0.98} & & \textbf{2.35} & & \textbf{0.29} & \\ \hline
\end{tabular}
}

\medskip

\small
\resizebox{!}{\tableheight}{
\begin{tabular}{|l|l|llll|llll|llll|llll|}
\hline
\multicolumn{1}{|c|}{\multirow{3}{*}{$k$}} & \multicolumn{1}{c|}{\multirow{3}{*}{$f^*$}} & \multicolumn{4}{c|}{K-means++} & \multicolumn{4}{c|}{CURE} & \multicolumn{4}{c|}{CluDataSE} & \multicolumn{4}{c|}{LW-coreset} \\ \cline{3-18}
\multicolumn{1}{|c|}{} & \multicolumn{1}{c|}{} & \multicolumn{2}{c|}{$\varepsilon$} & \multicolumn{2}{c|}{$t$} & \multicolumn{2}{c|}{$\varepsilon$} & \multicolumn{2}{c|}{$t$} & \multicolumn{2}{c|}{$\varepsilon$} & \multicolumn{2}{c|}{$t$} & \multicolumn{2}{c|}{$\varepsilon$} & \multicolumn{2}{c|}{$t$} \\ \cline{3-18}
\multicolumn{1}{|c|}{} & \multicolumn{1}{c|}{} & \multicolumn{1}{c|}{med} & \multicolumn{1}{c|}{std} & \multicolumn{1}{c|}{med} & \multicolumn{1}{c|}{std} & \multicolumn{1}{c|}{med} & \multicolumn{1}{c|}{std} & \multicolumn{1}{c|}{med} & \multicolumn{1}{c|}{std} & \multicolumn{1}{c|}{med} & \multicolumn{1}{c|}{std} & \multicolumn{1}{c|}{med} & \multicolumn{1}{c|}{std} & \multicolumn{1}{c|}{med} & \multicolumn{1}{c|}{std} & \multicolumn{1}{c|}{med} & \multicolumn{1}{c|}{std} \\ \hline
2 & 0.74513$^*$ & 0.0 & 0.0 & 0.22 & 0.05 & 17.73 & 6.96 & 7.12 & 0.62 & 0.0 & 0.0 & 0.35 & 0.02 & 0.02 & 0.01 & 0.11 & 0.02 \\
3 & 0.50215$^*$ & 0.0 & 0.0 & 0.3 & 0.05 & 29.78 & 24.69 & 6.24 & 0.33 & 0.0 & 0.0 & 0.42 & 0.06 & 0.03 & 0.02 & 0.1 & 0.02 \\
5 & 0.3456$^*$ & -0.0 & 0.0 & 0.75 & 0.17 & 32.44 & 15.24 & 6.89 & 0.51 & -0.0 & 0.0 & 1.21 & 0.25 & 0.05 & 0.05 & 0.2 & 0.05 \\
10 & 0.21763$^*$ & 1.88 & 1.25 & 2.25 & 0.79 & 62.49 & 16.29 & 7.98 & 0.77 & -0.01 & 1.17 & 2.83 & 0.67 & 2.71 & 2.2 & 0.32 & 0.14 \\
15 & 0.17608$^*$ & 1.35 & 0.77 & 4.5 & 2.06 & 49.94 & 11.76 & 10.29 & 0.76 & 1.7 & 1.58 & 6.19 & 2.56 & 3.47 & 1.64 & 0.5 & 0.13 \\
20 & 0.15383$^*$ & 1.14 & 1.31 & 7.44 & 2.49 & 44.23 & 13.8 & 14.43 & 1.27 & 2.1 & 1.36 & 10.78 & 2.7 & 3.12 & 2.01 & 0.76 & 0.35 \\
25 & 0.14109$^*$ & 0.67 & 1.23 & 11.75 & 3.21 & 39.19 & 6.69 & 10.01 & 0.72 & 4.94 & 1.79 & 16.19 & 5.31 & 2.87 & 2.07 & 1.07 & 0.34 \\
\hline
\multicolumn{2}{|c|}{Mean:} & \textbf{0.72} & & \textbf{3.89} & & \textbf{39.4} & & \textbf{8.99} & & \textbf{1.25} & & \textbf{5.42} & & \textbf{1.75} & & \textbf{0.44} & \\ \hline
\end{tabular}
}

\bigskip

\caption{Clustering details with MFCCs for Speech Emotion Recognition}
\label{TabDetailsD9}
\resizebox{\linewidth}{!}{
\begin{tabular}{|l|l|lllll|l|l|ll|l|ll|lllll|ll|}
\hline
\multicolumn{1}{|c|}{\multirow{2}{*}{$k$}} & \multicolumn{1}{c|}{\multirow{2}{*}{$n_{exec}$}} & \multicolumn{5}{c|}{Big-means} & \multicolumn{1}{c|}{IK-means} & \multicolumn{1}{c|}{BDCSM} & \multicolumn{2}{c|}{Minibatch K-means} & \multicolumn{1}{c|}{K-means++} & \multicolumn{2}{c|}{CURE} & \multicolumn{5}{c|}{CluDataSE} & \multicolumn{2}{c|}{LW-coreset} \\ \cline{3-21}
\multicolumn{1}{|c|}{} & \multicolumn{1}{c|}{} & \multicolumn{1}{c|}{$s$} & \multicolumn{1}{c|}{$n_{s}$} & \multicolumn{1}{c|}{$T_1$} & \multicolumn{1}{c|}{$T_2$} & \multicolumn{1}{c|}{$n_{d}$} & \multicolumn{1}{c|}{$n_{d}$} & \multicolumn{1}{c|}{$n_{d}$} & \multicolumn{1}{c|}{$n_{s}$} & \multicolumn{1}{c|}{$n_{d}$} & \multicolumn{1}{c|}{$n_{d}$} & \multicolumn{1}{c|}{$s$} & \multicolumn{1}{c|}{$n_{d}$} & \multicolumn{1}{c|}{$s$} & \multicolumn{1}{c|}{$eps$} & \multicolumn{1}{c|}{$min\_pts$} & \multicolumn{1}{c|}{$sf$} & \multicolumn{1}{c|}{$n_{d}$} & \multicolumn{1}{c|}{$s$} & \multicolumn{1}{c|}{$n_{d}$} \\
\hline
2 & 20 & 12000 & 612 & 0.77 & 0.23 & 6.6E+07 & 6.8E+05 & 3.3E+06 & 25 & 6.0E+05 & 3.9E+06 & 12000 & 6.2E+07 & 12000 & 25.0 & 16 & 0.5 & 1.5E+08 & 12000 & 8.0E+05 \\
3 & 20 & 12000 & 226 & 0.2 & 0.8 & 8.1E+07 & 1.0E+06 & 4.8E+06 & 36 & 1.3E+06 & 5.6E+06 & 12000 & 5.2E+07 & 12000 & 25.0 & 16 & 0.5 & 1.5E+08 & 12000 & 1.1E+06 \\
5 & 20 & 12000 & 144 & 0.83 & 0.17 & 9.6E+07 & 1.7E+06 & 1.5E+07 & 36 & 2.2E+06 & 1.6E+07 & 12000 & 4.4E+07 & 12000 & 25.0 & 16 & 0.5 & 1.6E+08 & 12000 & 2.5E+06 \\
10 & 20 & 12000 & 58 & 0.97 & 0.03 & 1.1E+08 & 3.4E+06 & 4.8E+07 & 37 & 4.4E+06 & 5.4E+07 & 12000 & 4.0E+07 & 12000 & 25.0 & 16 & 0.5 & 2.1E+08 & 12000 & 7.3E+06 \\
15 & 20 & 12000 & 14 & 0.03 & 0.97 & 8.0E+07 & 5.1E+06 & 9.0E+07 & 42 & 7.6E+06 & 1.2E+08 & 12000 & 4.2E+07 & 12000 & 25.0 & 16 & 0.5 & 3.0E+08 & 12000 & 1.3E+07 \\
20 & 20 & 12000 & 12 & 0.9 & 0.1 & 1.1E+08 & 7.9E+06 & 1.8E+08 & 40 & 9.5E+06 & 1.9E+08 & 12000 & 4.6E+07 & 12000 & 25.0 & 16 & 0.5 & 4.2E+08 & 12000 & 2.0E+07 \\
25 & 20 & 12000 & 14 & 0.83 & 0.17 & 1.3E+08 & 1.0E+07 & 2.1E+08 & 32 & 9.4E+06 & 3.0E+08 & 12000 & 4.3E+07 & 12000 & 25.0 & 16 & 0.5 & 5.6E+08 & 12000 & 2.9E+07 \\
\hline
\end{tabular}
}

\end{table}

\newpage


\subsection{ISOLET}
Dimensions: $m$ = 7797, $n$ = 617.
\par
Description: data set of patterns for spoken letter recognition which contains the spectral coefficients and other additional features.

\begin{table}[!htbp]
\centering

\caption{Summary of the results with ISOLET ($\times10^{5}$)}
\label{TabResultsD10}
\small
\resizebox{!}{\tableheight}{
\begin{tabular}{|l|l|llll|llll|llll|llll|}
\hline
\multicolumn{1}{|c|}{\multirow{3}{*}{$k$}} & \multicolumn{1}{c|}{\multirow{3}{*}{$f^*$}} & \multicolumn{4}{c|}{Big-means} & \multicolumn{4}{c|}{IK-means} & \multicolumn{4}{c|}{BDCSM} & \multicolumn{4}{c|}{Minibatch K-means} \\ \cline{3-18}
\multicolumn{1}{|c|}{} & \multicolumn{1}{c|}{} & \multicolumn{2}{c|}{$\varepsilon$} & \multicolumn{2}{c|}{$t$} & \multicolumn{2}{c|}{$\varepsilon$} & \multicolumn{2}{c|}{$t$} & \multicolumn{2}{c|}{$\varepsilon$} & \multicolumn{2}{c|}{$t$} & \multicolumn{2}{c|}{$\varepsilon$} & \multicolumn{2}{c|}{$t$} \\ \cline{3-18}
\multicolumn{1}{|c|}{} & \multicolumn{1}{c|}{} & \multicolumn{1}{c|}{med} & \multicolumn{1}{c|}{std} & \multicolumn{1}{c|}{med} & \multicolumn{1}{c|}{std} & \multicolumn{1}{c|}{med} & \multicolumn{1}{c|}{std} & \multicolumn{1}{c|}{med} & \multicolumn{1}{c|}{std} & \multicolumn{1}{c|}{med} & \multicolumn{1}{c|}{std} & \multicolumn{1}{c|}{med} & \multicolumn{1}{c|}{std} & \multicolumn{1}{c|}{med} & \multicolumn{1}{c|}{std} & \multicolumn{1}{c|}{med} & \multicolumn{1}{c|}{std} \\ \hline
2 & 7.2194 & 0.03 & 0.01 & 1.65 & 1.15 & 0.09 & 4.02 & 0.07 & 0.01 & 0.02 & 0.01 & 0.05 & 0.02 & 0.03 & 0.07 & 0.18 & 0.04 \\
3 & 6.78782 & 0.04 & 0.01 & 3.33 & 1.03 & 1.17 & 0.94 & 0.11 & 0.01 & 0.59 & 0.26 & 0.14 & 0.05 & 0.6 & 0.62 & 0.21 & 0.07 \\
5 & 6.13651 & 0.07 & 0.14 & 2.58 & 1.25 & 2.83 & 1.4 & 0.17 & 0.02 & 0.47 & 0.56 & 0.35 & 0.2 & 1.92 & 1.21 & 0.28 & 0.06 \\
10 & 5.28577 & 0.31 & 0.12 & 2.99 & 1.28 & 3.15 & 1.86 & 0.32 & 0.01 & 0.47 & 1.44 & 0.57 & 0.2 & 2.14 & 1.1 & 0.38 & 0.06 \\
15 & 4.87391 & 0.81 & 0.38 & 4.0 & 1.06 & 3.94 & 1.18 & 0.55 & 0.14 & 1.82 & 1.02 & 0.84 & 0.4 & 2.93 & 0.86 & 0.39 & 0.1 \\
20 & 4.60857 & 0.66 & 0.45 & 4.41 & 1.11 & 4.04 & 1.41 & 0.64 & 0.01 & 1.96 & 0.83 & 1.08 & 0.37 & 2.47 & 0.68 & 0.34 & 0.07 \\
25 & 4.44323 & 0.19 & 0.26 & 4.2 & 1.2 & 3.54 & 1.13 & 0.79 & 0.01 & 1.8 & 0.62 & 1.35 & 0.59 & 2.17 & 0.81 & 0.35 & 0.06 \\
\hline
\multicolumn{2}{|c|}{Mean:} & \textbf{0.3} & & \textbf{3.31} & & \textbf{2.68} & & \textbf{0.38} & & \textbf{1.02} & & \textbf{0.63} & & \textbf{1.75} & & \textbf{0.3} & \\ \hline
\end{tabular}
}

\medskip

\small
\resizebox{!}{\tableheight}{
\begin{tabular}{|l|l|llll|llll|llll|llll|}
\hline
\multicolumn{1}{|c|}{\multirow{3}{*}{$k$}} & \multicolumn{1}{c|}{\multirow{3}{*}{$f^*$}} & \multicolumn{4}{c|}{K-means++} & \multicolumn{4}{c|}{CURE} & \multicolumn{4}{c|}{CluDataSE} & \multicolumn{4}{c|}{LW-coreset} \\ \cline{3-18}
\multicolumn{1}{|c|}{} & \multicolumn{1}{c|}{} & \multicolumn{2}{c|}{$\varepsilon$} & \multicolumn{2}{c|}{$t$} & \multicolumn{2}{c|}{$\varepsilon$} & \multicolumn{2}{c|}{$t$} & \multicolumn{2}{c|}{$\varepsilon$} & \multicolumn{2}{c|}{$t$} & \multicolumn{2}{c|}{$\varepsilon$} & \multicolumn{2}{c|}{$t$} \\ \cline{3-18}
\multicolumn{1}{|c|}{} & \multicolumn{1}{c|}{} & \multicolumn{1}{c|}{med} & \multicolumn{1}{c|}{std} & \multicolumn{1}{c|}{med} & \multicolumn{1}{c|}{std} & \multicolumn{1}{c|}{med} & \multicolumn{1}{c|}{std} & \multicolumn{1}{c|}{med} & \multicolumn{1}{c|}{std} & \multicolumn{1}{c|}{med} & \multicolumn{1}{c|}{std} & \multicolumn{1}{c|}{med} & \multicolumn{1}{c|}{std} & \multicolumn{1}{c|}{med} & \multicolumn{1}{c|}{std} & \multicolumn{1}{c|}{med} & \multicolumn{1}{c|}{std} \\ \hline
2 & 7.2194 & -0.0 & 0.0 & 0.22 & 0.06 & 19.2 & 3.27 & 1.32 & 0.06 & -0.0 & 0.0 & 0.24 & 0.03 & 0.05 & 0.02 & 0.18 & 0.02 \\
3 & 6.78782 & 0.55 & 0.52 & 0.49 & 0.39 & 24.29 & 4.43 & 1.45 & 0.08 & 0.55 & 0.26 & 0.5 & 0.39 & 0.13 & 0.53 & 0.22 & 0.08 \\
5 & 6.13651 & 0.39 & 0.49 & 0.78 & 0.18 & 24.07 & 7.12 & 1.72 & 0.11 & 0.39 & 0.48 & 1.01 & 0.17 & 0.56 & 1.15 & 0.39 & 0.15 \\
10 & 5.28577 & 1.06 & 1.17 & 1.36 & 0.78 & 50.87 & 12.37 & 2.79 & 0.16 & 1.39 & 0.68 & 2.12 & 0.85 & 2.34 & 1.32 & 0.78 & 0.27 \\
15 & 4.87391 & 0.82 & 0.46 & 2.18 & 1.04 & 57.14 & 6.7 & 2.55 & 0.11 & 2.23 & 0.87 & 2.11 & 0.73 & 2.14 & 1.24 & 1.08 & 0.48 \\
20 & 4.60857 & 0.93 & 0.94 & 3.52 & 1.28 & 54.25 & 4.81 & 2.07 & 0.11 & 2.07 & 0.69 & 3.81 & 1.19 & 2.64 & 1.32 & 1.18 & 0.39 \\
25 & 4.44323 & 0.66 & 0.56 & 4.92 & 1.52 & 43.09 & 6.57 & 1.69 & 0.06 & 2.76 & 1.08 & 4.68 & 1.42 & 2.29 & 0.94 & 1.52 & 0.29 \\
\hline
\multicolumn{2}{|c|}{Mean:} & \textbf{0.63} & & \textbf{1.92} & & \textbf{38.99} & & \textbf{1.94} & & \textbf{1.34} & & \textbf{2.07} & & \textbf{1.45} & & \textbf{0.77} & \\ \hline
\end{tabular}
}

\bigskip

\caption{Clustering details with ISOLET}
\label{TabDetailsD10}
\resizebox{\linewidth}{!}{
\begin{tabular}{|l|l|lllll|l|l|ll|l|ll|lllll|ll|}
\hline
\multicolumn{1}{|c|}{\multirow{2}{*}{$k$}} & \multicolumn{1}{c|}{\multirow{2}{*}{$n_{exec}$}} & \multicolumn{5}{c|}{Big-means} & \multicolumn{1}{c|}{IK-means} & \multicolumn{1}{c|}{BDCSM} & \multicolumn{2}{c|}{Minibatch K-means} & \multicolumn{1}{c|}{K-means++} & \multicolumn{2}{c|}{CURE} & \multicolumn{5}{c|}{CluDataSE} & \multicolumn{2}{c|}{LW-coreset} \\ \cline{3-21}
\multicolumn{1}{|c|}{} & \multicolumn{1}{c|}{} & \multicolumn{1}{c|}{$s$} & \multicolumn{1}{c|}{$n_{s}$} & \multicolumn{1}{c|}{$T_1$} & \multicolumn{1}{c|}{$T_2$} & \multicolumn{1}{c|}{$n_{d}$} & \multicolumn{1}{c|}{$n_{d}$} & \multicolumn{1}{c|}{$n_{d}$} & \multicolumn{1}{c|}{$n_{s}$} & \multicolumn{1}{c|}{$n_{d}$} & \multicolumn{1}{c|}{$n_{d}$} & \multicolumn{1}{c|}{$s$} & \multicolumn{1}{c|}{$n_{d}$} & \multicolumn{1}{c|}{$s$} & \multicolumn{1}{c|}{$eps$} & \multicolumn{1}{c|}{$min\_pts$} & \multicolumn{1}{c|}{$sf$} & \multicolumn{1}{c|}{$n_{d}$} & \multicolumn{1}{c|}{$s$} & \multicolumn{1}{c|}{$n_{d}$} \\
\hline
2 & 15 & 4000 & 622 & 1.17 & 3.83 & 5.1E+07 & 6.3E+04 & 1.0E+05 & 20 & 1.6E+05 & 1.9E+05 & 4000 & 2.5E+06 & 4000 & 9.0 & 16 & 0.5 & 1.6E+07 & 4000 & 1.2E+05 \\
3 & 15 & 4000 & 936 & 1.67 & 3.33 & 5.5E+07 & 1.0E+05 & 2.8E+05 & 22 & 2.6E+05 & 5.1E+05 & 4000 & 2.3E+06 & 4000 & 9.0 & 16 & 0.5 & 1.6E+07 & 4000 & 2.8E+05 \\
5 & 15 & 4000 & 419 & 1.83 & 3.17 & 5.8E+07 & 1.7E+05 & 6.0E+05 & 33 & 6.6E+05 & 1.1E+06 & 4000 & 2.9E+06 & 4000 & 9.0 & 16 & 0.5 & 1.7E+07 & 4000 & 5.9E+05 \\
10 & 15 & 4000 & 143 & 4.67 & 0.33 & 6.0E+07 & 3.3E+05 & 1.1E+06 & 33 & 1.3E+06 & 1.9E+06 & 4000 & 5.3E+06 & 4000 & 9.0 & 16 & 0.5 & 1.9E+07 & 4000 & 1.2E+06 \\
15 & 15 & 4000 & 101 & 1.17 & 3.83 & 5.3E+07 & 5.0E+05 & 1.5E+06 & 29 & 1.7E+06 & 3.0E+06 & 4000 & 4.7E+06 & 4000 & 9.0 & 16 & 0.5 & 1.9E+07 & 4000 & 1.8E+06 \\
20 & 15 & 4000 & 81 & 3.33 & 1.67 & 5.4E+07 & 6.8E+05 & 2.2E+06 & 33 & 2.6E+06 & 4.8E+06 & 4000 & 3.8E+06 & 4000 & 9.0 & 16 & 0.5 & 2.2E+07 & 4000 & 2.1E+06 \\
25 & 15 & 4000 & 47 & 3.17 & 1.83 & 5.3E+07 & 8.5E+05 & 2.7E+06 & 27 & 2.7E+06 & 7.2E+06 & 4000 & 3.2E+06 & 4000 & 9.0 & 16 & 0.5 & 2.4E+07 & 4000 & 2.7E+06 \\
\hline
\end{tabular}
}

\end{table}

\newpage


\newpage

\subsection{Sensorless Drive Diagnosis}
Dimensions: $m$ = 58509, $n$ = 48.
\par
Description: a data set for sensorless drive diagnosis with features extracted from motor current.

\begin{table}[!htbp]
\centering

\caption{Summary of the results with Sensorless Drive Diagnosis ($\times10^{7}$)}
\label{TabResultsD11}
\small
\resizebox{!}{\tableheight}{
\begin{tabular}{|l|l|llll|llll|llll|llll|}
\hline
\multicolumn{1}{|c|}{\multirow{3}{*}{$k$}} & \multicolumn{1}{c|}{\multirow{3}{*}{$f^*$}} & \multicolumn{4}{c|}{Big-means} & \multicolumn{4}{c|}{IK-means} & \multicolumn{4}{c|}{BDCSM} & \multicolumn{4}{c|}{Minibatch K-means} \\ \cline{3-18}
\multicolumn{1}{|c|}{} & \multicolumn{1}{c|}{} & \multicolumn{2}{c|}{$\varepsilon$} & \multicolumn{2}{c|}{$t$} & \multicolumn{2}{c|}{$\varepsilon$} & \multicolumn{2}{c|}{$t$} & \multicolumn{2}{c|}{$\varepsilon$} & \multicolumn{2}{c|}{$t$} & \multicolumn{2}{c|}{$\varepsilon$} & \multicolumn{2}{c|}{$t$} \\ \cline{3-18}
\multicolumn{1}{|c|}{} & \multicolumn{1}{c|}{} & \multicolumn{1}{c|}{med} & \multicolumn{1}{c|}{std} & \multicolumn{1}{c|}{med} & \multicolumn{1}{c|}{std} & \multicolumn{1}{c|}{med} & \multicolumn{1}{c|}{std} & \multicolumn{1}{c|}{med} & \multicolumn{1}{c|}{std} & \multicolumn{1}{c|}{med} & \multicolumn{1}{c|}{std} & \multicolumn{1}{c|}{med} & \multicolumn{1}{c|}{std} & \multicolumn{1}{c|}{med} & \multicolumn{1}{c|}{std} & \multicolumn{1}{c|}{med} & \multicolumn{1}{c|}{std} \\ \hline
2 & 3.88116 & -0.0 & 2.25 & 0.68 & 0.29 & 100.86 & 1.12 & 0.11 & 0.03 & 100.19 & 0.0 & 0.14 & 0.05 & 102.98 & 1.85 & 0.25 & 0.09 \\
3 & 2.91313 & -0.0 & 0.55 & 0.69 & 0.24 & 159.87 & 3.01 & 0.16 & 0.03 & 155.19 & 72.12 & 0.31 & 0.15 & 162.74 & 3.9 & 0.36 & 0.11 \\
5 & 1.93651 & 0.02 & 1.05 & 0.78 & 0.23 & 273.99 & 6.38 & 0.25 & 0.01 & 37.86 & 35.66 & 0.46 & 0.08 & 271.65 & 5.39 & 0.34 & 0.12 \\
10 & 0.98472 & -2.4 & 0.93 & 1.39 & 0.46 & 596.17 & 121.42 & 0.47 & 0.05 & 127.2 & 0.04 & 1.8 & 0.4 & 589.72 & 6.23 & 0.45 & 0.13 \\
15 & 0.62816 & 0.02 & 1.27 & 1.58 & 0.59 & 959.36 & 260.82 & 0.69 & 0.13 & 235.44 & 0.51 & 1.78 & 0.54 & 951.77 & 16.69 & 0.36 & 0.08 \\
20 & 0.49884 & -0.08 & 0.44 & 2.06 & 0.67 & 1221.22 & 296.82 & 0.91 & 0.09 & 309.27 & 24.05 & 3.74 & 1.28 & 1208.06 & 33.43 & 0.48 & 0.12 \\
25 & 0.42225 & 0.95 & 2.34 & 2.65 & 0.51 & 1398.7 & 364.82 & 1.14 & 0.02 & 315.17 & 12.88 & 6.61 & 1.54 & 1435.83 & 38.46 & 0.44 & 0.07 \\
\hline
\multicolumn{2}{|c|}{Mean:} & \textbf{-0.21} & & \textbf{1.4} & & \textbf{672.88} & & \textbf{0.53} & & \textbf{182.9} & & \textbf{2.12} & & \textbf{674.68} & & \textbf{0.38} & \\ \hline
\end{tabular}
}

\medskip

\small
\resizebox{!}{\tableheight}{
\begin{tabular}{|l|l|llll|llll|llll|llll|}
\hline
\multicolumn{1}{|c|}{\multirow{3}{*}{$k$}} & \multicolumn{1}{c|}{\multirow{3}{*}{$f^*$}} & \multicolumn{4}{c|}{K-means++} & \multicolumn{4}{c|}{CURE} & \multicolumn{4}{c|}{CluDataSE} & \multicolumn{4}{c|}{LW-coreset} \\ \cline{3-18}
\multicolumn{1}{|c|}{} & \multicolumn{1}{c|}{} & \multicolumn{2}{c|}{$\varepsilon$} & \multicolumn{2}{c|}{$t$} & \multicolumn{2}{c|}{$\varepsilon$} & \multicolumn{2}{c|}{$t$} & \multicolumn{2}{c|}{$\varepsilon$} & \multicolumn{2}{c|}{$t$} & \multicolumn{2}{c|}{$\varepsilon$} & \multicolumn{2}{c|}{$t$} \\ \cline{3-18}
\multicolumn{1}{|c|}{} & \multicolumn{1}{c|}{} & \multicolumn{1}{c|}{med} & \multicolumn{1}{c|}{std} & \multicolumn{1}{c|}{med} & \multicolumn{1}{c|}{std} & \multicolumn{1}{c|}{med} & \multicolumn{1}{c|}{std} & \multicolumn{1}{c|}{med} & \multicolumn{1}{c|}{std} & \multicolumn{1}{c|}{med} & \multicolumn{1}{c|}{std} & \multicolumn{1}{c|}{med} & \multicolumn{1}{c|}{std} & \multicolumn{1}{c|}{med} & \multicolumn{1}{c|}{std} & \multicolumn{1}{c|}{med} & \multicolumn{1}{c|}{std} \\ \hline
2 & 3.88116 & -0.0 & 45.91 & 0.04 & 0.1 & 107.02 & 4.22 & 2.43 & 0.11 & 100.19 & 0.0 & 0.4 & 0.05 & 0.01 & 0.01 & 0.03 & 0.0 \\
3 & 2.91313 & 10.86 & 39.12 & 0.2 & 0.13 & 168.74 & 4.83 & 2.74 & 0.1 & 10.86 & 71.42 & 0.87 & 0.23 & 10.91 & 0.03 & 0.05 & 0.01 \\
5 & 1.93651 & 25.53 & 16.58 & 0.43 & 0.22 & 289.24 & 7.4 & 3.5 & 0.6 & 37.86 & 74.9 & 1.06 & 0.14 & 37.97 & 8.29 & 0.07 & 0.02 \\
10 & 0.98472 & 14.54 & 9.69 & 1.28 & 0.3 & 650.79 & 9.22 & 5.07 & 0.4 & 127.2 & 0.04 & 2.59 & 0.47 & 108.26 & 24.43 & 0.17 & 0.06 \\
15 & 0.62816 & 17.38 & 15.85 & 1.83 & 0.76 & 1069.0 & 10.48 & 6.74 & 0.24 & 235.52 & 0.45 & 2.29 & 0.44 & 198.12 & 63.83 & 0.25 & 0.09 \\
20 & 0.49884 & 3.61 & 2.59 & 2.68 & 0.68 & 1368.43 & 9.32 & 7.37 & 0.38 & 309.34 & 22.34 & 4.12 & 1.65 & 260.43 & 66.21 & 0.33 & 0.08 \\
25 & 0.42225 & 3.42 & 6.86 & 3.49 & 1.11 & 1634.51 & 12.01 & 5.93 & 0.28 & 315.71 & 40.71 & 7.48 & 1.58 & 284.68 & 101.69 & 0.41 & 0.1 \\
\hline
\multicolumn{2}{|c|}{Mean:} & \textbf{10.76} & & \textbf{1.42} & & \textbf{755.39} & & \textbf{4.82} & & \textbf{162.38} & & \textbf{2.69} & & \textbf{128.63} & & \textbf{0.19} & \\ \hline
\end{tabular}
}

\bigskip

\caption{Clustering details with Sensorless Drive Diagnosis}
\label{TabDetailsD11}
\resizebox{\linewidth}{!}{
\begin{tabular}{|l|l|lllll|l|l|ll|l|ll|lllll|ll|}
\hline
\multicolumn{1}{|c|}{\multirow{2}{*}{$k$}} & \multicolumn{1}{c|}{\multirow{2}{*}{$n_{exec}$}} & \multicolumn{5}{c|}{Big-means} & \multicolumn{1}{c|}{IK-means} & \multicolumn{1}{c|}{BDCSM} & \multicolumn{2}{c|}{Minibatch K-means} & \multicolumn{1}{c|}{K-means++} & \multicolumn{2}{c|}{CURE} & \multicolumn{5}{c|}{CluDataSE} & \multicolumn{2}{c|}{LW-coreset} \\ \cline{3-21}
\multicolumn{1}{|c|}{} & \multicolumn{1}{c|}{} & \multicolumn{1}{c|}{$s$} & \multicolumn{1}{c|}{$n_{s}$} & \multicolumn{1}{c|}{$T_1$} & \multicolumn{1}{c|}{$T_2$} & \multicolumn{1}{c|}{$n_{d}$} & \multicolumn{1}{c|}{$n_{d}$} & \multicolumn{1}{c|}{$n_{d}$} & \multicolumn{1}{c|}{$n_{s}$} & \multicolumn{1}{c|}{$n_{d}$} & \multicolumn{1}{c|}{$n_{d}$} & \multicolumn{1}{c|}{$s$} & \multicolumn{1}{c|}{$n_{d}$} & \multicolumn{1}{c|}{$s$} & \multicolumn{1}{c|}{$eps$} & \multicolumn{1}{c|}{$min\_pts$} & \multicolumn{1}{c|}{$sf$} & \multicolumn{1}{c|}{$n_{d}$} & \multicolumn{1}{c|}{$s$} & \multicolumn{1}{c|}{$n_{d}$} \\
\hline
2 & 40 & 58500 & 237 & 0.27 & 0.73 & 8.9E+07 & 4.7E+05 & 4.0E+06 & 17 & 2.0E+06 & 4.7E+05 & 10000 & 1.5E+07 & 10000 & 4.0 & 16 & 0.5 & 1.0E+08 & 10000 & 2.9E+05 \\
3 & 40 & 58500 & 151 & 0.2 & 0.8 & 1.0E+08 & 7.8E+05 & 9.8E+06 & 20 & 3.5E+06 & 5.4E+06 & 10000 & 1.4E+07 & 10000 & 4.0 & 16 & 0.5 & 1.2E+08 & 10000 & 9.2E+05 \\
5 & 40 & 58500 & 72 & 0.83 & 0.17 & 1.1E+08 & 1.3E+06 & 1.5E+07 & 17 & 5.0E+06 & 1.2E+07 & 10000 & 1.4E+07 & 10000 & 4.0 & 16 & 0.5 & 1.2E+08 & 10000 & 1.7E+06 \\
10 & 40 & 58500 & 12 & 0.63 & 0.37 & 1.2E+08 & 2.5E+06 & 6.0E+07 & 20 & 1.1E+07 & 3.9E+07 & 10000 & 1.8E+07 & 10000 & 4.0 & 16 & 0.5 & 1.6E+08 & 10000 & 5.2E+06 \\
15 & 40 & 58500 & 3 & 0.4 & 0.6 & 2.2E+08 & 3.8E+06 & 6.1E+07 & 20 & 1.8E+07 & 5.6E+07 & 10000 & 2.2E+07 & 10000 & 4.0 & 16 & 0.5 & 1.6E+08 & 10000 & 8.3E+06 \\
20 & 40 & 58500 & 4 & 0.53 & 0.47 & 2.7E+08 & 5.0E+06 & 1.2E+08 & 18 & 2.0E+07 & 8.5E+07 & 10000 & 2.3E+07 & 10000 & 4.0 & 16 & 0.5 & 2.2E+08 & 10000 & 1.1E+07 \\
25 & 40 & 58500 & 5 & 0.77 & 0.23 & 3.5E+08 & 6.3E+06 & 2.2E+08 & 18 & 2.6E+07 & 1.1E+08 & 10000 & 2.1E+07 & 10000 & 4.0 & 16 & 0.5 & 3.4E+08 & 10000 & 1.4E+07 \\
\hline
\end{tabular}
}

\end{table}

\newpage


\subsection{Sensorless Drive Diagnosis (normalized)}
Dimensions: $m$ = 58509, $n$ = 48.
\par
Description: a data set for sensorless drive diagnosis with features extracted from motor current. Min-max scaling was used for normalization of data set values for better clusterization.

\begin{table}[!htbp]
\centering

\caption{Summary of the results with Sensorless Drive Diagnosis (normalized) ($\times10^{3}$)}
\label{TabResultsD12}
\small
\resizebox{!}{\tableheight}{
\begin{tabular}{|l|l|llll|llll|llll|llll|}
\hline
\multicolumn{1}{|c|}{\multirow{3}{*}{$k$}} & \multicolumn{1}{c|}{\multirow{3}{*}{$f^*$}} & \multicolumn{4}{c|}{Big-means} & \multicolumn{4}{c|}{IK-means} & \multicolumn{4}{c|}{BDCSM} & \multicolumn{4}{c|}{Minibatch K-means} \\ \cline{3-18}
\multicolumn{1}{|c|}{} & \multicolumn{1}{c|}{} & \multicolumn{2}{c|}{$\varepsilon$} & \multicolumn{2}{c|}{$t$} & \multicolumn{2}{c|}{$\varepsilon$} & \multicolumn{2}{c|}{$t$} & \multicolumn{2}{c|}{$\varepsilon$} & \multicolumn{2}{c|}{$t$} & \multicolumn{2}{c|}{$\varepsilon$} & \multicolumn{2}{c|}{$t$} \\ \cline{3-18}
\multicolumn{1}{|c|}{} & \multicolumn{1}{c|}{} & \multicolumn{1}{c|}{med} & \multicolumn{1}{c|}{std} & \multicolumn{1}{c|}{med} & \multicolumn{1}{c|}{std} & \multicolumn{1}{c|}{med} & \multicolumn{1}{c|}{std} & \multicolumn{1}{c|}{med} & \multicolumn{1}{c|}{std} & \multicolumn{1}{c|}{med} & \multicolumn{1}{c|}{std} & \multicolumn{1}{c|}{med} & \multicolumn{1}{c|}{std} & \multicolumn{1}{c|}{med} & \multicolumn{1}{c|}{std} & \multicolumn{1}{c|}{med} & \multicolumn{1}{c|}{std} \\ \hline
2 & 15.64798$^*$ & 0.1 & 0.04 & 0.16 & 0.09 & 0.0 & 16.58 & 0.11 & 0.02 & 0.0 & 0.88 & 0.01 & 0.0 & 0.03 & 0.13 & 0.14 & 0.05 \\
3 & 12.19375$^*$ & 0.11 & 0.99 & 0.21 & 0.09 & 3.57 & 3.22 & 0.18 & 0.07 & 1.78 & 21.07 & 0.03 & 0.01 & 3.6 & 2.53 & 0.11 & 0.06 \\
5 & 7.85054$^*$ & 0.31 & 0.23 & 0.14 & 0.07 & 3.7 & 7.55 & 0.29 & 0.04 & 4.02 & 9.62 & 0.05 & 0.01 & 0.82 & 2.9 & 0.13 & 0.05 \\
10 & 4.71275$^*$ & 0.65 & 1.11 & 0.26 & 0.08 & 7.95 & 7.17 & 0.68 & 0.12 & 12.71 & 11.38 & 0.12 & 0.03 & 6.1 & 2.66 & 0.13 & 0.04 \\
15 & 3.62541$^*$ & 1.34 & 1.04 & 0.29 & 0.08 & 9.08 & 6.52 & 1.15 & 0.21 & 16.44 & 8.86 & 0.21 & 0.04 & 6.43 & 2.86 & 0.15 & 0.03 \\
20 & 2.971$^*$ & 2.39 & 1.49 & 0.26 & 0.06 & 13.94 & 7.85 & 1.61 & 0.13 & 18.27 & 7.85 & 0.27 & 0.05 & 6.22 & 2.74 & 0.15 & 0.04 \\
25 & 2.60929$^*$ & 2.97 & 1.44 & 0.26 & 0.07 & 19.58 & 5.65 & 2.15 & 0.24 & 21.43 & 6.85 & 0.31 & 0.04 & 5.15 & 2.18 & 0.15 & 0.11 \\
\hline
\multicolumn{2}{|c|}{Mean:} & \textbf{1.12} & & \textbf{0.23} & & \textbf{8.26} & & \textbf{0.88} & & \textbf{10.66} & & \textbf{0.14} & & \textbf{4.05} & & \textbf{0.14} & \\ \hline
\end{tabular}
}

\medskip

\small
\resizebox{!}{\tableheight}{
\begin{tabular}{|l|l|llll|llll|llll|llll|}
\hline
\multicolumn{1}{|c|}{\multirow{3}{*}{$k$}} & \multicolumn{1}{c|}{\multirow{3}{*}{$f^*$}} & \multicolumn{4}{c|}{K-means++} & \multicolumn{4}{c|}{CURE} & \multicolumn{4}{c|}{CluDataSE} & \multicolumn{4}{c|}{LW-coreset} \\ \cline{3-18}
\multicolumn{1}{|c|}{} & \multicolumn{1}{c|}{} & \multicolumn{2}{c|}{$\varepsilon$} & \multicolumn{2}{c|}{$t$} & \multicolumn{2}{c|}{$\varepsilon$} & \multicolumn{2}{c|}{$t$} & \multicolumn{2}{c|}{$\varepsilon$} & \multicolumn{2}{c|}{$t$} & \multicolumn{2}{c|}{$\varepsilon$} & \multicolumn{2}{c|}{$t$} \\ \cline{3-18}
\multicolumn{1}{|c|}{} & \multicolumn{1}{c|}{} & \multicolumn{1}{c|}{med} & \multicolumn{1}{c|}{std} & \multicolumn{1}{c|}{med} & \multicolumn{1}{c|}{std} & \multicolumn{1}{c|}{med} & \multicolumn{1}{c|}{std} & \multicolumn{1}{c|}{med} & \multicolumn{1}{c|}{std} & \multicolumn{1}{c|}{med} & \multicolumn{1}{c|}{std} & \multicolumn{1}{c|}{med} & \multicolumn{1}{c|}{std} & \multicolumn{1}{c|}{med} & \multicolumn{1}{c|}{std} & \multicolumn{1}{c|}{med} & \multicolumn{1}{c|}{std} \\ \hline
2 & 15.64798$^*$ & 0.0 & 11.81 & 0.06 & 0.02 & 72.6 & 28.17 & 0.91 & 0.05 & 0.0 & 19.93 & 0.08 & 0.02 & 0.05 & 20.18 & 0.03 & 0.0 \\
3 & 12.19375$^*$ & 0.98 & 2.94 & 0.14 & 0.06 & 38.3 & 39.72 & 1.37 & 0.16 & 0.98 & 2.15 & 0.15 & 0.12 & 1.05 & 2.36 & 0.03 & 0.0 \\
5 & 7.85054$^*$ & 0.5 & 2.05 & 0.24 & 0.15 & 50.8 & 33.14 & 1.3 & 0.12 & 0.54 & 4.73 & 0.16 & 0.13 & 0.81 & 6.36 & 0.03 & 0.0 \\
10 & 4.71275$^*$ & 5.51 & 2.93 & 0.57 & 0.34 & 36.15 & 18.53 & 1.9 & 0.13 & 6.58 & 3.65 & 0.61 & 0.31 & 8.87 & 11.11 & 0.05 & 0.01 \\
15 & 3.62541$^*$ & 6.07 & 3.43 & 1.11 & 0.45 & 43.43 & 24.48 & 1.63 & 0.16 & 8.29 & 3.7 & 1.15 & 0.41 & 17.39 & 12.37 & 0.06 & 0.02 \\
20 & 2.971$^*$ & 7.88 & 3.68 & 1.49 & 0.63 & 51.39 & 20.4 & 1.27 & 0.09 & 12.01 & 4.0 & 1.61 & 0.59 & 26.03 & 15.3 & 0.07 & 0.02 \\
25 & 2.60929$^*$ & 8.98 & 4.14 & 2.06 & 0.83 & 38.42 & 10.9 & 1.09 & 0.09 & 14.03 & 5.16 & 1.69 & 0.84 & 26.57 & 11.61 & 0.08 & 0.02 \\
\hline
\multicolumn{2}{|c|}{Mean:} & \textbf{4.27} & & \textbf{0.81} & & \textbf{47.3} & & \textbf{1.35} & & \textbf{6.06} & & \textbf{0.78} & & \textbf{11.54} & & \textbf{0.05} & \\ \hline
\end{tabular}
}

\bigskip

\caption{Clustering details with Sensorless Drive Diagnosis (normalized)}
\label{TabDetailsD12}
\resizebox{\linewidth}{!}{
\begin{tabular}{|l|l|lllll|l|l|ll|l|ll|lllll|ll|}
\hline
\multicolumn{1}{|c|}{\multirow{2}{*}{$k$}} & \multicolumn{1}{c|}{\multirow{2}{*}{$n_{exec}$}} & \multicolumn{5}{c|}{Big-means} & \multicolumn{1}{c|}{IK-means} & \multicolumn{1}{c|}{BDCSM} & \multicolumn{2}{c|}{Minibatch K-means} & \multicolumn{1}{c|}{K-means++} & \multicolumn{2}{c|}{CURE} & \multicolumn{5}{c|}{CluDataSE} & \multicolumn{2}{c|}{LW-coreset} \\ \cline{3-21}
\multicolumn{1}{|c|}{} & \multicolumn{1}{c|}{} & \multicolumn{1}{c|}{$s$} & \multicolumn{1}{c|}{$n_{s}$} & \multicolumn{1}{c|}{$T_1$} & \multicolumn{1}{c|}{$T_2$} & \multicolumn{1}{c|}{$n_{d}$} & \multicolumn{1}{c|}{$n_{d}$} & \multicolumn{1}{c|}{$n_{d}$} & \multicolumn{1}{c|}{$n_{s}$} & \multicolumn{1}{c|}{$n_{d}$} & \multicolumn{1}{c|}{$n_{d}$} & \multicolumn{1}{c|}{$s$} & \multicolumn{1}{c|}{$n_{d}$} & \multicolumn{1}{c|}{$s$} & \multicolumn{1}{c|}{$eps$} & \multicolumn{1}{c|}{$min\_pts$} & \multicolumn{1}{c|}{$sf$} & \multicolumn{1}{c|}{$n_{d}$} & \multicolumn{1}{c|}{$s$} & \multicolumn{1}{c|}{$n_{d}$} \\
\hline
2 & 40 & 3500 & 364 & 0.16 & 0.14 & 1.5E+07 & 5.6E+05 & 1.1E+06 & 36 & 2.5E+05 & 1.2E+06 & 3500 & 8.7E+06 & 3500 & 0.12 & 16 & 0.5 & 1.3E+07 & 3500 & 3.0E+05 \\
3 & 40 & 3500 & 383 & 0.13 & 0.17 & 2.1E+07 & 9.3E+05 & 2.7E+06 & 34 & 3.6E+05 & 3.0E+06 & 3500 & 8.8E+06 & 3500 & 0.12 & 16 & 0.5 & 1.5E+07 & 3500 & 4.2E+05 \\
5 & 40 & 3500 & 169 & 0.01 & 0.29 & 2.7E+07 & 1.6E+06 & 5.3E+06 & 41 & 7.2E+05 & 6.7E+06 & 3500 & 9.4E+06 & 3500 & 0.12 & 16 & 0.5 & 1.6E+07 & 3500 & 6.9E+05 \\
10 & 40 & 3500 & 160 & 0.23 & 0.07 & 3.4E+07 & 3.7E+06 & 1.4E+07 & 44 & 1.5E+06 & 1.7E+07 & 3500 & 1.2E+07 & 3500 & 0.12 & 16 & 0.5 & 2.8E+07 & 3500 & 1.4E+06 \\
15 & 40 & 3500 & 78 & 0.26 & 0.04 & 3.6E+07 & 6.2E+06 & 2.3E+07 & 46 & 2.4E+06 & 3.5E+07 & 3500 & 1.1E+07 & 3500 & 0.12 & 16 & 0.5 & 4.5E+07 & 3500 & 2.2E+06 \\
20 & 40 & 3500 & 37 & 0.1 & 0.2 & 3.6E+07 & 8.8E+06 & 2.9E+07 & 44 & 3.1E+06 & 4.6E+07 & 3500 & 9.9E+06 & 3500 & 0.12 & 16 & 0.5 & 6.0E+07 & 3500 & 2.8E+06 \\
25 & 40 & 3500 & 22 & 0.1 & 0.2 & 3.5E+07 & 1.2E+07 & 3.6E+07 & 40 & 3.5E+06 & 6.3E+07 & 3500 & 9.2E+06 & 3500 & 0.12 & 16 & 0.5 & 6.8E+07 & 3500 & 3.5E+06 \\
\hline
\end{tabular}
}

\end{table}

\newpage


\newpage

\subsection{Online News Popularity}
Dimensions: $m$ = 39644, $n$ = 58.
\par
Description: this dataset summarizes a heterogeneous set of features about articles published by Mashable in a period of two years for predicting the number of shares in social networks (popularity).

\begin{table}[!htbp]
\centering

\caption{Summary of the results with Online News Popularity ($\times10^{14}$)}
\label{TabResultsD13}
\small
\resizebox{!}{\tableheight}{
\begin{tabular}{|l|l|llll|llll|llll|llll|}
\hline
\multicolumn{1}{|c|}{\multirow{3}{*}{$k$}} & \multicolumn{1}{c|}{\multirow{3}{*}{$f^*$}} & \multicolumn{4}{c|}{Big-means} & \multicolumn{4}{c|}{IK-means} & \multicolumn{4}{c|}{BDCSM} & \multicolumn{4}{c|}{Minibatch K-means} \\ \cline{3-18}
\multicolumn{1}{|c|}{} & \multicolumn{1}{c|}{} & \multicolumn{2}{c|}{$\varepsilon$} & \multicolumn{2}{c|}{$t$} & \multicolumn{2}{c|}{$\varepsilon$} & \multicolumn{2}{c|}{$t$} & \multicolumn{2}{c|}{$\varepsilon$} & \multicolumn{2}{c|}{$t$} & \multicolumn{2}{c|}{$\varepsilon$} & \multicolumn{2}{c|}{$t$} \\ \cline{3-18}
\multicolumn{1}{|c|}{} & \multicolumn{1}{c|}{} & \multicolumn{1}{c|}{med} & \multicolumn{1}{c|}{std} & \multicolumn{1}{c|}{med} & \multicolumn{1}{c|}{std} & \multicolumn{1}{c|}{med} & \multicolumn{1}{c|}{std} & \multicolumn{1}{c|}{med} & \multicolumn{1}{c|}{std} & \multicolumn{1}{c|}{med} & \multicolumn{1}{c|}{std} & \multicolumn{1}{c|}{med} & \multicolumn{1}{c|}{std} & \multicolumn{1}{c|}{med} & \multicolumn{1}{c|}{std} & \multicolumn{1}{c|}{med} & \multicolumn{1}{c|}{std} \\ \hline
2 & 9.53913 & 0.02 & 0.01 & 0.36 & 0.17 & 54.81 & 50.35 & 0.08 & 0.01 & 0.0 & 0.0 & 0.01 & 0.0 & 0.0 & 0.34 & 0.11 & 0.03 \\
3 & 5.91077 & 0.05 & 0.04 & 0.36 & 0.14 & 27.82 & 53.54 & 0.08 & 0.0 & 0.01 & 29.47 & 0.06 & 0.02 & 0.13 & 0.77 & 0.12 & 0.04 \\
5 & 3.09885 & 0.08 & 0.02 & 0.4 & 0.21 & 57.38 & 42.11 & 0.19 & 0.04 & 53.35 & 38.44 & 0.11 & 0.04 & 0.87 & 7.9 & 0.13 & 0.03 \\
10 & 1.17247 & 1.07 & 1.28 & 0.56 & 0.15 & 109.15 & 62.97 & 0.26 & 0.08 & 45.13 & 56.84 & 0.47 & 0.18 & 38.73 & 22.55 & 0.2 & 0.06 \\
15 & 0.77637 & 2.4 & 3.95 & 0.59 & 0.2 & 147.05 & 59.73 & 0.39 & 0.09 & 46.83 & 32.05 & 0.98 & 0.21 & 19.9 & 27.7 & 0.19 & 0.04 \\
20 & 0.59809 & 3.04 & 1.24 & 0.57 & 0.18 & 147.28 & 50.87 & 0.51 & 0.11 & 44.99 & 9.32 & 1.86 & 0.34 & 23.75 & 6.54 & 0.23 & 0.07 \\
25 & 0.49616 & 3.88 & 2.81 & 0.63 & 0.33 & 127.89 & 63.92 & 0.64 & 0.05 & 46.05 & 13.31 & 2.32 & 0.51 & 32.08 & 3.87 & 0.28 & 0.04 \\
\hline
\multicolumn{2}{|c|}{Mean:} & \textbf{1.51} & & \textbf{0.49} & & \textbf{95.91} & & \textbf{0.31} & & \textbf{33.77} & & \textbf{0.83} & & \textbf{16.5} & & \textbf{0.18} & \\ \hline
\end{tabular}
}

\medskip

\small
\resizebox{!}{\tableheight}{
\begin{tabular}{|l|l|llll|llll|llll|llll|}
\hline
\multicolumn{1}{|c|}{\multirow{3}{*}{$k$}} & \multicolumn{1}{c|}{\multirow{3}{*}{$f^*$}} & \multicolumn{4}{c|}{K-means++} & \multicolumn{4}{c|}{CURE} & \multicolumn{4}{c|}{CluDataSE} & \multicolumn{4}{c|}{LW-coreset} \\ \cline{3-18}
\multicolumn{1}{|c|}{} & \multicolumn{1}{c|}{} & \multicolumn{2}{c|}{$\varepsilon$} & \multicolumn{2}{c|}{$t$} & \multicolumn{2}{c|}{$\varepsilon$} & \multicolumn{2}{c|}{$t$} & \multicolumn{2}{c|}{$\varepsilon$} & \multicolumn{2}{c|}{$t$} & \multicolumn{2}{c|}{$\varepsilon$} & \multicolumn{2}{c|}{$t$} \\ \cline{3-18}
\multicolumn{1}{|c|}{} & \multicolumn{1}{c|}{} & \multicolumn{1}{c|}{med} & \multicolumn{1}{c|}{std} & \multicolumn{1}{c|}{med} & \multicolumn{1}{c|}{std} & \multicolumn{1}{c|}{med} & \multicolumn{1}{c|}{std} & \multicolumn{1}{c|}{med} & \multicolumn{1}{c|}{std} & \multicolumn{1}{c|}{med} & \multicolumn{1}{c|}{std} & \multicolumn{1}{c|}{med} & \multicolumn{1}{c|}{std} & \multicolumn{1}{c|}{med} & \multicolumn{1}{c|}{std} & \multicolumn{1}{c|}{med} & \multicolumn{1}{c|}{std} \\ \hline
2 & 9.53913 & -0.0 & 0.0 & 0.02 & 0.02 & 2.03 & 0.2 & 5.92 & 0.43 & -0.0 & 0.0 & 0.15 & 0.01 & 0.01 & 54.01 & 0.03 & 0.0 \\
3 & 5.91077 & 0.0 & 12.85 & 0.16 & 0.06 & 28.56 & 11.9 & 5.87 & 0.44 & 0.0 & 12.75 & 0.33 & 0.08 & 34.61 & 19.8 & 0.03 & 0.02 \\
5 & 3.09885 & 12.07 & 15.65 & 0.15 & 0.17 & 32.96 & 20.18 & 5.08 & 0.33 & 12.07 & 9.38 & 0.52 & 0.1 & 23.25 & 30.06 & 0.06 & 0.03 \\
10 & 1.17247 & 3.69 & 4.73 & 0.57 & 0.49 & 153.44 & 37.58 & 6.12 & 0.48 & 49.41 & 23.4 & 1.27 & 0.65 & 16.23 & 12.92 & 0.15 & 0.06 \\
15 & 0.77637 & 9.9 & 5.16 & 1.07 & 0.42 & 213.96 & 62.36 & 7.93 & 0.62 & 14.63 & 2.82 & 2.7 & 0.89 & 14.38 & 11.13 & 0.21 & 0.12 \\
20 & 0.59809 & 13.25 & 6.25 & 1.51 & 0.29 & 194.3 & 79.4 & 9.21 & 0.52 & 20.87 & 2.65 & 3.04 & 1.15 & 17.17 & 9.72 & 0.38 & 0.16 \\
25 & 0.49616 & 8.03 & 7.49 & 2.05 & 0.86 & 339.74 & 78.75 & 6.95 & 0.29 & 27.01 & 2.38 & 6.32 & 2.31 & 12.01 & 6.97 & 0.4 & 0.21 \\
\hline
\multicolumn{2}{|c|}{Mean:} & \textbf{6.71} & & \textbf{0.79} & & \textbf{137.86} & & \textbf{6.73} & & \textbf{17.71} & & \textbf{2.05} & & \textbf{16.81} & & \textbf{0.18} & \\ \hline
\end{tabular}
}

\bigskip

\caption{Clustering details with Online News Popularity}
\label{TabDetailsD13}
\resizebox{\linewidth}{!}{
\begin{tabular}{|l|l|lllll|l|l|ll|l|ll|lllll|ll|}
\hline
\multicolumn{1}{|c|}{\multirow{2}{*}{$k$}} & \multicolumn{1}{c|}{\multirow{2}{*}{$n_{exec}$}} & \multicolumn{5}{c|}{Big-means} & \multicolumn{1}{c|}{IK-means} & \multicolumn{1}{c|}{BDCSM} & \multicolumn{2}{c|}{Minibatch K-means} & \multicolumn{1}{c|}{K-means++} & \multicolumn{2}{c|}{CURE} & \multicolumn{5}{c|}{CluDataSE} & \multicolumn{2}{c|}{LW-coreset} \\ \cline{3-21}
\multicolumn{1}{|c|}{} & \multicolumn{1}{c|}{} & \multicolumn{1}{c|}{$s$} & \multicolumn{1}{c|}{$n_{s}$} & \multicolumn{1}{c|}{$T_1$} & \multicolumn{1}{c|}{$T_2$} & \multicolumn{1}{c|}{$n_{d}$} & \multicolumn{1}{c|}{$n_{d}$} & \multicolumn{1}{c|}{$n_{d}$} & \multicolumn{1}{c|}{$n_{s}$} & \multicolumn{1}{c|}{$n_{d}$} & \multicolumn{1}{c|}{$n_{d}$} & \multicolumn{1}{c|}{$s$} & \multicolumn{1}{c|}{$n_{d}$} & \multicolumn{1}{c|}{$s$} & \multicolumn{1}{c|}{$eps$} & \multicolumn{1}{c|}{$min\_pts$} & \multicolumn{1}{c|}{$sf$} & \multicolumn{1}{c|}{$n_{d}$} & \multicolumn{1}{c|}{$s$} & \multicolumn{1}{c|}{$n_{d}$} \\
\hline
2 & 20 & 10000 & 617 & 0.63 & 0.07 & 5.0E+07 & 3.2E+05 & 4.8E+05 & 19 & 3.8E+05 & 3.2E+05 & 10000 & 6.4E+07 & 10000 & 10000.0 & 16 & 0.5 & 1.0E+08 & 10000 & 2.2E+05 \\
3 & 20 & 10000 & 236 & 0.14 & 0.56 & 6.4E+07 & 4.8E+05 & 2.0E+06 & 18 & 5.6E+05 & 3.9E+06 & 10000 & 5.7E+07 & 10000 & 10000.0 & 16 & 0.5 & 1.0E+08 & 10000 & 4.1E+05 \\
5 & 20 & 10000 & 186 & 0.47 & 0.23 & 6.7E+07 & 7.9E+05 & 4.1E+06 & 21 & 1.0E+06 & 3.1E+06 & 10000 & 4.8E+07 & 10000 & 10000.0 & 16 & 0.5 & 1.1E+08 & 10000 & 1.2E+06 \\
10 & 20 & 10000 & 87 & 0.49 & 0.21 & 7.1E+07 & 1.6E+06 & 1.6E+07 & 30 & 3.0E+06 & 1.4E+07 & 10000 & 4.3E+07 & 10000 & 10000.0 & 16 & 0.5 & 1.2E+08 & 10000 & 3.8E+06 \\
15 & 20 & 10000 & 14 & 0.05 & 0.65 & 6.2E+07 & 2.4E+06 & 3.5E+07 & 22 & 3.3E+06 & 2.8E+07 & 10000 & 4.3E+07 & 10000 & 10000.0 & 16 & 0.5 & 1.7E+08 & 10000 & 5.8E+06 \\
20 & 20 & 10000 & 18 & 0.63 & 0.07 & 8.2E+07 & 3.2E+06 & 6.9E+07 & 24 & 4.8E+06 & 3.6E+07 & 10000 & 4.3E+07 & 10000 & 10000.0 & 16 & 0.5 & 1.7E+08 & 10000 & 7.7E+06 \\
25 & 20 & 10000 & 12 & 0.65 & 0.05 & 8.8E+07 & 4.0E+06 & 8.9E+07 & 25 & 6.2E+06 & 5.3E+07 & 10000 & 4.2E+07 & 10000 & 10000.0 & 16 & 0.5 & 2.6E+08 & 10000 & 1.1E+07 \\
\hline
\end{tabular}
}

\end{table}

\newpage


\subsection{Gas Sensor Array Drift}
Dimensions: $m$ = 13910, $n$ = 128.
\par
Description: this data set contains measurements from chemical sensors utilized in simulations for drift compensation in a discrimination task of different gases at various levels of concentrations.

\begin{table}[!htbp]
\centering

\caption{Summary of the results with Gas Sensor Array Drift ($\times10^{13}$)}
\label{TabResultsD14}
\small
\resizebox{!}{\tableheight}{
\begin{tabular}{|l|l|llll|llll|llll|llll|}
\hline
\multicolumn{1}{|c|}{\multirow{3}{*}{$k$}} & \multicolumn{1}{c|}{\multirow{3}{*}{$f^*$}} & \multicolumn{4}{c|}{Big-means} & \multicolumn{4}{c|}{IK-means} & \multicolumn{4}{c|}{BDCSM} & \multicolumn{4}{c|}{Minibatch K-means} \\ \cline{3-18}
\multicolumn{1}{|c|}{} & \multicolumn{1}{c|}{} & \multicolumn{2}{c|}{$\varepsilon$} & \multicolumn{2}{c|}{$t$} & \multicolumn{2}{c|}{$\varepsilon$} & \multicolumn{2}{c|}{$t$} & \multicolumn{2}{c|}{$\varepsilon$} & \multicolumn{2}{c|}{$t$} & \multicolumn{2}{c|}{$\varepsilon$} & \multicolumn{2}{c|}{$t$} \\ \cline{3-18}
\multicolumn{1}{|c|}{} & \multicolumn{1}{c|}{} & \multicolumn{1}{c|}{med} & \multicolumn{1}{c|}{std} & \multicolumn{1}{c|}{med} & \multicolumn{1}{c|}{std} & \multicolumn{1}{c|}{med} & \multicolumn{1}{c|}{std} & \multicolumn{1}{c|}{med} & \multicolumn{1}{c|}{std} & \multicolumn{1}{c|}{med} & \multicolumn{1}{c|}{std} & \multicolumn{1}{c|}{med} & \multicolumn{1}{c|}{std} & \multicolumn{1}{c|}{med} & \multicolumn{1}{c|}{std} & \multicolumn{1}{c|}{med} & \multicolumn{1}{c|}{std} \\ \hline
2 & 7.91186 & 0.12 & 0.08 & 0.68 & 0.57 & 14.06 & 18.21 & 0.03 & 0.0 & 0.01 & 0.04 & 0.03 & 0.01 & 0.17 & 0.21 & 0.13 & 0.04 \\
3 & 5.02412 & 0.15 & 0.1 & 0.7 & 0.55 & 42.42 & 29.16 & 0.04 & 0.0 & 0.02 & 0.07 & 0.06 & 0.02 & 0.97 & 2.93 & 0.14 & 0.05 \\
5 & 3.22394 & 0.22 & 3.39 & 1.27 & 0.5 & 21.81 & 33.61 & 0.07 & 0.0 & 8.15 & 0.39 & 0.13 & 0.03 & 8.45 & 4.99 & 0.11 & 0.03 \\
10 & 1.65524 & 0.43 & 1.5 & 1.76 & 0.54 & 77.48 & 41.68 & 0.14 & 0.0 & 44.3 & 16.52 & 0.39 & 0.16 & 30.65 & 13.92 & 0.17 & 0.05 \\
15 & 1.13801 & -0.08 & 1.36 & 1.29 & 0.55 & 101.91 & 45.96 & 0.21 & 0.0 & 29.22 & 24.59 & 0.66 & 0.31 & 8.67 & 11.99 & 0.47 & 0.2 \\
20 & 0.87916 & 1.46 & 1.01 & 1.62 & 0.34 & 135.38 & 51.73 & 0.28 & 0.03 & 46.2 & 9.82 & 1.07 & 0.43 & 13.95 & 5.77 & 0.6 & 0.23 \\
25 & 0.72274 & 2.31 & 1.36 & 1.81 & 0.48 & 151.15 & 41.8 & 0.33 & 0.08 & 54.01 & 13.61 & 1.95 & 0.78 & 14.12 & 5.3 & 0.23 & 0.05 \\
\hline
\multicolumn{2}{|c|}{Mean:} & \textbf{0.66} & & \textbf{1.3} & & \textbf{77.74} & & \textbf{0.16} & & \textbf{25.99} & & \textbf{0.61} & & \textbf{11.0} & & \textbf{0.27} & \\ \hline
\end{tabular}
}

\medskip

\small
\resizebox{!}{\tableheight}{
\begin{tabular}{|l|l|llll|llll|llll|llll|}
\hline
\multicolumn{1}{|c|}{\multirow{3}{*}{$k$}} & \multicolumn{1}{c|}{\multirow{3}{*}{$f^*$}} & \multicolumn{4}{c|}{K-means++} & \multicolumn{4}{c|}{CURE} & \multicolumn{4}{c|}{CluDataSE} & \multicolumn{4}{c|}{LW-coreset} \\ \cline{3-18}
\multicolumn{1}{|c|}{} & \multicolumn{1}{c|}{} & \multicolumn{2}{c|}{$\varepsilon$} & \multicolumn{2}{c|}{$t$} & \multicolumn{2}{c|}{$\varepsilon$} & \multicolumn{2}{c|}{$t$} & \multicolumn{2}{c|}{$\varepsilon$} & \multicolumn{2}{c|}{$t$} & \multicolumn{2}{c|}{$\varepsilon$} & \multicolumn{2}{c|}{$t$} \\ \cline{3-18}
\multicolumn{1}{|c|}{} & \multicolumn{1}{c|}{} & \multicolumn{1}{c|}{med} & \multicolumn{1}{c|}{std} & \multicolumn{1}{c|}{med} & \multicolumn{1}{c|}{std} & \multicolumn{1}{c|}{med} & \multicolumn{1}{c|}{std} & \multicolumn{1}{c|}{med} & \multicolumn{1}{c|}{std} & \multicolumn{1}{c|}{med} & \multicolumn{1}{c|}{std} & \multicolumn{1}{c|}{med} & \multicolumn{1}{c|}{std} & \multicolumn{1}{c|}{med} & \multicolumn{1}{c|}{std} & \multicolumn{1}{c|}{med} & \multicolumn{1}{c|}{std} \\ \hline
2 & 7.91186 & -0.0 & 0.0 & 0.07 & 0.02 & 10.17 & 45.46 & 5.83 & 0.41 & -0.0 & 0.0 & 0.2 & 0.01 & 0.02 & 0.02 & 0.06 & 0.01 \\
3 & 5.02412 & -0.0 & 12.28 & 0.08 & 0.03 & 15.92 & 15.71 & 5.81 & 0.52 & -0.0 & 0.0 & 0.28 & 0.04 & 0.05 & 16.16 & 0.07 & 0.02 \\
5 & 3.22394 & 3.46 & 5.24 & 0.19 & 0.1 & 34.32 & 7.78 & 5.6 & 0.17 & 8.1 & 0.59 & 0.35 & 0.09 & 6.99 & 6.4 & 0.13 & 0.03 \\
10 & 1.65524 & 4.02 & 10.5 & 0.37 & 0.15 & 74.31 & 14.14 & 6.96 & 0.28 & 32.31 & 13.99 & 0.81 & 0.24 & 6.44 & 8.46 & 0.29 & 0.13 \\
15 & 1.13801 & 5.25 & 2.78 & 0.79 & 0.26 & 113.63 & 15.68 & 9.26 & 3.62 & 20.78 & 7.25 & 1.26 & 0.43 & 9.6 & 5.08 & 0.48 & 0.17 \\
20 & 0.87916 & 6.25 & 5.75 & 1.13 & 0.48 & 148.13 & 21.46 & 14.68 & 3.73 & 21.78 & 8.56 & 2.18 & 0.46 & 14.85 & 7.76 & 0.67 & 0.25 \\
25 & 0.72274 & 5.55 & 2.67 & 1.21 & 0.37 & 179.99 & 17.23 & 8.71 & 1.17 & 29.99 & 6.8 & 3.27 & 0.74 & 18.87 & 8.68 & 0.8 & 0.34 \\
\hline
\multicolumn{2}{|c|}{Mean:} & \textbf{3.5} & & \textbf{0.55} & & \textbf{82.36} & & \textbf{8.12} & & \textbf{16.14} & & \textbf{1.19} & & \textbf{8.12} & & \textbf{0.36} & \\ \hline
\end{tabular}
}

\bigskip

\caption{Clustering details with Gas Sensor Array Drift}
\label{TabDetailsD14}
\resizebox{\linewidth}{!}{
\begin{tabular}{|l|l|lllll|l|l|ll|l|ll|lllll|ll|}
\hline
\multicolumn{1}{|c|}{\multirow{2}{*}{$k$}} & \multicolumn{1}{c|}{\multirow{2}{*}{$n_{exec}$}} & \multicolumn{5}{c|}{Big-means} & \multicolumn{1}{c|}{IK-means} & \multicolumn{1}{c|}{BDCSM} & \multicolumn{2}{c|}{Minibatch K-means} & \multicolumn{1}{c|}{K-means++} & \multicolumn{2}{c|}{CURE} & \multicolumn{5}{c|}{CluDataSE} & \multicolumn{2}{c|}{LW-coreset} \\ \cline{3-21}
\multicolumn{1}{|c|}{} & \multicolumn{1}{c|}{} & \multicolumn{1}{c|}{$s$} & \multicolumn{1}{c|}{$n_{s}$} & \multicolumn{1}{c|}{$T_1$} & \multicolumn{1}{c|}{$T_2$} & \multicolumn{1}{c|}{$n_{d}$} & \multicolumn{1}{c|}{$n_{d}$} & \multicolumn{1}{c|}{$n_{d}$} & \multicolumn{1}{c|}{$n_{s}$} & \multicolumn{1}{c|}{$n_{d}$} & \multicolumn{1}{c|}{$n_{d}$} & \multicolumn{1}{c|}{$s$} & \multicolumn{1}{c|}{$n_{d}$} & \multicolumn{1}{c|}{$s$} & \multicolumn{1}{c|}{$eps$} & \multicolumn{1}{c|}{$min\_pts$} & \multicolumn{1}{c|}{$sf$} & \multicolumn{1}{c|}{$n_{d}$} & \multicolumn{1}{c|}{$s$} & \multicolumn{1}{c|}{$n_{d}$} \\
\hline
2 & 30 & 9000 & 430 & 1.0 & 1.0 & 9.4E+07 & 1.1E+05 & 2.8E+05 & 20 & 3.7E+05 & 4.5E+05 & 9000 & 4.2E+07 & 9000 & 3000.0 & 16 & 0.5 & 8.1E+07 & 9000 & 2.7E+05 \\
3 & 30 & 9000 & 350 & 0.87 & 1.13 & 9.7E+07 & 1.7E+05 & 6.2E+05 & 18 & 4.7E+05 & 6.3E+05 & 9000 & 4.2E+07 & 9000 & 3000.0 & 16 & 0.5 & 8.2E+07 & 9000 & 3.9E+05 \\
5 & 30 & 9000 & 346 & 0.87 & 1.13 & 1.0E+08 & 2.8E+05 & 1.4E+06 & 22 & 9.7E+05 & 1.8E+06 & 9000 & 4.3E+07 & 9000 & 3000.0 & 16 & 0.5 & 8.3E+07 & 9000 & 9.7E+05 \\
10 & 30 & 9000 & 210 & 1.4 & 0.6 & 1.2E+08 & 5.6E+05 & 4.1E+06 & 20 & 1.8E+06 & 3.3E+06 & 9000 & 4.6E+07 & 9000 & 3000.0 & 16 & 0.5 & 8.7E+07 & 9000 & 2.6E+06 \\
15 & 30 & 9000 & 80 & 1.87 & 0.13 & 1.3E+08 & 8.3E+05 & 6.8E+06 & 23 & 3.1E+06 & 6.4E+06 & 9000 & 5.1E+07 & 9000 & 3000.0 & 16 & 0.5 & 9.0E+07 & 9000 & 4.0E+06 \\
20 & 30 & 9000 & 43 & 0.2 & 1.8 & 1.0E+08 & 1.1E+06 & 1.1E+07 & 25 & 4.5E+06 & 1.0E+07 & 9000 & 5.1E+07 & 9000 & 3000.0 & 16 & 0.5 & 9.9E+07 & 9000 & 5.8E+06 \\
25 & 30 & 9000 & 22 & 0.8 & 1.2 & 8.5E+07 & 1.4E+06 & 2.0E+07 & 27 & 6.1E+06 & 1.1E+07 & 9000 & 4.7E+07 & 9000 & 3000.0 & 16 & 0.5 & 1.1E+08 & 9000 & 7.5E+06 \\
\hline
\end{tabular}
}

\end{table}

\newpage


\newpage

\subsection{3D Road Network}
Dimensions: $m$ = 434874, $n$ = 3.
\par
Description: 3D road network from Denmark with highly accurate elevation information which contains longitude, latitude and altitude for each road segment or edge in the graph. Usually this data set used in eco-routing and fuel/Co2-estimation routing algorithms.

\begin{table}[!htbp]
\centering

\caption{Summary of the results with 3D Road Network ($\times10^{6}$)}
\label{TabResultsD15}
\small
\resizebox{!}{\tableheight}{
\begin{tabular}{|l|l|llll|llll|llll|llll|}
\hline
\multicolumn{1}{|c|}{\multirow{3}{*}{$k$}} & \multicolumn{1}{c|}{\multirow{3}{*}{$f^*$}} & \multicolumn{4}{c|}{Big-means} & \multicolumn{4}{c|}{IK-means} & \multicolumn{4}{c|}{BDCSM} & \multicolumn{4}{c|}{Minibatch K-means} \\ \cline{3-18}
\multicolumn{1}{|c|}{} & \multicolumn{1}{c|}{} & \multicolumn{2}{c|}{$\varepsilon$} & \multicolumn{2}{c|}{$t$} & \multicolumn{2}{c|}{$\varepsilon$} & \multicolumn{2}{c|}{$t$} & \multicolumn{2}{c|}{$\varepsilon$} & \multicolumn{2}{c|}{$t$} & \multicolumn{2}{c|}{$\varepsilon$} & \multicolumn{2}{c|}{$t$} \\ \cline{3-18}
\multicolumn{1}{|c|}{} & \multicolumn{1}{c|}{} & \multicolumn{1}{c|}{med} & \multicolumn{1}{c|}{std} & \multicolumn{1}{c|}{med} & \multicolumn{1}{c|}{std} & \multicolumn{1}{c|}{med} & \multicolumn{1}{c|}{std} & \multicolumn{1}{c|}{med} & \multicolumn{1}{c|}{std} & \multicolumn{1}{c|}{med} & \multicolumn{1}{c|}{std} & \multicolumn{1}{c|}{med} & \multicolumn{1}{c|}{std} & \multicolumn{1}{c|}{med} & \multicolumn{1}{c|}{std} & \multicolumn{1}{c|}{med} & \multicolumn{1}{c|}{std} \\ \hline
2 & 49.13298 & 0.01 & 0.01 & 0.24 & 0.13 & 1.9 & 2.16 & 0.38 & 0.06 & 0.0 & 0.0 & 0.06 & 0.0 & 0.18 & 0.42 & 0.37 & 0.13 \\
3 & 22.77818 & 0.01 & 0.01 & 0.28 & 0.13 & 13.59 & 27.4 & 0.54 & 0.01 & 0.0 & 74.5 & 0.1 & 0.01 & 0.8 & 0.85 & 0.41 & 0.11 \\
5 & 8.82574 & 0.03 & 0.02 & 0.34 & 0.13 & 33.32 & 55.9 & 0.84 & 0.12 & 77.07 & 46.21 & 0.24 & 0.03 & 1.3 & 2.54 & 0.49 & 0.1 \\
10 & 2.56661 & 0.19 & 0.22 & 0.43 & 0.18 & 111.62 & 127.83 & 2.11 & 0.4 & 52.14 & 53.8 & 1.65 & 0.26 & 2.02 & 2.47 & 0.78 & 0.25 \\
15 & 1.27069 & 0.23 & 0.29 & 0.51 & 0.38 & 205.67 & 165.09 & 3.67 & 0.44 & 59.12 & 27.43 & 2.54 & 0.27 & 4.42 & 3.43 & 0.66 & 0.1 \\
20 & 0.80865 & 0.59 & 0.59 & 0.66 & 0.42 & 173.25 & 112.73 & 5.2 & 0.6 & 45.69 & 23.7 & 4.64 & 0.67 & 5.17 & 3.14 & 0.78 & 0.11 \\
25 & 0.59259 & 0.58 & 0.52 & 0.79 & 0.61 & 224.75 & 131.93 & 6.56 & 0.66 & 50.01 & 21.16 & 6.75 & 0.8 & 5.4 & 4.09 & 0.83 & 0.12 \\
\hline
\multicolumn{2}{|c|}{Mean:} & \textbf{0.24} & & \textbf{0.46} & & \textbf{109.16} & & \textbf{2.76} & & \textbf{40.57} & & \textbf{2.28} & & \textbf{2.76} & & \textbf{0.62} & \\ \hline
\end{tabular}
}

\medskip

\small
\resizebox{!}{\tableheight}{
\begin{tabular}{|l|l|llll|llll|llll|llll|}
\hline
\multicolumn{1}{|c|}{\multirow{3}{*}{$k$}} & \multicolumn{1}{c|}{\multirow{3}{*}{$f^*$}} & \multicolumn{4}{c|}{K-means++} & \multicolumn{4}{c|}{CURE} & \multicolumn{4}{c|}{CluDataSE} & \multicolumn{4}{c|}{LW-coreset} \\ \cline{3-18}
\multicolumn{1}{|c|}{} & \multicolumn{1}{c|}{} & \multicolumn{2}{c|}{$\varepsilon$} & \multicolumn{2}{c|}{$t$} & \multicolumn{2}{c|}{$\varepsilon$} & \multicolumn{2}{c|}{$t$} & \multicolumn{2}{c|}{$\varepsilon$} & \multicolumn{2}{c|}{$t$} & \multicolumn{2}{c|}{$\varepsilon$} & \multicolumn{2}{c|}{$t$} \\ \cline{3-18}
\multicolumn{1}{|c|}{} & \multicolumn{1}{c|}{} & \multicolumn{1}{c|}{med} & \multicolumn{1}{c|}{std} & \multicolumn{1}{c|}{med} & \multicolumn{1}{c|}{std} & \multicolumn{1}{c|}{med} & \multicolumn{1}{c|}{std} & \multicolumn{1}{c|}{med} & \multicolumn{1}{c|}{std} & \multicolumn{1}{c|}{med} & \multicolumn{1}{c|}{std} & \multicolumn{1}{c|}{med} & \multicolumn{1}{c|}{std} & \multicolumn{1}{c|}{med} & \multicolumn{1}{c|}{std} & \multicolumn{1}{c|}{med} & \multicolumn{1}{c|}{std} \\ \hline
2 & 49.13298 & 0.0 & 0.0 & 0.11 & 0.03 & 11.76 & 11.65 & 19.41 & 1.36 & 0.0 & 0.0 & 0.33 & 0.05 & 0.03 & 0.09 & 0.01 & 0.0 \\
3 & 22.77818 & 0.0 & 0.0 & 0.17 & 0.07 & 20.61 & 21.84 & 15.74 & 1.43 & 0.0 & 0.0 & 0.4 & 0.05 & 0.04 & 0.08 & 0.01 & 0.0 \\
5 & 8.82574 & 0.0 & 0.0 & 0.51 & 0.1 & 32.07 & 24.5 & 14.88 & 0.98 & 0.0 & 0.0 & 0.83 & 0.06 & 0.12 & 0.17 & 0.02 & 0.0 \\
10 & 2.56661 & 0.01 & 0.0 & 4.98 & 1.52 & 60.63 & 22.45 & 15.78 & 0.92 & 0.01 & 0.0 & 6.82 & 0.15 & 0.92 & 0.81 & 0.07 & 0.02 \\
15 & 1.27069 & 0.0 & 0.0 & 4.94 & 2.47 & 123.99 & 34.74 & 16.93 & 1.24 & 0.0 & 1.88 & 9.35 & 0.69 & 2.5 & 3.25 & 0.1 & 0.02 \\
20 & 0.80865 & 0.01 & 0.0 & 14.38 & 5.61 & 178.43 & 64.98 & 17.3 & 0.95 & 0.01 & 2.21 & 29.08 & 1.26 & 9.44 & 5.35 & 0.13 & 0.04 \\
25 & 0.59259 & 0.07 & 1.03 & 12.62 & 5.03 & 275.78 & 73.71 & 16.03 & 0.94 & 1.62 & 1.57 & 31.19 & 1.34 & 19.7 & 9.55 & 0.15 & 0.07 \\
\hline
\multicolumn{2}{|c|}{Mean:} & \textbf{0.01} & & \textbf{5.39} & & \textbf{100.47} & & \textbf{16.58} & & \textbf{0.23} & & \textbf{11.14} & & \textbf{4.68} & & \textbf{0.07} & \\ \hline
\end{tabular}
}

\bigskip

\caption{Clustering details with 3D Road Network}
\label{TabDetailsD15}
\resizebox{\linewidth}{!}{
\begin{tabular}{|l|l|lllll|l|l|ll|l|ll|lllll|ll|}
\hline
\multicolumn{1}{|c|}{\multirow{2}{*}{$k$}} & \multicolumn{1}{c|}{\multirow{2}{*}{$n_{exec}$}} & \multicolumn{5}{c|}{Big-means} & \multicolumn{1}{c|}{IK-means} & \multicolumn{1}{c|}{BDCSM} & \multicolumn{2}{c|}{Minibatch K-means} & \multicolumn{1}{c|}{K-means++} & \multicolumn{2}{c|}{CURE} & \multicolumn{5}{c|}{CluDataSE} & \multicolumn{2}{c|}{LW-coreset} \\ \cline{3-21}
\multicolumn{1}{|c|}{} & \multicolumn{1}{c|}{} & \multicolumn{1}{c|}{$s$} & \multicolumn{1}{c|}{$n_{s}$} & \multicolumn{1}{c|}{$T_1$} & \multicolumn{1}{c|}{$T_2$} & \multicolumn{1}{c|}{$n_{d}$} & \multicolumn{1}{c|}{$n_{d}$} & \multicolumn{1}{c|}{$n_{d}$} & \multicolumn{1}{c|}{$n_{s}$} & \multicolumn{1}{c|}{$n_{d}$} & \multicolumn{1}{c|}{$n_{d}$} & \multicolumn{1}{c|}{$s$} & \multicolumn{1}{c|}{$n_{d}$} & \multicolumn{1}{c|}{$s$} & \multicolumn{1}{c|}{$eps$} & \multicolumn{1}{c|}{$min\_pts$} & \multicolumn{1}{c|}{$sf$} & \multicolumn{1}{c|}{$n_{d}$} & \multicolumn{1}{c|}{$s$} & \multicolumn{1}{c|}{$n_{d}$} \\
\hline
2 & 40 & 100000 & 90 & 0.03 & 0.47 & 1.2E+08 & 3.5E+06 & 1.8E+07 & 27 & 5.4E+06 & 2.2E+07 & 10000 & 1.1E+08 & 10000 & 0.05 & 10 & 0.5 & 1.2E+08 & 10000 & 2.0E+06 \\
3 & 40 & 100000 & 99 & 0.47 & 0.03 & 1.8E+08 & 5.2E+06 & 4.3E+07 & 29 & 8.7E+06 & 4.2E+07 & 10000 & 1.1E+08 & 10000 & 0.05 & 10 & 0.5 & 1.5E+08 & 10000 & 2.9E+06 \\
5 & 40 & 100000 & 80 & 0.12 & 0.38 & 3.2E+08 & 8.7E+06 & 1.3E+08 & 32 & 1.6E+07 & 1.4E+08 & 10000 & 1.1E+08 & 10000 & 0.05 & 10 & 0.5 & 2.7E+08 & 10000 & 5.7E+06 \\
10 & 40 & 100000 & 10 & 0.15 & 0.35 & 4.5E+08 & 2.1E+07 & 1.2E+09 & 35 & 3.5E+07 & 1.8E+09 & 10000 & 1.1E+08 & 10000 & 0.05 & 8 & 0.5 & 2.5E+09 & 10000 & 1.9E+07 \\
15 & 40 & 100000 & 14 & 0.45 & 0.05 & 8.5E+08 & 3.4E+07 & 2.3E+09 & 34 & 5.0E+07 & 1.9E+09 & 10000 & 1.2E+08 & 10000 & 0.05 & 8 & 0.5 & 3.7E+09 & 10000 & 4.2E+07 \\
20 & 40 & 100000 & 7 & 0.33 & 0.17 & 9.7E+08 & 4.8E+07 & 4.8E+09 & 35 & 7.0E+07 & 6.0E+09 & 10000 & 1.2E+08 & 10000 & 0.05 & 8 & 0.5 & 1.2E+10 & 10000 & 5.5E+07 \\
25 & 40 & 100000 & 5 & 0.32 & 0.18 & 1.3E+09 & 5.8E+07 & 7.9E+09 & 32 & 8.0E+07 & 5.2E+09 & 10000 & 1.2E+08 & 10000 & 0.05 & 8 & 0.5 & 1.3E+10 & 10000 & 6.8E+07 \\
\hline
\end{tabular}
}

\end{table}

\newpage


\subsection{Skin Segmentation}
Dimensions: $m$ = 245057, $n$ = 3.
\par
Description: Skin and Nonskin dataset is generated using skin textures from face images of diversity of age, gender, and race people and constructed over B, G, R color space.

\begin{table}[!htbp]
\centering

\caption{Summary of the results with Skin Segmentation ($\times10^{9}$)}
\label{TabResultsD16}
\small
\resizebox{!}{\tableheight}{
\begin{tabular}{|l|l|llll|llll|llll|llll|}
\hline
\multicolumn{1}{|c|}{\multirow{3}{*}{$k$}} & \multicolumn{1}{c|}{\multirow{3}{*}{$f^*$}} & \multicolumn{4}{c|}{Big-means} & \multicolumn{4}{c|}{IK-means} & \multicolumn{4}{c|}{BDCSM} & \multicolumn{4}{c|}{Minibatch K-means} \\ \cline{3-18}
\multicolumn{1}{|c|}{} & \multicolumn{1}{c|}{} & \multicolumn{2}{c|}{$\varepsilon$} & \multicolumn{2}{c|}{$t$} & \multicolumn{2}{c|}{$\varepsilon$} & \multicolumn{2}{c|}{$t$} & \multicolumn{2}{c|}{$\varepsilon$} & \multicolumn{2}{c|}{$t$} & \multicolumn{2}{c|}{$\varepsilon$} & \multicolumn{2}{c|}{$t$} \\ \cline{3-18}
\multicolumn{1}{|c|}{} & \multicolumn{1}{c|}{} & \multicolumn{1}{c|}{med} & \multicolumn{1}{c|}{std} & \multicolumn{1}{c|}{med} & \multicolumn{1}{c|}{std} & \multicolumn{1}{c|}{med} & \multicolumn{1}{c|}{std} & \multicolumn{1}{c|}{med} & \multicolumn{1}{c|}{std} & \multicolumn{1}{c|}{med} & \multicolumn{1}{c|}{std} & \multicolumn{1}{c|}{med} & \multicolumn{1}{c|}{std} & \multicolumn{1}{c|}{med} & \multicolumn{1}{c|}{std} & \multicolumn{1}{c|}{med} & \multicolumn{1}{c|}{std} \\ \hline
2 & 1.32236 & 0.04 & 0.02 & 0.12 & 0.06 & 0.42 & 10.01 & 0.21 & 0.02 & -0.0 & 0.0 & 0.01 & 0.0 & 0.02 & 0.1 & 0.08 & 0.04 \\
3 & 0.89362 & 0.04 & 0.04 & 0.14 & 0.05 & 10.8 & 7.71 & 0.3 & 0.02 & 0.02 & 50.56 & 0.02 & 0.0 & 0.36 & 2.31 & 0.11 & 0.04 \\
5 & 0.50203 & 0.09 & 0.29 & 0.15 & 0.07 & 16.38 & 14.76 & 0.47 & 0.06 & 14.34 & 20.25 & 0.03 & 0.0 & 1.86 & 5.17 & 0.11 & 0.04 \\
10 & 0.25121 & 0.21 & 1.65 & 0.18 & 0.04 & 30.94 & 13.97 & 0.92 & 0.04 & 30.53 & 15.12 & 0.06 & 0.01 & 5.33 & 4.2 & 0.14 & 0.06 \\
15 & 0.16964 & 0.98 & 1.83 & 0.17 & 0.05 & 40.88 & 18.18 & 1.4 & 0.23 & 30.04 & 18.17 & 0.1 & 0.01 & 6.03 & 3.24 & 0.18 & 0.06 \\
20 & 0.12615 & 1.99 & 1.87 & 0.15 & 0.05 & 61.99 & 21.75 & 1.86 & 0.24 & 43.07 & 20.61 & 0.14 & 0.02 & 6.64 & 4.69 & 0.24 & 0.07 \\
25 & 0.10228 & 3.15 & 1.8 & 0.18 & 0.05 & 64.36 & 21.86 & 2.28 & 0.35 & 28.76 & 16.08 & 0.18 & 0.02 & 6.79 & 3.27 & 0.19 & 0.06 \\
\hline
\multicolumn{2}{|c|}{Mean:} & \textbf{0.93} & & \textbf{0.16} & & \textbf{32.25} & & \textbf{1.06} & & \textbf{20.97} & & \textbf{0.08} & & \textbf{3.86} & & \textbf{0.15} & \\ \hline
\end{tabular}
}

\medskip

\small
\resizebox{!}{\tableheight}{
\begin{tabular}{|l|l|llll|llll|llll|llll|}
\hline
\multicolumn{1}{|c|}{\multirow{3}{*}{$k$}} & \multicolumn{1}{c|}{\multirow{3}{*}{$f^*$}} & \multicolumn{4}{c|}{K-means++} & \multicolumn{4}{c|}{CURE} & \multicolumn{4}{c|}{CluDataSE} & \multicolumn{4}{c|}{LW-coreset} \\ \cline{3-18}
\multicolumn{1}{|c|}{} & \multicolumn{1}{c|}{} & \multicolumn{2}{c|}{$\varepsilon$} & \multicolumn{2}{c|}{$t$} & \multicolumn{2}{c|}{$\varepsilon$} & \multicolumn{2}{c|}{$t$} & \multicolumn{2}{c|}{$\varepsilon$} & \multicolumn{2}{c|}{$t$} & \multicolumn{2}{c|}{$\varepsilon$} & \multicolumn{2}{c|}{$t$} \\ \cline{3-18}
\multicolumn{1}{|c|}{} & \multicolumn{1}{c|}{} & \multicolumn{1}{c|}{med} & \multicolumn{1}{c|}{std} & \multicolumn{1}{c|}{med} & \multicolumn{1}{c|}{std} & \multicolumn{1}{c|}{med} & \multicolumn{1}{c|}{std} & \multicolumn{1}{c|}{med} & \multicolumn{1}{c|}{std} & \multicolumn{1}{c|}{med} & \multicolumn{1}{c|}{std} & \multicolumn{1}{c|}{med} & \multicolumn{1}{c|}{std} & \multicolumn{1}{c|}{med} & \multicolumn{1}{c|}{std} & \multicolumn{1}{c|}{med} & \multicolumn{1}{c|}{std} \\ \hline
2 & 1.32236 & -0.0 & 0.0 & 0.06 & 0.01 & 90.1 & 31.19 & 3.7 & 0.21 & -0.0 & 0.0 & 0.07 & 0.0 & 0.02 & 0.01 & 0.01 & 0.0 \\
3 & 0.89362 & -0.0 & 0.0 & 0.08 & 0.02 & 135.81 & 52.11 & 4.1 & 0.44 & -0.0 & 0.0 & 0.17 & 0.02 & 0.04 & 0.03 & 0.01 & 0.0 \\
5 & 0.50203 & 1.65 & 6.84 & 0.11 & 0.04 & 153.15 & 81.59 & 3.72 & 0.28 & 9.31 & 3.67 & 0.14 & 0.07 & 9.37 & 6.65 & 0.01 & 0.0 \\
10 & 0.25121 & 8.98 & 5.69 & 0.2 & 0.08 & 77.69 & 36.59 & 4.94 & 0.13 & 16.1 & 2.94 & 0.3 & 0.05 & 17.64 & 7.56 & 0.02 & 0.0 \\
15 & 0.16964 & 8.02 & 6.62 & 0.37 & 0.16 & 54.66 & 14.73 & 6.89 & 0.35 & 25.33 & 9.29 & 0.6 & 0.15 & 20.08 & 12.74 & 0.02 & 0.0 \\
20 & 0.12615 & 9.74 & 3.31 & 0.59 & 0.19 & 78.9 & 16.03 & 5.95 & 0.42 & 27.45 & 8.19 & 0.68 & 0.23 & 29.2 & 10.48 & 0.02 & 0.01 \\
25 & 0.10228 & 8.33 & 3.28 & 0.6 & 0.46 & 79.07 & 14.74 & 4.97 & 0.21 & 31.9 & 5.22 & 1.02 & 0.28 & 33.67 & 13.82 & 0.03 & 0.01 \\
\hline
\multicolumn{2}{|c|}{Mean:} & \textbf{5.24} & & \textbf{0.29} & & \textbf{95.62} & & \textbf{4.9} & & \textbf{15.73} & & \textbf{0.43} & & \textbf{15.72} & & \textbf{0.02} & \\ \hline
\end{tabular}
}

\bigskip

\caption{Clustering details with Skin Segmentation}
\label{TabDetailsD16}
\resizebox{\linewidth}{!}{
\begin{tabular}{|l|l|lllll|l|l|ll|l|ll|lllll|ll|}
\hline
\multicolumn{1}{|c|}{\multirow{2}{*}{$k$}} & \multicolumn{1}{c|}{\multirow{2}{*}{$n_{exec}$}} & \multicolumn{5}{c|}{Big-means} & \multicolumn{1}{c|}{IK-means} & \multicolumn{1}{c|}{BDCSM} & \multicolumn{2}{c|}{Minibatch K-means} & \multicolumn{1}{c|}{K-means++} & \multicolumn{2}{c|}{CURE} & \multicolumn{5}{c|}{CluDataSE} & \multicolumn{2}{c|}{LW-coreset} \\ \cline{3-21}
\multicolumn{1}{|c|}{} & \multicolumn{1}{c|}{} & \multicolumn{1}{c|}{$s$} & \multicolumn{1}{c|}{$n_{s}$} & \multicolumn{1}{c|}{$T_1$} & \multicolumn{1}{c|}{$T_2$} & \multicolumn{1}{c|}{$n_{d}$} & \multicolumn{1}{c|}{$n_{d}$} & \multicolumn{1}{c|}{$n_{d}$} & \multicolumn{1}{c|}{$n_{s}$} & \multicolumn{1}{c|}{$n_{d}$} & \multicolumn{1}{c|}{$n_{d}$} & \multicolumn{1}{c|}{$s$} & \multicolumn{1}{c|}{$n_{d}$} & \multicolumn{1}{c|}{$s$} & \multicolumn{1}{c|}{$eps$} & \multicolumn{1}{c|}{$min\_pts$} & \multicolumn{1}{c|}{$sf$} & \multicolumn{1}{c|}{$n_{d}$} & \multicolumn{1}{c|}{$s$} & \multicolumn{1}{c|}{$n_{d}$} \\
\hline
2 & 30 & 8000 & 119 & 0.05 & 0.15 & 1.1E+07 & 2.0E+06 & 5.4E+06 & 30 & 4.9E+05 & 6.6E+06 & 8000 & 4.7E+07 & 8000 & 2.0 & 16 & 0.5 & 6.9E+07 & 8000 & 1.1E+06 \\
3 & 30 & 8000 & 124 & 0.03 & 0.17 & 2.1E+07 & 2.9E+06 & 1.6E+07 & 47 & 1.1E+06 & 2.0E+07 & 8000 & 4.1E+07 & 8000 & 2.0 & 16 & 0.5 & 8.8E+07 & 8000 & 1.8E+06 \\
5 & 30 & 8000 & 138 & 0.15 & 0.05 & 3.3E+07 & 4.9E+06 & 2.4E+07 & 38 & 1.5E+06 & 2.8E+07 & 8000 & 3.8E+07 & 8000 & 2.0 & 16 & 0.5 & 8.8E+07 & 8000 & 2.6E+06 \\
10 & 30 & 8000 & 128 & 0.13 & 0.07 & 7.2E+07 & 9.8E+06 & 6.6E+07 & 48 & 3.9E+06 & 6.4E+07 & 8000 & 3.7E+07 & 8000 & 2.0 & 16 & 0.5 & 1.6E+08 & 8000 & 6.4E+06 \\
15 & 30 & 8000 & 88 & 0.15 & 0.05 & 1.2E+08 & 1.5E+07 & 1.3E+08 & 49 & 5.9E+06 & 1.3E+08 & 8000 & 4.0E+07 & 8000 & 2.0 & 16 & 0.5 & 2.8E+08 & 8000 & 9.3E+06 \\
20 & 30 & 8000 & 56 & 0.05 & 0.15 & 1.3E+08 & 2.0E+07 & 1.9E+08 & 60 & 9.5E+06 & 1.8E+08 & 8000 & 4.0E+07 & 8000 & 2.0 & 16 & 0.5 & 3.3E+08 & 8000 & 1.1E+07 \\
25 & 30 & 8000 & 62 & 0.16 & 0.04 & 1.6E+08 & 2.5E+07 & 2.3E+08 & 52 & 1.0E+07 & 2.3E+08 & 8000 & 3.9E+07 & 8000 & 2.0 & 16 & 0.5 & 4.8E+08 & 8000 & 1.6E+07 \\
\hline
\end{tabular}
}

\end{table}

\newpage


\newpage

\subsection{KEGG Metabolic Relation Network (Directed)}
Dimensions: $m$ = 53413, $n$ = 20.
\par
Description:

\begin{table}[!htbp]
\centering

\caption{Summary of the results with KEGG Metabolic Relation Network (Directed) ($\times10^{8}$)}
\label{TabResultsD17}
\small
\resizebox{!}{\tableheight}{
\begin{tabular}{|l|l|llll|llll|llll|llll|}
\hline
\multicolumn{1}{|c|}{\multirow{3}{*}{$k$}} & \multicolumn{1}{c|}{\multirow{3}{*}{$f^*$}} & \multicolumn{4}{c|}{Big-means} & \multicolumn{4}{c|}{IK-means} & \multicolumn{4}{c|}{BDCSM} & \multicolumn{4}{c|}{Minibatch K-means} \\ \cline{3-18}
\multicolumn{1}{|c|}{} & \multicolumn{1}{c|}{} & \multicolumn{2}{c|}{$\varepsilon$} & \multicolumn{2}{c|}{$t$} & \multicolumn{2}{c|}{$\varepsilon$} & \multicolumn{2}{c|}{$t$} & \multicolumn{2}{c|}{$\varepsilon$} & \multicolumn{2}{c|}{$t$} & \multicolumn{2}{c|}{$\varepsilon$} & \multicolumn{2}{c|}{$t$} \\ \cline{3-18}
\multicolumn{1}{|c|}{} & \multicolumn{1}{c|}{} & \multicolumn{1}{c|}{med} & \multicolumn{1}{c|}{std} & \multicolumn{1}{c|}{med} & \multicolumn{1}{c|}{std} & \multicolumn{1}{c|}{med} & \multicolumn{1}{c|}{std} & \multicolumn{1}{c|}{med} & \multicolumn{1}{c|}{std} & \multicolumn{1}{c|}{med} & \multicolumn{1}{c|}{std} & \multicolumn{1}{c|}{med} & \multicolumn{1}{c|}{std} & \multicolumn{1}{c|}{med} & \multicolumn{1}{c|}{std} & \multicolumn{1}{c|}{med} & \multicolumn{1}{c|}{std} \\ \hline
2 & 11.3853 & 0.24 & 0.11 & 0.63 & 0.28 & 29.27 & 4.48 & 0.06 & 0.64 & 18.85 & 0.0 & 0.03 & 0.0 & 34.75 & 8.46 & 0.24 & 0.09 \\
3 & 4.9006 & 0.56 & 0.26 & 0.5 & 0.23 & 174.75 & 19.91 & 0.09 & 0.0 & 124.79 & 0.0 & 0.08 & 0.0 & 141.74 & 7.79 & 0.3 & 0.09 \\
5 & 1.88367 & 0.01 & 0.69 & 0.66 & 0.27 & 536.66 & 53.13 & 0.14 & 0.01 & 0.0 & 8.01 & 0.19 & 0.01 & 458.97 & 21.01 & 0.42 & 0.23 \\
10 & 0.60513 & 0.04 & 1.83 & 0.68 & 0.19 & 1721.11 & 424.91 & 0.34 & 0.03 & 36.81 & 7.78 & 0.61 & 0.05 & 1508.43 & 54.07 & 0.51 & 0.14 \\
15 & 0.35393 & -0.03 & 4.41 & 0.9 & 0.2 & 2819.02 & 156.69 & 0.48 & 0.03 & 96.64 & 1.32 & 1.99 & 0.18 & 2556.76 & 106.8 & 0.39 & 0.1 \\
20 & 0.25027 & 0.09 & 0.59 & 1.04 & 0.25 & 3935.45 & 653.74 & 0.64 & 0.06 & 169.77 & 4.28 & 2.8 & 0.74 & 3619.14 & 186.14 & 0.58 & 0.12 \\
25 & 0.19289 & 1.04 & 0.88 & 1.17 & 0.23 & 4955.81 & 382.92 & 0.8 & 0.08 & 227.98 & 5.76 & 5.54 & 0.83 & 4694.61 & 217.03 & 0.55 & 0.31 \\
\hline
\multicolumn{2}{|c|}{Mean:} & \textbf{0.28} & & \textbf{0.8} & & \textbf{2024.58} & & \textbf{0.36} & & \textbf{96.41} & & \textbf{1.61} & & \textbf{1859.2} & & \textbf{0.43} & \\ \hline
\end{tabular}
}

\medskip

\small
\resizebox{!}{\tableheight}{
\begin{tabular}{|l|l|llll|llll|llll|llll|}
\hline
\multicolumn{1}{|c|}{\multirow{3}{*}{$k$}} & \multicolumn{1}{c|}{\multirow{3}{*}{$f^*$}} & \multicolumn{4}{c|}{K-means++} & \multicolumn{4}{c|}{CURE} & \multicolumn{4}{c|}{CluDataSE} & \multicolumn{4}{c|}{LW-coreset} \\ \cline{3-18}
\multicolumn{1}{|c|}{} & \multicolumn{1}{c|}{} & \multicolumn{2}{c|}{$\varepsilon$} & \multicolumn{2}{c|}{$t$} & \multicolumn{2}{c|}{$\varepsilon$} & \multicolumn{2}{c|}{$t$} & \multicolumn{2}{c|}{$\varepsilon$} & \multicolumn{2}{c|}{$t$} & \multicolumn{2}{c|}{$\varepsilon$} & \multicolumn{2}{c|}{$t$} \\ \cline{3-18}
\multicolumn{1}{|c|}{} & \multicolumn{1}{c|}{} & \multicolumn{1}{c|}{med} & \multicolumn{1}{c|}{std} & \multicolumn{1}{c|}{med} & \multicolumn{1}{c|}{std} & \multicolumn{1}{c|}{med} & \multicolumn{1}{c|}{std} & \multicolumn{1}{c|}{med} & \multicolumn{1}{c|}{std} & \multicolumn{1}{c|}{med} & \multicolumn{1}{c|}{std} & \multicolumn{1}{c|}{med} & \multicolumn{1}{c|}{std} & \multicolumn{1}{c|}{med} & \multicolumn{1}{c|}{std} & \multicolumn{1}{c|}{med} & \multicolumn{1}{c|}{std} \\ \hline
2 & 11.3853 & -0.0 & 9.0 & 0.01 & 0.68 & 56.76 & 10.82 & 2.84 & 10.63 & 18.85 & 0.0 & 0.22 & 0.07 & 0.0 & 0.0 & 0.01 & 3.63 \\
3 & 4.9006 & -0.0 & 50.11 & 0.06 & 0.02 & 243.76 & 22.11 & 3.51 & 0.54 & 124.79 & 0.0 & 0.32 & 0.05 & 0.01 & 0.03 & 0.02 & 0.0 \\
5 & 1.88367 & 0.0 & 16.48 & 0.13 & 0.04 & 791.95 & 68.61 & 3.62 & 0.88 & 0.0 & 0.0 & 0.47 & 0.1 & 0.1 & 0.07 & 0.04 & 0.01 \\
10 & 0.60513 & 8.43 & 15.9 & 0.3 & 0.11 & 2320.71 & 164.16 & 5.18 & 0.33 & 36.81 & 0.0 & 0.98 & 0.09 & 36.61 & 6.87 & 0.07 & 0.02 \\
15 & 0.35393 & 6.46 & 17.67 & 0.36 & 0.22 & 3975.36 & 147.37 & 7.45 & 0.41 & 96.64 & 0.69 & 2.33 & 0.23 & 91.8 & 21.36 & 0.09 & 0.05 \\
20 & 0.25027 & 10.65 & 25.17 & 0.64 & 0.29 & 5735.89 & 47.7 & 7.96 & 0.46 & 160.92 & 0.0 & 4.07 & 0.31 & 139.79 & 38.94 & 0.12 & 0.04 \\
25 & 0.19289 & 8.36 & 11.33 & 1.02 & 0.47 & 7484.65 & 320.47 & 7.51 & 0.6 & 221.85 & 1.86 & 5.78 & 0.38 & 180.45 & 57.4 & 0.17 & 0.05 \\
\hline
\multicolumn{2}{|c|}{Mean:} & \textbf{4.84} & & \textbf{0.36} & & \textbf{2944.16} & & \textbf{5.44} & & \textbf{94.27} & & \textbf{2.02} & & \textbf{64.11} & & \textbf{0.08} & \\ \hline
\end{tabular}
}

\bigskip

\caption{Clustering details with KEGG Metabolic Relation Network (Directed)}
\label{TabDetailsD17}
\resizebox{\linewidth}{!}{
\begin{tabular}{|l|l|lllll|l|l|ll|l|ll|lllll|ll|}
\hline
\multicolumn{1}{|c|}{\multirow{2}{*}{$k$}} & \multicolumn{1}{c|}{\multirow{2}{*}{$n_{exec}$}} & \multicolumn{5}{c|}{Big-means} & \multicolumn{1}{c|}{IK-means} & \multicolumn{1}{c|}{BDCSM} & \multicolumn{2}{c|}{Minibatch K-means} & \multicolumn{1}{c|}{K-means++} & \multicolumn{2}{c|}{CURE} & \multicolumn{5}{c|}{CluDataSE} & \multicolumn{2}{c|}{LW-coreset} \\ \cline{3-21}
\multicolumn{1}{|c|}{} & \multicolumn{1}{c|}{} & \multicolumn{1}{c|}{$s$} & \multicolumn{1}{c|}{$n_{s}$} & \multicolumn{1}{c|}{$T_1$} & \multicolumn{1}{c|}{$T_2$} & \multicolumn{1}{c|}{$n_{d}$} & \multicolumn{1}{c|}{$n_{d}$} & \multicolumn{1}{c|}{$n_{d}$} & \multicolumn{1}{c|}{$n_{s}$} & \multicolumn{1}{c|}{$n_{d}$} & \multicolumn{1}{c|}{$n_{d}$} & \multicolumn{1}{c|}{$s$} & \multicolumn{1}{c|}{$n_{d}$} & \multicolumn{1}{c|}{$s$} & \multicolumn{1}{c|}{$eps$} & \multicolumn{1}{c|}{$min\_pts$} & \multicolumn{1}{c|}{$sf$} & \multicolumn{1}{c|}{$n_{d}$} & \multicolumn{1}{c|}{$s$} & \multicolumn{1}{c|}{$n_{d}$} \\
\hline
2 & 20 & 53350 & 614 & 0.77 & 0.23 & 2.2E+08 & 4.3E+05 & 2.0E+06 & 25 & 2.7E+06 & 3.2E+05 & 10000 & 2.2E+07 & 10000 & 3.0 & 16 & 0.5 & 1.0E+08 & 10000 & 2.7E+05 \\
3 & 20 & 53350 & 355 & 0.97 & 0.03 & 2.7E+08 & 6.4E+05 & 5.4E+06 & 19 & 3.0E+06 & 2.7E+06 & 10000 & 1.8E+07 & 10000 & 3.0 & 16 & 0.5 & 1.1E+08 & 10000 & 6.9E+05 \\
5 & 20 & 53350 & 282 & 0.33 & 0.67 & 3.1E+08 & 1.1E+06 & 1.5E+07 & 16 & 4.4E+06 & 7.3E+06 & 10000 & 1.6E+07 & 10000 & 3.0 & 16 & 0.5 & 1.2E+08 & 10000 & 1.7E+06 \\
10 & 20 & 53350 & 110 & 0.87 & 0.13 & 3.5E+08 & 2.6E+06 & 5.3E+07 & 22 & 1.1E+07 & 2.1E+07 & 10000 & 2.0E+07 & 10000 & 3.0 & 16 & 0.5 & 1.6E+08 & 10000 & 4.3E+06 \\
15 & 20 & 53350 & 62 & 0.6 & 0.4 & 3.2E+08 & 3.7E+06 & 1.7E+08 & 19 & 1.5E+07 & 2.8E+07 & 10000 & 2.5E+07 & 10000 & 3.0 & 16 & 0.5 & 2.8E+08 & 10000 & 7.0E+06 \\
20 & 20 & 53350 & 14 & 0.1 & 0.9 & 2.5E+08 & 5.0E+06 & 2.5E+08 & 21 & 2.2E+07 & 4.9E+07 & 10000 & 2.6E+07 & 10000 & 3.0 & 16 & 0.5 & 4.3E+08 & 10000 & 9.7E+06 \\
25 & 20 & 53350 & 12 & 0.03 & 0.97 & 2.9E+08 & 6.2E+06 & 4.8E+08 & 20 & 2.7E+07 & 7.3E+07 & 10000 & 2.3E+07 & 10000 & 3.0 & 16 & 0.5 & 5.8E+08 & 10000 & 1.5E+07 \\
\hline
\end{tabular}
}

\end{table}

\newpage


\subsection{Shuttle Control}
Dimensions: $m$ = 58000, $n$ = 9.
\par
Description: each entity in the dataset contains several shuttle control attributes.

\begin{table}[!htbp]
\centering

\caption{Summary of the results with Shuttle Control ($\times10^{8}$)}
\label{TabResultsD18}
\small
\resizebox{!}{\tableheight}{
\begin{tabular}{|l|l|llll|llll|llll|llll|}
\hline
\multicolumn{1}{|c|}{\multirow{3}{*}{$k$}} & \multicolumn{1}{c|}{\multirow{3}{*}{$f^*$}} & \multicolumn{4}{c|}{Big-means} & \multicolumn{4}{c|}{IK-means} & \multicolumn{4}{c|}{BDCSM} & \multicolumn{4}{c|}{Minibatch K-means} \\ \cline{3-18}
\multicolumn{1}{|c|}{} & \multicolumn{1}{c|}{} & \multicolumn{2}{c|}{$\varepsilon$} & \multicolumn{2}{c|}{$t$} & \multicolumn{2}{c|}{$\varepsilon$} & \multicolumn{2}{c|}{$t$} & \multicolumn{2}{c|}{$\varepsilon$} & \multicolumn{2}{c|}{$t$} & \multicolumn{2}{c|}{$\varepsilon$} & \multicolumn{2}{c|}{$t$} \\ \cline{3-18}
\multicolumn{1}{|c|}{} & \multicolumn{1}{c|}{} & \multicolumn{1}{c|}{med} & \multicolumn{1}{c|}{std} & \multicolumn{1}{c|}{med} & \multicolumn{1}{c|}{std} & \multicolumn{1}{c|}{med} & \multicolumn{1}{c|}{std} & \multicolumn{1}{c|}{med} & \multicolumn{1}{c|}{std} & \multicolumn{1}{c|}{med} & \multicolumn{1}{c|}{std} & \multicolumn{1}{c|}{med} & \multicolumn{1}{c|}{std} & \multicolumn{1}{c|}{med} & \multicolumn{1}{c|}{std} & \multicolumn{1}{c|}{med} & \multicolumn{1}{c|}{std} \\ \hline
2 & 21.34329 & 0.0 & 2.33 & 0.89 & 0.54 & 51.16 & 1.14 & 0.06 & 0.0 & 51.11 & 0.03 & 0.03 & 0.01 & 51.82 & 0.67 & 0.21 & 0.06 \\
3 & 10.85415 & 3.66 & 1.45 & 0.8 & 0.34 & 195.45 & 10.97 & 0.09 & 0.0 & 100.52 & 32.99 & 0.04 & 0.02 & 196.64 & 1.8 & 0.21 & 0.07 \\
4 & 8.8691 & 8.54 & 7.05 & 0.85 & 0.44 & 260.04 & 37.5 & 0.11 & 0.02 & 143.41 & 56.3 & 0.04 & 0.02 & 261.76 & 0.86 & 0.23 & 0.1 \\
5 & 7.24479 & 0.18 & 7.31 & 0.84 & 0.31 & 299.22 & 37.03 & 0.14 & 0.05 & 38.69 & 35.03 & 0.06 & 0.02 & 341.24 & 4.43 & 0.21 & 0.05 \\
10 & 2.83216 & 0.74 & 25.95 & 0.75 & 0.38 & 720.46 & 109.45 & 0.31 & 0.09 & 135.63 & 34.19 & 0.16 & 0.1 & 998.49 & 16.18 & 0.23 & 0.07 \\
15 & 1.53154 & 4.46 & 2.41 & 1.27 & 0.44 & 1029.99 & 336.19 & 0.45 & 0.12 & 225.71 & 28.86 & 0.28 & 0.11 & 1883.71 & 66.41 & 0.25 & 0.06 \\
20 & 1.06012 & -0.45 & 1.36 & 1.13 & 0.28 & 823.88 & 245.56 & 0.59 & 0.04 & 307.27 & 31.55 & 0.26 & 0.07 & 2641.15 & 139.66 & 0.33 & 0.12 \\
25 & 0.77978 & 3.53 & 56.3 & 1.11 & 0.32 & 1089.67 & 615.73 & 0.76 & 0.17 & 392.57 & 28.66 & 0.5 & 0.19 & 3466.23 & 202.52 & 0.42 & 0.1 \\
\hline
\multicolumn{2}{|c|}{Mean:} & \textbf{2.58} & & \textbf{0.96} & & \textbf{558.74} & & \textbf{0.31} & & \textbf{174.37} & & \textbf{0.17} & & \textbf{1230.13} & & \textbf{0.26} & \\ \hline
\end{tabular}
}

\medskip

\small
\resizebox{!}{\tableheight}{
\begin{tabular}{|l|l|llll|llll|llll|llll|}
\hline
\multicolumn{1}{|c|}{\multirow{3}{*}{$k$}} & \multicolumn{1}{c|}{\multirow{3}{*}{$f^*$}} & \multicolumn{4}{c|}{K-means++} & \multicolumn{4}{c|}{CURE} & \multicolumn{4}{c|}{CluDataSE} & \multicolumn{4}{c|}{LW-coreset} \\ \cline{3-18}
\multicolumn{1}{|c|}{} & \multicolumn{1}{c|}{} & \multicolumn{2}{c|}{$\varepsilon$} & \multicolumn{2}{c|}{$t$} & \multicolumn{2}{c|}{$\varepsilon$} & \multicolumn{2}{c|}{$t$} & \multicolumn{2}{c|}{$\varepsilon$} & \multicolumn{2}{c|}{$t$} & \multicolumn{2}{c|}{$\varepsilon$} & \multicolumn{2}{c|}{$t$} \\ \cline{3-18}
\multicolumn{1}{|c|}{} & \multicolumn{1}{c|}{} & \multicolumn{1}{c|}{med} & \multicolumn{1}{c|}{std} & \multicolumn{1}{c|}{med} & \multicolumn{1}{c|}{std} & \multicolumn{1}{c|}{med} & \multicolumn{1}{c|}{std} & \multicolumn{1}{c|}{med} & \multicolumn{1}{c|}{std} & \multicolumn{1}{c|}{med} & \multicolumn{1}{c|}{std} & \multicolumn{1}{c|}{med} & \multicolumn{1}{c|}{std} & \multicolumn{1}{c|}{med} & \multicolumn{1}{c|}{std} & \multicolumn{1}{c|}{med} & \multicolumn{1}{c|}{std} \\ \hline
2 & 21.34329 & 5.04 & 15.24 & 0.01 & 0.0 & 54.09 & 0.7 & 4.57 & 0.36 & 51.07 & 0.03 & 0.36 & 0.04 & 0.01 & 2.02 & 0.01 & 0.0 \\
3 & 10.85415 & 0.0 & 43.14 & 0.01 & 0.01 & 197.47 & 1.24 & 5.95 & 0.33 & 100.56 & 37.99 & 0.4 & 0.05 & 84.8 & 44.12 & 0.01 & 0.0 \\
4 & 8.8691 & 15.19 & 40.54 & 0.02 & 0.01 & 263.16 & 0.58 & 5.65 & 0.42 & 143.53 & 43.61 & 0.46 & 0.04 & 15.28 & 49.49 & 0.01 & 0.0 \\
5 & 7.24479 & 13.63 & 5.74 & 0.02 & 0.01 & 343.78 & 0.87 & 5.72 & 0.25 & 175.82 & 67.7 & 0.42 & 0.04 & 29.49 & 68.11 & 0.01 & 0.0 \\
10 & 2.83216 & 22.05 & 30.77 & 0.06 & 0.03 & 1025.5 & 1.53 & 7.33 & 0.37 & 135.05 & 33.81 & 0.49 & 0.06 & 76.95 & 43.0 & 0.02 & 0.01 \\
15 & 1.53154 & 30.94 & 12.83 & 0.17 & 0.12 & 1974.2 & 2.06 & 9.8 & 0.88 & 216.54 & 10.79 & 0.63 & 0.08 & 60.86 & 69.01 & 0.02 & 0.01 \\
20 & 1.06012 & 26.77 & 17.9 & 0.19 & 0.09 & 2891.43 & 2.85 & 9.43 & 0.2 & 308.13 & 41.72 & 0.66 & 0.15 & 54.95 & 89.27 & 0.04 & 0.01 \\
25 & 0.77978 & 14.45 & 16.27 & 0.3 & 0.09 & 3962.48 & 2.09 & 8.06 & 0.48 & 385.48 & 21.99 & 1.22 & 0.29 & 38.41 & 105.71 & 0.04 & 0.02 \\
\hline
\multicolumn{2}{|c|}{Mean:} & \textbf{16.01} & & \textbf{0.1} & & \textbf{1339.01} & & \textbf{7.06} & & \textbf{189.52} & & \textbf{0.58} & & \textbf{45.09} & & \textbf{0.02} & \\ \hline
\end{tabular}
}

\bigskip

\caption{Clustering details with Shuttle Control}
\label{TabDetailsD18}
\resizebox{\linewidth}{!}{
\begin{tabular}{|l|l|lllll|l|l|ll|l|ll|lllll|ll|}
\hline
\multicolumn{1}{|c|}{\multirow{2}{*}{$k$}} & \multicolumn{1}{c|}{\multirow{2}{*}{$n_{exec}$}} & \multicolumn{5}{c|}{Big-means} & \multicolumn{1}{c|}{IK-means} & \multicolumn{1}{c|}{BDCSM} & \multicolumn{2}{c|}{Minibatch K-means} & \multicolumn{1}{c|}{K-means++} & \multicolumn{2}{c|}{CURE} & \multicolumn{5}{c|}{CluDataSE} & \multicolumn{2}{c|}{LW-coreset} \\ \cline{3-21}
\multicolumn{1}{|c|}{} & \multicolumn{1}{c|}{} & \multicolumn{1}{c|}{$s$} & \multicolumn{1}{c|}{$n_{s}$} & \multicolumn{1}{c|}{$T_1$} & \multicolumn{1}{c|}{$T_2$} & \multicolumn{1}{c|}{$n_{d}$} & \multicolumn{1}{c|}{$n_{d}$} & \multicolumn{1}{c|}{$n_{d}$} & \multicolumn{1}{c|}{$n_{s}$} & \multicolumn{1}{c|}{$n_{d}$} & \multicolumn{1}{c|}{$n_{d}$} & \multicolumn{1}{c|}{$s$} & \multicolumn{1}{c|}{$n_{d}$} & \multicolumn{1}{c|}{$s$} & \multicolumn{1}{c|}{$eps$} & \multicolumn{1}{c|}{$min\_pts$} & \multicolumn{1}{c|}{$sf$} & \multicolumn{1}{c|}{$n_{d}$} & \multicolumn{1}{c|}{$s$} & \multicolumn{1}{c|}{$n_{d}$} \\
\hline
2 & 15 & 57950 & 1416 & 0.75 & 0.75 & 5.5E+08 & 4.6E+05 & 3.5E+06 & 21 & 2.4E+06 & 4.6E+05 & 10000 & 3.8E+07 & 10000 & 3.0 & 16 & 0.5 & 1.0E+08 & 10000 & 3.3E+05 \\
3 & 15 & 57950 & 915 & 0.65 & 0.85 & 6.3E+08 & 7.0E+05 & 5.4E+06 & 19 & 3.3E+06 & 8.7E+05 & 10000 & 4.0E+07 & 10000 & 3.0 & 16 & 0.5 & 1.1E+08 & 10000 & 5.0E+05 \\
4 & 15 & 57950 & 904 & 0.6 & 0.9 & 7.1E+08 & 9.3E+05 & 7.0E+06 & 20 & 4.6E+06 & 1.2E+06 & 10000 & 4.2E+07 & 10000 & 3.0 & 16 & 0.5 & 1.1E+08 & 10000 & 6.7E+05 \\
5 & 15 & 57950 & 787 & 0.6 & 0.9 & 8.0E+08 & 1.2E+06 & 9.9E+06 & 19 & 5.5E+06 & 1.4E+06 & 10000 & 4.4E+07 & 10000 & 3.0 & 16 & 0.5 & 1.1E+08 & 10000 & 8.6E+05 \\
10 & 15 & 57950 & 359 & 0.1 & 1.4 & 9.4E+08 & 2.5E+06 & 3.4E+07 & 17 & 9.9E+06 & 7.5E+06 & 10000 & 4.9E+07 & 10000 & 3.0 & 16 & 0.5 & 1.3E+08 & 10000 & 2.3E+06 \\
15 & 15 & 57950 & 351 & 1.1 & 0.4 & 9.7E+08 & 3.9E+06 & 5.0E+07 & 17 & 1.5E+07 & 1.9E+07 & 10000 & 5.3E+07 & 10000 & 3.0 & 16 & 0.5 & 1.6E+08 & 10000 & 4.4E+06 \\
20 & 15 & 57950 & 188 & 0.6 & 0.9 & 1.0E+09 & 5.1E+06 & 5.9E+07 & 22 & 2.5E+07 & 3.5E+07 & 10000 & 5.5E+07 & 10000 & 3.0 & 16 & 0.5 & 1.6E+08 & 10000 & 7.7E+06 \\
25 & 15 & 57950 & 102 & 0.35 & 1.15 & 9.4E+08 & 6.5E+06 & 1.1E+08 & 23 & 3.3E+07 & 4.4E+07 & 10000 & 5.2E+07 & 10000 & 3.0 & 16 & 0.5 & 2.7E+08 & 10000 & 8.8E+06 \\
\hline
\end{tabular}
}

\end{table}

\newpage


\newpage

\subsection{Shuttle Control (normalized)}
Dimensions: $m$ = 58000, $n$ = 9.
\par
Description: each entity in the dataset contains several shuttle control attributes. Min-max scaling was used for normalization of data set values for better clusterization.

\begin{table}[!htbp]
\centering

\caption{Summary of the results with Shuttle Control (normalized) ($\times10^{1}$)}
\label{TabResultsD19}
\small
\resizebox{!}{\tableheight}{
\begin{tabular}{|l|l|llll|llll|llll|llll|}
\hline
\multicolumn{1}{|c|}{\multirow{3}{*}{$k$}} & \multicolumn{1}{c|}{\multirow{3}{*}{$f^*$}} & \multicolumn{4}{c|}{Big-means} & \multicolumn{4}{c|}{IK-means} & \multicolumn{4}{c|}{BDCSM} & \multicolumn{4}{c|}{Minibatch K-means} \\ \cline{3-18}
\multicolumn{1}{|c|}{} & \multicolumn{1}{c|}{} & \multicolumn{2}{c|}{$\varepsilon$} & \multicolumn{2}{c|}{$t$} & \multicolumn{2}{c|}{$\varepsilon$} & \multicolumn{2}{c|}{$t$} & \multicolumn{2}{c|}{$\varepsilon$} & \multicolumn{2}{c|}{$t$} & \multicolumn{2}{c|}{$\varepsilon$} & \multicolumn{2}{c|}{$t$} \\ \cline{3-18}
\multicolumn{1}{|c|}{} & \multicolumn{1}{c|}{} & \multicolumn{1}{c|}{med} & \multicolumn{1}{c|}{std} & \multicolumn{1}{c|}{med} & \multicolumn{1}{c|}{std} & \multicolumn{1}{c|}{med} & \multicolumn{1}{c|}{std} & \multicolumn{1}{c|}{med} & \multicolumn{1}{c|}{std} & \multicolumn{1}{c|}{med} & \multicolumn{1}{c|}{std} & \multicolumn{1}{c|}{med} & \multicolumn{1}{c|}{std} & \multicolumn{1}{c|}{med} & \multicolumn{1}{c|}{std} & \multicolumn{1}{c|}{med} & \multicolumn{1}{c|}{std} \\ \hline
2 & 104.41601 & 0.21 & 0.16 & 0.18 & 0.11 & 14.73 & 13.71 & 0.17 & 0.04 & 10.73 & 12.02 & 0.01 & 0.0 & 4.1 & 10.58 & 0.11 & 0.05 \\
3 & 73.28769 & 0.56 & 0.37 & 0.31 & 0.13 & 1.4 & 13.41 & 0.19 & 0.08 & 8.35 & 13.38 & 0.01 & 0.0 & 0.42 & 0.88 & 0.07 & 0.05 \\
4 & 50.076 & 0.57 & 0.39 & 0.25 & 0.1 & 0.02 & 6.6 & 0.27 & 0.08 & 0.54 & 22.36 & 0.01 & 0.0 & 6.76 & 13.43 & 0.12 & 0.05 \\
5 & 39.78043 & 1.47 & 0.79 & 0.32 & 0.1 & 0.86 & 9.7 & 0.32 & 0.07 & 12.94 & 26.62 & 0.01 & 0.0 & 1.67 & 1.58 & 0.11 & 0.04 \\
10 & 15.04997 & 0.79 & 1.08 & 0.31 & 0.11 & 34.16 & 22.8 & 0.71 & 0.14 & 63.36 & 28.42 & 0.02 & 0.0 & 4.37 & 2.78 & 0.08 & 0.05 \\
15 & 9.81804 & 2.67 & 1.97 & 0.21 & 0.11 & 88.68 & 38.77 & 1.28 & 0.24 & 32.71 & 40.53 & 0.03 & 0.0 & 7.63 & 3.03 & 0.08 & 0.04 \\
20 & 7.233 & 1.64 & 2.0 & 0.28 & 0.11 & 21.68 & 56.63 & 1.63 & 0.22 & 38.97 & 48.08 & 0.04 & 0.0 & 6.63 & 4.49 & 0.08 & 0.04 \\
25 & 5.86461 & 5.26 & 2.03 & 0.22 & 0.12 & 31.02 & 65.59 & 2.15 & 0.32 & 53.82 & 61.3 & 0.04 & 0.01 & 7.71 & 2.55 & 0.08 & 0.05 \\
\hline
\multicolumn{2}{|c|}{Mean:} & \textbf{1.65} & & \textbf{0.26} & & \textbf{24.07} & & \textbf{0.84} & & \textbf{27.68} & & \textbf{0.02} & & \textbf{4.91} & & \textbf{0.09} & \\ \hline
\end{tabular}
}

\medskip

\small
\resizebox{!}{\tableheight}{
\begin{tabular}{|l|l|llll|llll|llll|llll|}
\hline
\multicolumn{1}{|c|}{\multirow{3}{*}{$k$}} & \multicolumn{1}{c|}{\multirow{3}{*}{$f^*$}} & \multicolumn{4}{c|}{K-means++} & \multicolumn{4}{c|}{CURE} & \multicolumn{4}{c|}{CluDataSE} & \multicolumn{4}{c|}{LW-coreset} \\ \cline{3-18}
\multicolumn{1}{|c|}{} & \multicolumn{1}{c|}{} & \multicolumn{2}{c|}{$\varepsilon$} & \multicolumn{2}{c|}{$t$} & \multicolumn{2}{c|}{$\varepsilon$} & \multicolumn{2}{c|}{$t$} & \multicolumn{2}{c|}{$\varepsilon$} & \multicolumn{2}{c|}{$t$} & \multicolumn{2}{c|}{$\varepsilon$} & \multicolumn{2}{c|}{$t$} \\ \cline{3-18}
\multicolumn{1}{|c|}{} & \multicolumn{1}{c|}{} & \multicolumn{1}{c|}{med} & \multicolumn{1}{c|}{std} & \multicolumn{1}{c|}{med} & \multicolumn{1}{c|}{std} & \multicolumn{1}{c|}{med} & \multicolumn{1}{c|}{std} & \multicolumn{1}{c|}{med} & \multicolumn{1}{c|}{std} & \multicolumn{1}{c|}{med} & \multicolumn{1}{c|}{std} & \multicolumn{1}{c|}{med} & \multicolumn{1}{c|}{std} & \multicolumn{1}{c|}{med} & \multicolumn{1}{c|}{std} & \multicolumn{1}{c|}{med} & \multicolumn{1}{c|}{std} \\ \hline
2 & 104.41601 & 0.0 & 15.41 & 0.01 & 0.01 & 0.99 & 1.96 & 0.43 & 0.12 & 0.0 & 15.13 & 0.02 & 0.0 & 0.07 & 13.39 & 0.0 & 0.0 \\
3 & 73.28769 & 1.76 & 11.09 & 0.03 & 0.01 & 32.04 & 2.43 & 0.54 & 0.01 & 1.76 & 13.06 & 0.03 & 0.0 & 2.09 & 14.72 & 0.0 & 0.0 \\
4 & 50.076 & 12.41 & 17.03 & 0.03 & 0.01 & 36.92 & 14.09 & 0.62 & 0.03 & 0.0 & 12.01 & 0.03 & 0.0 & 30.69 & 29.04 & 0.0 & 0.0 \\
5 & 39.78043 & 2.72 & 9.08 & 0.04 & 0.02 & 33.66 & 13.6 & 0.71 & 0.04 & 0.83 & 2.45 & 0.05 & 0.01 & 25.43 & 41.88 & 0.01 & 0.0 \\
10 & 15.04997 & 6.45 & 16.3 & 0.08 & 0.03 & 46.5 & 22.29 & 0.81 & 0.04 & 2.87 & 23.99 & 0.05 & 0.03 & 51.65 & 58.52 & 0.01 & 0.0 \\
15 & 9.81804 & 10.69 & 7.06 & 0.13 & 0.05 & 53.8 & 63.43 & 0.64 & 0.04 & 7.82 & 17.34 & 0.08 & 0.04 & 44.92 & 40.22 & 0.01 & 0.0 \\
20 & 7.233 & 13.48 & 5.32 & 0.19 & 0.06 & 117.02 & 190.87 & 0.55 & 0.03 & 7.97 & 24.7 & 0.16 & 0.09 & 58.76 & 33.04 & 0.01 & 0.0 \\
25 & 5.86461 & 10.76 & 5.15 & 0.18 & 0.06 & 386.47 & 313.37 & 0.48 & 0.01 & 10.63 & 5.87 & 0.13 & 0.06 & 48.45 & 31.99 & 0.01 & 0.0 \\
\hline
\multicolumn{2}{|c|}{Mean:} & \textbf{7.28} & & \textbf{0.09} & & \textbf{88.43} & & \textbf{0.6} & & \textbf{3.98} & & \textbf{0.07} & & \textbf{32.76} & & \textbf{0.01} & \\ \hline
\end{tabular}
}

\bigskip

\caption{Clustering details with Shuttle Control (normalized)}
\label{TabDetailsD19}
\resizebox{\linewidth}{!}{
\begin{tabular}{|l|l|lllll|l|l|ll|l|ll|lllll|ll|}
\hline
\multicolumn{1}{|c|}{\multirow{2}{*}{$k$}} & \multicolumn{1}{c|}{\multirow{2}{*}{$n_{exec}$}} & \multicolumn{5}{c|}{Big-means} & \multicolumn{1}{c|}{IK-means} & \multicolumn{1}{c|}{BDCSM} & \multicolumn{2}{c|}{Minibatch K-means} & \multicolumn{1}{c|}{K-means++} & \multicolumn{2}{c|}{CURE} & \multicolumn{5}{c|}{CluDataSE} & \multicolumn{2}{c|}{LW-coreset} \\ \cline{3-21}
\multicolumn{1}{|c|}{} & \multicolumn{1}{c|}{} & \multicolumn{1}{c|}{$s$} & \multicolumn{1}{c|}{$n_{s}$} & \multicolumn{1}{c|}{$T_1$} & \multicolumn{1}{c|}{$T_2$} & \multicolumn{1}{c|}{$n_{d}$} & \multicolumn{1}{c|}{$n_{d}$} & \multicolumn{1}{c|}{$n_{d}$} & \multicolumn{1}{c|}{$n_{s}$} & \multicolumn{1}{c|}{$n_{d}$} & \multicolumn{1}{c|}{$n_{d}$} & \multicolumn{1}{c|}{$s$} & \multicolumn{1}{c|}{$n_{d}$} & \multicolumn{1}{c|}{$s$} & \multicolumn{1}{c|}{$eps$} & \multicolumn{1}{c|}{$min\_pts$} & \multicolumn{1}{c|}{$sf$} & \multicolumn{1}{c|}{$n_{d}$} & \multicolumn{1}{c|}{$s$} & \multicolumn{1}{c|}{$n_{d}$} \\
\hline
2 & 20 & 2000 & 872 & 0.04 & 0.36 & 1.7E+07 & 7.1E+05 & 1.4E+06 & 37 & 1.5E+05 & 5.8E+05 & 2000 & 4.1E+06 & 2000 & 0.0003 & 2 & 0.5 & 4.8E+06 & 2000 & 2.5E+05 \\
3 & 20 & 2000 & 1278 & 0.28 & 0.12 & 4.0E+07 & 1.3E+06 & 2.3E+06 & 27 & 1.6E+05 & 2.3E+06 & 2000 & 4.3E+06 & 2000 & 0.0003 & 2 & 0.5 & 5.6E+06 & 2000 & 3.3E+05 \\
4 & 20 & 2000 & 974 & 0.24 & 0.16 & 4.9E+07 & 1.7E+06 & 3.1E+06 & 42 & 3.3E+05 & 3.1E+06 & 2000 & 4.6E+06 & 2000 & 0.0003 & 2 & 0.5 & 6.6E+06 & 2000 & 4.3E+05 \\
5 & 20 & 2000 & 1144 & 0.19 & 0.21 & 7.0E+07 & 2.3E+06 & 4.8E+06 & 39 & 3.9E+05 & 4.5E+06 & 2000 & 4.8E+06 & 2000 & 0.0003 & 2 & 0.5 & 9.0E+06 & 2000 & 5.1E+05 \\
10 & 20 & 2000 & 924 & 0.16 & 0.24 & 1.1E+08 & 5.5E+06 & 1.2E+07 & 28 & 5.6E+05 & 1.2E+07 & 2000 & 5.3E+06 & 2000 & 0.0003 & 2 & 0.5 & 9.9E+06 & 2000 & 9.8E+05 \\
15 & 20 & 2000 & 370 & 0.32 & 0.08 & 1.5E+08 & 9.8E+06 & 2.0E+07 & 21 & 6.3E+05 & 1.7E+07 & 2000 & 4.9E+06 & 2000 & 0.0003 & 2 & 0.5 & 1.8E+07 & 2000 & 1.6E+06 \\
20 & 20 & 2000 & 343 & 0.25 & 0.15 & 1.8E+08 & 1.3E+07 & 2.6E+07 & 19 & 7.6E+05 & 3.2E+07 & 2000 & 4.8E+06 & 2000 & 0.0003 & 2 & 0.5 & 3.0E+07 & 2000 & 2.1E+06 \\
25 & 20 & 2000 & 229 & 0.08 & 0.32 & 2.0E+08 & 1.7E+07 & 3.3E+07 & 19 & 9.5E+05 & 3.1E+07 & 2000 & 4.5E+06 & 2000 & 0.0003 & 2 & 0.5 & 2.9E+07 & 2000 & 2.6E+06 \\
\hline
\end{tabular}
}

\end{table}

\newpage


\subsection{EEG Eye State}
Dimensions: $m$ = 14980, $n$ = 14.
\par
Description: the data set consists of 14 electroencephalogram (EEG) values for predicting the corresponding eye state.

\begin{table}[!htbp]
\centering

\caption{Summary of the results with EEG Eye State ($\times10^{8}$)}
\label{TabResultsD20}
\small
\resizebox{!}{\tableheight}{
\begin{tabular}{|l|l|llll|llll|llll|llll|}
\hline
\multicolumn{1}{|c|}{\multirow{3}{*}{$k$}} & \multicolumn{1}{c|}{\multirow{3}{*}{$f^*$}} & \multicolumn{4}{c|}{Big-means} & \multicolumn{4}{c|}{IK-means} & \multicolumn{4}{c|}{BDCSM} & \multicolumn{4}{c|}{Minibatch K-means} \\ \cline{3-18}
\multicolumn{1}{|c|}{} & \multicolumn{1}{c|}{} & \multicolumn{2}{c|}{$\varepsilon$} & \multicolumn{2}{c|}{$t$} & \multicolumn{2}{c|}{$\varepsilon$} & \multicolumn{2}{c|}{$t$} & \multicolumn{2}{c|}{$\varepsilon$} & \multicolumn{2}{c|}{$t$} & \multicolumn{2}{c|}{$\varepsilon$} & \multicolumn{2}{c|}{$t$} \\ \cline{3-18}
\multicolumn{1}{|c|}{} & \multicolumn{1}{c|}{} & \multicolumn{1}{c|}{med} & \multicolumn{1}{c|}{std} & \multicolumn{1}{c|}{med} & \multicolumn{1}{c|}{std} & \multicolumn{1}{c|}{med} & \multicolumn{1}{c|}{std} & \multicolumn{1}{c|}{med} & \multicolumn{1}{c|}{std} & \multicolumn{1}{c|}{med} & \multicolumn{1}{c|}{std} & \multicolumn{1}{c|}{med} & \multicolumn{1}{c|}{std} & \multicolumn{1}{c|}{med} & \multicolumn{1}{c|}{std} & \multicolumn{1}{c|}{med} & \multicolumn{1}{c|}{std} \\ \hline
2 & 7845.09934 & 4.25 & 0.0 & 0.89 & 0.45 & 97.25 & 33.08 & 0.02 & 0.0 & -0.0 & 0.93 & 0.0 & 0.0 & 98.06 & 0.24 & 0.08 & 0.02 \\
3 & 1833.88058 & 0.0 & 65.98 & 0.6 & 0.46 & 327.74 & 86.37 & 0.02 & 0.0 & 227.91 & 81.38 & 0.01 & 0.0 & 747.5 & 1.08 & 0.06 & 0.02 \\
4 & 2.23605 & 0.0 & 0.0 & 0.82 & 0.41 & 350642.24 & 88855.37 & 0.03 & 0.0 & 268809.8 & 128214.1 & 0.02 & 0.01 & 694466.19 & 499.31 & 0.05 & 0.01 \\
5 & 1.33858 & -0.0 & 6.52 & 0.93 & 0.35 & 585674.55 & 186919.07 & 0.04 & 0.0 & 449091.75 & 214188.85 & 0.03 & 0.01 & 1159957.25 & 4993.87 & 0.06 & 0.02 \\
10 & 0.4531 & 0.0 & 0.01 & 0.9 & 0.38 & 1119056.99 & 644551.55 & 0.09 & 0.0 & 663435.01 & 663330.92 & 0.1 & 0.02 & 3386305.55 & 61885.8 & 0.08 & 0.02 \\
15 & 0.34653 & 0.07 & 0.15 & 0.92 & 0.36 & 2261071.4 & 775757.28 & 0.11 & 0.01 & 1734685.37 & 862995.0 & 0.15 & 0.04 & 4344184.11 & 97469.29 & 0.13 & 0.02 \\
20 & 0.28986 & 0.05 & 0.23 & 1.42 & 0.43 & 630877.35 & 1068598.65 & 0.14 & 0.02 & 2073832.6 & 1015966.35 & 0.26 & 0.09 & 5305346.5 & 194325.45 & 0.11 & 0.02 \\
25 & 0.25989 & 0.12 & 0.07 & 1.06 & 0.34 & 702866.81 & 1017166.36 & 0.21 & 0.03 & 2312984.45 & 1150694.93 & 0.41 & 0.09 & 5857365.28 & 167835.2 & 0.13 & 0.03 \\
\hline
\multicolumn{2}{|c|}{Mean:} & \textbf{0.56} & & \textbf{0.94} & & \textbf{706326.79} & & \textbf{0.08} & & \textbf{937883.36} & & \textbf{0.12} & & \textbf{2593558.8} & & \textbf{0.09} & \\ \hline
\end{tabular}
}

\medskip

\small
\resizebox{!}{\tableheight}{
\begin{tabular}{|l|l|llll|llll|llll|llll|}
\hline
\multicolumn{1}{|c|}{\multirow{3}{*}{$k$}} & \multicolumn{1}{c|}{\multirow{3}{*}{$f^*$}} & \multicolumn{4}{c|}{K-means++} & \multicolumn{4}{c|}{CURE} & \multicolumn{4}{c|}{CluDataSE} & \multicolumn{4}{c|}{LW-coreset} \\ \cline{3-18}
\multicolumn{1}{|c|}{} & \multicolumn{1}{c|}{} & \multicolumn{2}{c|}{$\varepsilon$} & \multicolumn{2}{c|}{$t$} & \multicolumn{2}{c|}{$\varepsilon$} & \multicolumn{2}{c|}{$t$} & \multicolumn{2}{c|}{$\varepsilon$} & \multicolumn{2}{c|}{$t$} & \multicolumn{2}{c|}{$\varepsilon$} & \multicolumn{2}{c|}{$t$} \\ \cline{3-18}
\multicolumn{1}{|c|}{} & \multicolumn{1}{c|}{} & \multicolumn{1}{c|}{med} & \multicolumn{1}{c|}{std} & \multicolumn{1}{c|}{med} & \multicolumn{1}{c|}{std} & \multicolumn{1}{c|}{med} & \multicolumn{1}{c|}{std} & \multicolumn{1}{c|}{med} & \multicolumn{1}{c|}{std} & \multicolumn{1}{c|}{med} & \multicolumn{1}{c|}{std} & \multicolumn{1}{c|}{med} & \multicolumn{1}{c|}{std} & \multicolumn{1}{c|}{med} & \multicolumn{1}{c|}{std} & \multicolumn{1}{c|}{med} & \multicolumn{1}{c|}{std} \\ \hline
2 & 7845.09934 & 17.49 & 20.02 & 0.0 & 0.0 & 98.37 & 0.02 & 1.92 & 0.56 & -0.0 & 0.0 & 0.83 & 0.02 & 4.25 & 6.59 & 0.0 & 0.0 \\
3 & 1833.88058 & 0.0 & 122.29 & 0.0 & 0.0 & 748.33 & 0.04 & 2.26 & 0.46 & 227.91 & 0.0 & 0.83 & 0.02 & 123.03 & 126.44 & 0.0 & 0.0 \\
4 & 2.23605 & 0.0 & 0.0 & 0.01 & 0.0 & 695638.7 & 37.25 & 2.34 & 0.29 & 268809.8 & 0.0 & 0.84 & 0.02 & 0.02 & 63442.6 & 0.0 & 0.0 \\
5 & 1.33858 & 14.95 & 14.95 & 0.01 & 0.01 & 1162076.58 & 71.7 & 2.66 & 0.25 & 449091.75 & 0.0 & 0.89 & 0.06 & 30.0 & 12.0 & 0.01 & 0.0 \\
10 & 0.4531 & -0.01 & 0.32 & 0.07 & 0.03 & 3432933.34 & 115.55 & 3.94 & 0.1 & 1326870.19 & 289134.66 & 0.93 & 0.03 & 190.17 & 74.52 & 0.03 & 0.01 \\
15 & 0.34653 & 0.88 & 0.29 & 0.15 & 0.09 & 4488641.07 & 43.11 & 5.53 & 0.16 & 1734686.09 & 0.29 & 1.0 & 0.04 & 2.34 & 123.48 & 0.06 & 0.03 \\
20 & 0.28986 & 0.78 & 0.54 & 0.21 & 0.12 & 5366270.87 & 120.44 & 4.82 & 0.34 & 2073833.78 & 0.66 & 1.16 & 0.1 & 304.43 & 147.96 & 0.08 & 0.03 \\
25 & 0.25989 & 0.3 & 0.66 & 0.28 & 0.09 & 5985037.33 & 175.09 & 4.02 & 0.19 & 2312984.96 & 0.83 & 1.19 & 0.12 & 340.37 & 146.72 & 0.07 & 0.04 \\
\hline
\multicolumn{2}{|c|}{Mean:} & \textbf{4.3} & & \textbf{0.09} & & \textbf{2641430.58} & & \textbf{3.44} & & \textbf{1020813.06} & & \textbf{0.96} & & \textbf{124.33} & & \textbf{0.03} & \\ \hline
\end{tabular}
}

\bigskip

\caption{Clustering details with EEG Eye State}
\label{TabDetailsD20}
\resizebox{\linewidth}{!}{
\begin{tabular}{|l|l|lllll|l|l|ll|l|ll|lllll|ll|}
\hline
\multicolumn{1}{|c|}{\multirow{2}{*}{$k$}} & \multicolumn{1}{c|}{\multirow{2}{*}{$n_{exec}$}} & \multicolumn{5}{c|}{Big-means} & \multicolumn{1}{c|}{IK-means} & \multicolumn{1}{c|}{BDCSM} & \multicolumn{2}{c|}{Minibatch K-means} & \multicolumn{1}{c|}{K-means++} & \multicolumn{2}{c|}{CURE} & \multicolumn{5}{c|}{CluDataSE} & \multicolumn{2}{c|}{LW-coreset} \\ \cline{3-21}
\multicolumn{1}{|c|}{} & \multicolumn{1}{c|}{} & \multicolumn{1}{c|}{$s$} & \multicolumn{1}{c|}{$n_{s}$} & \multicolumn{1}{c|}{$T_1$} & \multicolumn{1}{c|}{$T_2$} & \multicolumn{1}{c|}{$n_{d}$} & \multicolumn{1}{c|}{$n_{d}$} & \multicolumn{1}{c|}{$n_{d}$} & \multicolumn{1}{c|}{$n_{s}$} & \multicolumn{1}{c|}{$n_{d}$} & \multicolumn{1}{c|}{$n_{d}$} & \multicolumn{1}{c|}{$s$} & \multicolumn{1}{c|}{$n_{d}$} & \multicolumn{1}{c|}{$s$} & \multicolumn{1}{c|}{$eps$} & \multicolumn{1}{c|}{$min\_pts$} & \multicolumn{1}{c|}{$sf$} & \multicolumn{1}{c|}{$n_{d}$} & \multicolumn{1}{c|}{$s$} & \multicolumn{1}{c|}{$n_{d}$} \\
\hline
2 & 20 & 14979 & 4504 & 0.7 & 0.8 & 4.8E+08 & 1.2E+05 & 1.8E+05 & 20 & 6.1E+05 & 9.0E+04 & 8000 & 9.3E+06 & 8000 & 25.0 & 16 & 0.5 & 6.4E+07 & 8000 & 1.1E+05 \\
3 & 20 & 14979 & 2504 & 0.05 & 1.45 & 5.7E+08 & 1.8E+05 & 4.0E+05 & 16 & 7.2E+05 & 1.3E+05 & 8000 & 8.8E+06 & 8000 & 25.0 & 16 & 0.5 & 6.5E+07 & 8000 & 1.5E+05 \\
4 & 20 & 14979 & 2835 & 0.3 & 1.2 & 6.5E+08 & 2.4E+05 & 1.6E+06 & 21 & 1.3E+06 & 1.8E+05 & 8000 & 9.8E+06 & 8000 & 25.0 & 16 & 0.5 & 6.6E+07 & 8000 & 1.9E+05 \\
5 & 20 & 14979 & 2783 & 0.85 & 0.65 & 6.9E+08 & 3.0E+05 & 3.4E+06 & 17 & 1.3E+06 & 3.4E+05 & 8000 & 1.1E+07 & 8000 & 25.0 & 16 & 0.5 & 6.8E+07 & 8000 & 7.4E+05 \\
10 & 20 & 14979 & 1413 & 0.85 & 0.65 & 7.9E+08 & 6.0E+05 & 9.9E+06 & 20 & 2.9E+06 & 6.9E+06 & 8000 & 1.7E+07 & 8000 & 25.0 & 16 & 0.5 & 7.8E+07 & 8000 & 2.9E+06 \\
15 & 20 & 14979 & 970 & 0.35 & 1.15 & 8.5E+08 & 9.0E+05 & 1.9E+07 & 21 & 4.7E+06 & 1.5E+07 & 8000 & 2.3E+07 & 8000 & 25.0 & 16 & 0.5 & 8.6E+07 & 8000 & 6.4E+06 \\
20 & 20 & 14979 & 1120 & 1.4 & 0.1 & 8.7E+08 & 1.2E+06 & 3.4E+07 & 20 & 6.1E+06 & 2.4E+07 & 8000 & 2.0E+07 & 8000 & 25.0 & 16 & 0.5 & 1.1E+08 & 8000 & 9.0E+06 \\
25 & 20 & 14979 & 454 & 0.2 & 1.3 & 7.9E+08 & 1.5E+06 & 4.8E+07 & 18 & 6.6E+06 & 3.2E+07 & 8000 & 1.7E+07 & 8000 & 25.0 & 16 & 0.5 & 1.1E+08 & 8000 & 8.1E+06 \\
\hline
\end{tabular}
}

\end{table}

\newpage


\newpage

\subsection{EEG Eye State (normalized)}
Dimensions: $m$ = 14980, $n$ = 14.
\par
Description: the data set consists of 14 electroencephalogram (EEG) values for predicting the corresponding eye state. Min-max scaling was used for normalization of data set values for better clusterization.

\begin{table}[!htbp]
\centering

\caption{Summary of the results with EEG Eye State (normalized) ($\times10^{1}$)}
\label{TabResultsD21}
\small
\resizebox{!}{\tableheight}{
\begin{tabular}{|l|l|llll|llll|llll|llll|}
\hline
\multicolumn{1}{|c|}{\multirow{3}{*}{$k$}} & \multicolumn{1}{c|}{\multirow{3}{*}{$f^*$}} & \multicolumn{4}{c|}{Big-means} & \multicolumn{4}{c|}{IK-means} & \multicolumn{4}{c|}{BDCSM} & \multicolumn{4}{c|}{Minibatch K-means} \\ \cline{3-18}
\multicolumn{1}{|c|}{} & \multicolumn{1}{c|}{} & \multicolumn{2}{c|}{$\varepsilon$} & \multicolumn{2}{c|}{$t$} & \multicolumn{2}{c|}{$\varepsilon$} & \multicolumn{2}{c|}{$t$} & \multicolumn{2}{c|}{$\varepsilon$} & \multicolumn{2}{c|}{$t$} & \multicolumn{2}{c|}{$\varepsilon$} & \multicolumn{2}{c|}{$t$} \\ \cline{3-18}
\multicolumn{1}{|c|}{} & \multicolumn{1}{c|}{} & \multicolumn{1}{c|}{med} & \multicolumn{1}{c|}{std} & \multicolumn{1}{c|}{med} & \multicolumn{1}{c|}{std} & \multicolumn{1}{c|}{med} & \multicolumn{1}{c|}{std} & \multicolumn{1}{c|}{med} & \multicolumn{1}{c|}{std} & \multicolumn{1}{c|}{med} & \multicolumn{1}{c|}{std} & \multicolumn{1}{c|}{med} & \multicolumn{1}{c|}{std} & \multicolumn{1}{c|}{med} & \multicolumn{1}{c|}{std} & \multicolumn{1}{c|}{med} & \multicolumn{1}{c|}{std} \\ \hline
2 & 1.15267 & 0.0 & 3.14 & 0.56 & 0.27 & 25.4 & 0.01 & 0.13 & 0.06 & 25.4 & 0.01 & 0.01 & 0.0 & 27.5 & 1.26 & 0.06 & 0.02 \\
3 & 0.82423 & 0.0 & 6.44 & 0.61 & 0.22 & 69.04 & 0.03 & 0.15 & 0.03 & 69.04 & 0.03 & 0.02 & 0.0 & 74.74 & 2.38 & 0.06 & 0.02 \\
4 & 0.5429 & 0.0 & 6.11 & 0.72 & 0.31 & 152.48 & 0.06 & 0.22 & 0.09 & 152.47 & 0.06 & 0.02 & 0.01 & 160.89 & 4.6 & 0.07 & 0.02 \\
5 & 0.28952 & 0.0 & 18.53 & 0.64 & 0.24 & 367.07 & 22.96 & 0.39 & 0.09 & 367.1 & 0.08 & 0.04 & 0.01 & 382.53 & 8.64 & 0.08 & 0.03 \\
10 & 0.10269 & -0.01 & 0.0 & 0.72 & 0.25 & 879.03 & 161.91 & 1.29 & 0.32 & 879.79 & 175.98 & 0.11 & 0.03 & 1188.62 & 3.96 & 0.09 & 0.03 \\
15 & 0.07469 & 0.04 & 0.06 & 0.66 & 0.19 & 853.05 & 313.05 & 1.7 & 0.36 & 852.99 & 263.04 & 0.15 & 0.05 & 1639.36 & 26.21 & 0.11 & 0.03 \\
20 & 0.06125 & 0.07 & 0.15 & 0.77 & 0.17 & 1044.53 & 390.42 & 2.68 & 0.61 & 1044.25 & 417.09 & 0.24 & 0.1 & 2007.07 & 52.77 & 0.11 & 0.03 \\
25 & 0.05385 & -0.2 & 0.11 & 0.97 & 0.13 & 1191.01 & 254.04 & 3.55 & 0.51 & 1187.56 & 339.48 & 0.3 & 0.07 & 2276.63 & 68.05 & 0.15 & 0.05 \\
\hline
\multicolumn{2}{|c|}{Mean:} & \textbf{-0.01} & & \textbf{0.71} & & \textbf{572.7} & & \textbf{1.26} & & \textbf{572.33} & & \textbf{0.11} & & \textbf{969.67} & & \textbf{0.09} & \\ \hline
\end{tabular}
}

\medskip

\small
\resizebox{!}{\tableheight}{
\begin{tabular}{|l|l|llll|llll|llll|llll|}
\hline
\multicolumn{1}{|c|}{\multirow{3}{*}{$k$}} & \multicolumn{1}{c|}{\multirow{3}{*}{$f^*$}} & \multicolumn{4}{c|}{K-means++} & \multicolumn{4}{c|}{CURE} & \multicolumn{4}{c|}{CluDataSE} & \multicolumn{4}{c|}{LW-coreset} \\ \cline{3-18}
\multicolumn{1}{|c|}{} & \multicolumn{1}{c|}{} & \multicolumn{2}{c|}{$\varepsilon$} & \multicolumn{2}{c|}{$t$} & \multicolumn{2}{c|}{$\varepsilon$} & \multicolumn{2}{c|}{$t$} & \multicolumn{2}{c|}{$\varepsilon$} & \multicolumn{2}{c|}{$t$} & \multicolumn{2}{c|}{$\varepsilon$} & \multicolumn{2}{c|}{$t$} \\ \cline{3-18}
\multicolumn{1}{|c|}{} & \multicolumn{1}{c|}{} & \multicolumn{1}{c|}{med} & \multicolumn{1}{c|}{std} & \multicolumn{1}{c|}{med} & \multicolumn{1}{c|}{std} & \multicolumn{1}{c|}{med} & \multicolumn{1}{c|}{std} & \multicolumn{1}{c|}{med} & \multicolumn{1}{c|}{std} & \multicolumn{1}{c|}{med} & \multicolumn{1}{c|}{std} & \multicolumn{1}{c|}{med} & \multicolumn{1}{c|}{std} & \multicolumn{1}{c|}{med} & \multicolumn{1}{c|}{std} & \multicolumn{1}{c|}{med} & \multicolumn{1}{c|}{std} \\ \hline
2 & 1.15267 & 11.4 & 10.29 & 0.0 & 0.01 & 25.64 & 1.19 & 1.99 & 0.1 & 25.4 & 0.0 & 0.61 & 0.09 & 6.1 & 4.82 & 0.0 & 0.0 \\
3 & 0.82423 & 27.59 & 22.03 & 0.01 & 0.01 & 73.56 & 1.84 & 2.11 & 0.07 & 69.04 & 0.04 & 0.66 & 0.09 & 38.62 & 6.92 & 0.01 & 0.0 \\
4 & 0.5429 & 37.84 & 27.91 & 0.02 & 0.01 & 159.64 & 1.72 & 2.36 & 0.07 & 152.35 & 0.05 & 0.6 & 0.08 & 90.57 & 14.93 & 0.01 & 0.0 \\
5 & 0.28952 & 105.16 & 54.36 & 0.03 & 0.01 & 381.15 & 2.61 & 2.75 & 0.27 & 367.07 & 0.11 & 0.67 & 0.2 & 261.69 & 37.78 & 0.02 & 0.01 \\
10 & 0.10269 & 0.7 & 0.56 & 0.13 & 0.13 & 1218.5 & 8.26 & 4.19 & 0.19 & 879.68 & 79.92 & 0.75 & 0.09 & 568.99 & 170.56 & 0.04 & 0.01 \\
15 & 0.07469 & 0.15 & 0.23 & 0.18 & 0.08 & 1685.07 & 13.69 & 6.05 & 0.2 & 853.24 & 170.74 & 0.76 & 0.1 & 413.56 & 258.62 & 0.08 & 0.03 \\
20 & 0.06125 & 0.5 & 1.26 & 0.22 & 0.13 & 2062.84 & 12.35 & 4.71 & 0.16 & 1044.43 & 1.67 & 0.9 & 0.16 & 502.54 & 291.59 & 0.07 & 0.02 \\
25 & 0.05385 & 0.7 & 0.61 & 0.29 & 0.1 & 2344.18 & 10.51 & 4.19 & 0.12 & 1191.05 & 110.44 & 0.97 & 0.14 & 565.67 & 344.68 & 0.09 & 0.03 \\
\hline
\multicolumn{2}{|c|}{Mean:} & \textbf{23.01} & & \textbf{0.11} & & \textbf{993.82} & & \textbf{3.54} & & \textbf{572.78} & & \textbf{0.74} & & \textbf{305.97} & & \textbf{0.04} & \\ \hline
\end{tabular}
}

\bigskip

\caption{Clustering details with EEG Eye State (normalized)}
\label{TabDetailsD21}
\resizebox{\linewidth}{!}{
\begin{tabular}{|l|l|lllll|l|l|ll|l|ll|lllll|ll|}
\hline
\multicolumn{1}{|c|}{\multirow{2}{*}{$k$}} & \multicolumn{1}{c|}{\multirow{2}{*}{$n_{exec}$}} & \multicolumn{5}{c|}{Big-means} & \multicolumn{1}{c|}{IK-means} & \multicolumn{1}{c|}{BDCSM} & \multicolumn{2}{c|}{Minibatch K-means} & \multicolumn{1}{c|}{K-means++} & \multicolumn{2}{c|}{CURE} & \multicolumn{5}{c|}{CluDataSE} & \multicolumn{2}{c|}{LW-coreset} \\ \cline{3-21}
\multicolumn{1}{|c|}{} & \multicolumn{1}{c|}{} & \multicolumn{1}{c|}{$s$} & \multicolumn{1}{c|}{$n_{s}$} & \multicolumn{1}{c|}{$T_1$} & \multicolumn{1}{c|}{$T_2$} & \multicolumn{1}{c|}{$n_{d}$} & \multicolumn{1}{c|}{$n_{d}$} & \multicolumn{1}{c|}{$n_{d}$} & \multicolumn{1}{c|}{$n_{s}$} & \multicolumn{1}{c|}{$n_{d}$} & \multicolumn{1}{c|}{$n_{d}$} & \multicolumn{1}{c|}{$s$} & \multicolumn{1}{c|}{$n_{d}$} & \multicolumn{1}{c|}{$s$} & \multicolumn{1}{c|}{$eps$} & \multicolumn{1}{c|}{$min\_pts$} & \multicolumn{1}{c|}{$sf$} & \multicolumn{1}{c|}{$n_{d}$} & \multicolumn{1}{c|}{$s$} & \multicolumn{1}{c|}{$n_{d}$} \\
\hline
2 & 30 & 14979 & 2454 & 0.23 & 0.77 & 2.7E+08 & 6.9E+05 & 9.9E+05 & 18 & 5.5E+05 & 9.0E+04 & 8000 & 1.4E+07 & 8000 & 0.003 & 16 & 0.5 & 6.5E+07 & 8000 & 1.1E+05 \\
3 & 30 & 14979 & 2034 & 0.53 & 0.47 & 3.2E+08 & 9.4E+05 & 1.5E+06 & 18 & 7.9E+05 & 5.2E+05 & 8000 & 1.3E+07 & 8000 & 0.003 & 16 & 0.5 & 6.6E+07 & 8000 & 6.3E+05 \\
4 & 30 & 14979 & 2214 & 0.93 & 0.07 & 3.6E+08 & 1.5E+06 & 2.4E+06 & 20 & 1.2E+06 & 1.4E+06 & 8000 & 1.3E+07 & 8000 & 0.003 & 16 & 0.5 & 6.6E+07 & 8000 & 7.0E+05 \\
5 & 30 & 14979 & 1580 & 0.73 & 0.27 & 3.8E+08 & 2.6E+06 & 4.0E+06 & 22 & 1.6E+06 & 2.5E+06 & 8000 & 1.4E+07 & 8000 & 0.003 & 16 & 0.5 & 6.9E+07 & 8000 & 1.3E+06 \\
10 & 30 & 14979 & 928 & 0.5 & 0.5 & 4.4E+08 & 8.7E+06 & 1.4E+07 & 20 & 3.0E+06 & 1.0E+07 & 8000 & 1.8E+07 & 8000 & 0.003 & 16 & 0.5 & 7.8E+07 & 8000 & 4.3E+06 \\
15 & 30 & 14979 & 518 & 0.33 & 0.67 & 4.6E+08 & 1.2E+07 & 2.0E+07 & 20 & 4.5E+06 & 1.7E+07 & 8000 & 2.3E+07 & 8000 & 0.003 & 16 & 0.5 & 8.7E+07 & 8000 & 8.2E+06 \\
20 & 30 & 14979 & 451 & 0.73 & 0.27 & 4.6E+08 & 1.9E+07 & 3.3E+07 & 18 & 5.4E+06 & 2.2E+07 & 8000 & 2.1E+07 & 8000 & 0.003 & 16 & 0.5 & 1.0E+08 & 8000 & 8.6E+06 \\
25 & 30 & 14979 & 418 & 0.93 & 0.07 & 4.7E+08 & 2.3E+07 & 4.0E+07 & 19 & 7.1E+06 & 3.7E+07 & 8000 & 1.8E+07 & 8000 & 0.003 & 16 & 0.5 & 1.1E+08 & 8000 & 1.1E+07 \\
\hline
\end{tabular}
}

\end{table}

\newpage


\subsection{Pla85900}
Dimensions: $m$ = 85900, $n$ = 2.
\par
Description: a data set contains cities coordinates for traveling salesman problem.

\begin{table}[!htbp]
\centering

\caption{Summary of the results with Pla85900 ($\times10^{15}$)}
\label{TabResultsD22}
\small
\resizebox{!}{\tableheight}{
\begin{tabular}{|l|l|llll|llll|llll|llll|}
\hline
\multicolumn{1}{|c|}{\multirow{3}{*}{$k$}} & \multicolumn{1}{c|}{\multirow{3}{*}{$f^*$}} & \multicolumn{4}{c|}{Big-means} & \multicolumn{4}{c|}{IK-means} & \multicolumn{4}{c|}{BDCSM} & \multicolumn{4}{c|}{Minibatch K-means} \\ \cline{3-18}
\multicolumn{1}{|c|}{} & \multicolumn{1}{c|}{} & \multicolumn{2}{c|}{$\varepsilon$} & \multicolumn{2}{c|}{$t$} & \multicolumn{2}{c|}{$\varepsilon$} & \multicolumn{2}{c|}{$t$} & \multicolumn{2}{c|}{$\varepsilon$} & \multicolumn{2}{c|}{$t$} & \multicolumn{2}{c|}{$\varepsilon$} & \multicolumn{2}{c|}{$t$} \\ \cline{3-18}
\multicolumn{1}{|c|}{} & \multicolumn{1}{c|}{} & \multicolumn{1}{c|}{med} & \multicolumn{1}{c|}{std} & \multicolumn{1}{c|}{med} & \multicolumn{1}{c|}{std} & \multicolumn{1}{c|}{med} & \multicolumn{1}{c|}{std} & \multicolumn{1}{c|}{med} & \multicolumn{1}{c|}{std} & \multicolumn{1}{c|}{med} & \multicolumn{1}{c|}{std} & \multicolumn{1}{c|}{med} & \multicolumn{1}{c|}{std} & \multicolumn{1}{c|}{med} & \multicolumn{1}{c|}{std} & \multicolumn{1}{c|}{med} & \multicolumn{1}{c|}{std} \\ \hline
2 & 3.74908 & 0.01 & 0.01 & 0.89 & 0.45 & 1.84 & 2.65 & 0.12 & 0.02 & 7.14 & 3.16 & 0.01 & 0.0 & 1.49 & 1.36 & 0.08 & 0.03 \\
3 & 2.28057 & 0.02 & 0.02 & 0.58 & 0.42 & 9.0 & 8.72 & 0.1 & 0.03 & 0.01 & 20.75 & 0.02 & 0.0 & 0.57 & 3.06 & 0.11 & 0.03 \\
5 & 1.33972 & 0.04 & 0.03 & 1.06 & 0.4 & 9.35 & 8.03 & 0.16 & 0.02 & 7.63 & 18.79 & 0.03 & 0.01 & 2.02 & 1.32 & 0.11 & 0.04 \\
10 & 0.68294 & 0.12 & 0.17 & 0.92 & 0.4 & 13.68 & 9.38 & 0.29 & 0.03 & 21.1 & 12.44 & 0.07 & 0.02 & 4.0 & 1.73 & 0.18 & 0.06 \\
15 & 0.46029 & 0.24 & 0.19 & 1.03 & 0.42 & 17.58 & 9.01 & 0.45 & 0.08 & 16.73 & 8.27 & 0.11 & 0.03 & 5.32 & 1.54 & 0.17 & 0.04 \\
20 & 0.34988 & 0.29 & 0.13 & 0.81 & 0.39 & 20.07 & 6.98 & 0.58 & 0.11 & 13.01 & 7.03 & 0.14 & 0.03 & 5.2 & 1.48 & 0.21 & 0.05 \\
25 & 0.28259 & 0.56 & 0.43 & 1.04 & 0.46 & 18.78 & 6.48 & 0.72 & 0.12 & 15.33 & 7.15 & 0.18 & 0.04 & 5.34 & 1.59 & 0.23 & 0.06 \\
\hline
\multicolumn{2}{|c|}{Mean:} & \textbf{0.18} & & \textbf{0.9} & & \textbf{12.9} & & \textbf{0.35} & & \textbf{11.56} & & \textbf{0.08} & & \textbf{3.42} & & \textbf{0.16} & \\ \hline
\end{tabular}
}

\medskip

\small
\resizebox{!}{\tableheight}{
\begin{tabular}{|l|l|llll|llll|llll|llll|}
\hline
\multicolumn{1}{|c|}{\multirow{3}{*}{$k$}} & \multicolumn{1}{c|}{\multirow{3}{*}{$f^*$}} & \multicolumn{4}{c|}{K-means++} & \multicolumn{4}{c|}{CURE} & \multicolumn{4}{c|}{CluDataSE} & \multicolumn{4}{c|}{LW-coreset} \\ \cline{3-18}
\multicolumn{1}{|c|}{} & \multicolumn{1}{c|}{} & \multicolumn{2}{c|}{$\varepsilon$} & \multicolumn{2}{c|}{$t$} & \multicolumn{2}{c|}{$\varepsilon$} & \multicolumn{2}{c|}{$t$} & \multicolumn{2}{c|}{$\varepsilon$} & \multicolumn{2}{c|}{$t$} & \multicolumn{2}{c|}{$\varepsilon$} & \multicolumn{2}{c|}{$t$} \\ \cline{3-18}
\multicolumn{1}{|c|}{} & \multicolumn{1}{c|}{} & \multicolumn{1}{c|}{med} & \multicolumn{1}{c|}{std} & \multicolumn{1}{c|}{med} & \multicolumn{1}{c|}{std} & \multicolumn{1}{c|}{med} & \multicolumn{1}{c|}{std} & \multicolumn{1}{c|}{med} & \multicolumn{1}{c|}{std} & \multicolumn{1}{c|}{med} & \multicolumn{1}{c|}{std} & \multicolumn{1}{c|}{med} & \multicolumn{1}{c|}{std} & \multicolumn{1}{c|}{med} & \multicolumn{1}{c|}{std} & \multicolumn{1}{c|}{med} & \multicolumn{1}{c|}{std} \\ \hline
2 & 3.74908 & 0.0 & 0.71 & 0.02 & 0.01 & 30.27 & 7.7 & 163.84 & 6.87 & 0.0 & 0.7 & 0.06 & 0.01 & 0.05 & 0.72 & 0.01 & 0.0 \\
3 & 2.28057 & 0.0 & 0.0 & 0.09 & 0.04 & 69.0 & 13.09 & 160.56 & 7.95 & 0.0 & 0.0 & 0.08 & 0.01 & 0.03 & 0.07 & 0.01 & 0.0 \\
5 & 1.33972 & 0.81 & 0.94 & 0.07 & 0.04 & 65.37 & 28.85 & 161.63 & 10.27 & 0.0 & 1.1 & 0.08 & 0.03 & 0.87 & 1.29 & 0.01 & 0.01 \\
10 & 0.68294 & 0.49 & 0.85 & 0.24 & 0.08 & 32.94 & 9.6 & 181.45 & 8.86 & 0.42 & 0.5 & 0.25 & 0.1 & 0.96 & 0.94 & 0.03 & 0.01 \\
15 & 0.46029 & 0.51 & 0.68 & 0.47 & 0.27 & 30.44 & 10.35 & 185.32 & 13.63 & 0.48 & 0.66 & 0.38 & 0.2 & 1.32 & 1.0 & 0.04 & 0.02 \\
20 & 0.34988 & 0.46 & 0.69 & 0.56 & 0.24 & 27.52 & 7.65 & 201.73 & 9.83 & 0.66 & 0.71 & 0.46 & 0.25 & 1.81 & 0.78 & 0.06 & 0.02 \\
25 & 0.28259 & 0.86 & 0.54 & 0.62 & 0.29 & 27.47 & 5.6 & 183.76 & 16.33 & 0.79 & 0.52 & 0.72 & 0.27 & 2.13 & 0.83 & 0.07 & 0.04 \\
\hline
\multicolumn{2}{|c|}{Mean:} & \textbf{0.45} & & \textbf{0.3} & & \textbf{40.43} & & \textbf{176.9} & & \textbf{0.34} & & \textbf{0.29} & & \textbf{1.02} & & \textbf{0.03} & \\ \hline
\end{tabular}
}

\bigskip

\caption{Clustering details with Pla85900}
\label{TabDetailsD22}
\resizebox{\linewidth}{!}{
\begin{tabular}{|l|l|lllll|l|l|ll|l|ll|lllll|ll|}
\hline
\multicolumn{1}{|c|}{\multirow{2}{*}{$k$}} & \multicolumn{1}{c|}{\multirow{2}{*}{$n_{exec}$}} & \multicolumn{5}{c|}{Big-means} & \multicolumn{1}{c|}{IK-means} & \multicolumn{1}{c|}{BDCSM} & \multicolumn{2}{c|}{Minibatch K-means} & \multicolumn{1}{c|}{K-means++} & \multicolumn{2}{c|}{CURE} & \multicolumn{5}{c|}{CluDataSE} & \multicolumn{2}{c|}{LW-coreset} \\ \cline{3-21}
\multicolumn{1}{|c|}{} & \multicolumn{1}{c|}{} & \multicolumn{1}{c|}{$s$} & \multicolumn{1}{c|}{$n_{s}$} & \multicolumn{1}{c|}{$T_1$} & \multicolumn{1}{c|}{$T_2$} & \multicolumn{1}{c|}{$n_{d}$} & \multicolumn{1}{c|}{$n_{d}$} & \multicolumn{1}{c|}{$n_{d}$} & \multicolumn{1}{c|}{$n_{s}$} & \multicolumn{1}{c|}{$n_{d}$} & \multicolumn{1}{c|}{$n_{d}$} & \multicolumn{1}{c|}{$s$} & \multicolumn{1}{c|}{$n_{d}$} & \multicolumn{1}{c|}{$s$} & \multicolumn{1}{c|}{$eps$} & \multicolumn{1}{c|}{$min\_pts$} & \multicolumn{1}{c|}{$sf$} & \multicolumn{1}{c|}{$n_{d}$} & \multicolumn{1}{c|}{$s$} & \multicolumn{1}{c|}{$n_{d}$} \\
\hline
2 & 40 & 14000 & 2394 & 1.0 & 0.5 & 3.9E+08 & 6.9E+05 & 4.0E+06 & 29 & 8.1E+05 & 4.8E+06 & 14000 & 2.6E+08 & 14000 & 2000.0 & 3 & 0.5 & 2.0E+08 & 14000 & 1.0E+06 \\
3 & 40 & 14000 & 1386 & 1.4 & 0.1 & 6.6E+08 & 1.0E+06 & 1.1E+07 & 32 & 1.3E+06 & 1.4E+07 & 14000 & 2.6E+08 & 14000 & 2000.0 & 3 & 0.5 & 2.1E+08 & 14000 & 2.2E+06 \\
5 & 40 & 14000 & 2182 & 1.05 & 0.45 & 1.0E+09 & 1.7E+06 & 1.7E+07 & 34 & 2.4E+06 & 1.8E+07 & 14000 & 2.6E+08 & 14000 & 2000.0 & 3 & 0.5 & 2.1E+08 & 14000 & 2.9E+06 \\
10 & 40 & 14000 & 1204 & 1.2 & 0.3 & 2.0E+09 & 3.4E+06 & 6.0E+07 & 39 & 5.5E+06 & 8.8E+07 & 14000 & 2.7E+08 & 14000 & 2000.0 & 3 & 0.5 & 2.8E+08 & 14000 & 9.2E+06 \\
15 & 40 & 14000 & 770 & 0.6 & 0.9 & 2.3E+09 & 5.2E+06 & 1.0E+08 & 38 & 8.1E+06 & 1.7E+08 & 14000 & 2.8E+08 & 14000 & 2000.0 & 3 & 0.5 & 3.4E+08 & 14000 & 1.6E+07 \\
20 & 40 & 14000 & 428 & 1.4 & 0.1 & 2.9E+09 & 6.9E+06 & 1.5E+08 & 40 & 1.1E+07 & 2.7E+08 & 14000 & 2.9E+08 & 14000 & 2000.0 & 3 & 0.5 & 4.1E+08 & 14000 & 2.5E+07 \\
25 & 40 & 14000 & 393 & 1.15 & 0.35 & 3.2E+09 & 8.6E+06 & 2.0E+08 & 43 & 1.5E+07 & 3.1E+08 & 14000 & 2.9E+08 & 14000 & 2000.0 & 3 & 0.5 & 5.3E+08 & 14000 & 3.0E+07 \\
\hline
\end{tabular}
}

\end{table}

\newpage


\newpage

\subsection{D15112}
Dimensions: $m$ = 15112, $n$ = 2.
\par
Description: a data set with German cities coordinates for travelling salesman problem.

\begin{table}[!htbp]
\centering

\caption{Summary of the results with D15112 ($\times10^{11}$)}
\label{TabResultsD23}
\small
\resizebox{!}{\tableheight}{
\begin{tabular}{|l|l|llll|llll|llll|llll|}
\hline
\multicolumn{1}{|c|}{\multirow{3}{*}{$k$}} & \multicolumn{1}{c|}{\multirow{3}{*}{$f^*$}} & \multicolumn{4}{c|}{Big-means} & \multicolumn{4}{c|}{IK-means} & \multicolumn{4}{c|}{BDCSM} & \multicolumn{4}{c|}{Minibatch K-means} \\ \cline{3-18}
\multicolumn{1}{|c|}{} & \multicolumn{1}{c|}{} & \multicolumn{2}{c|}{$\varepsilon$} & \multicolumn{2}{c|}{$t$} & \multicolumn{2}{c|}{$\varepsilon$} & \multicolumn{2}{c|}{$t$} & \multicolumn{2}{c|}{$\varepsilon$} & \multicolumn{2}{c|}{$t$} & \multicolumn{2}{c|}{$\varepsilon$} & \multicolumn{2}{c|}{$t$} \\ \cline{3-18}
\multicolumn{1}{|c|}{} & \multicolumn{1}{c|}{} & \multicolumn{1}{c|}{med} & \multicolumn{1}{c|}{std} & \multicolumn{1}{c|}{med} & \multicolumn{1}{c|}{std} & \multicolumn{1}{c|}{med} & \multicolumn{1}{c|}{std} & \multicolumn{1}{c|}{med} & \multicolumn{1}{c|}{std} & \multicolumn{1}{c|}{med} & \multicolumn{1}{c|}{std} & \multicolumn{1}{c|}{med} & \multicolumn{1}{c|}{std} & \multicolumn{1}{c|}{med} & \multicolumn{1}{c|}{std} & \multicolumn{1}{c|}{med} & \multicolumn{1}{c|}{std} \\ \hline
2 & 3.68403 & 0.02 & 0.01 & 0.65 & 0.41 & 2.2 & 12.38 & 0.01 & 0.0 & 0.01 & 0.01 & 0.0 & 0.0 & 0.1 & 0.61 & 0.06 & 0.02 \\
3 & 2.5324 & 0.04 & 0.02 & 1.03 & 0.43 & 8.51 & 7.61 & 0.03 & 0.0 & 0.03 & 0.03 & 0.0 & 0.0 & 0.25 & 4.11 & 0.06 & 0.02 \\
5 & 1.32707 & 0.05 & 0.02 & 1.13 & 0.4 & 27.62 & 19.48 & 0.03 & 0.0 & 0.05 & 6.27 & 0.0 & 0.0 & 0.6 & 3.28 & 0.08 & 0.01 \\
10 & 0.64491 & 0.1 & 0.21 & 1.08 & 0.37 & 25.11 & 14.92 & 0.11 & 0.01 & 0.77 & 3.3 & 0.02 & 0.01 & 4.61 & 3.1 & 0.11 & 0.04 \\
15 & 0.43136 & 0.24 & 0.16 & 0.93 & 0.49 & 20.27 & 10.23 & 0.08 & 0.0 & 4.44 & 2.8 & 0.02 & 0.01 & 5.63 & 2.64 & 0.15 & 0.03 \\
20 & 0.32177 & 0.55 & 0.33 & 1.25 & 0.41 & 18.5 & 4.75 & 0.2 & 0.04 & 1.97 & 1.0 & 0.02 & 0.01 & 6.31 & 1.78 & 0.16 & 0.03 \\
25 & 0.25308 & 0.35 & 0.36 & 0.94 & 0.33 & 22.65 & 4.86 & 0.14 & 0.0 & 3.86 & 3.1 & 0.02 & 0.01 & 6.83 & 2.69 & 0.12 & 0.03 \\
\hline
\multicolumn{2}{|c|}{Mean:} & \textbf{0.19} & & \textbf{1.0} & & \textbf{17.84} & & \textbf{0.09} & & \textbf{1.59} & & \textbf{0.01} & & \textbf{3.48} & & \textbf{0.11} & \\ \hline
\end{tabular}
}

\medskip

\small
\resizebox{!}{\tableheight}{
\begin{tabular}{|l|l|llll|llll|llll|llll|}
\hline
\multicolumn{1}{|c|}{\multirow{3}{*}{$k$}} & \multicolumn{1}{c|}{\multirow{3}{*}{$f^*$}} & \multicolumn{4}{c|}{K-means++} & \multicolumn{4}{c|}{CURE} & \multicolumn{4}{c|}{CluDataSE} & \multicolumn{4}{c|}{LW-coreset} \\ \cline{3-18}
\multicolumn{1}{|c|}{} & \multicolumn{1}{c|}{} & \multicolumn{2}{c|}{$\varepsilon$} & \multicolumn{2}{c|}{$t$} & \multicolumn{2}{c|}{$\varepsilon$} & \multicolumn{2}{c|}{$t$} & \multicolumn{2}{c|}{$\varepsilon$} & \multicolumn{2}{c|}{$t$} & \multicolumn{2}{c|}{$\varepsilon$} & \multicolumn{2}{c|}{$t$} \\ \cline{3-18}
\multicolumn{1}{|c|}{} & \multicolumn{1}{c|}{} & \multicolumn{1}{c|}{med} & \multicolumn{1}{c|}{std} & \multicolumn{1}{c|}{med} & \multicolumn{1}{c|}{std} & \multicolumn{1}{c|}{med} & \multicolumn{1}{c|}{std} & \multicolumn{1}{c|}{med} & \multicolumn{1}{c|}{std} & \multicolumn{1}{c|}{med} & \multicolumn{1}{c|}{std} & \multicolumn{1}{c|}{med} & \multicolumn{1}{c|}{std} & \multicolumn{1}{c|}{med} & \multicolumn{1}{c|}{std} & \multicolumn{1}{c|}{med} & \multicolumn{1}{c|}{std} \\ \hline
2 & 3.68403 & 0.0 & 0.0 & 0.0 & 0.0 & 2.45 & 20.97 & 10.62 & 0.72 & 0.0 & 0.0 & 0.03 & 0.0 & 0.03 & 0.02 & 0.0 & 0.0 \\
3 & 2.5324 & 0.0 & 0.0 & 0.01 & 0.0 & 30.34 & 15.04 & 10.42 & 0.71 & 0.0 & 0.0 & 0.06 & 0.0 & 0.07 & 0.04 & 0.01 & 0.0 \\
5 & 1.32707 & -0.0 & 0.0 & 0.01 & 0.01 & 52.77 & 26.89 & 11.11 & 0.65 & -0.0 & 0.0 & 0.04 & 0.0 & 0.09 & 0.07 & 0.01 & 0.0 \\
10 & 0.64491 & 0.63 & 1.38 & 0.03 & 0.01 & 40.22 & 13.27 & 12.59 & 0.63 & 3.83 & 1.46 & 0.08 & 0.01 & 1.7 & 2.67 & 0.01 & 0.01 \\
15 & 0.43136 & 0.59 & 1.47 & 0.04 & 0.03 & 32.88 & 7.82 & 15.28 & 0.66 & 6.89 & 2.4 & 0.11 & 0.04 & 4.14 & 2.34 & 0.02 & 0.0 \\
20 & 0.32177 & 1.99 & 1.56 & 0.07 & 0.02 & 36.48 & 8.62 & 13.69 & 0.74 & 1.37 & 3.24 & 0.09 & 0.01 & 3.77 & 1.94 & 0.02 & 0.01 \\
25 & 0.25308 & 2.01 & 1.79 & 0.08 & 0.05 & 36.09 & 8.91 & 12.01 & 0.66 & 7.88 & 2.51 & 0.12 & 0.03 & 5.67 & 2.64 & 0.02 & 0.01 \\
\hline
\multicolumn{2}{|c|}{Mean:} & \textbf{0.74} & & \textbf{0.03} & & \textbf{33.03} & & \textbf{12.25} & & \textbf{2.85} & & \textbf{0.08} & & \textbf{2.21} & & \textbf{0.01} & \\ \hline
\end{tabular}
}

\bigskip

\caption{Clustering details with D15112}
\label{TabDetailsD23}
\resizebox{\linewidth}{!}{
\begin{tabular}{|l|l|lllll|l|l|ll|l|ll|lllll|ll|}
\hline
\multicolumn{1}{|c|}{\multirow{2}{*}{$k$}} & \multicolumn{1}{c|}{\multirow{2}{*}{$n_{exec}$}} & \multicolumn{5}{c|}{Big-means} & \multicolumn{1}{c|}{IK-means} & \multicolumn{1}{c|}{BDCSM} & \multicolumn{2}{c|}{Minibatch K-means} & \multicolumn{1}{c|}{K-means++} & \multicolumn{2}{c|}{CURE} & \multicolumn{5}{c|}{CluDataSE} & \multicolumn{2}{c|}{LW-coreset} \\ \cline{3-21}
\multicolumn{1}{|c|}{} & \multicolumn{1}{c|}{} & \multicolumn{1}{c|}{$s$} & \multicolumn{1}{c|}{$n_{s}$} & \multicolumn{1}{c|}{$T_1$} & \multicolumn{1}{c|}{$T_2$} & \multicolumn{1}{c|}{$n_{d}$} & \multicolumn{1}{c|}{$n_{d}$} & \multicolumn{1}{c|}{$n_{d}$} & \multicolumn{1}{c|}{$n_{s}$} & \multicolumn{1}{c|}{$n_{d}$} & \multicolumn{1}{c|}{$n_{d}$} & \multicolumn{1}{c|}{$s$} & \multicolumn{1}{c|}{$n_{d}$} & \multicolumn{1}{c|}{$s$} & \multicolumn{1}{c|}{$eps$} & \multicolumn{1}{c|}{$min\_pts$} & \multicolumn{1}{c|}{$sf$} & \multicolumn{1}{c|}{$n_{d}$} & \multicolumn{1}{c|}{$s$} & \multicolumn{1}{c|}{$n_{d}$} \\
\hline
2 & 15 & 8000 & 6558 & 0.95 & 0.55 & 7.6E+08 & 1.2E+05 & 2.5E+05 & 25 & 4.0E+05 & 4.5E+05 & 8000 & 7.6E+07 & 8000 & 200.0 & 16 & 0.5 & 6.4E+07 & 8000 & 2.7E+05 \\
3 & 15 & 8000 & 8391 & 0.7 & 0.8 & 1.2E+09 & 1.8E+05 & 8.4E+05 & 25 & 6.0E+05 & 1.9E+06 & 8000 & 7.4E+07 & 8000 & 200.0 & 16 & 0.5 & 6.6E+07 & 8000 & 9.2E+05 \\
5 & 15 & 8000 & 7163 & 0.7 & 0.8 & 1.4E+09 & 3.0E+05 & 1.1E+06 & 25 & 1.0E+06 & 1.4E+06 & 8000 & 7.6E+07 & 8000 & 200.0 & 16 & 0.5 & 6.7E+07 & 8000 & 1.1E+06 \\
10 & 15 & 8000 & 3244 & 0.9 & 0.6 & 2.2E+09 & 6.0E+05 & 3.0E+06 & 35 & 2.8E+06 & 8.3E+06 & 8000 & 8.0E+07 & 8000 & 200.0 & 16 & 0.5 & 7.3E+07 & 8000 & 3.1E+06 \\
15 & 15 & 8000 & 1584 & 1.35 & 0.15 & 2.5E+09 & 9.1E+05 & 6.3E+06 & 34 & 4.1E+06 & 1.2E+07 & 8000 & 8.4E+07 & 8000 & 200.0 & 16 & 0.5 & 8.2E+07 & 8000 & 5.7E+06 \\
20 & 15 & 8000 & 1371 & 1.35 & 0.15 & 2.8E+09 & 1.2E+06 & 1.1E+07 & 33 & 5.3E+06 & 1.8E+07 & 8000 & 8.2E+07 & 8000 & 200.0 & 16 & 0.5 & 9.5E+07 & 8000 & 7.1E+06 \\
25 & 15 & 8000 & 946 & 0.85 & 0.65 & 3.1E+09 & 1.5E+06 & 1.2E+07 & 26 & 5.2E+06 & 2.4E+07 & 8000 & 8.0E+07 & 8000 & 200.0 & 16 & 0.5 & 1.1E+08 & 8000 & 9.2E+06 \\
\hline
\end{tabular}
}

\end{table}

\newpage


\newpage

\begin{figure}[htb]
\centering
\begin{subfigure}[b]{0.32\linewidth}
\centering
\begin{tikzpicture}[scale=0.6]
\begin{axis}[xlabel={No of clusters}, ylabel={No of dist. func. eval.}, legend pos=north west, legend style={nodes={scale=0.5, transform shape}}]
\addplot[plotStyle1] coordinates {
(2, 168751232.0)
(3, 218726848.0)
(5, 282326080.0)
(10, 360684160.0)
(15, 368002240.0)
(20, 350040320.0)
(25, 365678400.0)
};
\addlegendentry{Big-means}
\addplot[plotStyle2] coordinates {
(2, 4805506.0)
(3, 7851292.0)
(5, 12700145.0)
(10, 24803960.0)
(15, 37097749.0)
(20, 49241940.0)
(25, 61633209.0)
};
\addlegendentry{IK-means}
\addplot[plotStyle3] coordinates {
(2, 12271448.0)
(3, 50855334.0)
(5, 108440780.0)
(10, 437694160.0)
(15, 730454490.0)
(20, 1038631520.0)
(25, 1442280400.0)
};
\addlegendentry{BDCSM}
\addplot[plotStyle4] coordinates {
(2, 2688000.0)
(3, 6528000.0)
(5, 12960000.0)
(10, 22720000.0)
(15, 35040000.0)
(20, 55040000.0)
(25, 60000000.0)
};
\addlegendentry{Minibatch K-means}
\addplot[plotStyle5] coordinates {
(2, 14390784.0)
(3, 55764288.0)
(5, 161896320.0)
(10, 857450880.0)
(15, 1394107200.0)
(20, 1499040000.0)
(25, 3717619200.0)
};
\addlegendentry{K-means++}
\addplot[plotStyle6] coordinates {
(2, 71275281.0)
(3, 42066301.0)
(5, 36114351.0)
(10, 53390620.0)
(15, 75692527.0)
(20, 98305669.0)
(25, 121466233.0)
};
\addlegendentry{CURE}
\addplot[plotStyle7] coordinates {
(2, 1030443680.0)
(3, 1076959840.0)
(5, 1131112870.0)
(10, 1749236260.0)
(15, 2532904825.0)
(20, 2935631120.0)
(25, 3584340700.0)
};
\addlegendentry{CluDataSE}
\addplot[plotStyle8] coordinates {
(2, 2974464.0)
(3, 5782080.0)
(5, 9797312.0)
(10, 22875392.0)
(15, 66833472.0)
(20, 45191552.0)
(25, 98589632.0)
};
\addlegendentry{LW-coreset}
\end{axis}
\end{tikzpicture}
\caption{CORD-19 Embeddings}
\label{Fig1Exp2Ds1}
\end{subfigure}%
\begin{subfigure}[b]{0.32\linewidth}
\centering
\begin{tikzpicture}[scale=0.6]
\begin{axis}[xlabel={No of clusters}, ylabel={No of dist. func. eval.}, legend pos=north west, legend style={nodes={scale=0.5, transform shape}}]
\addplot[plotStyle1] coordinates {
(2, 113928000.0)
(3, 196364000.0)
(5, 374996000.0)
(10, 796456000.0)
(15, 1403516000.0)
(20, 1985616000.0)
(25, 2591076000.0)
};
\addlegendentry{Big-means}
\addplot[plotStyle2] coordinates {
(2, 96704795.0)
(3, 154550801.0)
(5, 244771597.0)
(10, 483431064.0)
(15, 719658304.0)
(20, 958805632.0)
(25, 1196993535.0)
};
\addlegendentry{IK-means}
\addplot[plotStyle3] coordinates {
(2, 405641312.0)
(3, 1161616428.0)
(5, 2716200500.0)
(10, 10681114800.0)
(15, 21848330400.0)
(20, 29234918400.0)
(25, 36882930000.0)
};
\addlegendentry{BDCSM}
\addplot[plotStyle4] coordinates {
(2, 42496000.0)
(3, 75648000.0)
(5, 164800000.0)
(10, 336000000.0)
(15, 549120000.0)
(20, 698880000.0)
(25, 817600000.0)
};
\addlegendentry{Minibatch K-means}
\addplot[plotStyle5] coordinates {
(2, 714000000.0)
(3, 1512000000.0)
(5, 4200000000.0)
(10, 24045000000.0)
(15, 44415000000.0)
(20, 82530000000.0)
(25, 88987500000.0)
};
\addlegendentry{K-means++}
\addplot[plotStyle6] coordinates {
(2, 185157361.0)
(3, 135854987.0)
(5, 135032876.0)
(10, 209608988.0)
(15, 294894749.0)
(20, 381674737.0)
(25, 466659250.0)
};
\addlegendentry{CURE}
\addplot[plotStyle7] coordinates {
(2, 4560408668.0)
(3, 5115217729.0)
(5, 6901011970.0)
(10, 17802382390.0)
(15, 38001626185.0)
(20, 55022458240.0)
(25, 82511260000.0)
};
\addlegendentry{CluDataSE}
\addplot[plotStyle8] coordinates {
(2, 44432000.0)
(3, 59604000.0)
(5, 85980000.0)
(10, 194480000.0)
(15, 290820000.0)
(20, 401240000.0)
(25, 521900000.0)
};
\addlegendentry{LW-coreset}
\end{axis}
\end{tikzpicture}
\caption{HEPMASS}
\label{Fig1Exp2Ds2}
\end{subfigure}%
\begin{subfigure}[b]{0.32\linewidth}
\centering
\begin{tikzpicture}[scale=0.6]
\begin{axis}[xlabel={No of clusters}, ylabel={No of dist. func. eval.}, legend pos=north west, legend style={nodes={scale=0.5, transform shape}}]
\addplot[plotStyle1] coordinates {
(2, 12806570.0)
(3, 19149855.0)
(5, 35385425.0)
(10, 87816850.0)
(15, 133888275.0)
(20, 177469700.0)
(25, 217061125.0)
};
\addlegendentry{Big-means}
\addplot[plotStyle2] coordinates {
(2, 19666284.0)
(3, 29499928.5)
(5, 55779471.0)
(10, 108644965.0)
(15, 160914233.0)
(20, 213442708.5)
(25, 262833274.5)
};
\addlegendentry{IK-means}
\addplot[plotStyle3] coordinates {
(2, 19051478.0)
(3, 44015310.0)
(5, 148284475.0)
(10, 506443735.0)
(15, 916509930.0)
(20, 1293403520.0)
(25, 1692716800.0)
};
\addlegendentry{BDCSM}
\addplot[plotStyle4] coordinates {
(2, 216000.0)
(3, 603000.0)
(5, 10485000.0)
(10, 18390000.0)
(15, 27045000.0)
(20, 50640000.0)
(25, 64950000.0)
};
\addlegendentry{Minibatch K-means}
\addplot[plotStyle5] coordinates {
(2, 14749710.0)
(3, 22124565.0)
(5, 159788525.0)
(10, 909565450.0)
(15, 1345911037.5)
(20, 2654947800.0)
(25, 3963984562.5)
};
\addlegendentry{K-means++}
\addplot[plotStyle6] coordinates {
(2, 21738223.0)
(3, 21805624.5)
(5, 26489249.0)
(10, 39944093.5)
(15, 53030813.5)
(20, 64240279.5)
(25, 75586268.5)
};
\addlegendentry{CURE}
\addplot[plotStyle7] coordinates {
(2, 45842854.0)
(3, 50764204.5)
(5, 146663175.0)
(10, 650758220.0)
(15, 1474349782.5)
(20, 2519192610.0)
(25, 4031177475.0)
};
\addlegendentry{CluDataSE}
\addplot[plotStyle8] coordinates {
(2, 9881140.0)
(3, 12363425.0)
(5, 17417995.0)
(10, 30279420.0)
(15, 42960845.0)
(20, 55822270.0)
(25, 68323695.0)
};
\addlegendentry{LW-coreset}
\end{axis}
\end{tikzpicture}
\caption{US Census Data 1990}
\label{Fig1Exp2Ds3}
\end{subfigure}%
\vskip\baselineskip%
\begin{subfigure}[b]{0.32\linewidth}
\centering
\begin{tikzpicture}[scale=0.6]
\begin{axis}[xlabel={No of clusters}, ylabel={No of dist. func. eval.}, legend pos=north west, legend style={nodes={scale=0.5, transform shape}}]
\addplot[plotStyle1] coordinates {
(2, 3747000.0)
(3, 4170500.0)
(5, 6307500.0)
(10, 16275000.0)
(15, 25992500.0)
(20, 33910000.0)
(25, 42677500.0)
};
\addlegendentry{Big-means}
\addplot[plotStyle2] coordinates {
(2, 108004.0)
(3, 162012.0)
(5, 270010.0)
(10, 540040.0)
(15, 810060.0)
(20, 1080100.0)
(25, 1350100.0)
};
\addlegendentry{IK-means}
\addplot[plotStyle3] coordinates {
(2, 447008.0)
(3, 1780518.0)
(5, 2867550.0)
(10, 5835200.0)
(15, 9652950.0)
(20, 13870800.0)
(25, 15088750.0)
};
\addlegendentry{BDCSM}
\addplot[plotStyle4] coordinates {
(2, 520000.0)
(3, 720000.0)
(5, 1550000.0)
(10, 2600000.0)
(15, 4500000.0)
(20, 5600000.0)
(25, 7500000.0)
};
\addlegendentry{Minibatch K-means}
\addplot[plotStyle5] coordinates {
(2, 702000.0)
(3, 1782000.0)
(5, 4590000.0)
(10, 10665000.0)
(15, 17212500.0)
(20, 21600000.0)
(25, 24300000.0)
};
\addlegendentry{K-means++}
\addplot[plotStyle6] coordinates {
(2, 5479856.0)
(3, 5514953.0)
(5, 7874094.0)
(10, 15452305.0)
(15, 23229295.0)
(20, 25692801.0)
(25, 20592396.0)
};
\addlegendentry{CURE}
\addplot[plotStyle7] coordinates {
(2, 101732800.0)
(3, 101995660.0)
(5, 108300580.0)
(10, 115485950.0)
(15, 116198830.0)
(20, 125751940.0)
(25, 133870375.0)
};
\addlegendentry{CluDataSE}
\addplot[plotStyle8] coordinates {
(2, 514000.0)
(3, 1777500.0)
(5, 2844500.0)
(10, 6362000.0)
(15, 7429500.0)
(20, 9297000.0)
(25, 11864500.0)
};
\addlegendentry{LW-coreset}
\end{axis}
\end{tikzpicture}
\caption{Gisette}
\label{Fig1Exp2Ds4}
\end{subfigure}%
\begin{subfigure}[b]{0.32\linewidth}
\centering
\begin{tikzpicture}[scale=0.6]
\begin{axis}[xlabel={No of clusters}, ylabel={No of dist. func. eval.}, legend pos=north west, legend style={nodes={scale=0.5, transform shape}}]
\addplot[plotStyle1] coordinates {
(2, 90189148.0)
(3, 102481722.0)
(5, 110296870.0)
(10, 116829740.0)
(15, 117827610.0)
(20, 113275480.0)
(25, 115143350.0)
};
\addlegendentry{Big-means}
\addplot[plotStyle2] coordinates {
(2, 852608.0)
(3, 1278912.0)
(5, 2131565.0)
(10, 4263280.0)
(15, 6394987.5)
(20, 8526870.0)
(25, 10658700.0)
};
\addlegendentry{IK-means}
\addplot[plotStyle3] coordinates {
(2, 3483352.0)
(3, 8357257.5)
(5, 16974570.0)
(10, 60535090.0)
(15, 111973035.0)
(20, 151227080.0)
(25, 195032787.5)
};
\addlegendentry{BDCSM}
\addplot[plotStyle4] coordinates {
(2, 450000.0)
(3, 594000.0)
(5, 1350000.0)
(10, 2580000.0)
(15, 3870000.0)
(20, 5160000.0)
(25, 4800000.0)
};
\addlegendentry{Minibatch K-means}
\addplot[plotStyle5] coordinates {
(2, 4902404.0)
(3, 7513467.0)
(5, 28508545.0)
(10, 117231400.0)
(15, 208618605.0)
(20, 318656260.0)
(25, 402316850.0)
};
\addlegendentry{K-means++}
\addplot[plotStyle6] coordinates {
(2, 10328789.5)
(3, 8877850.5)
(5, 8620519.5)
(10, 11031587.5)
(15, 13075023.5)
(20, 11584338.0)
(25, 10755138.0)
};
\addlegendentry{CURE}
\addplot[plotStyle7] coordinates {
(2, 40923324.0)
(3, 54290197.5)
(5, 73128920.0)
(10, 157951740.0)
(15, 356211345.0)
(20, 402076150.0)
(25, 611911975.0)
};
\addlegendentry{CluDataSE}
\addplot[plotStyle8] coordinates {
(2, 630296.0)
(3, 874870.0)
(5, 1871018.0)
(10, 4788888.0)
(15, 7256758.0)
(20, 10984628.0)
(25, 14727498.0)
};
\addlegendentry{LW-coreset}
\end{axis}
\end{tikzpicture}
\caption{Music Analysis}
\label{Fig1Exp2Ds5}
\end{subfigure}%
\begin{subfigure}[b]{0.32\linewidth}
\centering
\begin{tikzpicture}[scale=0.6]
\begin{axis}[xlabel={No of clusters}, ylabel={No of dist. func. eval.}, legend pos=north west, legend style={nodes={scale=0.5, transform shape}}]
\addplot[plotStyle1] coordinates {
(2, 228435502.0)
(3, 261621253.0)
(5, 291592755.0)
(10, 216161510.0)
(15, 290250265.0)
(20, 386179020.0)
(25, 450747775.0)
};
\addlegendentry{Big-means}
\addplot[plotStyle2] coordinates {
(2, 1166022.0)
(3, 1749069.0)
(5, 2915220.0)
(10, 5830640.0)
(15, 8746185.0)
(20, 11661860.0)
(25, 14577475.0)
};
\addlegendentry{IK-means}
\addplot[plotStyle3] coordinates {
(2, 5443518.0)
(3, 17741289.0)
(5, 41888905.0)
(10, 167778110.0)
(15, 323908065.0)
(20, 491238220.0)
(25, 707850025.0)
};
\addlegendentry{BDCSM}
\addplot[plotStyle4] coordinates {
(2, 2128000.0)
(3, 4536000.0)
(5, 6720000.0)
(10, 15680000.0)
(15, 25200000.0)
(20, 31360000.0)
(25, 51800000.0)
};
\addlegendentry{Minibatch K-means}
\addplot[plotStyle5] coordinates {
(2, 2040514.0)
(3, 11368578.0)
(5, 40810280.0)
(10, 174901200.0)
(15, 406645290.0)
(20, 606324160.0)
(25, 659523275.0)
};
\addlegendentry{K-means++}
\addplot[plotStyle6] coordinates {
(2, 365611131.0)
(3, 245035297.0)
(5, 172291809.0)
(10, 153440267.0)
(15, 168939815.0)
(20, 189850361.0)
(25, 210983073.0)
};
\addlegendentry{CURE}
\addplot[plotStyle7] coordinates {
(2, 3144361954.0)
(3, 3161170864.0)
(5, 3206503360.0)
(10, 3414107540.0)
(15, 3837574405.0)
(20, 4044552800.0)
(25, 4439824175.0)
};
\addlegendentry{CluDataSE}
\addplot[plotStyle8] coordinates {
(2, 1031004.0)
(3, 4088755.0)
(5, 15580257.0)
(10, 61669012.0)
(15, 125957767.0)
(20, 164486522.0)
(25, 211135277.0)
};
\addlegendentry{LW-coreset}
\end{axis}
\end{tikzpicture}
\caption{Protein Homology}
\label{Fig1Exp2Ds6}
\end{subfigure}%
\vskip\baselineskip%
\begin{subfigure}[b]{0.32\linewidth}
\centering
\begin{tikzpicture}[scale=0.6]
\begin{axis}[xlabel={No of clusters}, ylabel={No of dist. func. eval.}, legend pos=north west, legend style={nodes={scale=0.5, transform shape}}]
\addplot[plotStyle1] coordinates {
(2, 154180128.0)
(3, 238940192.0)
(5, 248170320.0)
(10, 346320640.0)
(15, 608270960.0)
(20, 785721280.0)
(25, 1099671600.0)
};
\addlegendentry{Big-means}
\addplot[plotStyle2] coordinates {
(2, 1040552.0)
(3, 1560927.0)
(5, 2601805.0)
(10, 5204670.0)
(15, 7808010.0)
(20, 10412980.0)
(25, 13017325.0)
};
\addlegendentry{IK-means}
\addplot[plotStyle3] coordinates {
(2, 2860136.0)
(3, 8580210.0)
(5, 35750370.0)
(10, 260000840.0)
(15, 546001410.0)
(20, 735802080.0)
(25, 1082252850.0)
};
\addlegendentry{BDCSM}
\addplot[plotStyle4] coordinates {
(2, 8320000.0)
(3, 5850000.0)
(5, 14950000.0)
(10, 24700000.0)
(15, 37050000.0)
(20, 49400000.0)
(25, 84500000.0)
};
\addlegendentry{Minibatch K-means}
\addplot[plotStyle5] coordinates {
(2, 780384.0)
(3, 7023456.0)
(5, 28614080.0)
(10, 212004320.0)
(15, 294594960.0)
(20, 499445760.0)
(25, 640565200.0)
};
\addlegendentry{K-means++}
\addplot[plotStyle6] coordinates {
(2, 444927161.0)
(3, 382389783.0)
(5, 332014906.0)
(10, 302351188.0)
(15, 306206860.0)
(20, 317267760.0)
(25, 325491189.0)
};
\addlegendentry{CURE}
\addplot[plotStyle7] coordinates {
(2, 1600643838.0)
(3, 1605297838.0)
(5, 1637880865.0)
(10, 2036828860.0)
(15, 2215004425.0)
(20, 2443995240.0)
(25, 2759346925.0)
};
\addlegendentry{CluDataSE}
\addplot[plotStyle8] coordinates {
(2, 760256.0)
(3, 2330320.0)
(5, 6710448.0)
(10, 39160768.0)
(15, 88611088.0)
(20, 95661408.0)
(25, 148511728.0)
};
\addlegendentry{LW-coreset}
\end{axis}
\end{tikzpicture}
\caption{MiniBooNE Particle Identification}
\label{Fig1Exp2Ds7}
\end{subfigure}%
\begin{subfigure}[b]{0.32\linewidth}
\centering
\begin{tikzpicture}[scale=0.6]
\begin{axis}[xlabel={No of clusters}, ylabel={No of dist. func. eval.}, legend pos=north west, legend style={nodes={scale=0.5, transform shape}}]
\addplot[plotStyle1] coordinates {
(2, 49604128.0)
(3, 77868192.0)
(5, 103838320.0)
(10, 112168640.0)
(15, 128838960.0)
(20, 112449280.0)
(25, 139559600.0)
};
\addlegendentry{Big-means}
\addplot[plotStyle2] coordinates {
(2, 1932981.0)
(3, 2860326.0)
(5, 5871362.0)
(10, 19322218.5)
(15, 35659530.5)
(20, 54099987.5)
(25, 71859363.5)
};
\addlegendentry{IK-means}
\addplot[plotStyle3] coordinates {
(2, 4088228.0)
(3, 7248462.0)
(5, 18201195.0)
(10, 79185640.0)
(15, 162973335.0)
(20, 229571280.0)
(25, 309167225.0)
};
\addlegendentry{BDCSM}
\addplot[plotStyle4] coordinates {
(2, 348000.0)
(3, 792000.0)
(5, 4410000.0)
(10, 5640000.0)
(15, 7470000.0)
(20, 11280000.0)
(25, 13350000.0)
};
\addlegendentry{Minibatch K-means}
\addplot[plotStyle5] coordinates {
(2, 780384.0)
(3, 8194032.0)
(5, 23086360.0)
(10, 124861440.0)
(15, 238992600.0)
(20, 465629120.0)
(25, 565778400.0)
};
\addlegendentry{K-means++}
\addplot[plotStyle6] coordinates {
(2, 39566878.5)
(3, 32804845.5)
(5, 30392157.5)
(10, 33413502.0)
(15, 38985876.0)
(20, 45184087.0)
(25, 40965428.5)
};
\addlegendentry{CURE}
\addplot[plotStyle7] coordinates {
(2, 144535550.0)
(3, 152271603.0)
(5, 165037812.5)
(10, 299601650.0)
(15, 444588630.0)
(20, 597350480.0)
(25, 887347362.5)
};
\addlegendentry{CluDataSE}
\addplot[plotStyle8] coordinates {
(2, 592256.0)
(3, 1226320.0)
(5, 2560448.0)
(10, 7020768.0)
(15, 15891088.0)
(20, 18461408.0)
(25, 26911728.0)
};
\addlegendentry{LW-coreset}
\end{axis}
\end{tikzpicture}
\caption{MiniBooNE Particle Identification (normalized)}
\label{Fig1Exp2Ds8}
\end{subfigure}%
\begin{subfigure}[b]{0.32\linewidth}
\centering
\begin{tikzpicture}[scale=0.6]
\begin{axis}[xlabel={No of clusters}, ylabel={No of dist. func. eval.}, legend pos=north west, legend style={nodes={scale=0.5, transform shape}}]
\addplot[plotStyle1] coordinates {
(2, 66266268.0)
(3, 81225402.0)
(5, 95693670.0)
(10, 108779340.0)
(15, 80195010.0)
(20, 112870680.0)
(25, 127186350.0)
};
\addlegendentry{Big-means}
\addplot[plotStyle2] coordinates {
(2, 681185.0)
(3, 1021996.5)
(5, 1703875.0)
(10, 3410050.0)
(15, 5120047.5)
(20, 7891892.5)
(25, 10080445.0)
};
\addlegendentry{IK-means}
\addplot[plotStyle3] coordinates {
(2, 3254352.0)
(3, 4791591.0)
(5, 14736195.0)
(10, 48374140.0)
(15, 89936460.0)
(20, 178719480.0)
(25, 208402412.5)
};
\addlegendentry{BDCSM}
\addplot[plotStyle4] coordinates {
(2, 600000.0)
(3, 1278000.0)
(5, 2190000.0)
(10, 4440000.0)
(15, 7560000.0)
(20, 9480000.0)
(25, 9450000.0)
};
\addlegendentry{Minibatch K-means}
\addplot[plotStyle5] coordinates {
(2, 3916164.0)
(3, 5618844.0)
(5, 16388295.0)
(10, 54060090.0)
(15, 116207910.0)
(20, 194105520.0)
(25, 297969000.0)
};
\addlegendentry{K-means++}
\addplot[plotStyle6] coordinates {
(2, 61844169.5)
(3, 51720903.0)
(5, 43615406.5)
(10, 39667315.5)
(15, 42168638.5)
(20, 46057854.5)
(25, 42635544.5)
};
\addlegendentry{CURE}
\addplot[plotStyle7] coordinates {
(2, 147604303.0)
(3, 149004373.5)
(5, 163492315.0)
(10, 208952865.0)
(15, 301551315.0)
(20, 419794180.0)
(25, 556889000.0)
};
\addlegendentry{CluDataSE}
\addplot[plotStyle8] coordinates {
(2, 796536.0)
(3, 1073670.0)
(5, 2515938.0)
(10, 7321608.0)
(15, 12787278.0)
(20, 19752948.0)
(25, 29298618.0)
};
\addlegendentry{LW-coreset}
\end{axis}
\end{tikzpicture}
\caption{MFCCs for Speech Emotion Recognition}
\label{Fig1Exp2Ds9}
\end{subfigure}%
\caption{Distance function evaluations. Set 1}
\end{figure}
\begin{figure}[htb]
\centering
\begin{subfigure}[b]{0.32\linewidth}
\centering
\begin{tikzpicture}[scale=0.6]
\begin{axis}[xlabel={No of clusters}, ylabel={No of dist. func. eval.}, legend pos=north west, legend style={nodes={scale=0.5, transform shape}}]
\addplot[plotStyle1] coordinates {
(2, 51087594.0)
(3, 55399391.0)
(5, 58454985.0)
(10, 60053970.0)
(15, 53032955.0)
(20, 53611940.0)
(25, 53430925.0)
};
\addlegendentry{Big-means}
\addplot[plotStyle2] coordinates {
(2, 62618.0)
(3, 103260.0)
(5, 168108.0)
(10, 332115.0)
(15, 503297.0)
(20, 679819.0)
(25, 850429.0)
};
\addlegendentry{IK-means}
\addplot[plotStyle3] coordinates {
(2, 103602.0)
(3, 275409.0)
(5, 599035.0)
(10, 1118170.0)
(15, 1497405.0)
(20, 2156740.0)
(25, 2696175.0)
};
\addlegendentry{BDCSM}
\addplot[plotStyle4] coordinates {
(2, 160000.0)
(3, 264000.0)
(5, 660000.0)
(10, 1320000.0)
(15, 1740000.0)
(20, 2640000.0)
(25, 2700000.0)
};
\addlegendentry{Minibatch K-means}
\addplot[plotStyle5] coordinates {
(2, 187128.0)
(3, 514602.0)
(5, 1052595.0)
(10, 1871280.0)
(15, 3040830.0)
(20, 4834140.0)
(25, 7212225.0)
};
\addlegendentry{K-means++}
\addplot[plotStyle6] coordinates {
(2, 2474520.0)
(3, 2335515.0)
(5, 2901403.0)
(10, 5347667.0)
(15, 4730936.0)
(20, 3773142.0)
(25, 3198486.0)
};
\addlegendentry{CURE}
\addplot[plotStyle7] coordinates {
(2, 16140936.0)
(3, 16487143.0)
(5, 17209210.0)
(10, 18940580.0)
(15, 19434130.0)
(20, 21646040.0)
(25, 23861925.0)
};
\addlegendentry{CluDataSE}
\addplot[plotStyle8] coordinates {
(2, 119188.0)
(3, 278985.0)
(5, 594579.0)
(10, 1213564.0)
(15, 1752549.0)
(20, 2091534.0)
(25, 2710519.0)
};
\addlegendentry{LW-coreset}
\end{axis}
\end{tikzpicture}
\caption{ISOLET}
\label{Fig1Exp2Ds10}
\end{subfigure}%
\begin{subfigure}[b]{0.32\linewidth}
\centering
\begin{tikzpicture}[scale=0.6]
\begin{axis}[xlabel={No of clusters}, ylabel={No of dist. func. eval.}, legend pos=north west, legend style={nodes={scale=0.5, transform shape}}]
\addplot[plotStyle1] coordinates {
(2, 88627518.0)
(3, 99713277.0)
(5, 111969045.0)
(10, 120451590.0)
(15, 221071635.0)
(20, 274599180.0)
(25, 348601725.0)
};
\addlegendentry{Big-means}
\addplot[plotStyle2] coordinates {
(2, 469261.0)
(3, 781724.0)
(5, 1286628.5)
(10, 2525383.0)
(15, 3774604.5)
(20, 5025899.0)
(25, 6312276.0)
};
\addlegendentry{IK-means}
\addplot[plotStyle3] coordinates {
(2, 4036526.0)
(3, 9828045.0)
(5, 15063845.0)
(10, 59962790.0)
(15, 60548085.0)
(20, 122850980.0)
(25, 220838975.0)
};
\addlegendentry{BDCSM}
\addplot[plotStyle4] coordinates {
(2, 1989000.0)
(3, 3510000.0)
(5, 4972500.0)
(10, 11407500.0)
(15, 17550000.0)
(20, 20475000.0)
(25, 25593750.0)
};
\addlegendentry{Minibatch K-means}
\addplot[plotStyle5] coordinates {
(2, 468072.0)
(3, 5441337.0)
(5, 12286890.0)
(10, 39493575.0)
(15, 56168640.0)
(20, 85423140.0)
(25, 110435737.5)
};
\addlegendentry{K-means++}
\addplot[plotStyle6] coordinates {
(2, 15261239.0)
(3, 14072928.5)
(5, 14345407.0)
(10, 17971595.5)
(15, 22050038.5)
(20, 23396257.5)
(25, 20558038.5)
};
\addlegendentry{CURE}
\addplot[plotStyle7] coordinates {
(2, 104248855.0)
(3, 116919002.5)
(5, 119776855.0)
(10, 161514800.0)
(15, 157189555.0)
(20, 223708070.0)
(25, 344255337.5)
};
\addlegendentry{CluDataSE}
\addplot[plotStyle8] coordinates {
(2, 294036.0)
(3, 922545.0)
(5, 1709563.0)
(10, 5152108.0)
(15, 8269653.0)
(20, 10887198.0)
(25, 13829743.0)
};
\addlegendentry{LW-coreset}
\end{axis}
\end{tikzpicture}
\caption{Sensorless Drive Diagnosis}
\label{Fig1Exp2Ds11}
\end{subfigure}%
\begin{subfigure}[b]{0.32\linewidth}
\centering
\begin{tikzpicture}[scale=0.6]
\begin{axis}[xlabel={No of clusters}, ylabel={No of dist. func. eval.}, legend pos=north west, legend style={nodes={scale=0.5, transform shape}}]
\addplot[plotStyle1] coordinates {
(2, 15258018.0)
(3, 21376777.0)
(5, 27151545.0)
(10, 34024090.0)
(15, 36127885.0)
(20, 35519180.0)
(25, 35006725.0)
};
\addlegendentry{Big-means}
\addplot[plotStyle2] coordinates {
(2, 556030.0)
(3, 929986.5)
(5, 1635021.5)
(10, 3706743.0)
(15, 6234048.0)
(20, 8837694.0)
(25, 11728384.0)
};
\addlegendentry{IK-means}
\addplot[plotStyle3] coordinates {
(2, 1079646.0)
(3, 2711925.0)
(5, 5299145.0)
(10, 13775390.0)
(15, 22590735.0)
(20, 29494980.0)
(25, 36017725.0)
};
\addlegendentry{BDCSM}
\addplot[plotStyle4] coordinates {
(2, 248500.0)
(3, 357000.0)
(5, 717500.0)
(10, 1522500.0)
(15, 2441250.0)
(20, 3080000.0)
(25, 3500000.0)
};
\addlegendentry{Minibatch K-means}
\addplot[plotStyle5] coordinates {
(2, 1170180.0)
(3, 2983959.0)
(5, 6728535.0)
(10, 16967610.0)
(15, 35105400.0)
(20, 46222110.0)
(25, 62897175.0)
};
\addlegendentry{K-means++}
\addplot[plotStyle6] coordinates {
(2, 8709429.0)
(3, 8759230.5)
(5, 9437697.5)
(10, 12372892.0)
(15, 10716468.0)
(20, 9890972.0)
(25, 9175229.0)
};
\addlegendentry{CURE}
\addplot[plotStyle7] coordinates {
(2, 13250757.0)
(3, 15238381.0)
(5, 15776880.0)
(10, 28070075.0)
(15, 45258032.5)
(20, 59677800.0)
(25, 67933300.0)
};
\addlegendentry{CluDataSE}
\addplot[plotStyle8] coordinates {
(2, 297036.0)
(3, 418545.0)
(5, 689563.0)
(10, 1384608.0)
(15, 2202153.0)
(20, 2792198.0)
(25, 3504743.0)
};
\addlegendentry{LW-coreset}
\end{axis}
\end{tikzpicture}
\caption{Sensorless Drive Diagnosis (normalized)}
\label{Fig1Exp2Ds12}
\end{subfigure}%
\vskip\baselineskip%
\begin{subfigure}[b]{0.32\linewidth}
\centering
\begin{tikzpicture}[scale=0.6]
\begin{axis}[xlabel={No of clusters}, ylabel={No of dist. func. eval.}, legend pos=north west, legend style={nodes={scale=0.5, transform shape}}]
\addplot[plotStyle1] coordinates {
(2, 50449288.0)
(3, 64473932.0)
(5, 66813220.0)
(10, 71236440.0)
(15, 62384660.0)
(20, 82432880.0)
(25, 88331100.0)
};
\addlegendentry{Big-means}
\addplot[plotStyle2] coordinates {
(2, 317156.0)
(3, 475734.0)
(5, 792890.0)
(10, 1585780.0)
(15, 2378670.0)
(20, 3171560.0)
(25, 3964450.0)
};
\addlegendentry{IK-means}
\addplot[plotStyle3] coordinates {
(2, 479312.0)
(3, 2039013.0)
(5, 4123482.5)
(10, 15697640.0)
(15, 35097022.5)
(20, 69299480.0)
(25, 88501412.5)
};
\addlegendentry{BDCSM}
\addplot[plotStyle4] coordinates {
(2, 380000.0)
(3, 555000.0)
(5, 1050000.0)
(10, 2950000.0)
(15, 3300000.0)
(20, 4800000.0)
(25, 6250000.0)
};
\addlegendentry{Minibatch K-means}
\addplot[plotStyle5] coordinates {
(2, 317152.0)
(3, 3865290.0)
(5, 3072410.0)
(10, 14271840.0)
(15, 27949020.0)
(20, 36076040.0)
(25, 52528300.0)
};
\addlegendentry{K-means++}
\addplot[plotStyle6] coordinates {
(2, 63773335.5)
(3, 56504686.5)
(5, 47786154.5)
(10, 42738730.5)
(15, 42738633.0)
(20, 43126808.0)
(25, 42449579.0)
};
\addlegendentry{CURE}
\addplot[plotStyle7] coordinates {
(2, 100199997.0)
(3, 103903558.0)
(5, 107352210.0)
(10, 122649225.0)
(15, 166340140.0)
(20, 168905560.0)
(25, 263621950.0)
};
\addlegendentry{CluDataSE}
\addplot[plotStyle8] coordinates {
(2, 218576.0)
(3, 408220.0)
(5, 1177508.0)
(10, 3825728.0)
(15, 5773948.0)
(20, 7672168.0)
(25, 11195388.0)
};
\addlegendentry{LW-coreset}
\end{axis}
\end{tikzpicture}
\caption{Online News Popularity}
\label{Fig1Exp2Ds13}
\end{subfigure}%
\begin{subfigure}[b]{0.32\linewidth}
\centering
\begin{tikzpicture}[scale=0.6]
\begin{axis}[xlabel={No of clusters}, ylabel={No of dist. func. eval.}, legend pos=north west, legend style={nodes={scale=0.5, transform shape}}]
\addplot[plotStyle1] coordinates {
(2, 93987820.0)
(3, 97097730.0)
(5, 104865550.0)
(10, 123745100.0)
(15, 125209650.0)
(20, 102104200.0)
(25, 85141250.0)
};
\addlegendentry{Big-means}
\addplot[plotStyle2] coordinates {
(2, 111284.0)
(3, 166926.0)
(5, 278210.0)
(10, 556420.0)
(15, 834630.0)
(20, 1112840.0)
(25, 1391050.0)
};
\addlegendentry{IK-means}
\addplot[plotStyle3] coordinates {
(2, 279828.0)
(3, 622248.0)
(5, 1374600.0)
(10, 4054300.0)
(15, 6824100.0)
(20, 10989000.0)
(25, 19699000.0)
};
\addlegendentry{BDCSM}
\addplot[plotStyle4] coordinates {
(2, 369000.0)
(3, 472500.0)
(5, 967500.0)
(10, 1755000.0)
(15, 3105000.0)
(20, 4500000.0)
(25, 6075000.0)
};
\addlegendentry{Minibatch K-means}
\addplot[plotStyle5] coordinates {
(2, 445120.0)
(3, 625950.0)
(5, 1773525.0)
(10, 3338400.0)
(15, 6363825.0)
(20, 10015200.0)
(25, 10606375.0)
};
\addlegendentry{K-means++}
\addplot[plotStyle6] coordinates {
(2, 42006385.5)
(3, 42391484.0)
(5, 43351373.5)
(10, 45843623.5)
(15, 51076461.5)
(20, 50508272.5)
(25, 46848739.0)
};
\addlegendentry{CURE}
\addplot[plotStyle7] coordinates {
(2, 81237035.0)
(3, 81752862.0)
(5, 82550937.5)
(10, 86853405.0)
(15, 90085200.0)
(20, 99329680.0)
(25, 109471700.0)
};
\addlegendentry{CluDataSE}
\addplot[plotStyle8] coordinates {
(2, 271640.0)
(3, 393550.0)
(5, 974870.0)
(10, 2551920.0)
(15, 4016470.0)
(20, 5796020.0)
(25, 7463070.0)
};
\addlegendentry{LW-coreset}
\end{axis}
\end{tikzpicture}
\caption{Gas Sensor Array Drift}
\label{Fig1Exp2Ds14}
\end{subfigure}%
\begin{subfigure}[b]{0.32\linewidth}
\centering
\begin{tikzpicture}[scale=0.6]
\begin{axis}[xlabel={No of clusters}, ylabel={No of dist. func. eval.}, legend pos=north west, legend style={nodes={scale=0.5, transform shape}}]
\addplot[plotStyle1] coordinates {
(2, 118069748.0)
(3, 180904622.0)
(5, 315824370.0)
(10, 446248740.0)
(15, 847173110.0)
(20, 973097480.0)
(25, 1311771850.0)
};
\addlegendentry{Big-means}
\addplot[plotStyle2] coordinates {
(2, 3481045.0)
(3, 5228974.5)
(5, 8730542.5)
(10, 21063545.5)
(15, 34156884.5)
(20, 47883644.0)
(25, 58471388.0)
};
\addlegendentry{IK-means}
\addplot[plotStyle3] coordinates {
(2, 17669780.0)
(3, 42854694.0)
(5, 126424620.0)
(10, 1156849940.0)
(15, 2334526710.0)
(20, 4775709480.0)
(25, 7867130600.0)
};
\addlegendentry{BDCSM}
\addplot[plotStyle4] coordinates {
(2, 5400000.0)
(3, 8700000.0)
(5, 16250000.0)
(10, 35000000.0)
(15, 50250000.0)
(20, 70000000.0)
(25, 80000000.0)
};
\addlegendentry{Minibatch K-means}
\addplot[plotStyle5] coordinates {
(2, 21743700.0)
(3, 42400215.0)
(5, 144595605.0)
(10, 1804727100.0)
(15, 1917794340.0)
(20, 6023004900.0)
(25, 5202180225.0)
};
\addlegendentry{K-means++}
\addplot[plotStyle6] coordinates {
(2, 110110613.5)
(3, 106947152.0)
(5, 107069300.0)
(10, 112288265.5)
(15, 120101527.5)
(20, 123884942.5)
(25, 121932107.0)
};
\addlegendentry{CURE}
\addplot[plotStyle7] coordinates {
(2, 121743718.0)
(3, 146966419.0)
(5, 273949682.5)
(10, 2480935760.0)
(15, 3674665292.5)
(20, 11745928520.0)
(25, 12656989562.5)
};
\addlegendentry{CluDataSE}
\addplot[plotStyle8] coordinates {
(2, 2039496.0)
(3, 2894370.0)
(5, 5719118.0)
(10, 18568488.0)
(15, 41667858.0)
(20, 55467228.0)
(25, 67616598.0)
};
\addlegendentry{LW-coreset}
\end{axis}
\end{tikzpicture}
\caption{3D Road Network}
\label{Fig1Exp2Ds15}
\end{subfigure}%
\vskip\baselineskip%
\begin{subfigure}[b]{0.32\linewidth}
\centering
\begin{tikzpicture}[scale=0.6]
\begin{axis}[xlabel={No of clusters}, ylabel={No of dist. func. eval.}, legend pos=north west, legend style={nodes={scale=0.5, transform shape}}]
\addplot[plotStyle1] coordinates {
(2, 11218114.0)
(3, 20863171.0)
(5, 33077285.0)
(10, 71842570.0)
(15, 120427855.0)
(20, 133973140.0)
(25, 164298425.0)
};
\addlegendentry{Big-means}
\addplot[plotStyle2] coordinates {
(2, 1960677.0)
(3, 2941252.5)
(5, 4903830.0)
(10, 9810460.0)
(15, 14740275.0)
(20, 19647060.0)
(25, 24626612.0)
};
\addlegendentry{IK-means}
\addplot[plotStyle3] coordinates {
(2, 5362474.0)
(3, 16144116.0)
(5, 24050517.5)
(10, 65944020.0)
(15, 125079945.0)
(20, 187494510.0)
(25, 233293475.0)
};
\addlegendentry{BDCSM}
\addplot[plotStyle4] coordinates {
(2, 488000.0)
(3, 1128000.0)
(5, 1500000.0)
(10, 3880000.0)
(15, 5880000.0)
(20, 9520000.0)
(25, 10400000.0)
};
\addlegendentry{Minibatch K-means}
\addplot[plotStyle5] coordinates {
(2, 6616539.0)
(3, 19849617.0)
(5, 27568912.5)
(10, 63714820.0)
(15, 126816997.5)
(20, 181342180.0)
(25, 226677725.0)
};
\addlegendentry{K-means++}
\addplot[plotStyle6] coordinates {
(2, 47188509.0)
(3, 41205464.0)
(5, 37510181.5)
(10, 36820043.5)
(15, 39556443.0)
(20, 39510864.0)
(25, 39442870.0)
};
\addlegendentry{CURE}
\addplot[plotStyle7] coordinates {
(2, 69407823.0)
(3, 87573748.0)
(5, 88002477.5)
(10, 157287685.0)
(15, 282937765.0)
(20, 329278190.0)
(25, 475075350.0)
};
\addlegendentry{CluDataSE}
\addplot[plotStyle8] coordinates {
(2, 1140228.0)
(3, 1825285.0)
(5, 2575399.0)
(10, 6380684.0)
(15, 9265969.0)
(20, 11231254.0)
(25, 16316539.0)
};
\addlegendentry{LW-coreset}
\end{axis}
\end{tikzpicture}
\caption{Skin Segmentation}
\label{Fig1Exp2Ds16}
\end{subfigure}%
\begin{subfigure}[b]{0.32\linewidth}
\centering
\begin{tikzpicture}[scale=0.6]
\begin{axis}[xlabel={No of clusters}, ylabel={No of dist. func. eval.}, legend pos=north west, legend style={nodes={scale=0.5, transform shape}}]
\addplot[plotStyle1] coordinates {
(2, 221135876.0)
(3, 265149689.0)
(5, 308576715.0)
(10, 345121780.0)
(15, 324849095.0)
(20, 249892660.0)
(25, 286571100.0)
};
\addlegendentry{Big-means}
\addplot[plotStyle2] coordinates {
(2, 427361.0)
(3, 641502.0)
(5, 1070122.5)
(10, 2557290.0)
(15, 3743038.0)
(20, 4985800.0)
(25, 6220323.0)
};
\addlegendentry{IK-means}
\addplot[plotStyle3] coordinates {
(2, 2027434.0)
(3, 5441907.0)
(5, 15205115.0)
(10, 52550580.0)
(15, 174455895.0)
(20, 251280560.0)
(25, 482153450.0)
};
\addlegendentry{BDCSM}
\addplot[plotStyle4] coordinates {
(2, 2667500.0)
(3, 3040950.0)
(5, 4401375.0)
(10, 11470250.0)
(15, 15204750.0)
(20, 22407000.0)
(25, 27341875.0)
};
\addlegendentry{Minibatch K-means}
\addplot[plotStyle5] coordinates {
(2, 320478.0)
(3, 2724063.0)
(5, 7344287.5)
(10, 20831070.0)
(15, 27641227.5)
(20, 49139960.0)
(25, 72775212.5)
};
\addlegendentry{K-means++}
\addplot[plotStyle6] coordinates {
(2, 22232501.0)
(3, 18146966.0)
(5, 16264768.5)
(10, 20185547.5)
(15, 24781004.0)
(20, 26321931.5)
(25, 22798739.0)
};
\addlegendentry{CURE}
\addplot[plotStyle7] coordinates {
(2, 101868238.0)
(3, 105452845.0)
(5, 115499845.0)
(10, 156787400.0)
(15, 278827540.0)
(20, 428917410.0)
(25, 576368450.0)
};
\addlegendentry{CluDataSE}
\addplot[plotStyle8] coordinates {
(2, 273652.0)
(3, 687065.0)
(5, 1748891.0)
(10, 4340956.0)
(15, 6983021.0)
(20, 9675086.0)
(25, 14692151.0)
};
\addlegendentry{LW-coreset}
\end{axis}
\end{tikzpicture}
\caption{KEGG Metabolic Relation Network (Directed)}
\label{Fig1Exp2Ds17}
\end{subfigure}%
\begin{subfigure}[b]{0.32\linewidth}
\centering
\begin{tikzpicture}[scale=0.6]
\begin{axis}[xlabel={No of clusters}, ylabel={No of dist. func. eval.}, legend pos=north west, legend style={nodes={scale=0.5, transform shape}}]
\addplot[plotStyle1] coordinates {
(2, 548438900.0)
(3, 632060800.0)
(4, 712553400.0)
(5, 804578050.0)
(10, 941340300.0)
(15, 969156550.0)
(20, 1028265800.0)
(25, 940761550.0)
};
\addlegendentry{Big-means}
\addplot[plotStyle2] coordinates {
(2, 464100.0)
(3, 696879.0)
(4, 930040.0)
(5, 1163230.0)
(10, 2537347.0)
(15, 3860701.0)
(20, 5125338.0)
(25, 6522789.0)
};
\addlegendentry{IK-means}
\addplot[plotStyle3] coordinates {
(2, 3477108.0)
(3, 5389518.0)
(4, 6954232.0)
(5, 9851800.0)
(10, 33611700.0)
(15, 50417700.0)
(20, 59110800.0)
(25, 111556250.0)
};
\addlegendentry{BDCSM}
\addplot[plotStyle4] coordinates {
(2, 2433900.0)
(3, 3303150.0)
(4, 4636000.0)
(5, 5505250.0)
(10, 9851500.0)
(15, 14777250.0)
(20, 25498000.0)
(25, 33321250.0)
};
\addlegendentry{Minibatch K-means}
\addplot[plotStyle5] coordinates {
(2, 464000.0)
(3, 870000.0)
(4, 1160000.0)
(5, 1450000.0)
(10, 7540000.0)
(15, 19140000.0)
(20, 34800000.0)
(25, 43500000.0)
};
\addlegendentry{K-means++}
\addplot[plotStyle6] coordinates {
(2, 38108128.0)
(3, 40352883.0)
(4, 42318387.0)
(5, 44088540.0)
(10, 48686302.0)
(15, 53035481.0)
(20, 55180818.0)
(25, 51724677.0)
};
\addlegendentry{CURE}
\addplot[plotStyle7] coordinates {
(2, 102558552.0)
(3, 108188464.0)
(4, 106510760.0)
(5, 109318700.0)
(10, 129045900.0)
(15, 155726680.0)
(20, 162739840.0)
(25, 268374250.0)
};
\addlegendentry{CluDataSE}
\addplot[plotStyle8] coordinates {
(2, 332000.0)
(3, 500000.0)
(4, 668000.0)
(5, 856000.0)
(10, 2296000.0)
(15, 4436000.0)
(20, 7676000.0)
(25, 8816000.0)
};
\addlegendentry{LW-coreset}
\end{axis}
\end{tikzpicture}
\caption{Shuttle Control}
\label{Fig1Exp2Ds18}
\end{subfigure}%
\caption{Distance function evaluations. Set 2}
\end{figure}
\begin{figure}[htb]
\centering
\begin{subfigure}[b]{0.32\linewidth}
\centering
\begin{tikzpicture}[scale=0.6]
\begin{axis}[xlabel={No of clusters}, ylabel={No of dist. func. eval.}, legend pos=north west, legend style={nodes={scale=0.5, transform shape}}]
\addplot[plotStyle1] coordinates {
(2, 16760000.0)
(3, 40054000.0)
(4, 48916000.0)
(5, 69503000.0)
(10, 113798000.0)
(15, 153898000.0)
(20, 183028000.0)
(25, 202372000.0)
};
\addlegendentry{Big-means}
\addplot[plotStyle2] coordinates {
(2, 706563.5)
(3, 1265287.5)
(4, 1743580.0)
(5, 2309376.5)
(10, 5456622.5)
(15, 9781654.5)
(20, 13059448.0)
(25, 17007059.5)
};
\addlegendentry{IK-means}
\addplot[plotStyle3] coordinates {
(2, 1386406.0)
(3, 2287044.0)
(4, 3085392.0)
(5, 4783987.5)
(10, 11524800.0)
(15, 19899825.0)
(20, 26465600.0)
(25, 32646562.5)
};
\addlegendentry{BDCSM}
\addplot[plotStyle4] coordinates {
(2, 148000.0)
(3, 162000.0)
(4, 332000.0)
(5, 390000.0)
(10, 560000.0)
(15, 630000.0)
(20, 760000.0)
(25, 950000.0)
};
\addlegendentry{Minibatch K-means}
\addplot[plotStyle5] coordinates {
(2, 580000.0)
(3, 2262000.0)
(4, 3132000.0)
(5, 4495000.0)
(10, 11600000.0)
(15, 17400000.0)
(20, 31900000.0)
(25, 31175000.0)
};
\addlegendentry{K-means++}
\addplot[plotStyle6] coordinates {
(2, 4088026.5)
(3, 4291394.5)
(4, 4600206.5)
(5, 4809956.0)
(10, 5280161.5)
(15, 4914909.0)
(20, 4775542.0)
(25, 4507370.0)
};
\addlegendentry{CURE}
\addplot[plotStyle7] coordinates {
(2, 4758754.0)
(3, 5579209.0)
(4, 6575360.0)
(5, 8973875.0)
(10, 9892555.0)
(15, 17599120.0)
(20, 30239440.0)
(25, 28914112.5)
};
\addlegendentry{CluDataSE}
\addplot[plotStyle8] coordinates {
(2, 248000.0)
(3, 332000.0)
(4, 428000.0)
(5, 506000.0)
(10, 976000.0)
(15, 1616000.0)
(20, 2096000.0)
(25, 2591000.0)
};
\addlegendentry{LW-coreset}
\end{axis}
\end{tikzpicture}
\caption{Shuttle Control (normalized)}
\label{Fig1Exp2Ds19}
\end{subfigure}%
\begin{subfigure}[b]{0.32\linewidth}
\centering
\begin{tikzpicture}[scale=0.6]
\begin{axis}[xlabel={No of clusters}, ylabel={No of dist. func. eval.}, legend pos=north west, legend style={nodes={scale=0.5, transform shape}}]
\addplot[plotStyle1] coordinates {
(2, 483851660.0)
(3, 570010869.0)
(4, 653084404.0)
(5, 685648751.0)
(10, 790651546.0)
(15, 852552268.5)
(20, 868991726.0)
(25, 791962223.5)
};
\addlegendentry{Big-means}
\addplot[plotStyle2] coordinates {
(2, 119844.0)
(3, 179844.0)
(4, 240032.0)
(5, 300122.5)
(10, 601015.0)
(15, 902655.0)
(20, 1203640.0)
(25, 1505575.0)
};
\addlegendentry{IK-means}
\addplot[plotStyle3] coordinates {
(2, 179758.0)
(3, 404454.0)
(4, 1617768.0)
(5, 3445225.0)
(10, 9886350.0)
(15, 18986347.5)
(20, 34452520.0)
(25, 48495787.5)
};
\addlegendentry{BDCSM}
\addplot[plotStyle4] coordinates {
(2, 614139.0)
(3, 718992.0)
(4, 1258236.0)
(5, 1273215.0)
(10, 2920905.0)
(15, 4718385.0)
(20, 6141390.0)
(25, 6553312.5)
};
\addlegendentry{Minibatch K-means}
\addplot[plotStyle5] coordinates {
(2, 89880.0)
(3, 134820.0)
(4, 179760.0)
(5, 337050.0)
(10, 6890800.0)
(15, 14830200.0)
(20, 24267600.0)
(25, 32394250.0)
};
\addlegendentry{K-means++}
\addplot[plotStyle6] coordinates {
(2, 9290452.0)
(3, 8827285.5)
(4, 9765739.5)
(5, 10790468.5)
(10, 16700335.0)
(15, 22513246.0)
(20, 19968968.0)
(25, 16571751.5)
};
\addlegendentry{CURE}
\addplot[plotStyle7] coordinates {
(2, 64220416.0)
(3, 64611050.5)
(4, 66088628.0)
(5, 67882285.0)
(10, 78441690.0)
(15, 86143832.5)
(20, 106742180.0)
(25, 109993687.5)
};
\addlegendentry{CluDataSE}
\addplot[plotStyle8] coordinates {
(2, 107920.0)
(3, 146900.0)
(4, 185880.0)
(5, 744860.0)
(10, 2859760.0)
(15, 6434660.0)
(20, 9049560.0)
(25, 8104460.0)
};
\addlegendentry{LW-coreset}
\end{axis}
\end{tikzpicture}
\caption{EEG Eye State}
\label{Fig1Exp2Ds20}
\end{subfigure}%
\begin{subfigure}[b]{0.32\linewidth}
\centering
\begin{tikzpicture}[scale=0.6]
\begin{axis}[xlabel={No of clusters}, ylabel={No of dist. func. eval.}, legend pos=north west, legend style={nodes={scale=0.5, transform shape}}]
\addplot[plotStyle1] coordinates {
(2, 274040807.0)
(3, 318947850.0)
(4, 361712896.0)
(5, 376894113.5)
(10, 441416161.0)
(15, 462162081.0)
(20, 458716916.0)
(25, 470100961.0)
};
\addlegendentry{Big-means}
\addplot[plotStyle2] coordinates {
(2, 686195.5)
(3, 939524.5)
(4, 1502599.5)
(5, 2613947.0)
(10, 8682622.5)
(15, 12005836.5)
(20, 18760421.0)
(25, 23082353.0)
};
\addlegendentry{IK-means}
\addplot[plotStyle3] coordinates {
(2, 988624.0)
(3, 1482942.0)
(4, 2426634.0)
(5, 3969490.0)
(10, 14155365.0)
(15, 19885087.5)
(20, 33104410.0)
(25, 40444575.0)
};
\addlegendentry{BDCSM}
\addplot[plotStyle4] coordinates {
(2, 554223.0)
(3, 786397.5)
(4, 1168362.0)
(5, 1647690.0)
(10, 2995800.0)
(15, 4493700.0)
(20, 5392440.0)
(25, 7115025.0)
};
\addlegendentry{Minibatch K-means}
\addplot[plotStyle5] coordinates {
(2, 89880.0)
(3, 516810.0)
(4, 1408120.0)
(5, 2509150.0)
(10, 10111500.0)
(15, 16515450.0)
(20, 21721000.0)
(25, 36513750.0)
};
\addlegendentry{K-means++}
\addplot[plotStyle6] coordinates {
(2, 13661334.0)
(3, 12738416.5)
(4, 12967857.0)
(5, 13892659.0)
(10, 18292681.0)
(15, 22935686.5)
(20, 20565197.5)
(25, 17953463.5)
};
\addlegendentry{CURE}
\addplot[plotStyle7] coordinates {
(2, 65047072.0)
(3, 65699036.5)
(4, 66431698.0)
(5, 69185697.5)
(10, 78289430.0)
(15, 87022757.5)
(20, 99659630.0)
(25, 108331000.0)
};
\addlegendentry{CluDataSE}
\addplot[plotStyle8] coordinates {
(2, 107920.0)
(3, 626900.0)
(4, 697880.0)
(5, 1344860.0)
(10, 4259760.0)
(15, 8234660.0)
(20, 8569560.0)
(25, 10604460.0)
};
\addlegendentry{LW-coreset}
\end{axis}
\end{tikzpicture}
\caption{EEG Eye State (normalized)}
\label{Fig1Exp2Ds21}
\end{subfigure}%
\vskip\baselineskip%
\begin{subfigure}[b]{0.32\linewidth}
\centering
\begin{tikzpicture}[scale=0.6]
\begin{axis}[xlabel={No of clusters}, ylabel={No of dist. func. eval.}, legend pos=north west, legend style={nodes={scale=0.5, transform shape}}]
\addplot[plotStyle1] coordinates {
(2, 388293800.0)
(3, 660987700.0)
(5, 1024130500.0)
(10, 2046385000.0)
(15, 2339624500.0)
(20, 2908174000.0)
(25, 3204598500.0)
};
\addlegendentry{Big-means}
\addplot[plotStyle2] coordinates {
(2, 687204.0)
(3, 1030806.0)
(5, 1718010.0)
(10, 3436020.0)
(15, 5154030.0)
(20, 6872040.0)
(25, 8590050.0)
};
\addlegendentry{IK-means}
\addplot[plotStyle3] coordinates {
(2, 4021848.0)
(3, 11303862.0)
(5, 17160025.0)
(10, 60082000.0)
(15, 102095250.0)
(20, 146359600.0)
(25, 195020000.0)
};
\addlegendentry{BDCSM}
\addplot[plotStyle4] coordinates {
(2, 812000.0)
(3, 1344000.0)
(5, 2380000.0)
(10, 5460000.0)
(15, 8085000.0)
(20, 11060000.0)
(25, 15050000.0)
};
\addlegendentry{Minibatch K-means}
\addplot[plotStyle5] coordinates {
(2, 4810400.0)
(3, 14302350.0)
(5, 18468500.0)
(10, 87618000.0)
(15, 167505000.0)
(20, 269726000.0)
(25, 308166250.0)
};
\addlegendentry{K-means++}
\addplot[plotStyle6] coordinates {
(2, 258238035.5)
(3, 258273938.0)
(5, 262225782.0)
(10, 271164935.5)
(15, 282926083.0)
(20, 294189528.5)
(25, 290787312.5)
};
\addlegendentry{CURE}
\addplot[plotStyle7] coordinates {
(2, 201095007.0)
(3, 207398158.0)
(5, 211970570.0)
(10, 282202520.0)
(15, 343306922.5)
(20, 411252900.0)
(25, 527596025.0)
};
\addlegendentry{CluDataSE}
\addplot[plotStyle8] coordinates {
(2, 1043600.0)
(3, 2193500.0)
(5, 2946300.0)
(10, 9220800.0)
(15, 15950300.0)
(20, 25409800.0)
(25, 29969300.0)
};
\addlegendentry{LW-coreset}
\end{axis}
\end{tikzpicture}
\caption{Pla85900}
\label{Fig1Exp2Ds22}
\end{subfigure}%
\begin{subfigure}[b]{0.32\linewidth}
\centering
\begin{tikzpicture}[scale=0.6]
\begin{axis}[xlabel={No of clusters}, ylabel={No of dist. func. eval.}, legend pos=north west, legend style={nodes={scale=0.5, transform shape}}]
\addplot[plotStyle1] coordinates {
(2, 758174224.0)
(3, 1158469336.0)
(5, 1409947560.0)
(10, 2228183120.0)
(15, 2490098680.0)
(20, 2820654240.0)
(25, 3114849800.0)
};
\addlegendentry{Big-means}
\addplot[plotStyle2] coordinates {
(2, 120900.0)
(3, 181350.0)
(5, 302250.0)
(10, 604520.0)
(15, 906780.0)
(20, 1209040.0)
(25, 1511325.0)
};
\addlegendentry{IK-means}
\addplot[plotStyle3] coordinates {
(2, 254232.0)
(3, 837354.0)
(5, 1075610.0)
(10, 3031320.0)
(15, 6347130.0)
(20, 10703040.0)
(25, 11579050.0)
};
\addlegendentry{BDCSM}
\addplot[plotStyle4] coordinates {
(2, 400000.0)
(3, 600000.0)
(5, 1000000.0)
(10, 2800000.0)
(15, 4080000.0)
(20, 5280000.0)
(25, 5200000.0)
};
\addlegendentry{Minibatch K-means}
\addplot[plotStyle5] coordinates {
(2, 453360.0)
(3, 1858776.0)
(5, 1360080.0)
(10, 8311600.0)
(15, 11560680.0)
(20, 18436640.0)
(25, 23801400.0)
};
\addlegendentry{K-means++}
\addplot[plotStyle6] coordinates {
(2, 76411953.0)
(3, 74047212.0)
(5, 76132357.0)
(10, 80016728.0)
(15, 83950959.0)
(20, 81898931.0)
(25, 80164853.0)
};
\addlegendentry{CURE}
\addplot[plotStyle7] coordinates {
(2, 64424888.0)
(3, 66497248.0)
(5, 66503910.0)
(10, 73238830.0)
(15, 81702040.0)
(20, 94867480.0)
(25, 108255400.0)
};
\addlegendentry{CluDataSE}
\addplot[plotStyle8] coordinates {
(2, 268448.0)
(3, 915560.0)
(5, 1065784.0)
(10, 3141344.0)
(15, 5656904.0)
(20, 7052464.0)
(25, 9208024.0)
};
\addlegendentry{LW-coreset}
\end{axis}
\end{tikzpicture}
\caption{D15112}
\label{Fig1Exp2Ds23}
\end{subfigure}%
\caption{Distance function evaluations. Set 3}
\end{figure}

\end{landscape}

\end{appendices}

\bibliography{main}

\end{document}